\documentclass[10pt,twocolumn,letterpaper]{article}

\usepackage{iccv}
\usepackage{times}
\usepackage{epsfig}
\usepackage{graphicx}
\usepackage{amsmath}
\usepackage{amssymb}
\usepackage{booktabs}
\usepackage{xcolor}
\usepackage{multirow}
\usepackage{caption}
\usepackage{subcaption}
\usepackage{placeins}
\usepackage{enumitem}
\usepackage{tablefootnote}

\usepackage{hyperref}
\hypersetup{pagebackref=true,breaklinks=true,letterpaper=true,colorlinks,bookmarks=false}

\usepackage[capitalize]{cleveref}
\crefname{section}{Sec.}{Secs.}
\Crefname{section}{Section}{Sections}
\Crefname{table}{Table}{Tables}
\crefname{table}{Tab.}{Tabs.}

\newcommand{\nnscores}{\mathbf{S}^{\text{N}}}
\newcommand{\parscores}{\mathbf{S}^{\text{P}}}
\newcommand{\combscores}{\mathbf{S}^{\text{C}}}

\newcommand\blfootnote[1]{%
  \begingroup
  \renewcommand\thefootnote{}\footnote{#1}%
  \addtocounter{footnote}{-1}%
  \endgroup
}

\iccvfinalcopy 

\def\iccvPaperID{7} 

\ificcvfinal\fi

\begin{document}

\title{Far Away in the Deep Space:\\Dense Nearest-Neighbor-Based Out-of-Distribution Detection}

\author{Silvio Galesso\\
University of Freiburg\\
{\tt\small galessos@cs.uni-freiburg.de}
\and
Max Argus\\
University of Freiburg\\
{}
\and
Thomas Brox\\
University of Freiburg\\
{}
}
\maketitle

\begin{abstract}
The key to out-of-distribution detection is density estimation of the in-distribution data or of its feature representations.
This is particularly challenging for dense anomaly detection in domains where the in-distribution data has a complex underlying structure. 
Nearest-Neighbors approaches have been shown to work well in object-centric data domains, such as industrial inspection and image classification.
In this paper, we show that nearest-neighbor approaches also yield state-of-the-art results on dense novelty detection in complex driving scenes when working with an appropriate feature representation. 
In particular, we find that transformer-based architectures produce representations that yield much better similarity metrics for the task. We identify the multi-head structure of these models as one of the reasons, and demonstrate a way to transfer some of the improvements to CNNs.
Ultimately, the approach is simple and non-invasive, i.e., it does not affect the primary segmentation performance, refrains from training on examples of anomalies, and achieves state-of-the-art results on RoadAnomaly, StreetHazards, and SegmentMeIfYouCan-Anomaly.\blfootnote{\href{https://github.com/silviogalesso/dense-ood-knns}{{https://github.com/silviogalesso/dense-ood-knns}}}

\end{abstract}



\section{Introduction}

\begin{figure}[h!]
    \centering
    \begin{subfigure}[b]{0.15\textwidth}
        \centering
        \includegraphics[width=\textwidth]{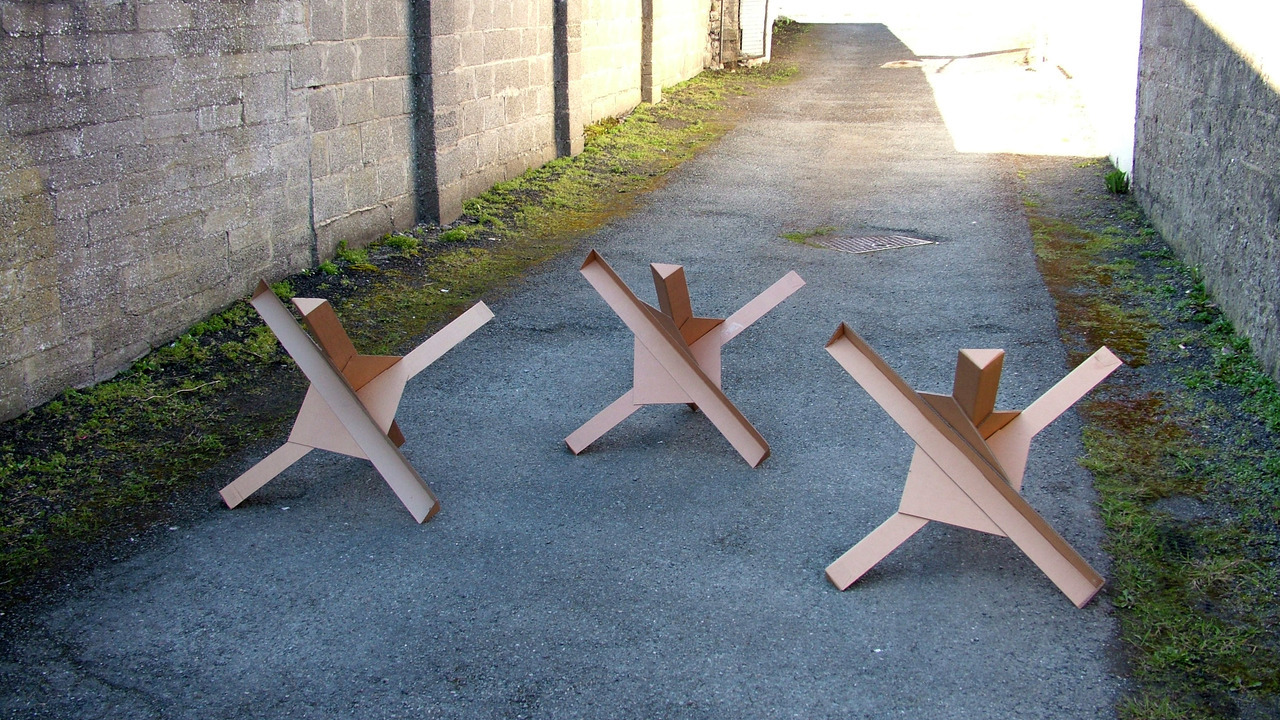}
    \end{subfigure}
    \begin{subfigure}[b]{0.15\textwidth}
        \centering
        \includegraphics[width=\textwidth]{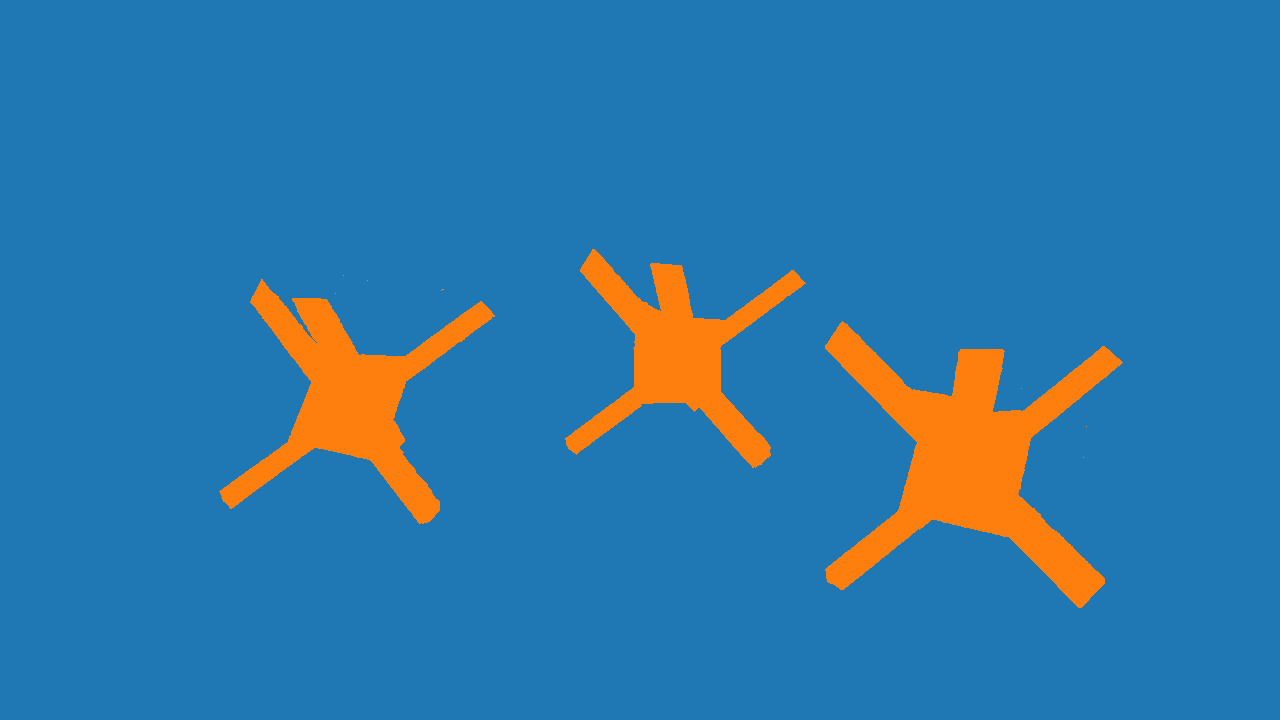}
    \end{subfigure}
    \begin{subfigure}[b]{0.15\textwidth}
        \centering
        \includegraphics[width=\textwidth]{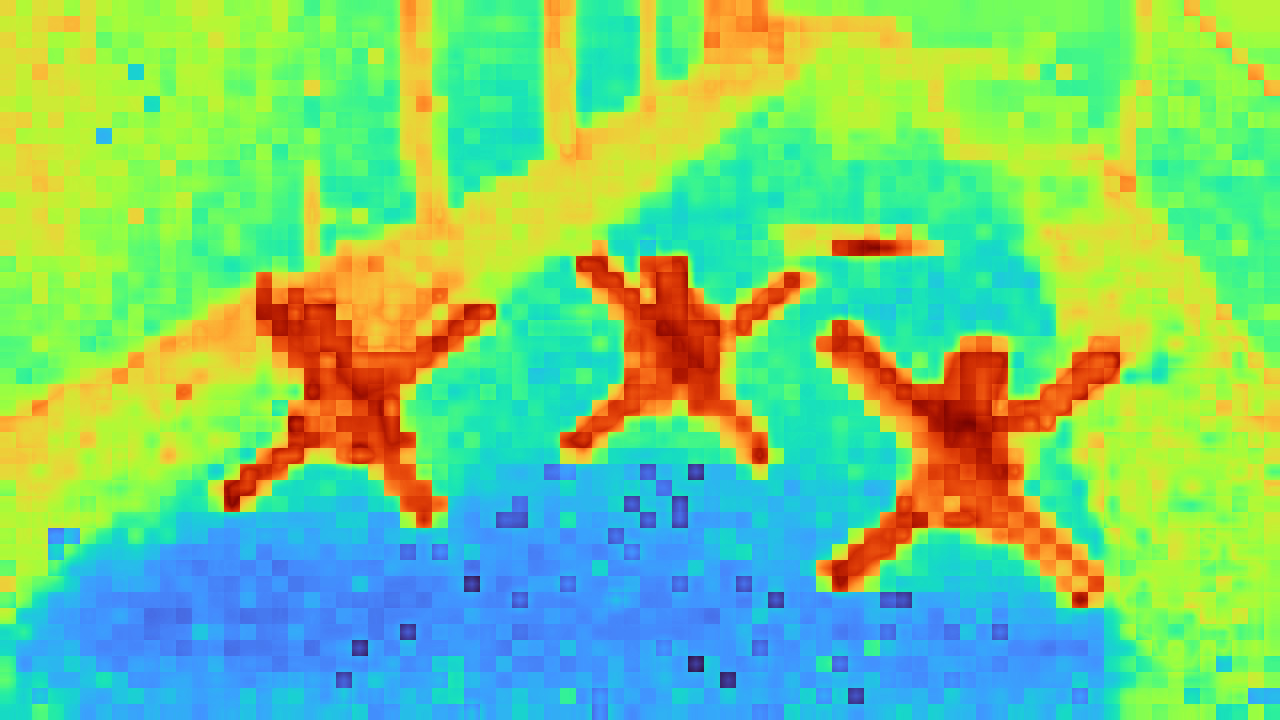}
    \end{subfigure}
    
    \begin{subfigure}[b]{0.15\textwidth}
        \centering
        \includegraphics[width=\textwidth]{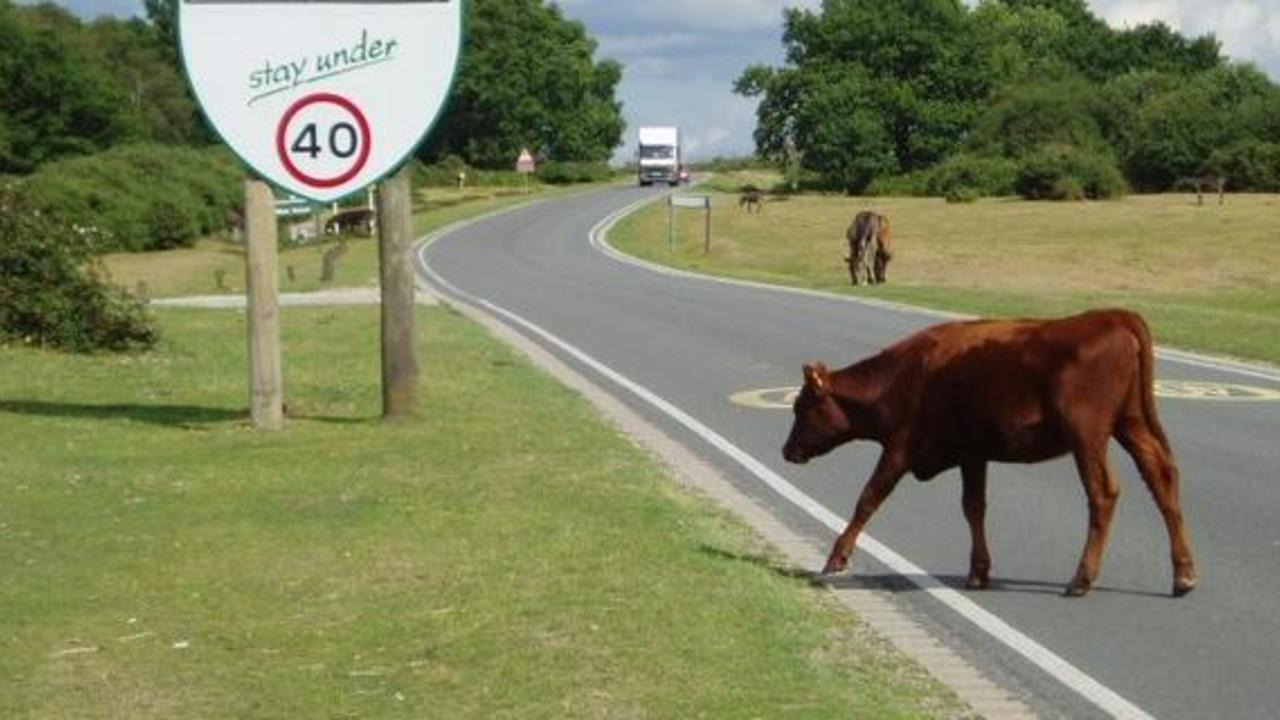}
    \end{subfigure}
    \begin{subfigure}[b]{0.15\textwidth}
        \centering
        \includegraphics[width=\textwidth]{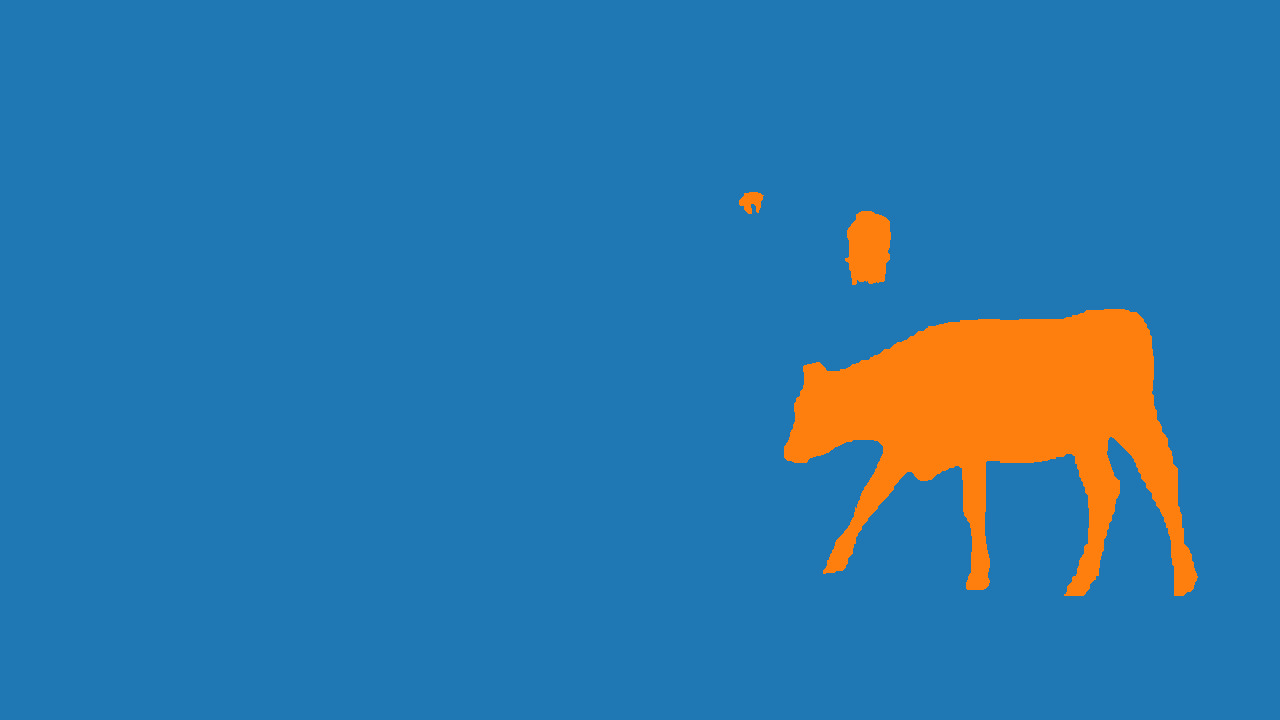}
    \end{subfigure}
    \begin{subfigure}[b]{0.15\textwidth}
        \centering
        \includegraphics[width=\textwidth]{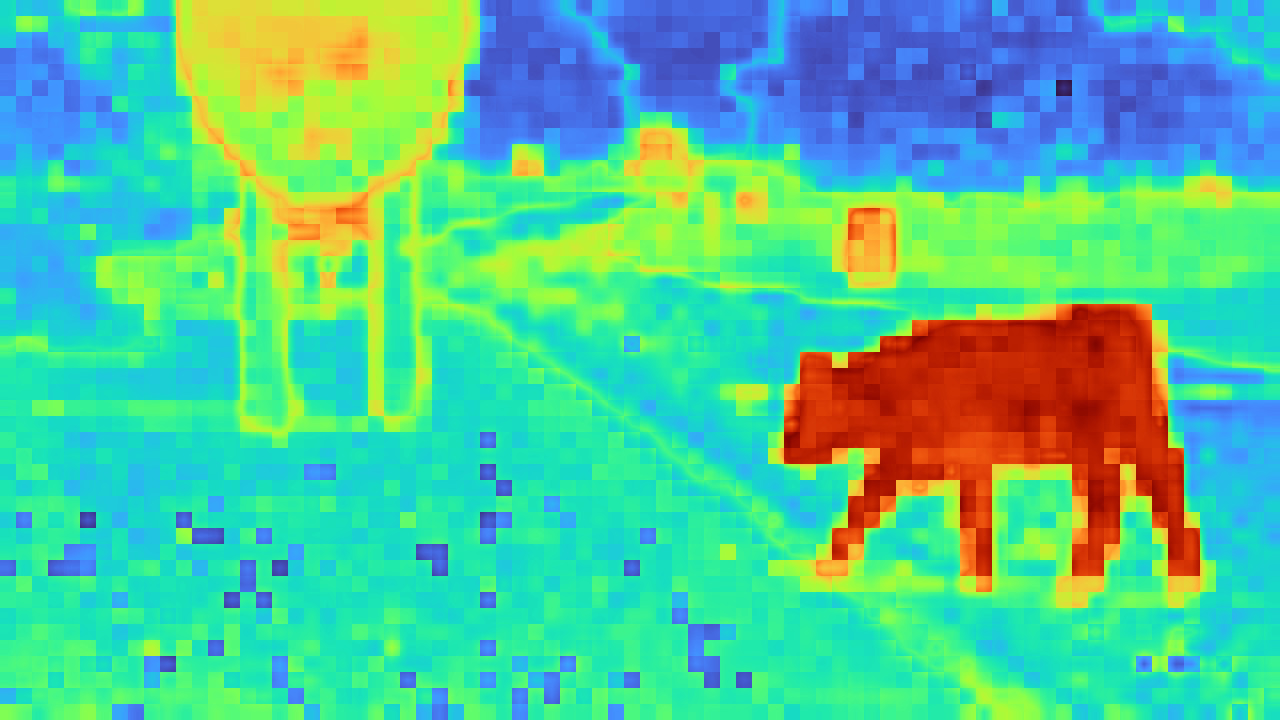}
    \end{subfigure}
    
    \begin{subfigure}[b]{0.15\textwidth}
        \centering
        \includegraphics[width=\textwidth]{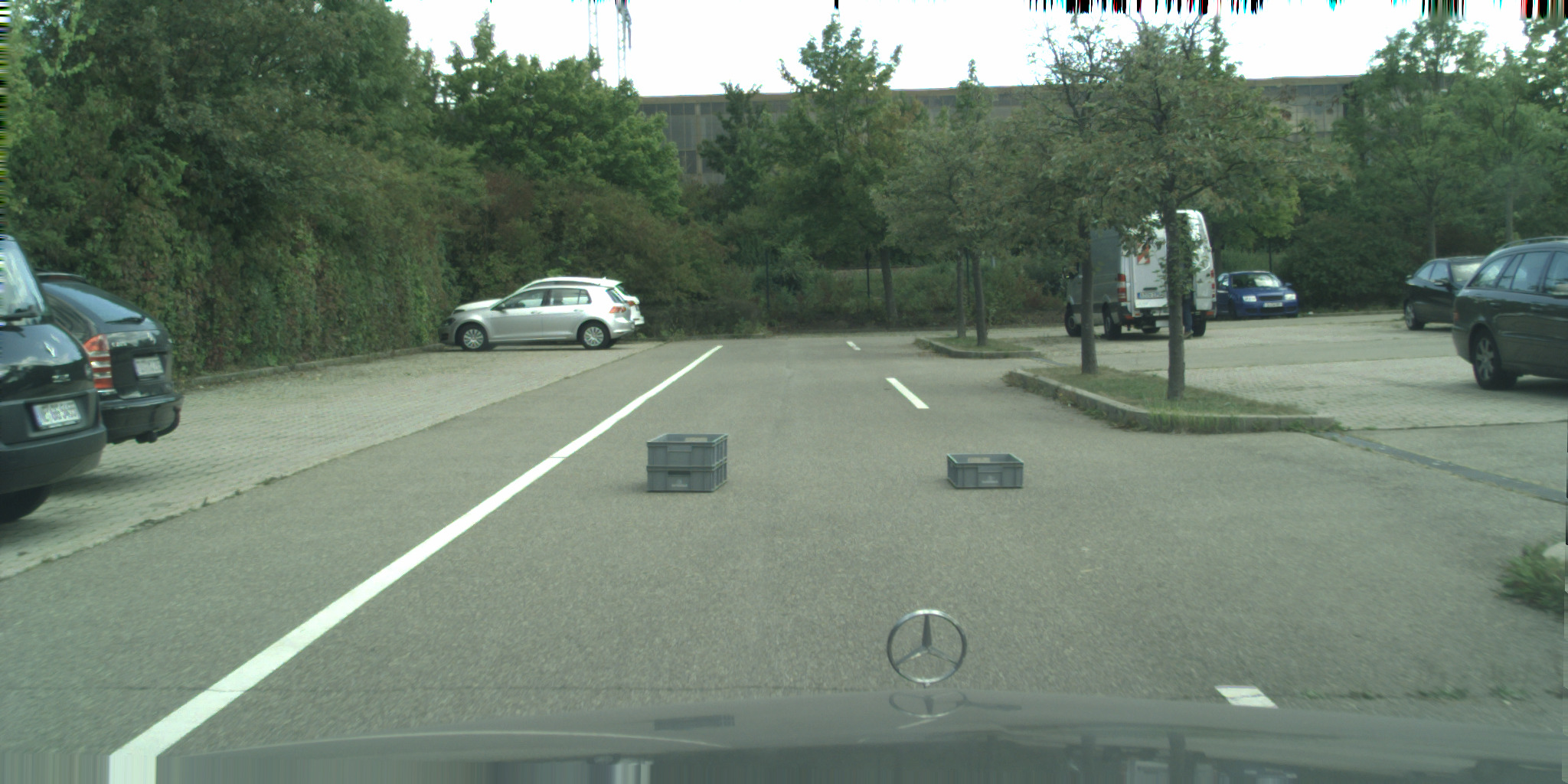}
    \end{subfigure}
    \begin{subfigure}[b]{0.15\textwidth}
        \centering
        \includegraphics[width=\textwidth]{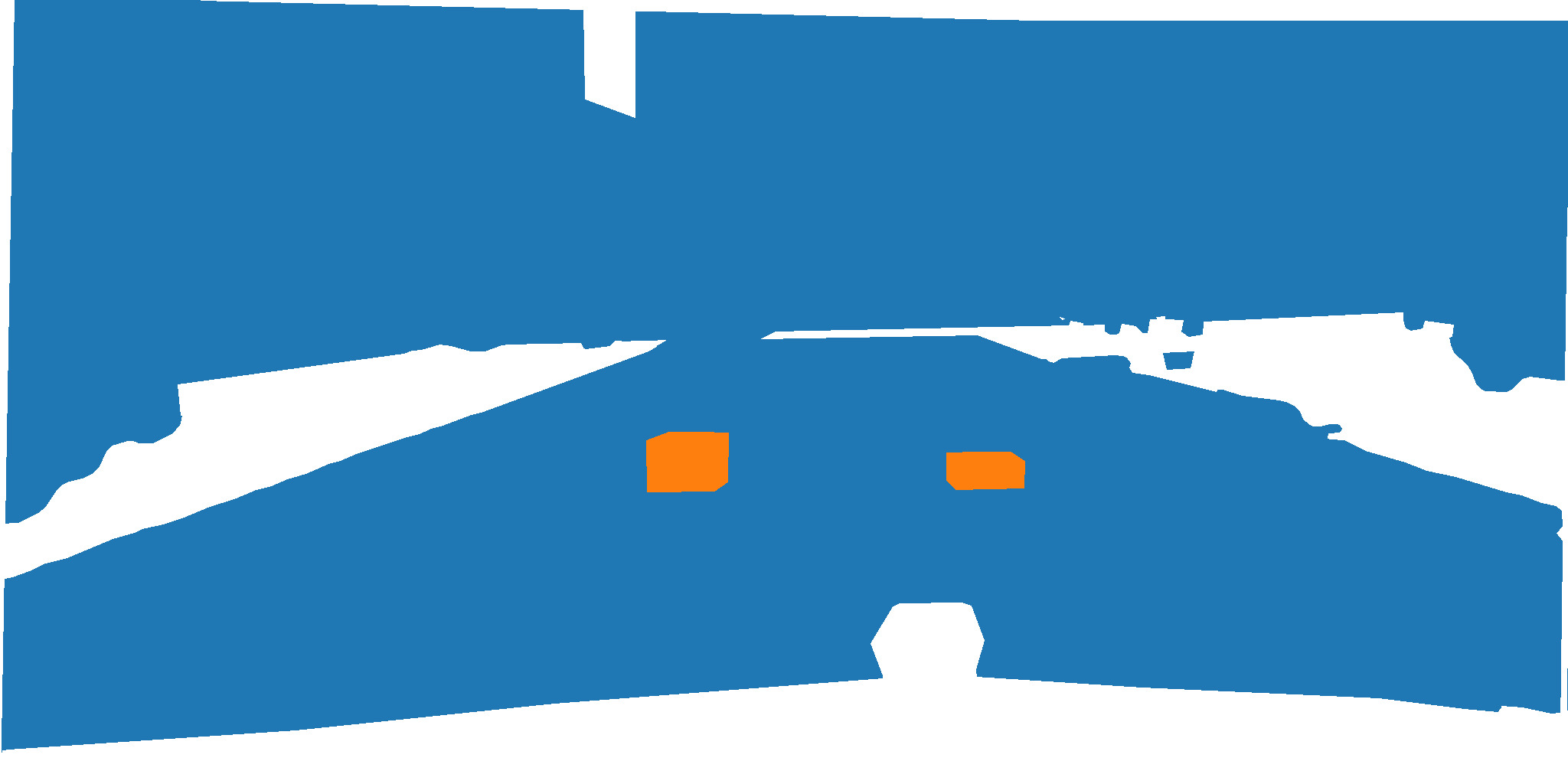}
    \end{subfigure}
    \begin{subfigure}[b]{0.15\textwidth}
        \centering
        \includegraphics[width=\textwidth]{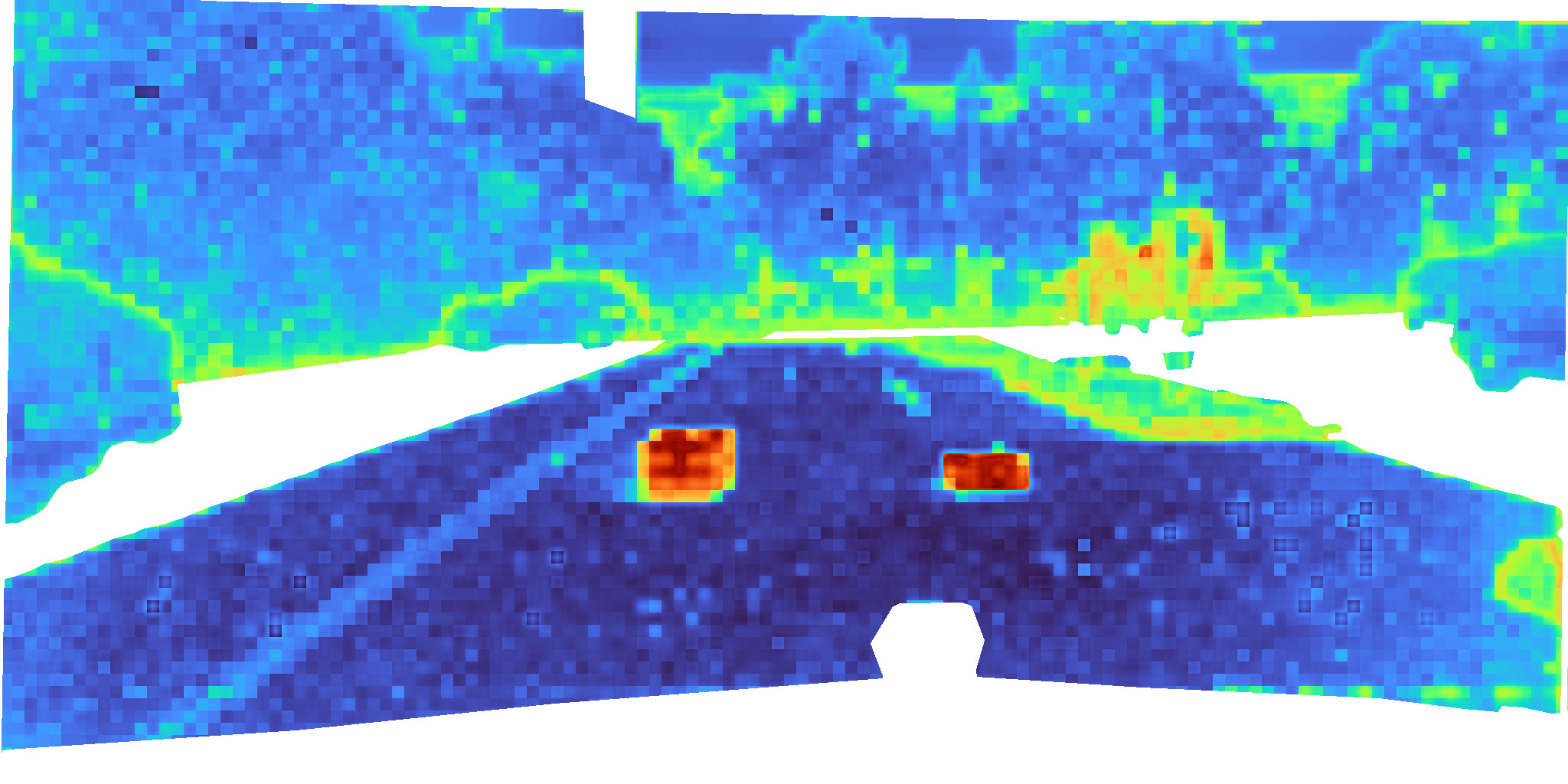}
    \end{subfigure}
    
    \begin{subfigure}[b]{0.15\textwidth}
        \centering
        \includegraphics[width=\textwidth]{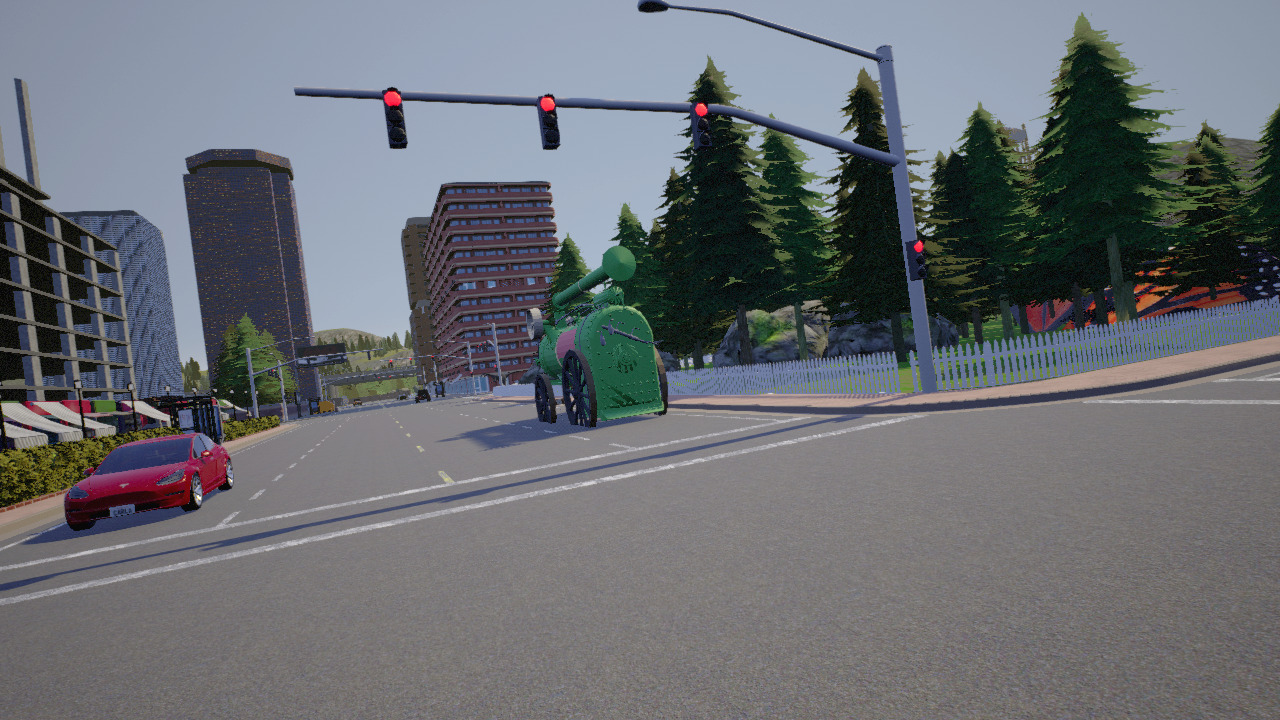}
        \caption*{Image}
    \end{subfigure}
    \begin{subfigure}[b]{0.15\textwidth}
        \centering
        \includegraphics[width=\textwidth]{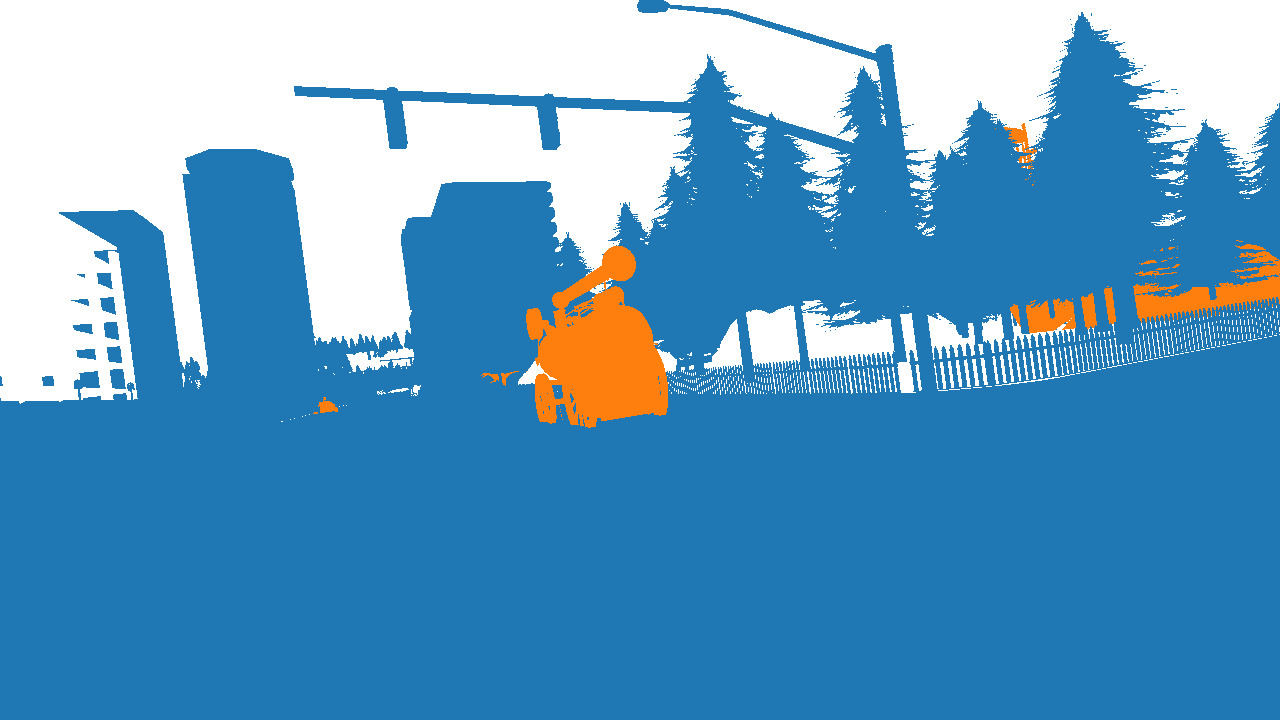}
        \caption*{Ground truth}
    \end{subfigure}
    \begin{subfigure}[b]{0.15\textwidth}
        \centering
        \includegraphics[width=\textwidth]{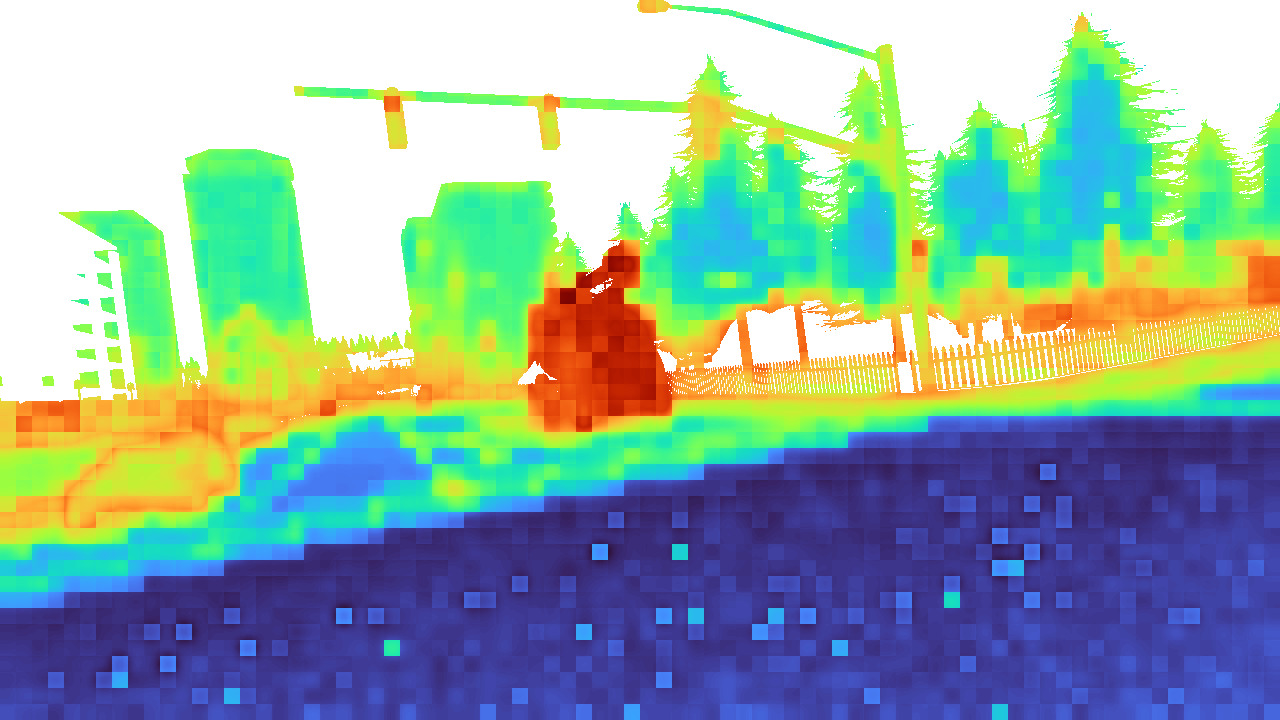}
        \caption*{OoD scores}
    \end{subfigure}
    \caption{Our approach uses a combination of conventional parametric anomaly detection and k-nearest-neighbors. The resulting anomaly scores can be used to identify semantic anomalies and obstacles (orange in the ground truth) in road scenes and achieve state-of-the-art anomaly detection performance on common benchmarks such as RoadAnomaly, StreetHazards, and SegmentMeIfYouCan-Anomaly.}
    \label{fig:teaser}
\end{figure}

Deep learning models can achieve remarkably good performance on a large number of tasks. However, when these models are evaluated on data outside of the training distribution, their performance usually deteriorates substantially~\cite{hendrycks2018benchmarking}. Even worse, the models often do not realize that they are out of distribution and make wrong predictions with high confidence~\cite{guo2017calibration}. For the safe deployment of machine learning systems in the open world, where there is no control over the distribution of the input data, the ability of a model to detect out-of-distribution (OoD) samples becomes crucial.
Since such a system should be able to identify \emph{all} unforeseen deviations from the training data, it cannot learn the distribution of the novel samples, but must base its decision on a model of the inlier distribution. This makes novelty detection a particularly challenging task, with an increasing number of research contributions in the last years.

In this work, we focus on segmentation datasets for autonomous driving, and we aim to detect and localize objects of unknown categories in the image. This requires spatially resolved outputs, rather than an accumulated decision for the whole image. Moreover, driving scenes comprise diverse patterns and multiple objects (see Figure~\ref{fig:teaser}), which makes modelling this complex inlier distribution challenging. The problem has been approached by several works already~\cite{Jung_2021_ICCV, Tian2021, Xia2020, Grcic2022}, and a number of accepted evaluation benchmarks exist~\cite{a_benchmark, blum2019fishyscapes, segmentmeifyoucan2021}.

Inspired by the success of non-parametric nearest-neighbor methods in the scope of industrial anomaly detection~\cite{SPADE, reiss2021panda, patchcore} and image recognition~\cite{sun2022knnood}, we investigate if and how these approaches can be modified to detect anomalous objects in driving scenes. The settings are very different.
In industrial inspection, objects are observed under similar conditions, which limits the scope of the in-distribution data, which is advantageous for nearest-neighbors. 
In OoD image recognition, while the data is more diverse than in industrial inspection, it is still object-centric, and decisions can be based on the global embedding vector.
Novelty detection in driving scenes must deal with diverse in-distribution data and decisions must be made locally. 
Segmentation embeddings are local, and therefore spatially correlated, have high intra-class diversity, and are known to be prone to false positives at class boundaries~\cite{Jung_2021_ICCV}.
Consequently, non-parametric approaches appear unfit to model the in-distribution data effectively, and indeed to the best of our knowledge no successful nearest-neighbor approach has so far been proposed in this setting.

Motivated by recent work showing the outstanding properties of attention-based representations~\cite{caron2021emerging,melas2022deep,zadaianchuk2022unsupervised}, we apply k-nearest-neighbor based OoD detection on the representations produced by transformers models for semantic segmentation.
As a key insight of this work, we find that transformer representations are a game changer for dense anomaly detection in driving scenes, outperforming their convolutional counterparts and state-of-the-art approaches. 
In particular, we find that the multiple heads of the transformer architecture play an important role. We show further evidence of the benefit of multiple heads by extending the idea to CNNs, improving their performance on OoD detection. We see a connection between this finding and the theoretical argument that nearest-neighbor approaches should fail in high-dimensional features spaces (``curse of dimensionality''). Nearest-neighbor approaches have been shown to work well in other settings~\cite{patchcore, sun2022knnood}, but in complex road scenes they need the right representations to succeed.

To summarize, our contributions are the following:
\begin{itemize}
    \item We show that k-nearest-neighbors (kNNs) with deep supervised segmentation features can achieve state-of-the-art performance on all common benchmarks (StreetHazards, RoadAnomaly, SegmentMeIfYouCan, FS Lost\&Found), at a favorable computational tradeoff and \emph{without} training on out-of-distribution data.
    \item In contrast to~\cite{sun2022knnood}, we investigate and find major differences in the performance of several feature encoders. We find attention-based models to be substantially and consistently superior to CNNs.
    \item We investigate the effects of the ``curse of dimensionality'' on the performance of kNNs and identify the multi-head design of transformers as a key to their success, demonstrating its beneficial effects on CNNs too.
\end{itemize}





\section{Related Work}

\subsection{Out-of-Distribution (OoD) Detection}
In computer vision, detecting anomalous patterns is a task with several applications. One is visual inspection in industrial manufacturing, with the goal of identifying production defects~\cite{mvtec,Zou2022}. In this scenario, many examples of healthy items are easily available, whereas the distribution of the possible faults is unknown. The aim is to identify the presence and location of the defect~\cite{SPADE, reiss2021panda, patchcore}. 

A more academic evaluation setting for OoD detection is image classification~\cite{hendrycks2021natural, zhang2020hybrid, yang2022openood}, where the normal data distribution is divided into classes, which can in turn have nuanced appearances and subtypes. The goal is to reliably identify images that do not belong to the semantic classes of a training set, such as CIFAR10~\cite{krizhevsky2009learning} or ImageNet~\cite{deng2009imagenet}. For this purpose several scoring functions have been proposed, mostly based on discriminative parametric models~\cite{mahalanobis, openhybrid, huang2021mos, sun2021react, huang2021on, haoqi2022vim, morteza2022provable}, optionally aided by the use of third-party outliers (outlier exposure)~\cite{hendrycks2018deep, NEURIPS2020_f5496252, pmlr-v162-ming22a, pmlr-v162-katz-samuels22a}.

While anomalous image recognition is an essential research problem, safety-critical real world applications such as autonomous driving deal with complex multi-object scenes and require accurate localization of unrecognized objects~\cite{a_benchmark, blum2019fishyscapes, segmentmeifyoucan2021}. This is the setting we address in this work, where we aim to identify the individual pixels that correspond to unknown entities.
As in image recognition, the data includes several semantic categories; however, each sample does not correspond to a whole object but to part of an object in relation with its context. 
Novelty detection in semantic segmentation has been studied less, most methods rely on scoring functions operating on the output of pre-trained segmentation models~\cite{a_benchmark, Jung_2021_ICCV, moose, cen2021deep, liang2022gmmseg}. Recent approaches~\cite{Grcic_2023_CVPR, RbA} make clever use of the mask transformer architecture~\cite{mask2former} for segmentation, which decouples mask and class predictions, for improved OoD detection. 

A trend in recent works for dense OoD detection is to use of outlier exposure~\cite{Chan2021EntropyMA, Tian2021, Grcic2022}, mentioned above.

\subsection{OoD Detection with k-Nearest-Neighbors}
Unknown patterns in the inference data can be detected by retrieval and comparison with available in-distribution samples. The particular nature and structure of the in-distribution data determines the difficulty and scalability of the task, i.e. how many in-distribution samples are needed to effectively represent the data distribution and distinguish it from anomalous entities, and how easy it is to compute a suitable feature representation.

Retrieval based approaches are very successful at detecting defects in products~\cite{SPADE, reiss2021panda, patchcore, Zou2022}, precisely because they can rely on abundant images of healthy samples with little variation in appearance~\cite{mvtec}. These methods use various types of learned deep features in combination with kNNs.

A kNN-based method has recently been proven successful in the image recognition setting, too~\cite{sun2022knnood}. Here the training data is more diverse, sometimes featuring 1k categories, but methods can rely on class labels and object-centric, single-instance images, which help the feature extraction process~\cite{demystifying_contrastive}. The evaluation setting in this line of research often uses different datasets for in- and out-of-distribution samples~\cite{huang2021mos,sun2022knnood}.

In dense out-of-distribution detection, each pixel is a sample and a potential anomaly.
High correlation with surrounding context further complicates the task, since anomalous pixels are correlated to nearby in-distribution pixels~\cite{a_benchmark, Jung_2021_ICCV}.
The outliers are not from another dataset, but part of the scene, coexisting and interacting with the in-distribution objects.
The training data ontologies have fewer categories~\cite{Cordts2016Cityscapes, a_benchmark}, but higher intra-class variation. Results for an embedded kNN approach similar to ours are reported in~\cite{blum2019fishyscapes}, but without the right representation choice its performance is inferior to most baselines.
\section{Deep Neighbor Proximity for OoD Detection}
At the core of our approach for out-of-distribution detection are deep k-Nearest-Neighbors.
As illustrated in Figure~\ref{fig:teaser}, our method relies on the computation of distances between feature representations produced by the encoder of a semantic segmentation network. 
At test time we collect the distances between the local representation maps of the test sample and a library of \textit{reference features} obtained from the in-distribution dataset -- i.e. the training set for the segmentation network.

More formally, consider a matrix of in-distribution reference features as $\mathbf{R}\in\mathbb{R}^{N{\times}C}$, where $N$ is the number of reference features and $C$ is the dimensionality of each feature vector. The computation of $\mathbf{R}$ will be described in Sections~\ref{sec:method_feature_selection} and ~\ref{sec:method_subsampling}.
For a test image, we extract the feature representation $\mathbf{T}\in\mathbb{R}^{H{\cdot}W{\times}C}$, and ``flatten" it in the spatial dimensions $H, W$.
We first compute the matrix $\mathbf{D}\in\mathbb{R}^{H{\cdot}W{\times}N}$ of distances between each possible combination of samples $\mathbf{t}$ and $\mathbf{r}$ in the feature sets:
\begin{equation}
    d_{i,j} = {dist}({t}_{j}, {r}_{i}) \hspace{10px} \forall ~ j\in\{1..H{\cdot}W\},~ i\in\{1..N\}
\end{equation}
where $dist$ is the euclidean distance.
Then, for each test feature $j$ we compute the OoD score as the average of the distances to the closest $k$ neighbors:
\begin{equation}
    s_{j}^{\text{N}}=\frac{1}{k}\cdot\min_{\substack{D'_i\subset D_i\\|D'_i|=k}}\hspace{5px}\sum_{d\in D'_i} d,
\label{eq:avg_nn_scores}
\end{equation}
and successively reshape them into the original feature shape, to obtain: $\mathbf{S}^{\text{N}}\in\mathbb{R}^{H{\times}W}$.

Although different distance functions were tried, $L_2$ was found to perform best. See Appendix for details.

\subsection{Out-of-Distribution Scores}
\label{sec:methods_ood_scores}
The distances obtained with the procedure described above can be directly used as anomaly scores, however they can also be combined with those obtained from the model predictions (i.e. \textit{parametric} scores).

In order to combine $\nnscores$ with the parametric scores $\parscores$, we first bring them to the same resolution, by upsampling $\nnscores$ to the original image size.
Subsequently, we simply scale both to the same range using their respective extrema estimated on the training set:
\begin{align}
    & \overline{\nnscores} = \nnscores/ \max{\nnscores_{\text{train}}} \\
    & \overline{\parscores} = (\parscores - \min{\parscores_{\text{train}}})/(\max{\parscores_{\text{train}}} - \min{\parscores_{\text{train}}})
\end{align}
and finally compute the combined scores:
\begin{equation}
    \combscores = \overline{\nnscores}+\overline{\parscores}.
\end{equation}

In the following text, we will use the abbreviations DNP (Deep Neighbor Proximity) and cDNP (combined Deep Neighbor Proximity) to refer to the approaches and results based on $\nnscores$ and $\combscores$ respectively.

In terms of parametric scores we considered all the best options in recent literature for dense OoD detection. While more results are available in the Appendix, here we report those obtained with \textit{LogSumExp} operator, which performed slightly better. LogSumExp can be interpreted as an energy functional built on a discriminative model's logits~\cite{Grathwohl2020Your,Tian2021}.

\subsection{Model Architectures and Feature Extraction}
\label{sec:archs}
To assess the versatility of our method, we apply it to four different feature extraction architectures: ResNet~\cite{he2016deep}, ConvNeXt~\cite{liu2022convnet}, MiT~\cite{xie2021segformer} and ViT~\cite{dosovitskiy2020image}.

\textbf{ResNet} and \textbf{ConvNeXt} are both CNNs consisting of a cascade of 4 computational stages. We can extract convolutional features at the end of each stage. Earlier stage features have higher resolution but less semantic content. Both are designed such that the 3rd stage contains more internal layers than the others stages.

\textbf{MiT} is also a hierarchical 4-stage architecture, but it uses alternating multi-head self-attention blocks and convolutional layers. Here we can test the representations from the output of each stage, as above, but also the internal features of the self-attention mechanism: queries, keys, and values.

\textbf{ViT} is a ``pure" transformer, as it is entirely composed of self-attention blocks that output a constant number of patch features, corresponding to a constant resolution. For this architecture we test the features taken from the output of different transformer blocks, as well as the queries, keys, and values produced by the attention mechanism.

In practice we use the networks above as encoders in semantic segmentation models, and thereby learn the feature embeddings as part of the standard supervised segmentation training procedure.
For ResNet and ConvNeXt we use UPerNet~\cite{xiao2018unified}, for MiT we use SegFormer~\cite{xie2021segformer}, and for ViT we use Segmenter~\cite{strudel2021segmenter} and SETR~\cite{zheng2021rethinking}, following established practices in semantic segmentation literature.

\textbf{Feature Selection:}
\label{sec:method_feature_selection}
For each encoder architecture there are many choices of features to extract for computing neighbor distances, from different depth levels to different functional layers. We evaluate the ones described above for each model type, so as to identify the most suitable for the task, and present the results in Section \ref{sec:ablation_fts}.

\subsection{Reference Feature Subsampling}
\label{sec:method_subsampling}
In order to have a tractable amount of reference features we sub-sample the representations obtained from the training set.
For this stage we evaluate three options: random sub-sampling, greedy coreset reduction (GCS), and per-class greedy coreset reduction (PC-GCS). GCS has been used by PatchCore~\cite{patchcore}, and consists in a 
greedy selection procedure aiming to find the subset of $\mathbb{R}$ with the closest solution to the nearest neighbor problem.
In~\cite{patchcore}, GCS is found to greatly outperform random sampling.

PC-GCS is our proposed variant of GCS applied separately to each category present in the segmentation dataset. PC-GCS makes sense in this setting because industrial inspection images, on which PatchCore is originally applied, are single-class and less diverse in appearance than the segmentation data we use. An application to coherent sub-components, such as semantic categories, is closer to its original intended scenario. PC-GCS also preserves the balance between classes of the original dataset.

\section{Experiments}
\label{sec:experiments}
Our evaluation investigates the influence of: the choice of features (\ref{sec:ablation_fts}), subsampling strategy (\ref{sec:ablation_subsampling}), and number of neighbors (\ref{sec:ablation_k}). Section~\ref{sec:baseline}, shows how our proposed nearest-neighbor-based approaches (Section~\ref{sec:methods_ood_scores}) perform in comparison with the parametric baseline, and finally how they compare with current state-of-the-art approaches in Section \ref{sec:sota}. 
Unless otherwise stated, we use $N=100$k reference features, and $k=3$ neighbors.

\subsection{Evaluation Benchmarks and Metrics}
Several benchmarks for the evaluation of dense OoD performance exist, and while they all revolve on semantic segmentation data for autonomous driving, they are quite different in nature. 

\textbf{StreetHazards}~\cite{a_benchmark} is a synthetic dataset and benchmark, featuring 12 in-distribution categories in the annotated training/validation sets, and 250 diverse OoD objects, annotated as one category in the test set. Its size (1500 test images), variety of OoD objects and locations makes it an important benchmark for research.
\textbf{RoadAnomaly}~\cite{lis2019detecting} is a benchmark made of images downloaded from the web, which features objects, such as animals or vehicles, with categories alien to the typical driving ontology e.g. Cityscapes~\cite{Cordts2016Cityscapes} or BDD100k~\cite{yu2018bdd100k}.
\textbf{SegmentMeIfYouCan - Anomaly}~\cite{segmentmeifyoucan2021} is an extension of RoadAnomaly, containing mostly images for which the OoD ground truth is undisclosed.
\textbf{Fishyscapes Lost\&Found}~\cite{blum2019fishyscapes} is a benchmark for the detection of road obstacles (e.g. lost cargo, small objects), originally designed as an extension of Cityscapes.

Models trained on Cityscapes are used for all benchmarks, except for StreetHazards, which comes with its own training set.
Our ablation experiments use RoadAnomaly and StreetHazards. We use the same model checkpoints for all Cityscapes-based benchmarks.

The most important metric for the task at hand is the Average Precision (\textbf{AP}), which is a holistic metric, averaged over several threshold values. Secondly, we report the False Positive Rate at 95\% True Positive Rate (\textbf{FPR}$_{95}$), which measures the performance at a high detection threshold - relevant for safety critical applications.

\subsection{Training the Feature Extractors}
We follow standard training procedures for each of our semantic segmentation models, optimizing for the cross entropy objective using exclusively the respective training dataset.
We use the \texttt{mmsegmentation}~\cite{mmsegmentation} framework and adhere to the default optimization settings when available (see Appendix).
For comparability we select network snapshots based on segmentation performance -- i.e. after full convergence -- even though this might negatively impact OoD detection results. In fact, OoD detection performance is observed to vary greatly over the epochs, and tends to reach its peak before full convergence of the segmentation loss, before declining again due to overconfidence~\cite{guo2017calibration}.

All encoders are initialized with the respective publicly available ImageNet~\cite{deng2009imagenet} pre-trained parameters. For fair comparison, on our ViT models we use the DeiT~\cite{deit} weights, instead of the original ones trained on a larger undisclosed dataset.

\subsection{Choosing the Best Reference Features}
\label{sec:ablation_fts}

The efficacy of our approach with different feature representations is evaluated here, as anticipated in Section~\ref{sec:method_feature_selection}.
For both CNN architectures, we observe the best suited representations are those extracted at the third stage (out of four). These features likely strike the right balance between resolution and semantic abstraction.

For the transformer encoders, MiT and ViT, which include self-attention layers, from which \textit{query}, \textit{key} and \textit{value} representations can be extracted, as well as ``end-of-the-block'' representations~\cite{NIPS2017_3f5ee243}.
Our results in the Appendix indicate a clear superiority of the self-attention features, which perform approximately equally and outperform the block features by $\sim$20\% and $\sim$55\% for ViT and MiT respectively. Based on performance, we choose the last layer representations for both transformer-based encoders.

In the following sections, we will use the \textit{keys} as default features for transformer backbones, in accordance with other works~\cite{LOST}, but we observed no clear superiority of one over the other.

\subsection{Comparing Parametric Scores, DNP, and cDNP}
\label{sec:baseline}
The three out-of-distribution scores defined in Section~\ref{sec:methods_ood_scores}, i.e. the model's own parametric scores (LogSumExp) and the nearest-neighbor based approaches: DNP and cDNP are compared in Table~\ref{tab:arch_scores_comparison} for the four feature extractors on RoadAnomaly and StreetHazards.
Consistently for all benchmarks, architectures, and metrics, cDNP performs better than its two component scores LogSumExp and DNP. However, the extent of its superiority changes for different architectures and is much greater for the attention-based encoders,  particularly for ViT. 

In fact, while LogSumExp is better than DNP for ResNet, for the other models the performance of DNP -- the pure kNN method -- is superior to that of LogSumExp, and performs almost as well as cDNP in the case of ViT. The performance gains of cDNP are particularly noticeable in terms of FPR$_{95}$, where they are substantial for all architectures.

\begin{table}
\scriptsize
\centering
\setlength\tabcolsep{6pt} 
\begin{tabular}{llcccc}
\toprule
                                                                             &        & \multicolumn{2}{c}{RoadAnomaly} & \multicolumn{2}{c}{StreetHazards}  \\
Model                                                                        & Method & AP$\uparrow$   & FPR$_{95}\downarrow$               & AP $\uparrow$  & FPR$_{95}\downarrow$          \\
\midrule
\multirow{3}{*}{\begin{tabular}[c]{@{}l@{}}UPerNet\\ResNet50\end{tabular}}   & LogSumExp     & 28.58 & 62.98      &     23.87 & 19.31  \\
                                                                             & DNP     & 27.81 & 54.72      &     15.30 & 25.88  \\
                                                                             & cDNP    & \underline{33.55} & \underline{42.14}  &     \underline{25.09} & \underline{14.30} \\
\midrule                                                                             
\multirow{3}{*}{\begin{tabular}[c]{@{}l@{}}UPerNet\\ConvNeXt-T\end{tabular}} & LogSumExp     &  40.04	& 59.43        & 15.79 & 25.19 \\ 
                                                                             & DNP     &  40.89 & \underline{40.02}         & 20.30 & 18.66 \\
                                                                             & cDNP    &  \underline{46.74} & 41.72         & \underline{26.55} & \underline{13.94} \\
\midrule
\multirow{3}{*}{\begin{tabular}[c]{@{}l@{}}SegFormer\\MiT-B2\end{tabular}}   & LogSumExp     & 69.62  & 27.74           & 16.11 & 25.33 \\
                                                                             & DNP     & 72.59  & 18.38            & 37.33 & 20.44 \\
                                                                             & cDNP    & \underline{77.92}  & \textbf{15.97}            & \underline{37.44} & \underline{16.75} \\
\midrule
\multirow{3}{*}{\begin{tabular}[c]{@{}l@{}}Segmenter\\ViT-S\end{tabular}}    & LogSumExp     & 56.39	& 34.54              & 18.16 & 28.94 \\
                                                                             & DNP     & 79.47	& 19.75              & \textbf{45.69} & 18.37 \\
                                                                             & cDNP    & \textbf{79.78}	& \underline{18.18}               & 43.89 & \textbf{15.69} \\  
\bottomrule
\end{tabular}
\caption{Comparison between the the parametric (LogSumExp) baseline, DNP, and their combination cDNP, for four encoder/decoder architectures, on RoadAnomaly and StreetHazards. Best results per model are underlined, best overall are bold. DNP/cDNP always outperform LogSumExp, cDNP is the best approach overall.}
\label{tab:arch_scores_comparison}
\end{table}

\subsection{Comparison to the State of the Art}

\label{sec:sota}
A comparison with the current state-of-the-art on RoadAnomaly, StreetHazards, SegmentMeIfYouCan (SMIYC) Anomaly, and Fishyscapes Lost\&Found is shown in the Tables~\ref{tab:sota_all}(a-c).

While our comparison is focused on methods that do not use outlier exposure for training (OE) -- as this technique breaks the interpretation of out-of-distribution data as completely unknown -- for completeness we include results for PEBAL~\cite{Tian2021} and DenseHybrid~\cite{Grcic2022}, two recent approaches based on OE.
Other notable results are those of M2F-EAM~\cite{Grcic_2023_CVPR}, and the concurrent RbA~\cite{RbA}: both cleverly exploit the Mask2Former segmentation model.

Along cDNP-Segmenter-ViT-B and cDNP-ConvNeXt-S, we include results for cDNP-SETR (based on ViT-L) We added the last model to show that the approach can work with different encoder sizes and segmentation heads, and we used an official snapshot, to test the method with high accuracy off-the-shelf parameters.

On RoadAnomaly, StreetHazards, and SMIYC-Anomaly -- whose ground truth is undisclosed, cDNP performs best, followed by RbA and M2F-EAM.


\begin{table*}[!htb]
\scriptsize
\setlength{\tabcolsep}{5px}
    \begin{subtable}{.35\linewidth}
    \setlength{\tabcolsep}{3px}
        \begin{tabular}{l|c|cccc}
        \toprule
             & & \multicolumn{2}{c}{RoadAnomaly} & \multicolumn{2}{c}{StreetHazards}\\
            Method & OE & AP & FPR$_{95}$ & AP & FPR$_{95}$\\
            \midrule
            DML\cite{cen2021deep} & & 37.0 & 37.0 & 14.7 & 17.3 \\
            MOoSe\cite{moose} &  & 43.6 & 32.1 & 15.2 & 17.6\\
            PEBAL\cite{Tian2021} & \checkmark & 62.4 & 28.3 & - & - \\
            DenseHybrid\cite{Grcic2022} & \checkmark & - & - & 30.2 & 13.0 \\
            M2F-EAM\cite{Grcic_2023_CVPR} & & 66.7 & 13.4 & - & - \\
            RbA\cite{RbA} & & 78.5 & 11.8 & - & - \\
            \midrule
            cDNP-Segmenter-B & & {85.6} & \textbf{9.8} & \textbf{46.2} & \textbf{14.9} \\
            cDNP-SETR-L &   & \textbf{85.9} & {13.8} & - & - \\
            \bottomrule
    \end{tabular}
    \label{tab:sota_ra_sh}
    \caption{}
    \end{subtable}%
    \hfill
    \begin{subtable}{.29\linewidth}
        \begin{tabular}{l|c|rr}
        \toprule
         & & \multicolumn{2}{c}{SMIYC-Anomaly}\\
        Method      & OE         & AP     & FPR$_{95}$ \\
        \midrule
        Resynth.~\cite{lis2019detecting} & & 52.3 & 25.9 \\
        PEBAL\cite{Tian2021}& \checkmark & 49.1 & 40.8 \\
        NFlowJS\cite{Grcic2021DenseAD}     &            & 56.9 & 34.7 \\
        ObsNet\cite{Besnier2021}      &            & 75.4 & 26.7 \\
        M2F-EAM\cite{Grcic_2023_CVPR} &             & 76.3 & 93.9 \\
        DenseHybrid\cite{Grcic2022} & \checkmark & 78.0 & {9.8} \\
        RbA\cite{RbA} & & 86.1 & 15.9 \\
        \midrule
        cDNP-Segmenter-B &            & \textbf{88.9}  & \textbf{11.4} \\
        \bottomrule
        \end{tabular}
        \label{tab:sota_smiyc}
        \caption{}
    \end{subtable} 
    \hfill
    \begin{subtable}{.29\linewidth}%
    \begin{tabular}{l|c|rr}
    \toprule
        & & \multicolumn{2}{c}{FS-Lost\&Found}\\
        Method & OE & \multicolumn{1}{c}{AP} & \multicolumn{1}{c}{FPR$_{95}$}\\
        \midrule
        M2F-EAM~\cite{Grcic_2023_CVPR} & & 9.4 & 41.5 \\
        NFlowJS & & 39.4 & 9.0 \\
        DenseHybrid\cite{Grcic2022} & \checkmark & 43.9 & 6.2 \\
        PEBAL~\cite{Tian2021} & \checkmark & 44.2 & 7.6 \\
        FlowEneDet & & 50.2 & \textbf{5.2} \\
        GMMSeg~\cite{liang2022gmmseg} & & 55.6 & 6.6 \\
        \midrule
        cDNP-Segmenter-B & & \textbf{62.2} & 8.9 \\
        cDNP-Segmenter-B & \checkmark & 69.8 & 7.5\\
    \bottomrule
    \end{tabular}
    \label{tab:laf}
    \caption{}
    \end{subtable}
    \caption{Comparison with the state-of-the art on RoadAnomaly\,(a), StreetHazards\,(a), SMIYC-Anomaly\,(b), and Fishyscapes-Lost\&Found-test\,(c). OE denotes the use of outlier exposure, according to each specific approach. Best results \textit{without} OE are shown in bold. Our approach performs best overall, except for Lost\&Found, where it has higher FPR$_{95}$.}
    \label{tab:sota_all}
\end{table*}





In Table~\ref{tab:sota_all}(c) we report results for Fishyscapes Lost\&Found-test. This benchmark focuses on road obstacles which are smaller and with less semantic variation. In this case we also include our method's results using a model trained with outlier exposure, following the protocol described in~\cite{Chan2021EntropyMA}. Here cDNP outperforms the other methods in terms of AP, but has a higher FPR$_{95}$.

The results on Fishyscapes Lost\&Found identify a potential limitation of our method in its current formulation, i.e. comparatively worse performance in detecting very small anomalies. This is due to the lower resolution of the transformer patches and kNN scores.
The Appendix has results for SMIYC-Obstacle, another benchmark with small obstacles.

\subsection{Qualitative Results}

\begin{figure*}[h!]
\centering
    \begin{subfigure}[b]{0.11\textwidth}
        \centering
        \includegraphics[width=\textwidth]{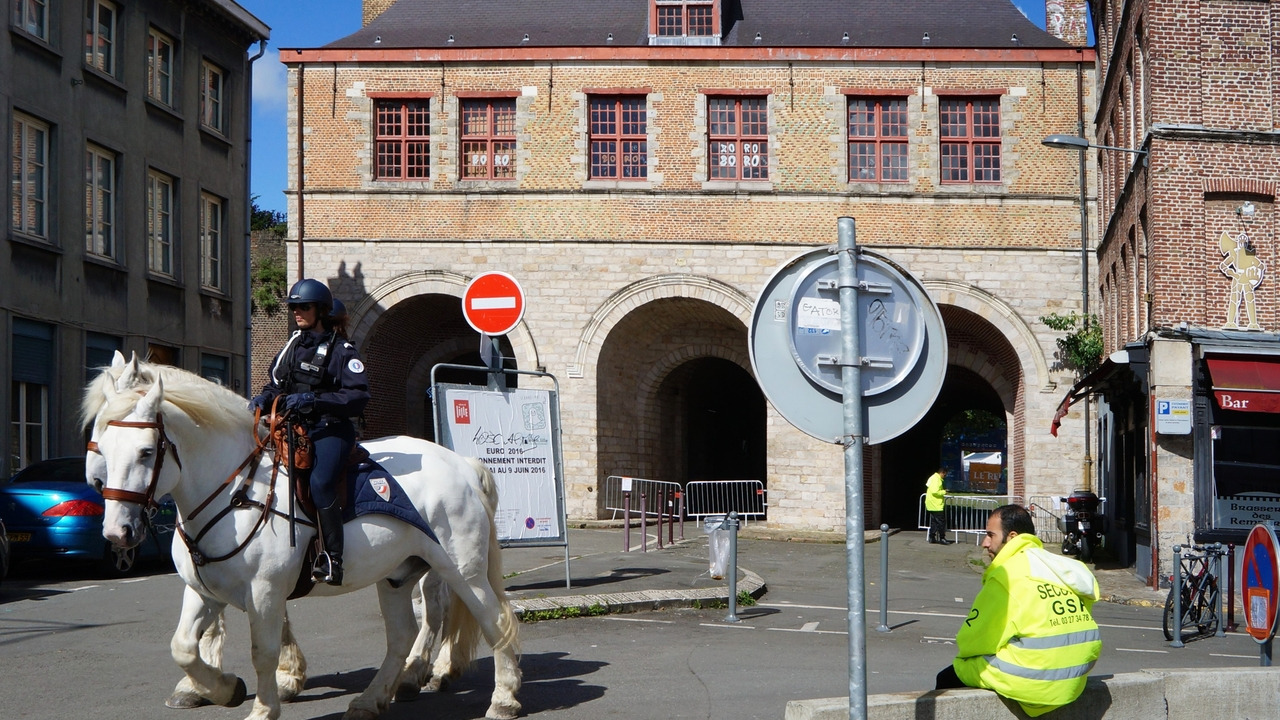}
    \end{subfigure}
    \begin{subfigure}[b]{0.11\textwidth}
        \centering
        \includegraphics[width=\textwidth]{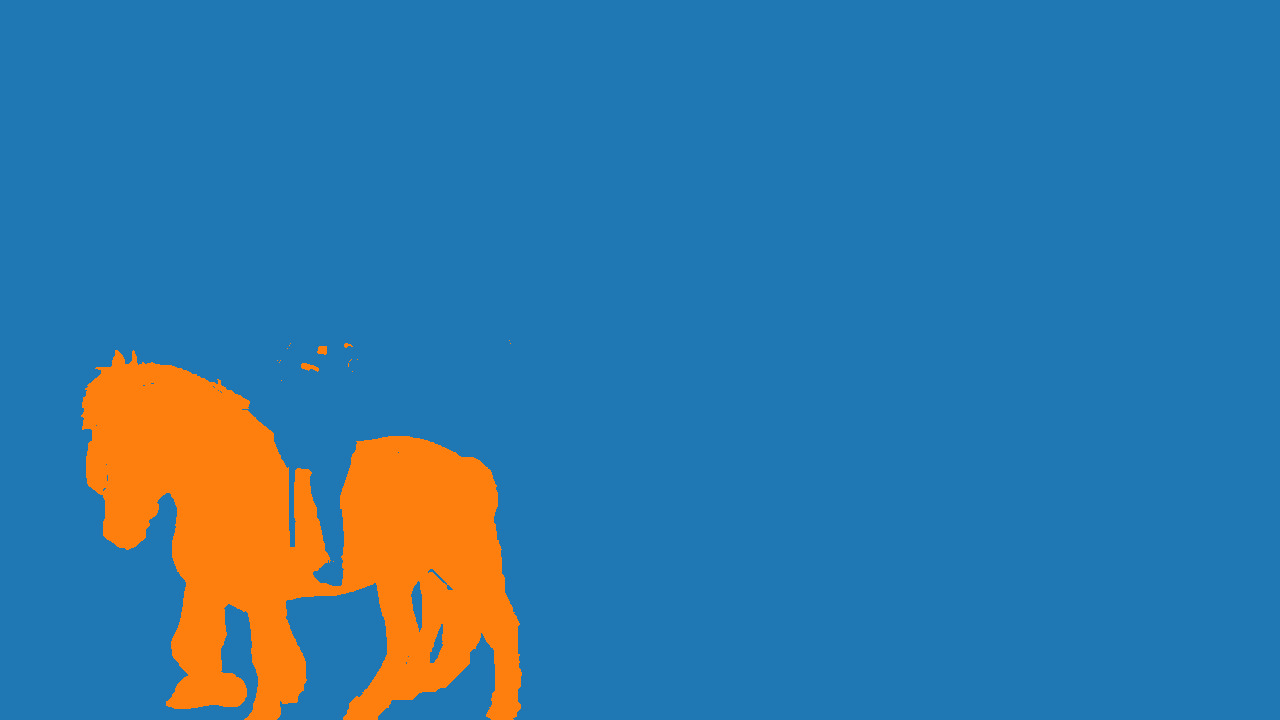}
    \end{subfigure}
    \begin{subfigure}[b]{0.11\textwidth}
        \centering
        \includegraphics[width=\textwidth]{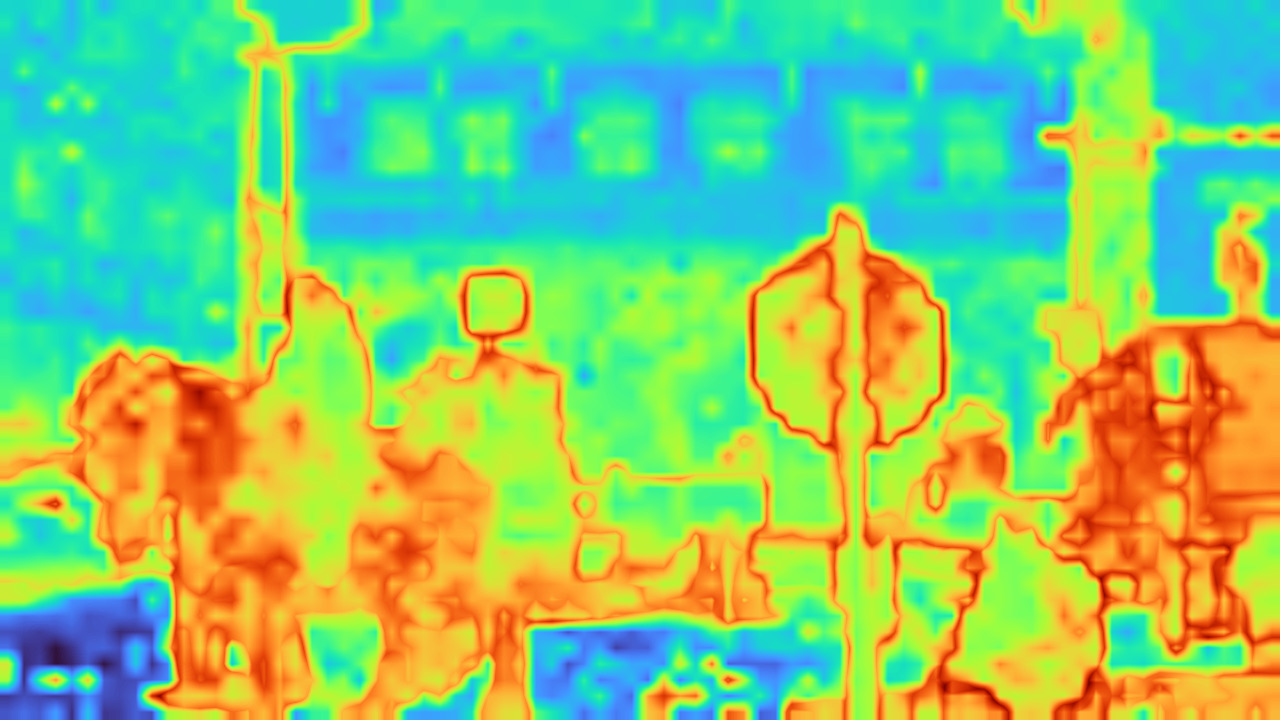}
    \end{subfigure}
    \begin{subfigure}[b]{0.11\textwidth}
        \centering
        \includegraphics[width=\textwidth]{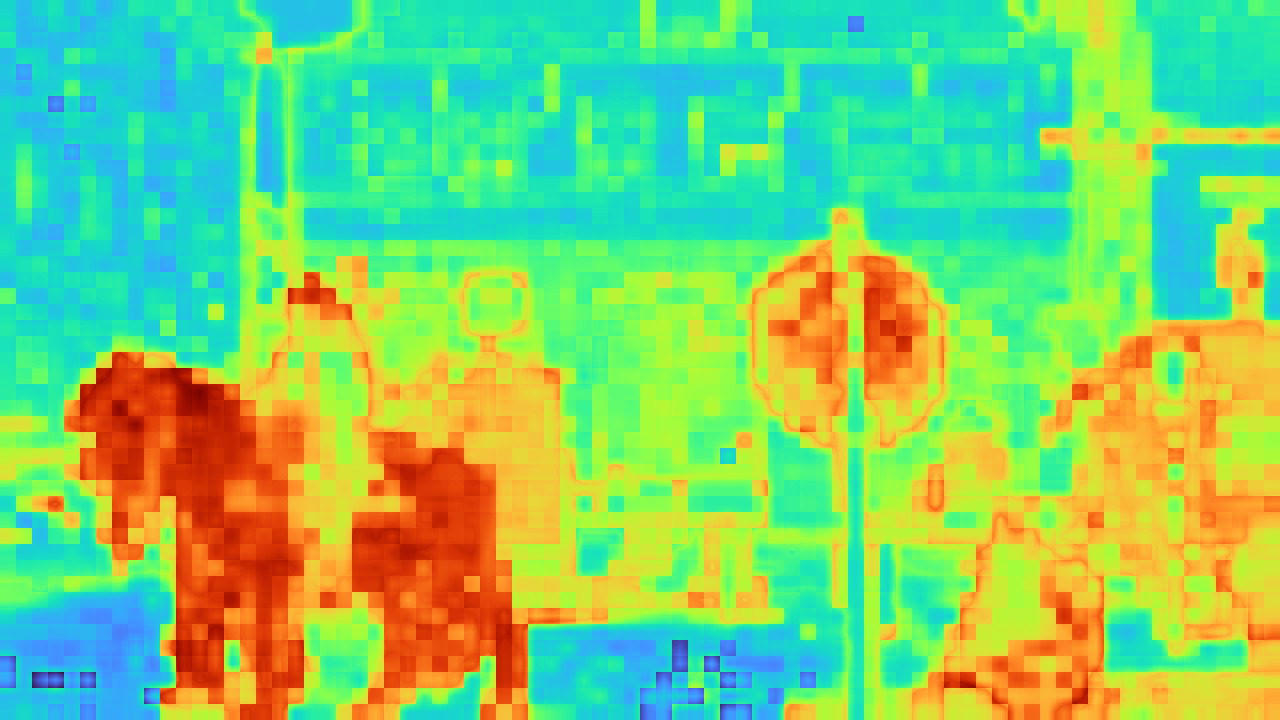}
    \end{subfigure}
    \hspace{10px}
    \begin{subfigure}[b]{0.11\textwidth}
        \centering
        \includegraphics[width=\textwidth]{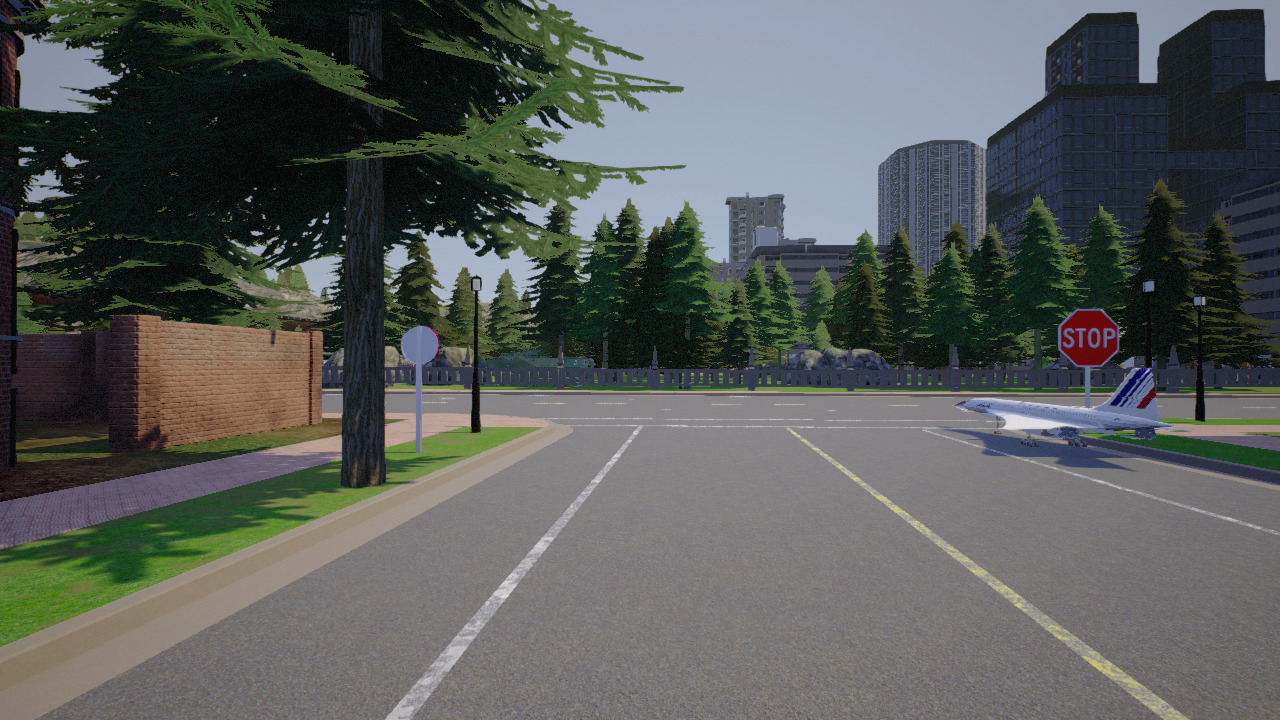}
    \end{subfigure}
    \begin{subfigure}[b]{0.11\textwidth}
        \centering
        \includegraphics[width=\textwidth]{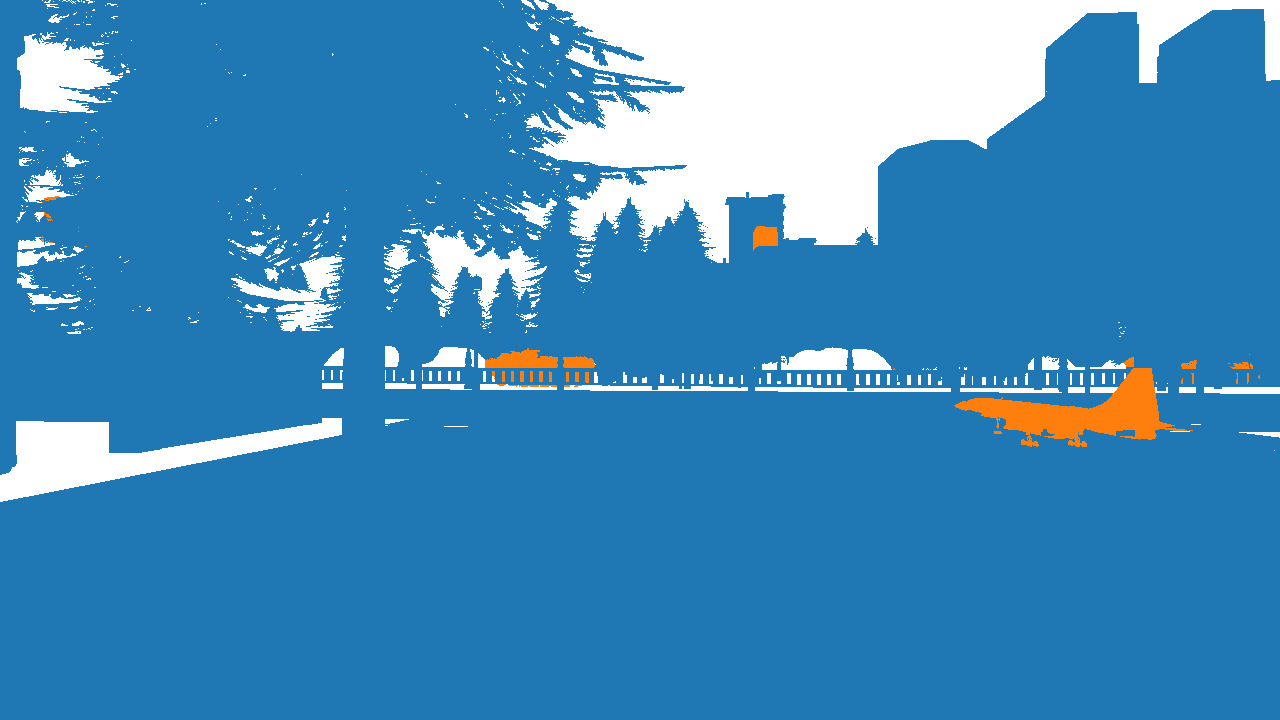}
    \end{subfigure}
    \begin{subfigure}[b]{0.11\textwidth}
        \centering
        \includegraphics[width=\textwidth]{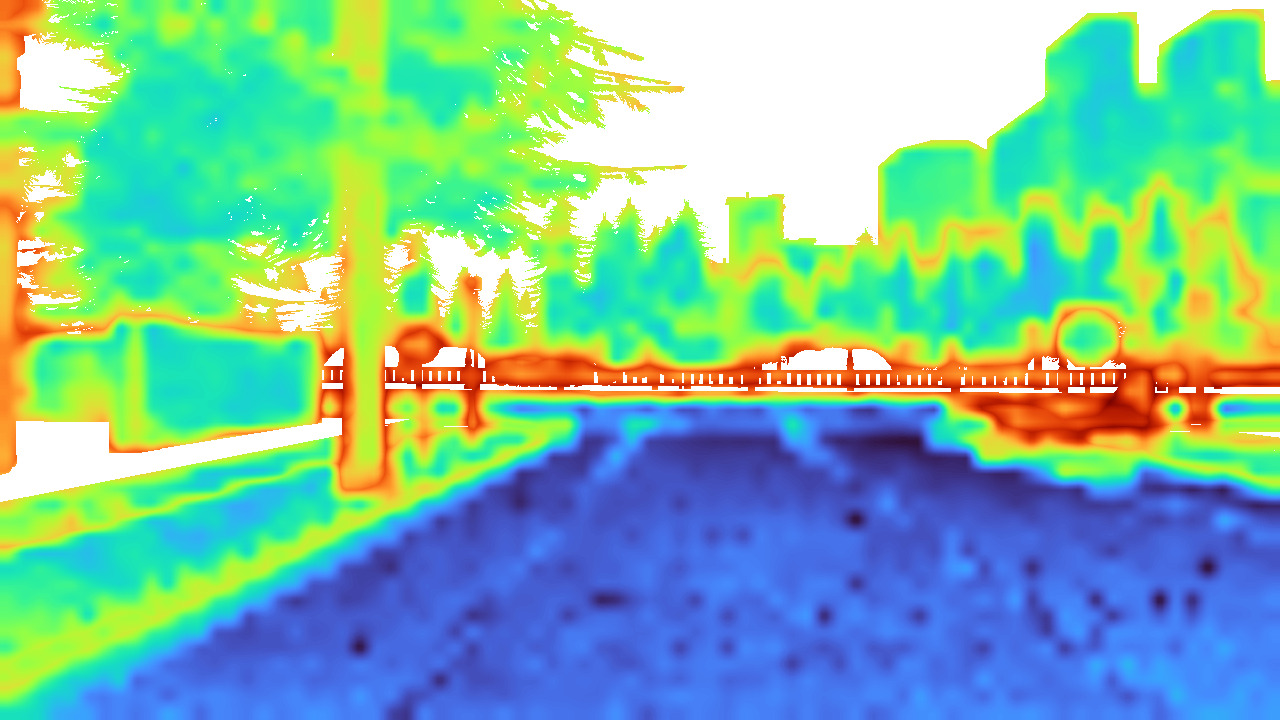}
    \end{subfigure}
    \begin{subfigure}[b]{0.11\textwidth}
        \centering
        \includegraphics[width=\textwidth]{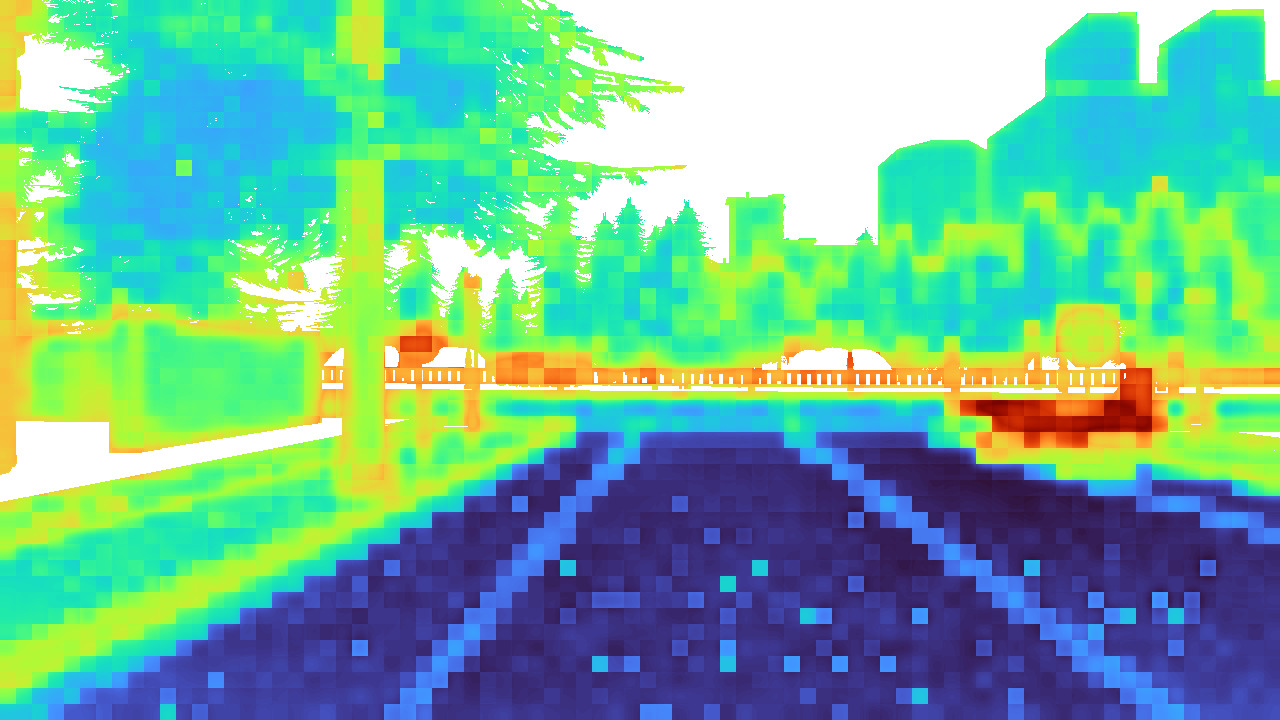}
    \end{subfigure}

    \begin{subfigure}[b]{0.11\textwidth}
        \centering
        \includegraphics[width=\textwidth]{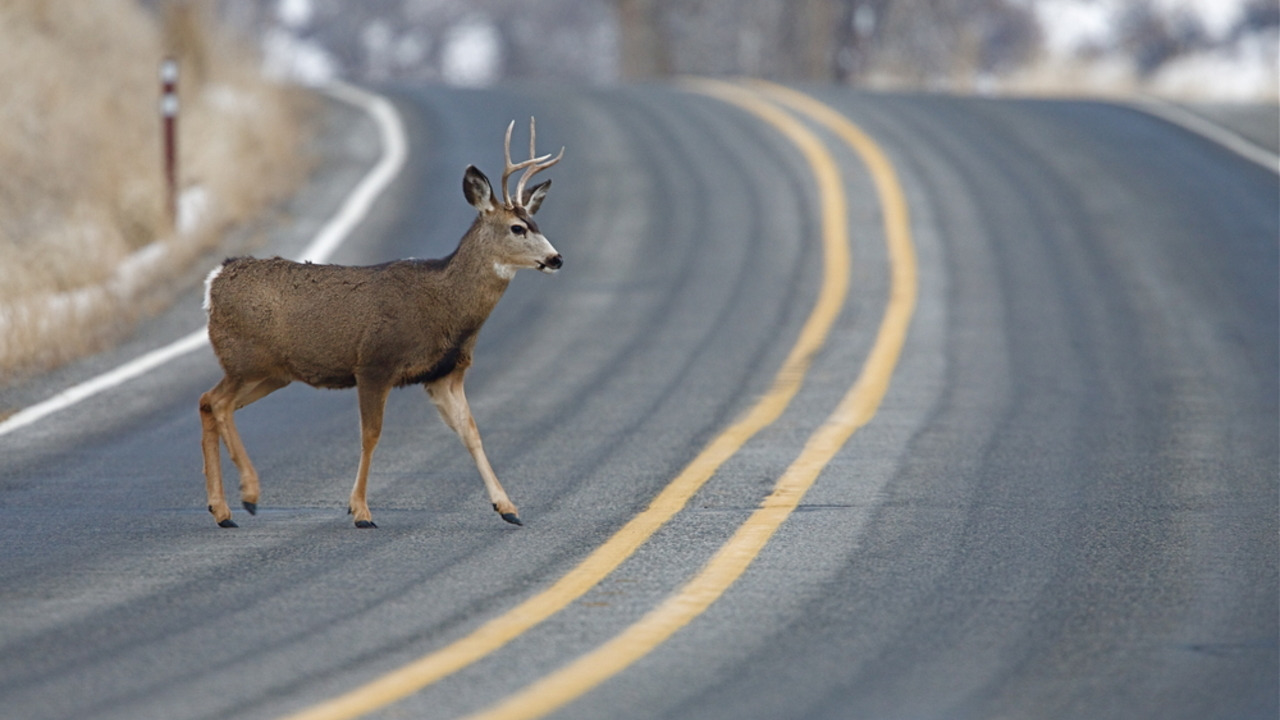}
        \caption*{\scriptsize Image}
    \end{subfigure}
    \begin{subfigure}[b]{0.11\textwidth}
        \centering
        \includegraphics[width=\textwidth]{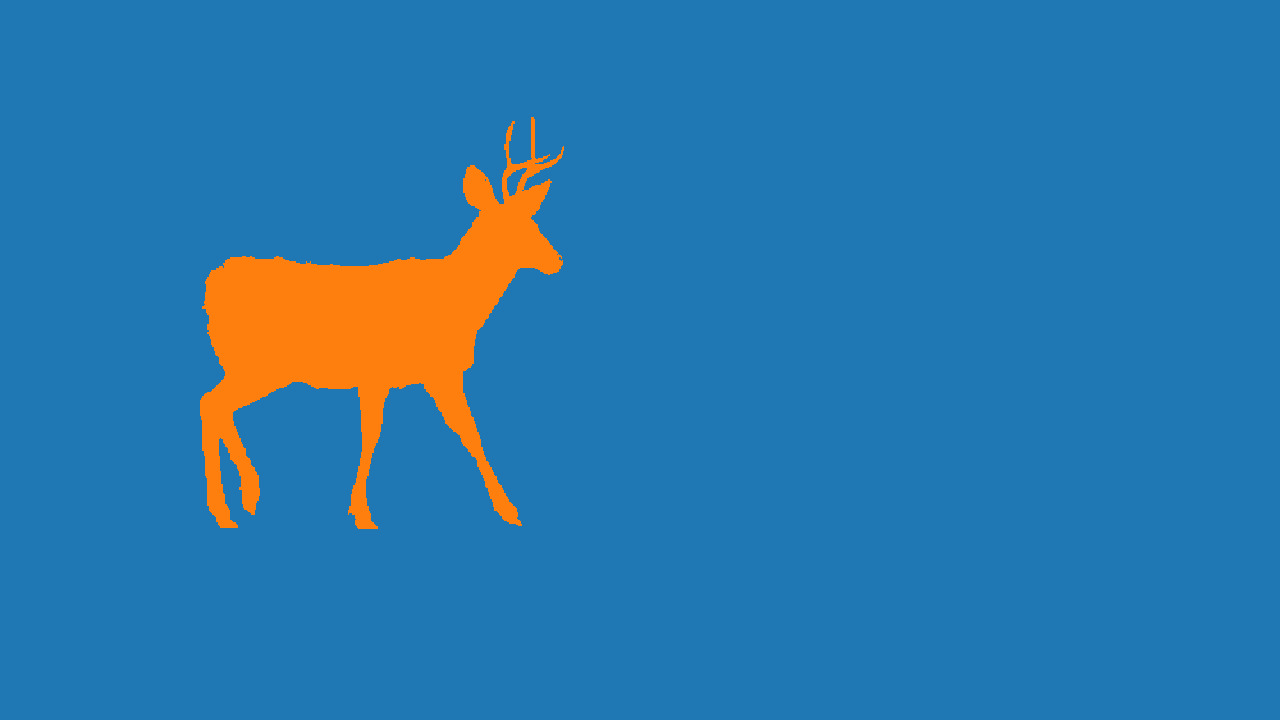}
        \caption*{\scriptsize Ground truth}
    \end{subfigure}
    \begin{subfigure}[b]{0.11\textwidth}
        \centering
        \includegraphics[width=\textwidth]{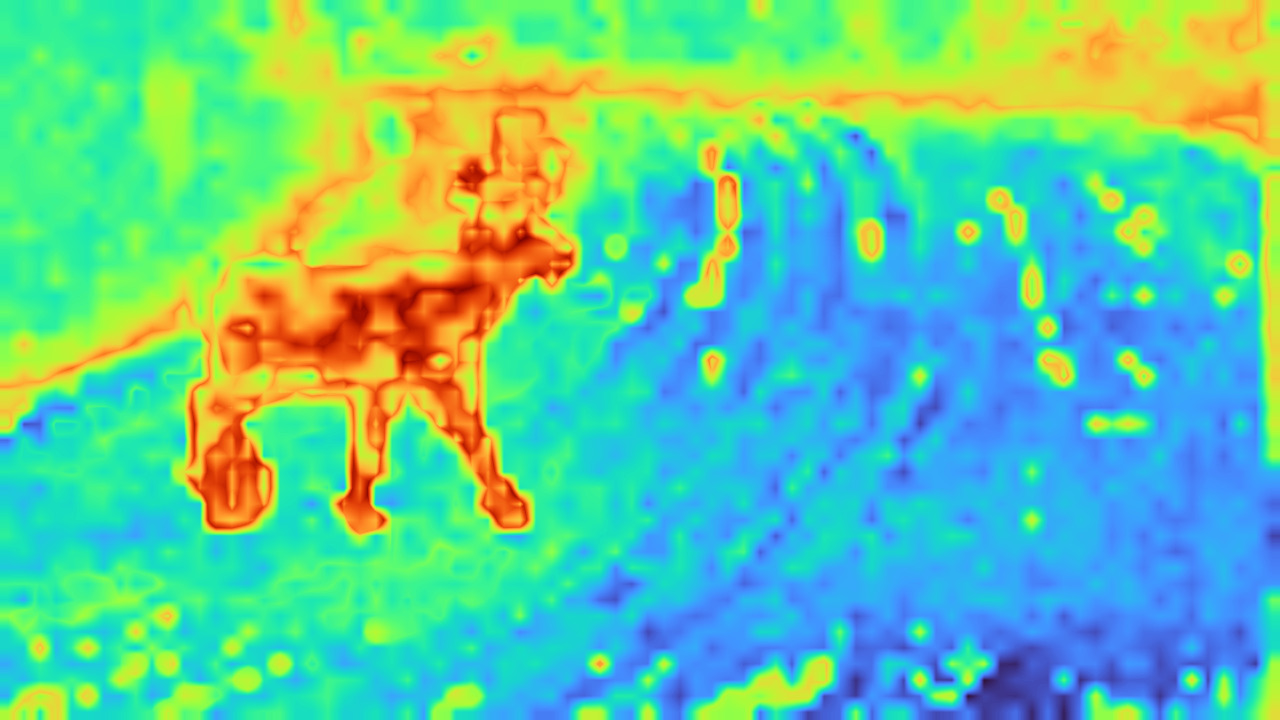}
        \caption*{\scriptsize Param.}
    \end{subfigure}
    \begin{subfigure}[b]{0.11\textwidth}
        \centering
        \includegraphics[width=\textwidth]{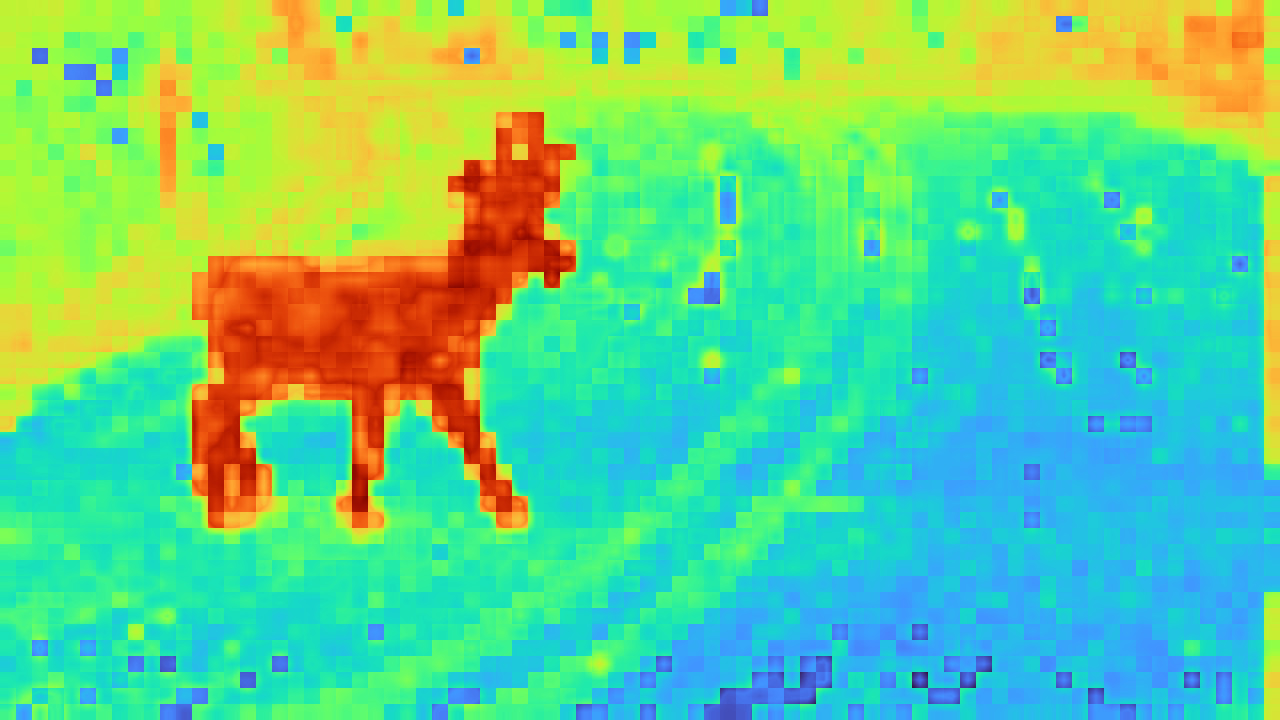}
        \caption*{\scriptsize cDNP}
    \end{subfigure}
    \hspace{10px}
    \begin{subfigure}[b]{0.11\textwidth}
        \centering
        \includegraphics[width=\textwidth]{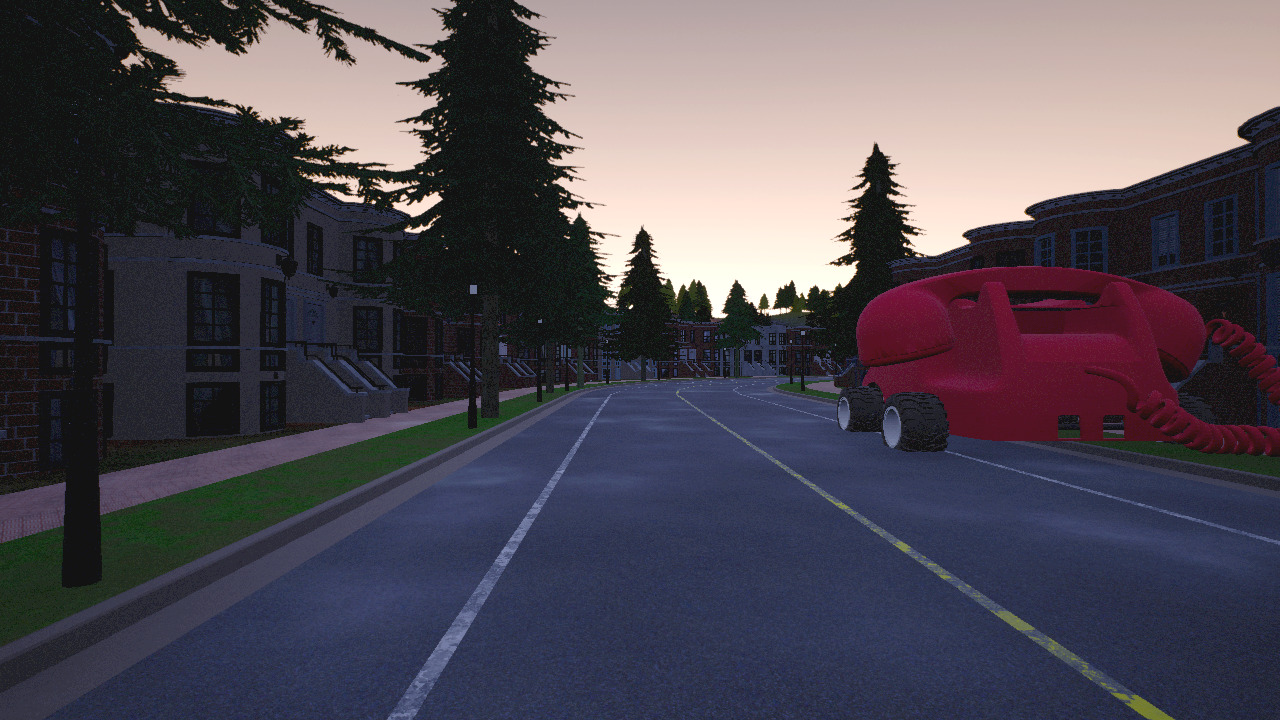}
        \caption*{\scriptsize Image}
    \end{subfigure}
    \begin{subfigure}[b]{0.11\textwidth}
        \centering
        \includegraphics[width=\textwidth]{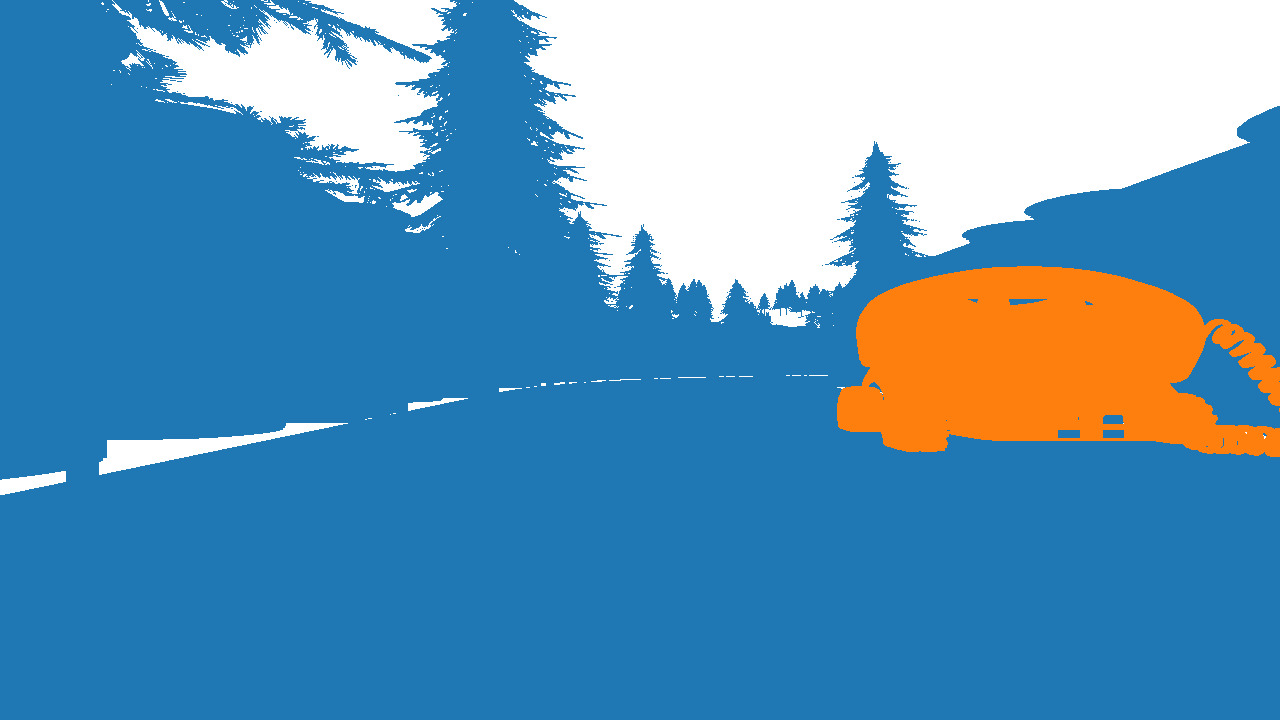}
        \caption*{\scriptsize Ground truth}
    \end{subfigure}
    \begin{subfigure}[b]{0.11\textwidth}
        \centering
        \includegraphics[width=\textwidth]{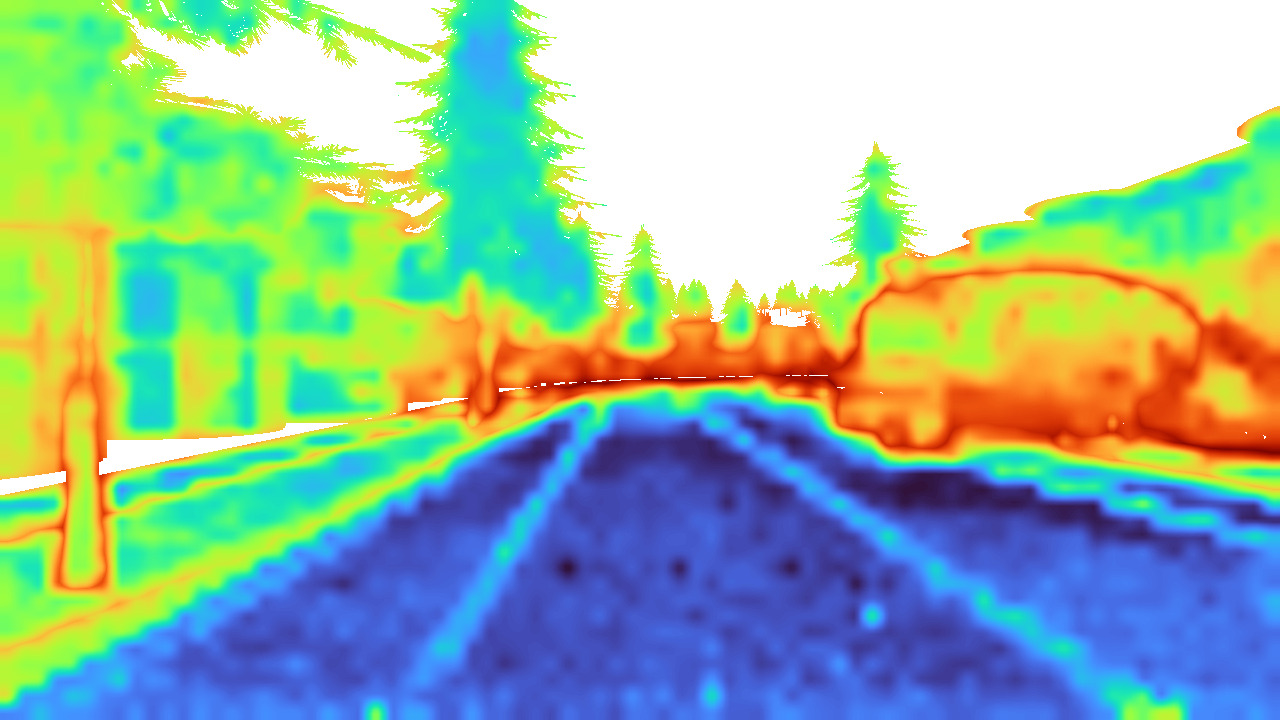}
        \caption*{\scriptsize Param.}
    \end{subfigure}
    \begin{subfigure}[b]{0.11\textwidth}
        \centering
        \includegraphics[width=\textwidth]{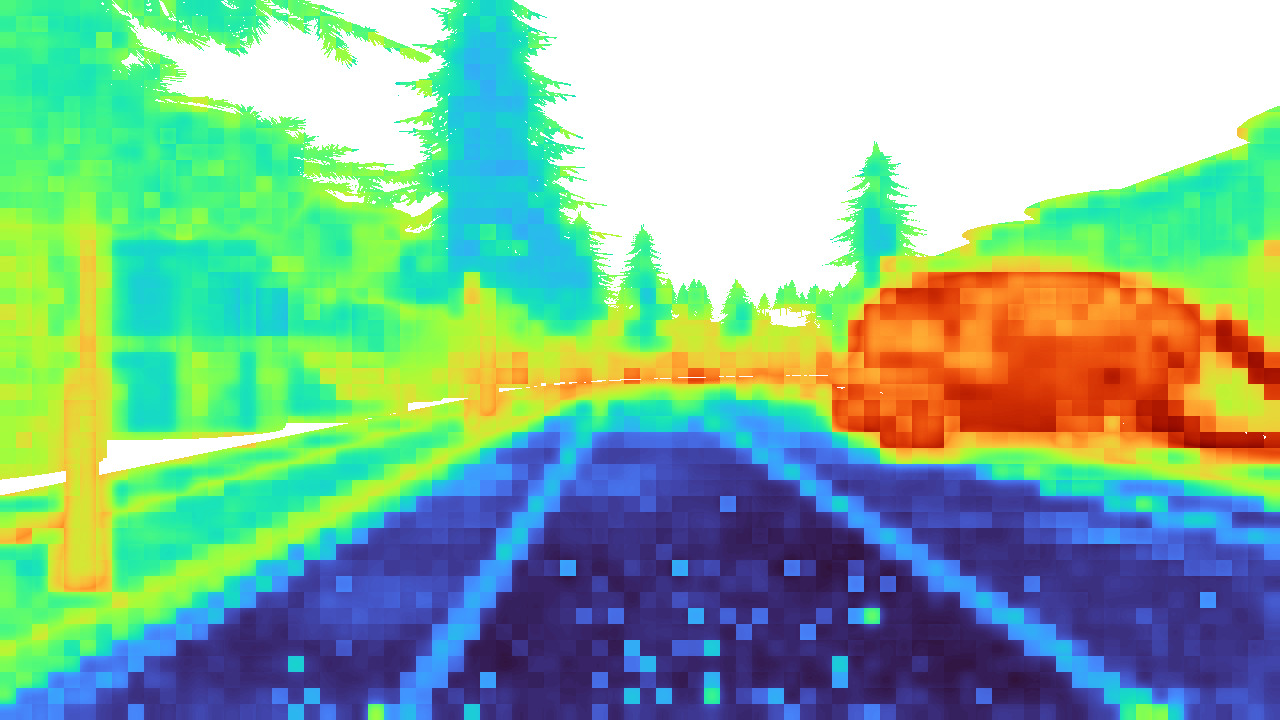}
        \caption*{\scriptsize cDNP}
    \end{subfigure}
    
    \caption{
    Qualitative results for Segmenter-ViT-B on RoadAnomaly (left) and StreetHazards (right). OoD objects are indicated in orange in the ground truth. It can be observed how cDNP scores are better markers for anomalous entities than the parametric ones. The latter also wrongly mark object boundaries and distant objects as anomalous, more often than cDNP.
    }
    \label{fig:quali_ra_sh}
\end{figure*}

Qualitative examples of our method on RoadAnomaly and StreetHazards, using Segmenter-ViT-B are shown in Figure~\ref{fig:quali_ra_sh}.
In both cases, the anomaly score-maps show the superiority of the kNN-based scores compared to the parametric ones: cDNP has both fewer false negatives (it marks most anomalous locations) and false positives (especially at class boundaries and for distant objects).

\begin{figure*}[h!]
    \centering
    \begin{subfigure}[b]{0.11\textwidth}
        \centering
        \includegraphics[width=\textwidth]{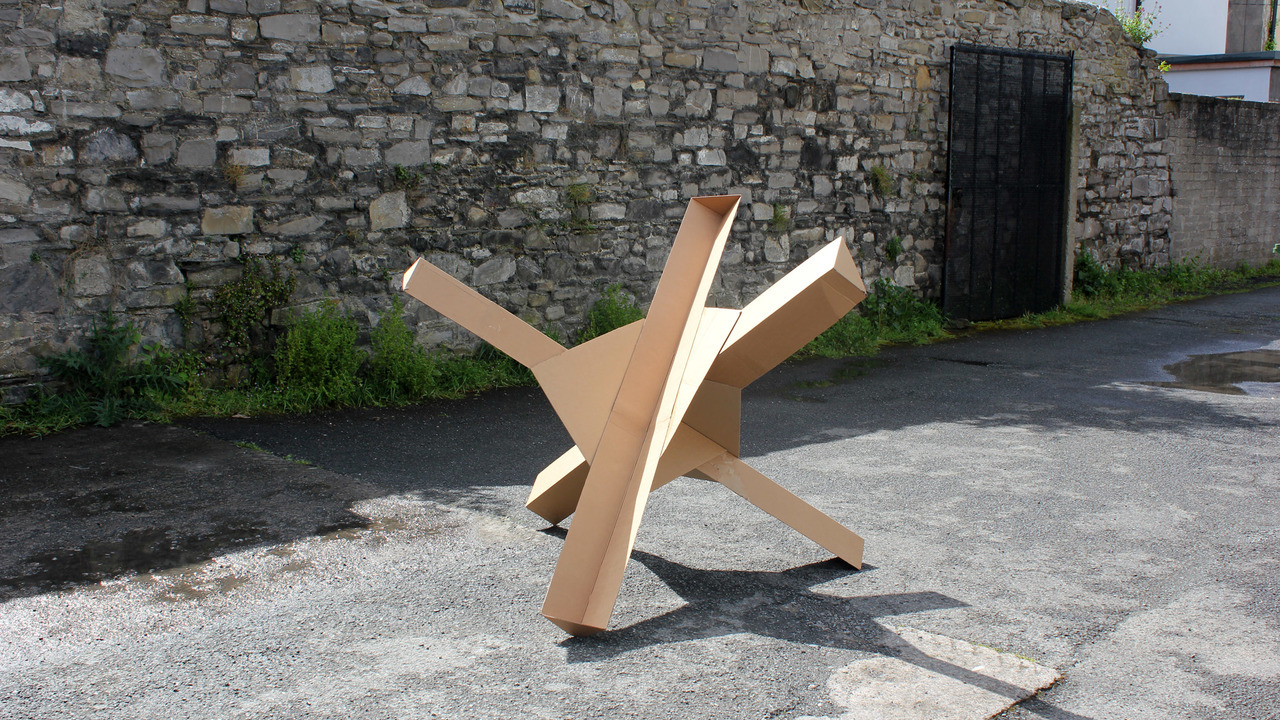}
    \end{subfigure}
    \begin{subfigure}[b]{0.11\textwidth}
        \centering
        \includegraphics[width=\textwidth]{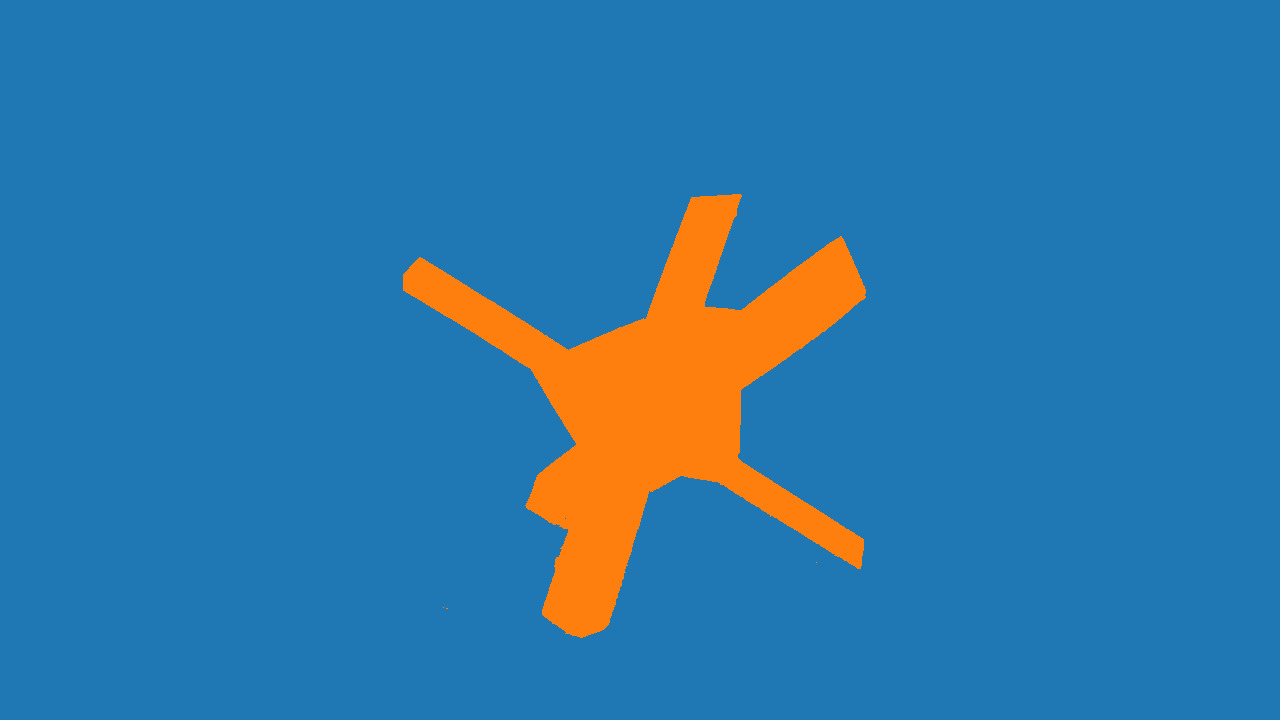}
    \end{subfigure}
    \hspace{3px}
    \begin{subfigure}[b]{0.11\textwidth}
        \centering
        \includegraphics[width=\textwidth]{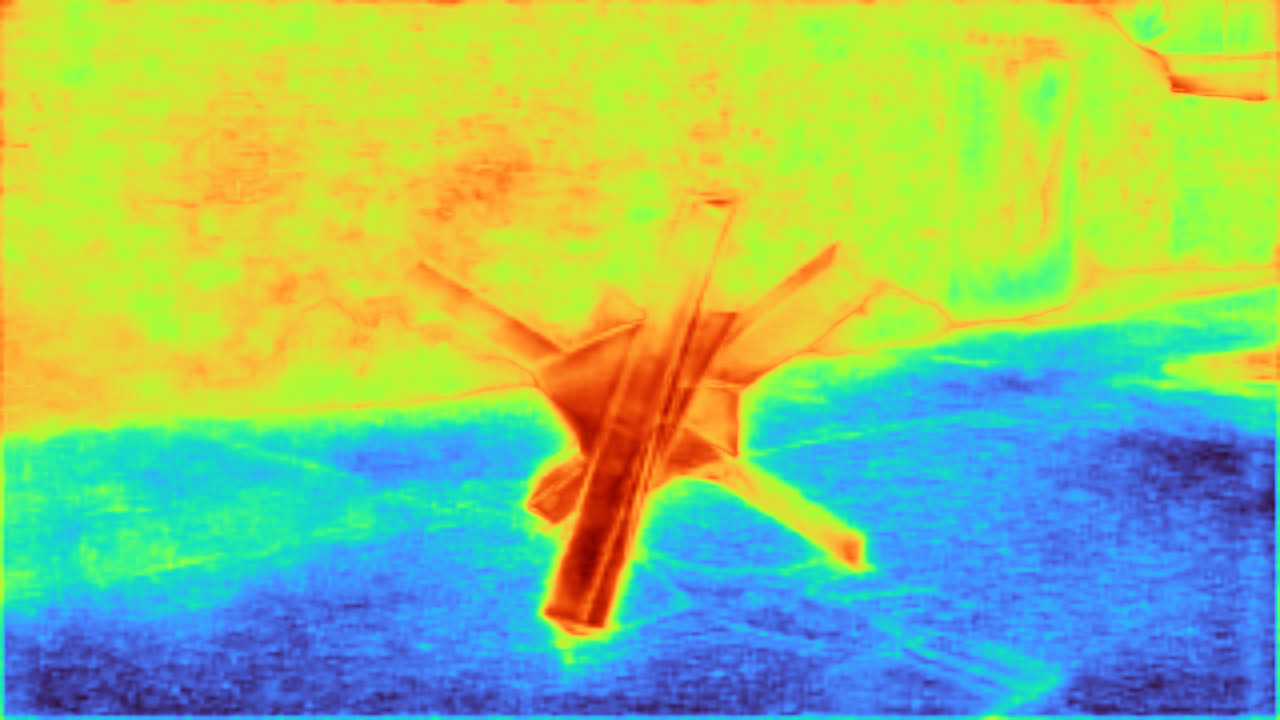}
    \end{subfigure}
    \begin{subfigure}[b]{0.11\textwidth}
        \centering
        \includegraphics[width=\textwidth]{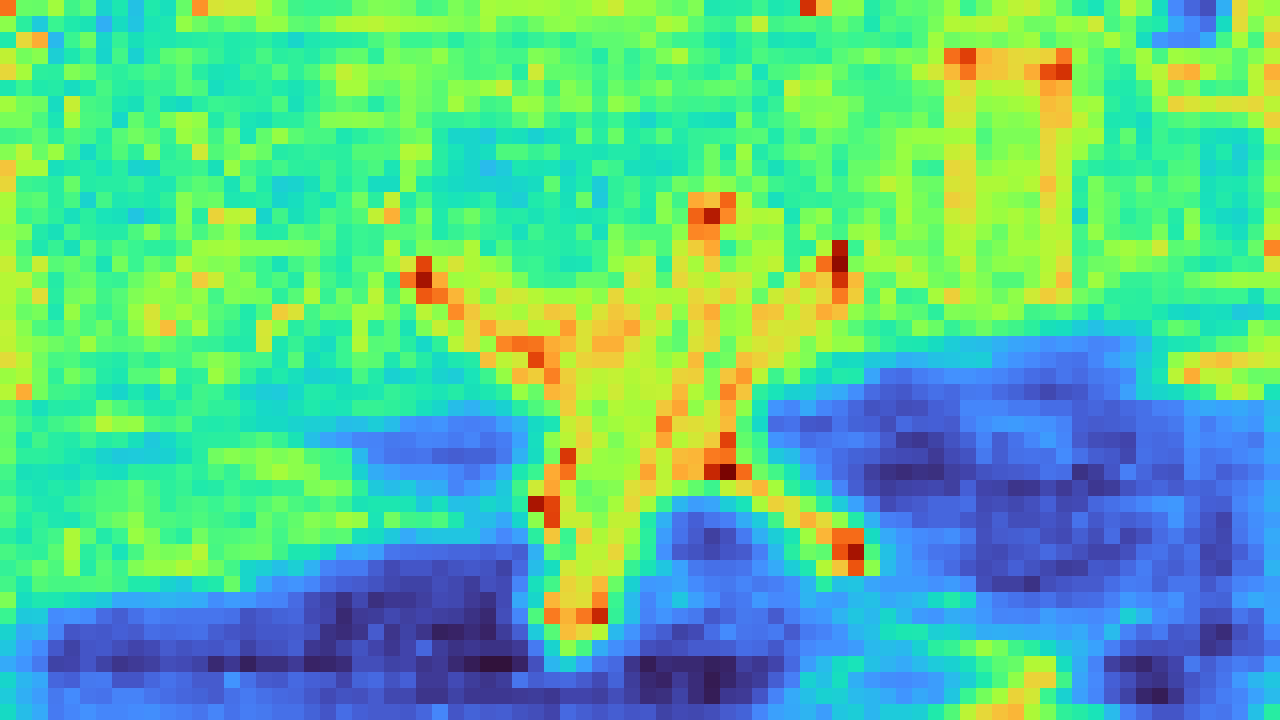}
    \end{subfigure}
    \begin{subfigure}[b]{0.11\textwidth}
        \centering
        \includegraphics[width=\textwidth]{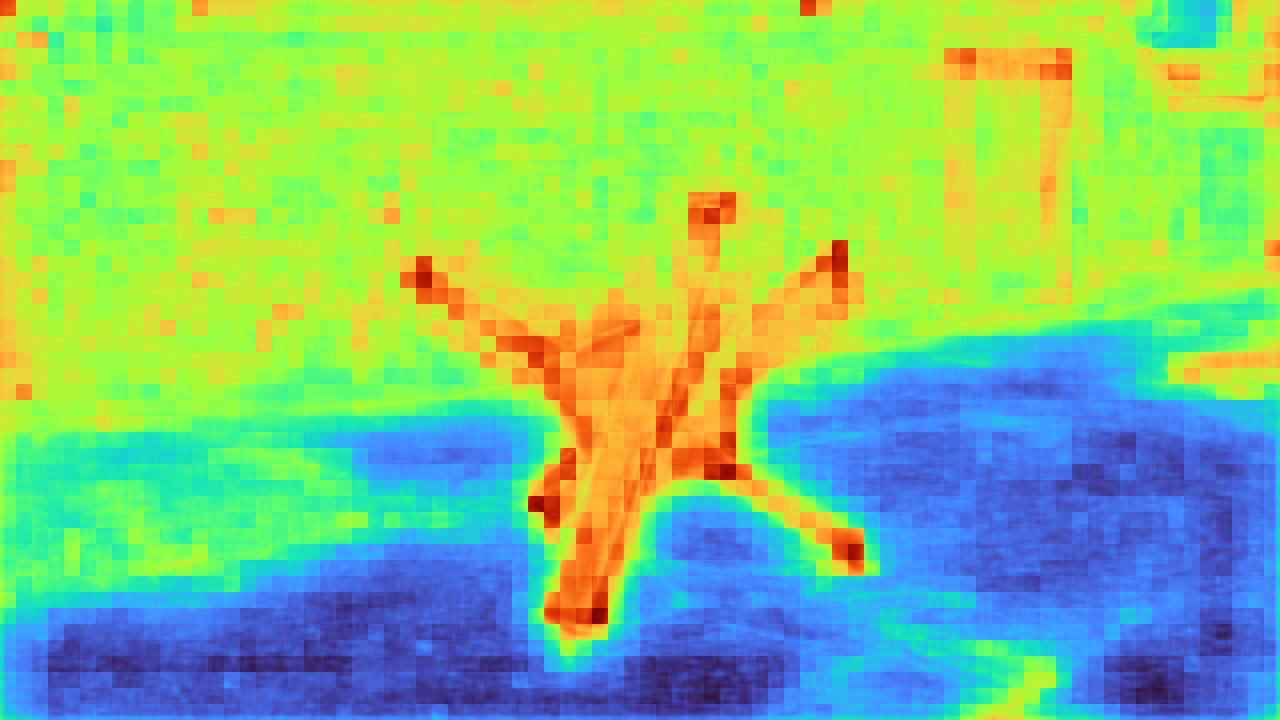}
    \end{subfigure}
    \hspace{3px}
    \begin{subfigure}[b]{0.11\textwidth}
        \centering
        \includegraphics[width=\textwidth]{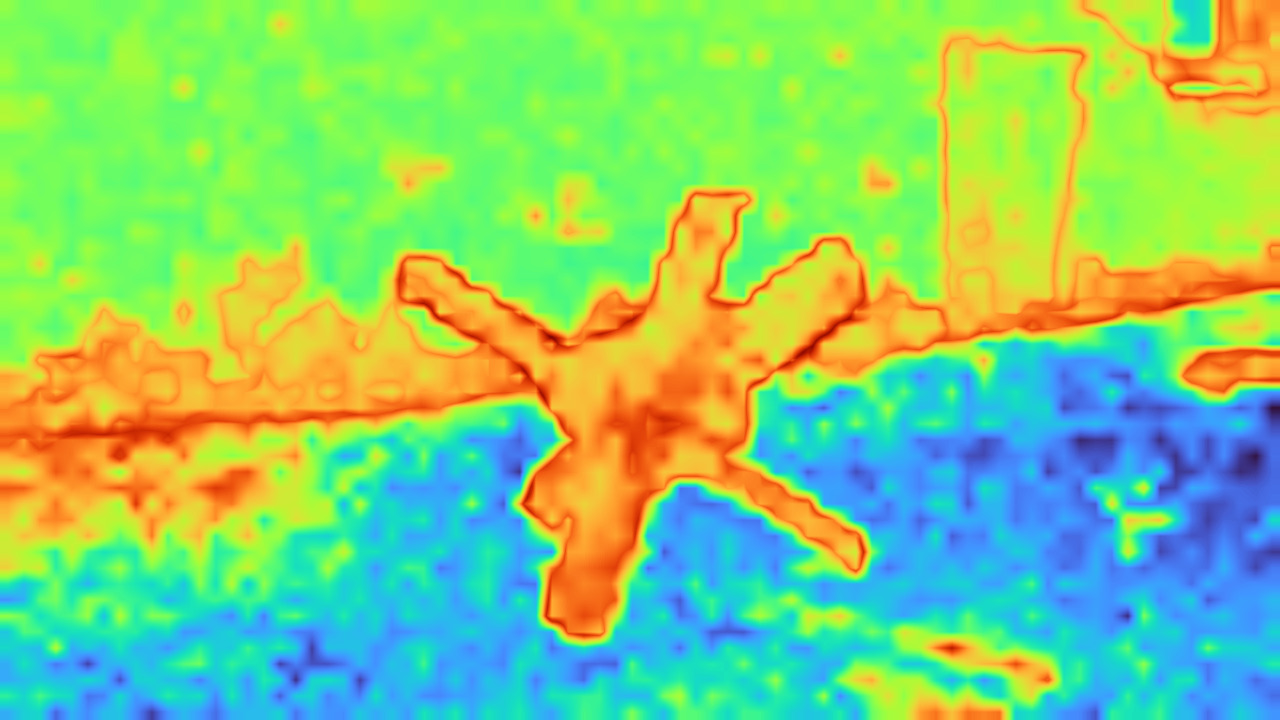}
    \end{subfigure}
    \begin{subfigure}[b]{0.11\textwidth}
        \centering
        \includegraphics[width=\textwidth]{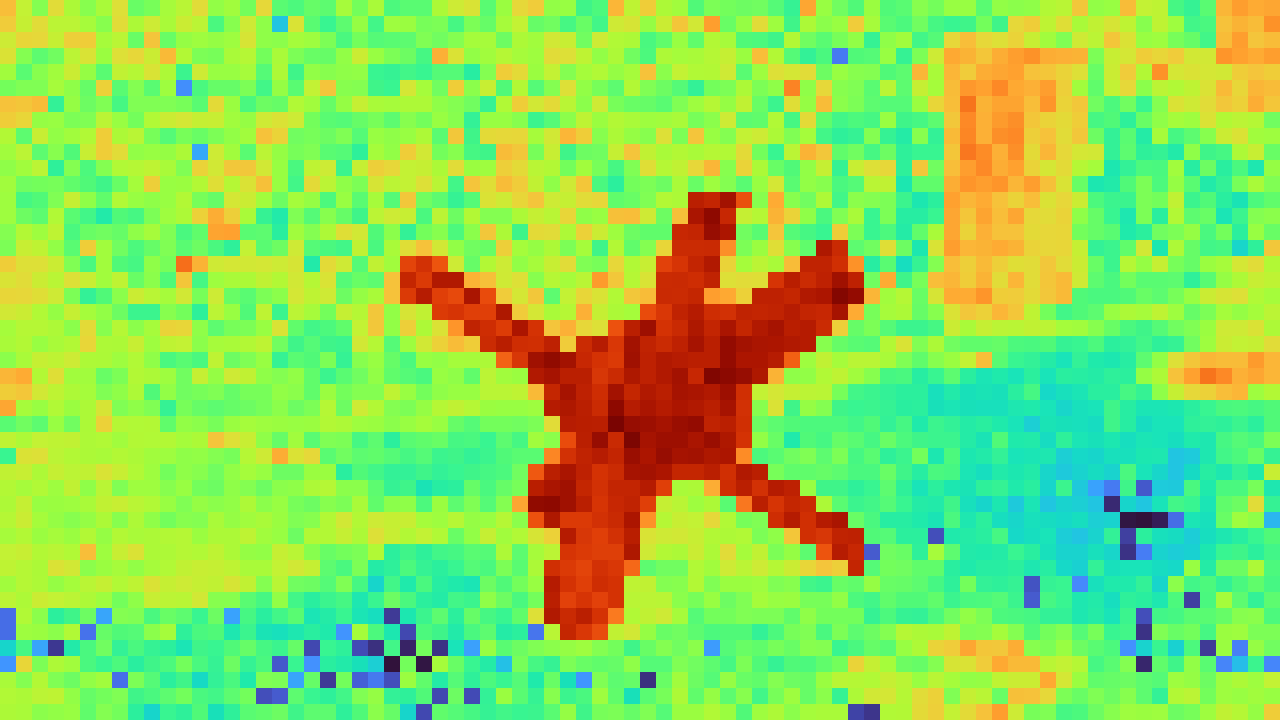}
    \end{subfigure}
    \begin{subfigure}[b]{0.11\textwidth}
        \centering
        \includegraphics[width=\textwidth]{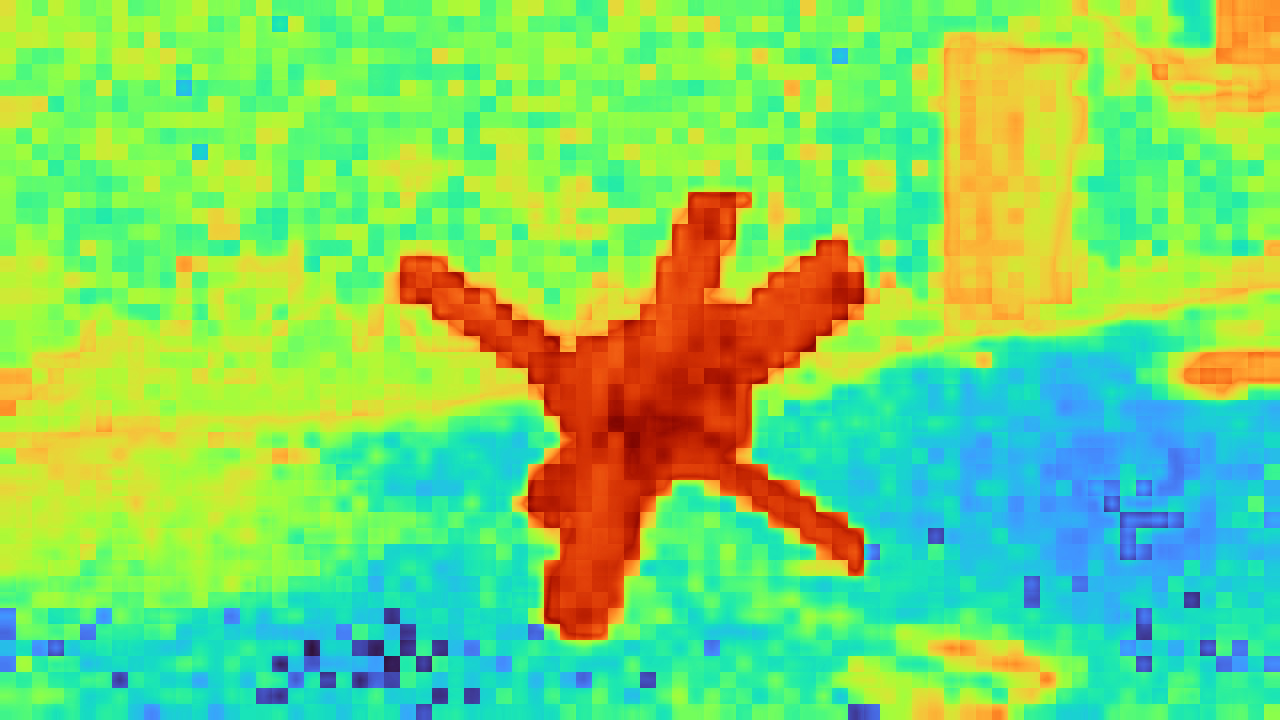}
    \end{subfigure}
    
    
    \begin{subfigure}[b]{0.11\textwidth}
        \centering
        \includegraphics[width=\textwidth]{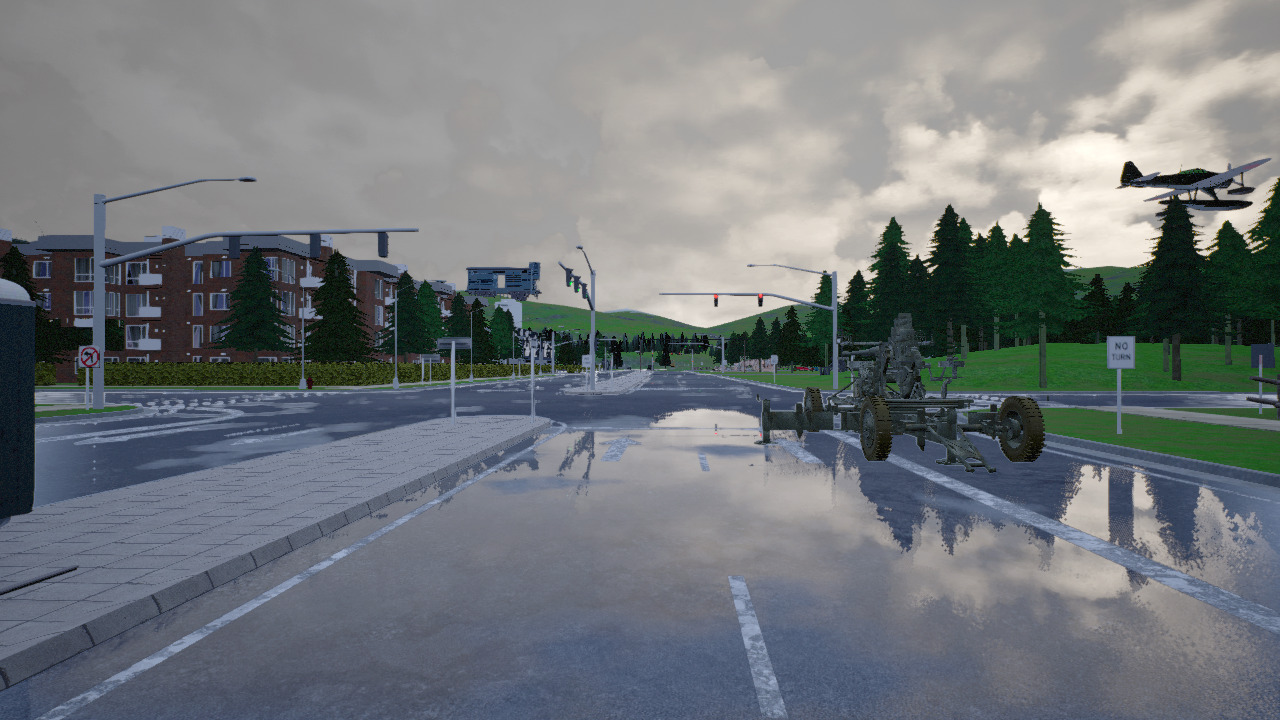}
    \end{subfigure}
    \begin{subfigure}[b]{0.11\textwidth}
        \centering
        \includegraphics[width=\textwidth]{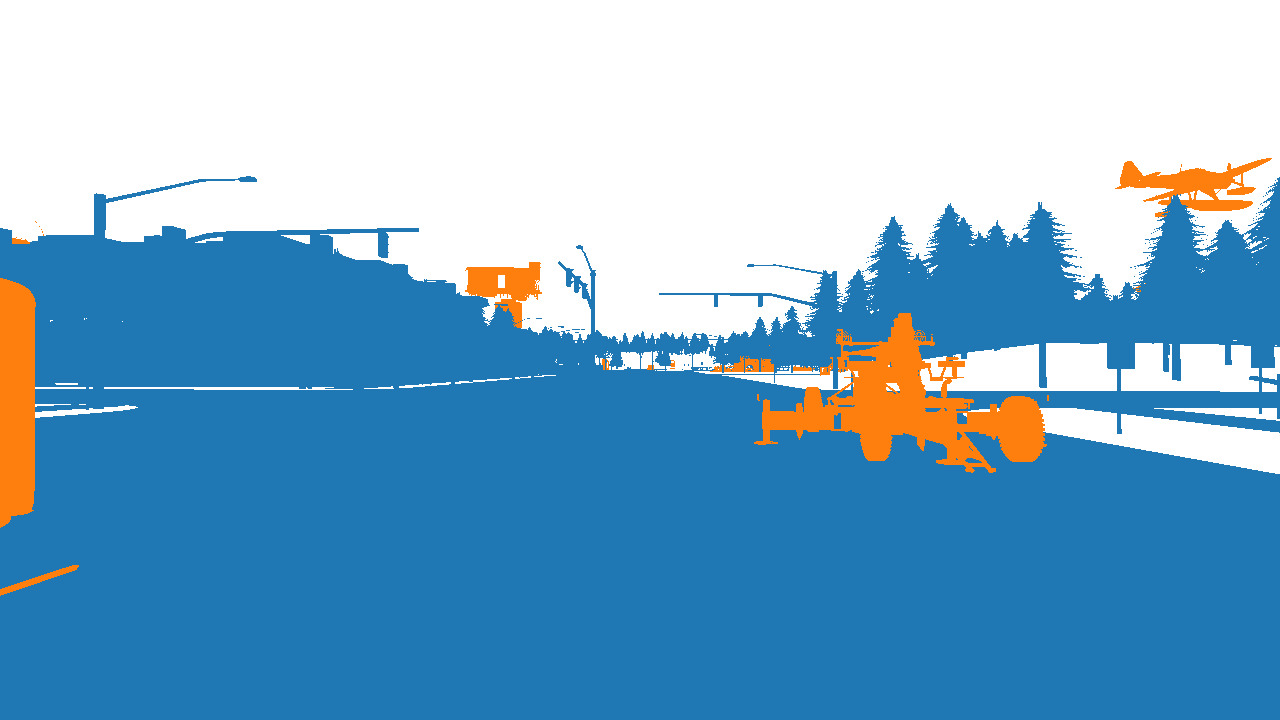}
    \end{subfigure}
    \hspace{3px}
    \begin{subfigure}[b]{0.11\textwidth}
        \centering
        \includegraphics[width=\textwidth]{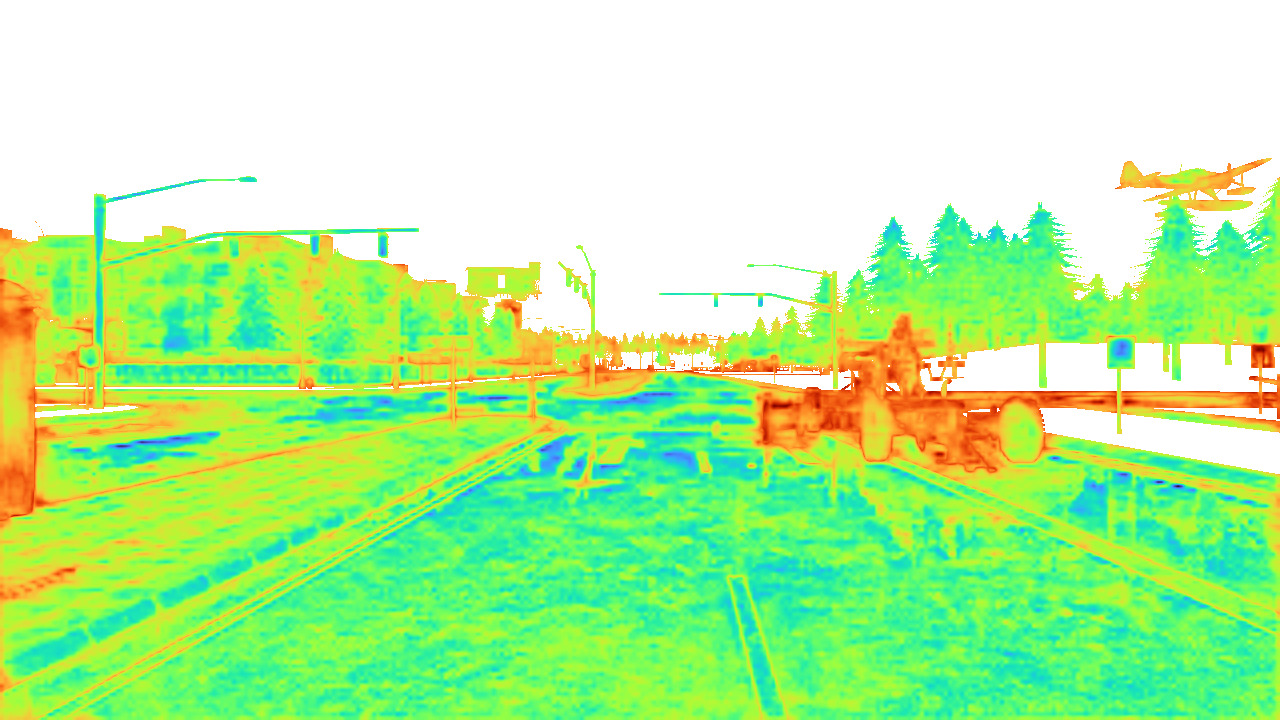}
    \end{subfigure}
    \begin{subfigure}[b]{0.11\textwidth}
        \centering
        \includegraphics[width=\textwidth]{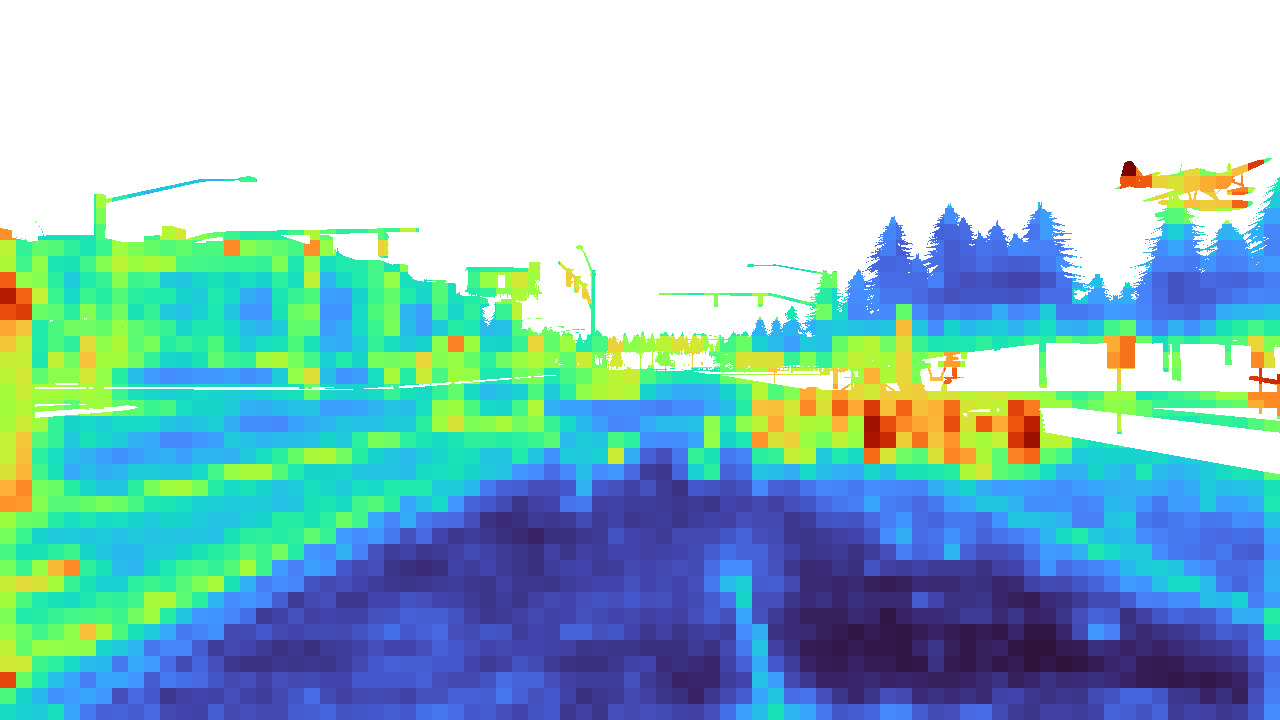}
    \end{subfigure}
    \begin{subfigure}[b]{0.11\textwidth}
        \centering
        \includegraphics[width=\textwidth]{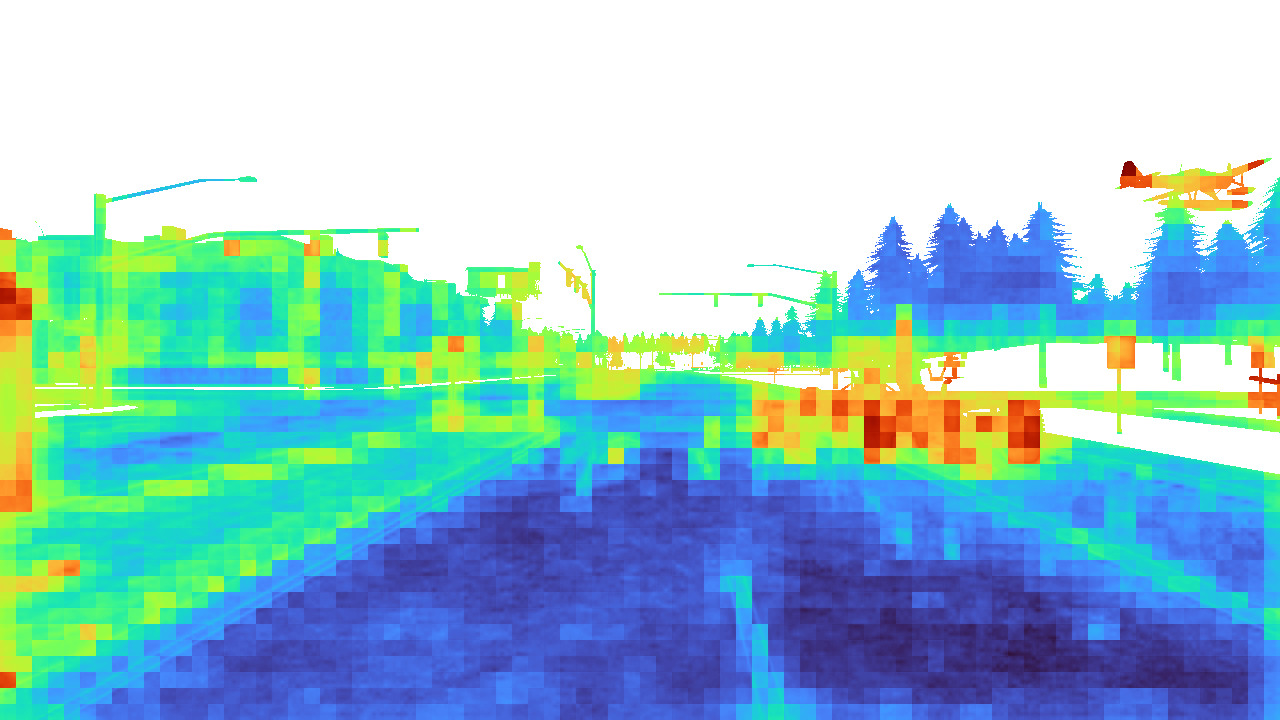}
    \end{subfigure}
    \hspace{3px}
    \begin{subfigure}[b]{0.11\textwidth}
        \centering
        \includegraphics[width=\textwidth]{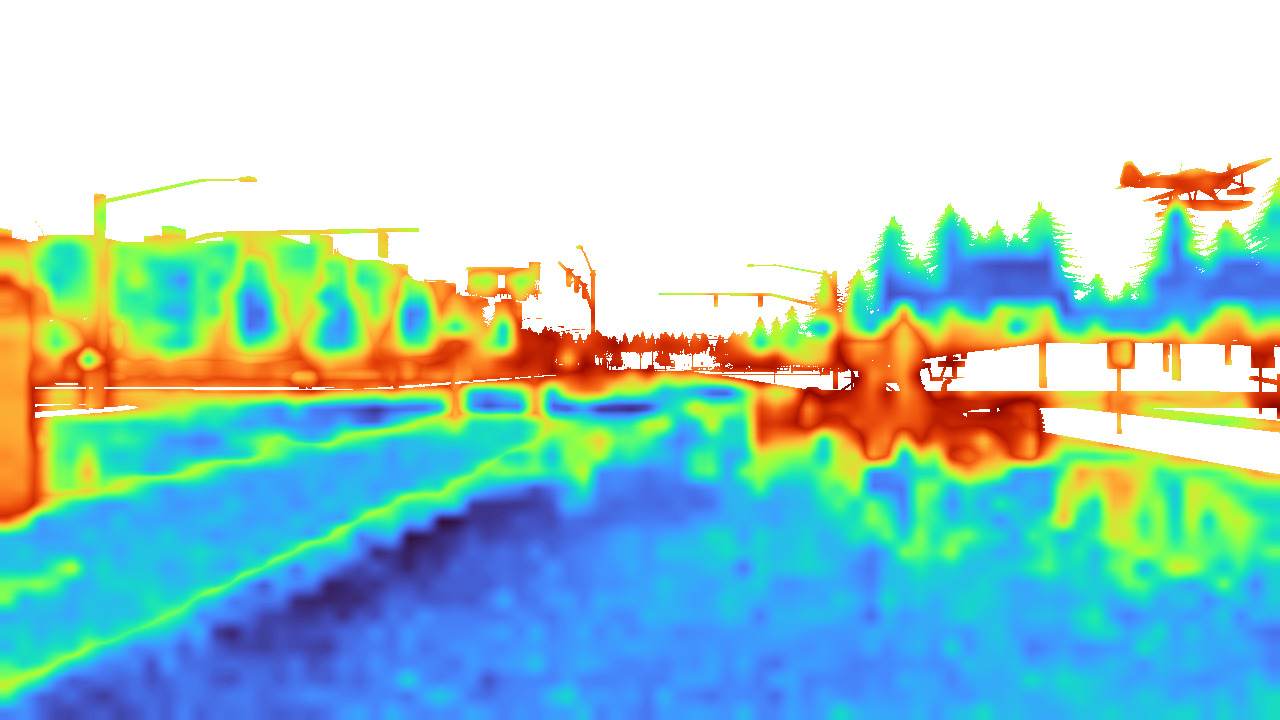}
    \end{subfigure}
    \begin{subfigure}[b]{0.11\textwidth}
        \centering
        \includegraphics[width=\textwidth]{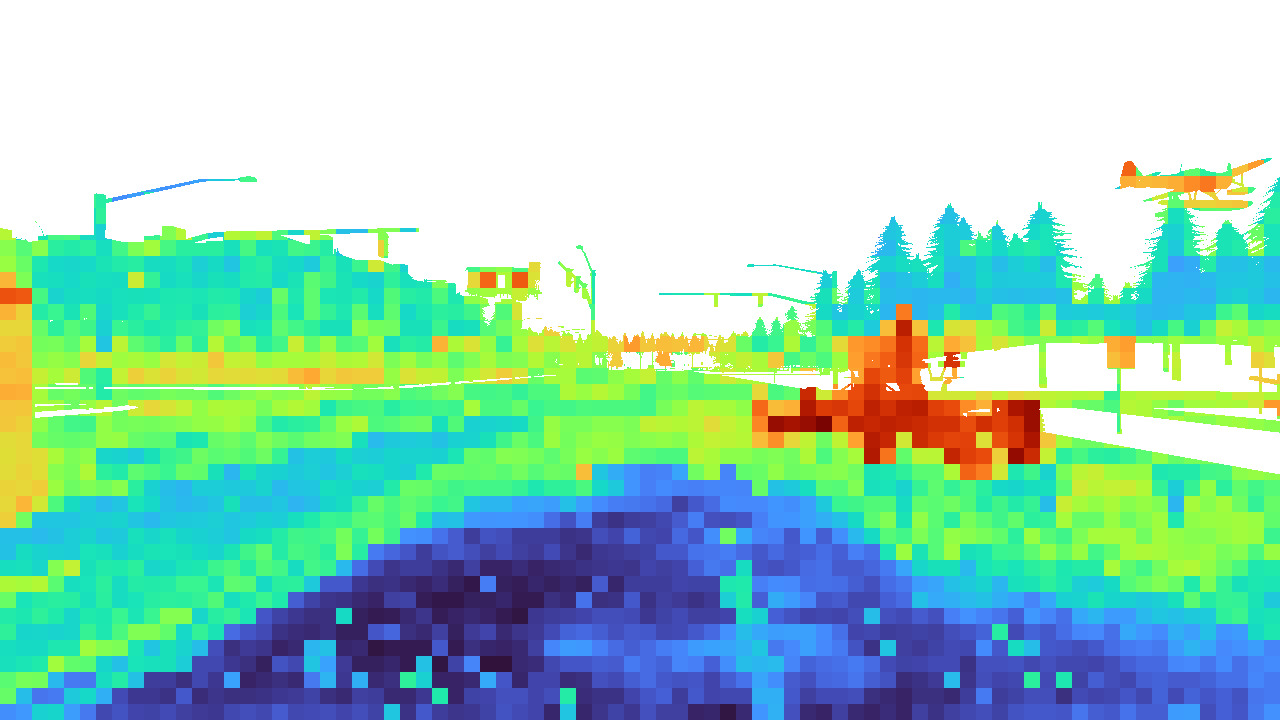}
    \end{subfigure}
    \begin{subfigure}[b]{0.11\textwidth}
        \centering
        \includegraphics[width=\textwidth]{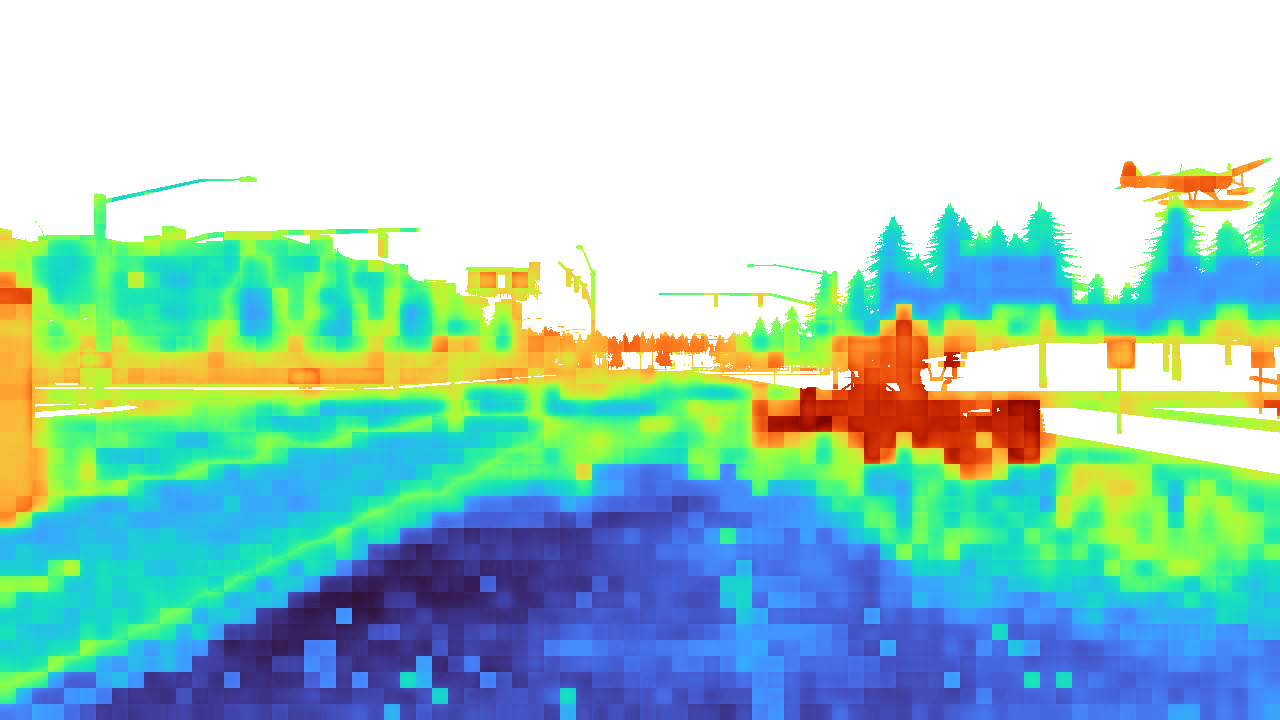}
    \end{subfigure}
    
    \begin{subfigure}[b]{0.11\textwidth}
        \centering
        \includegraphics[width=\textwidth]{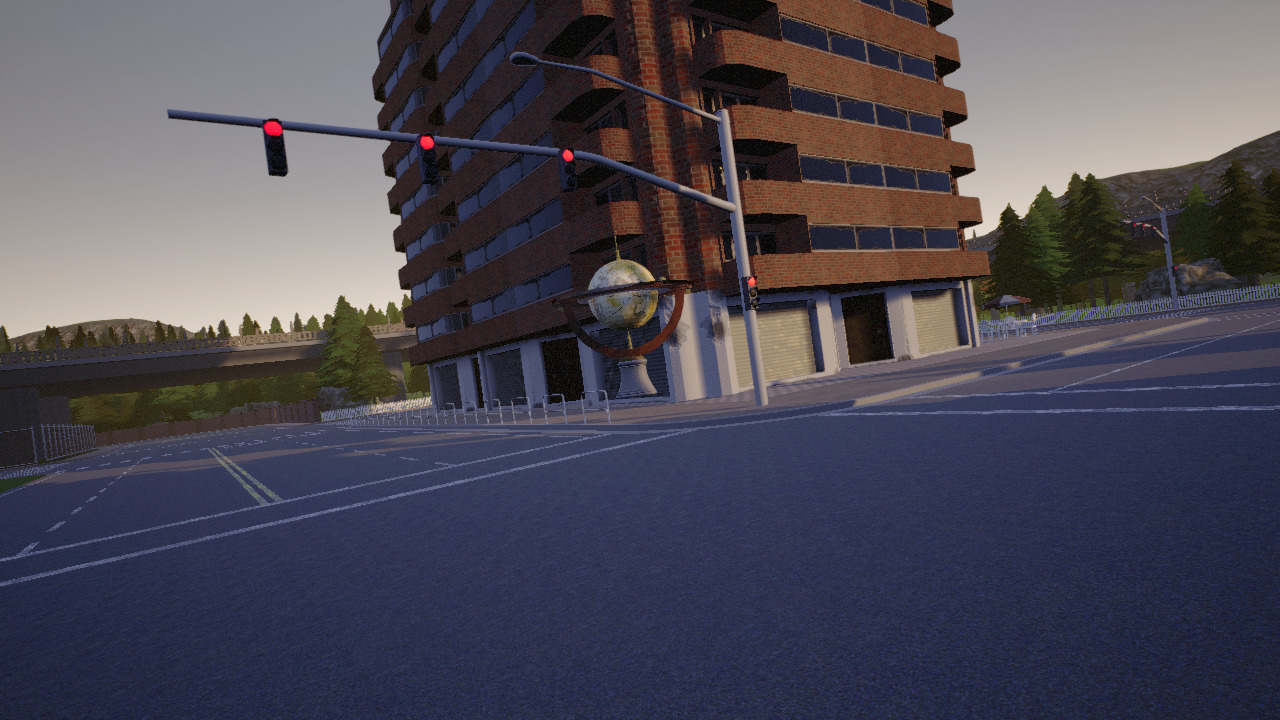}
        \caption*{\scriptsize Image}
    \end{subfigure}
    \begin{subfigure}[b]{0.11\textwidth}
        \centering
        \includegraphics[width=\textwidth]{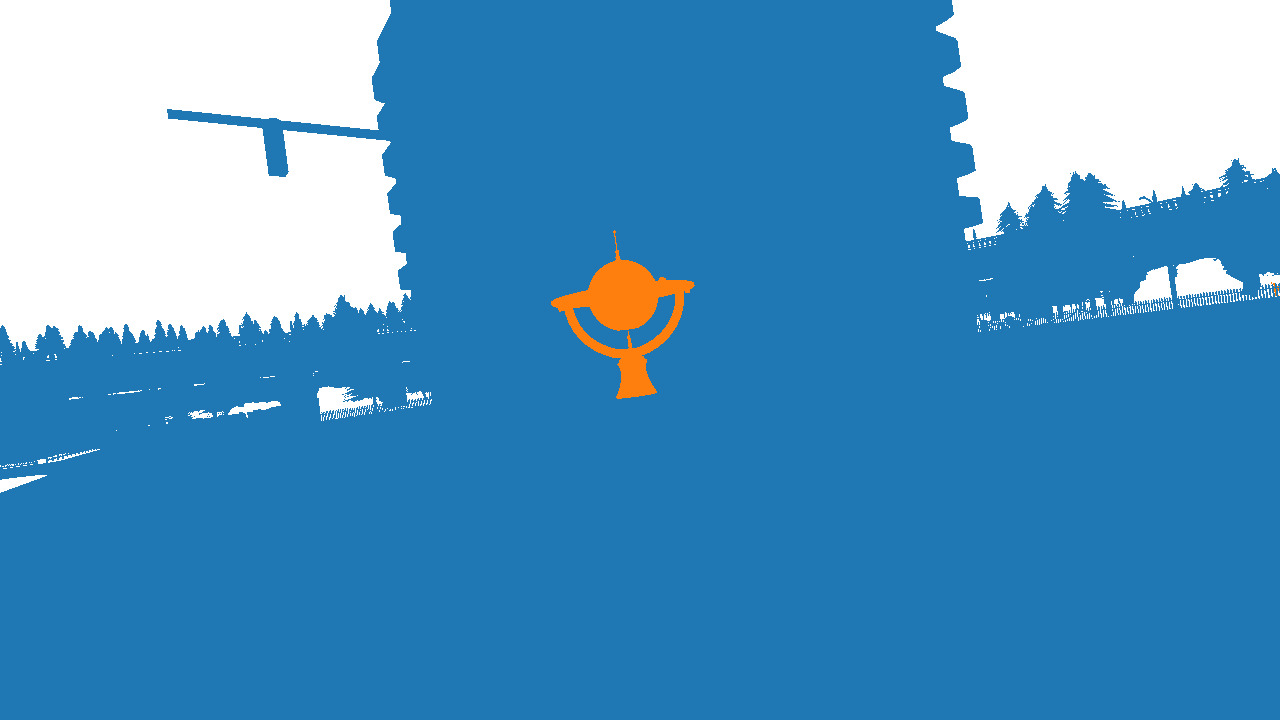}
        \caption*{\scriptsize Ground truth}
    \end{subfigure}
    \hspace{3px}
    \begin{subfigure}[b]{0.11\textwidth}
        \centering
        \includegraphics[width=\textwidth]{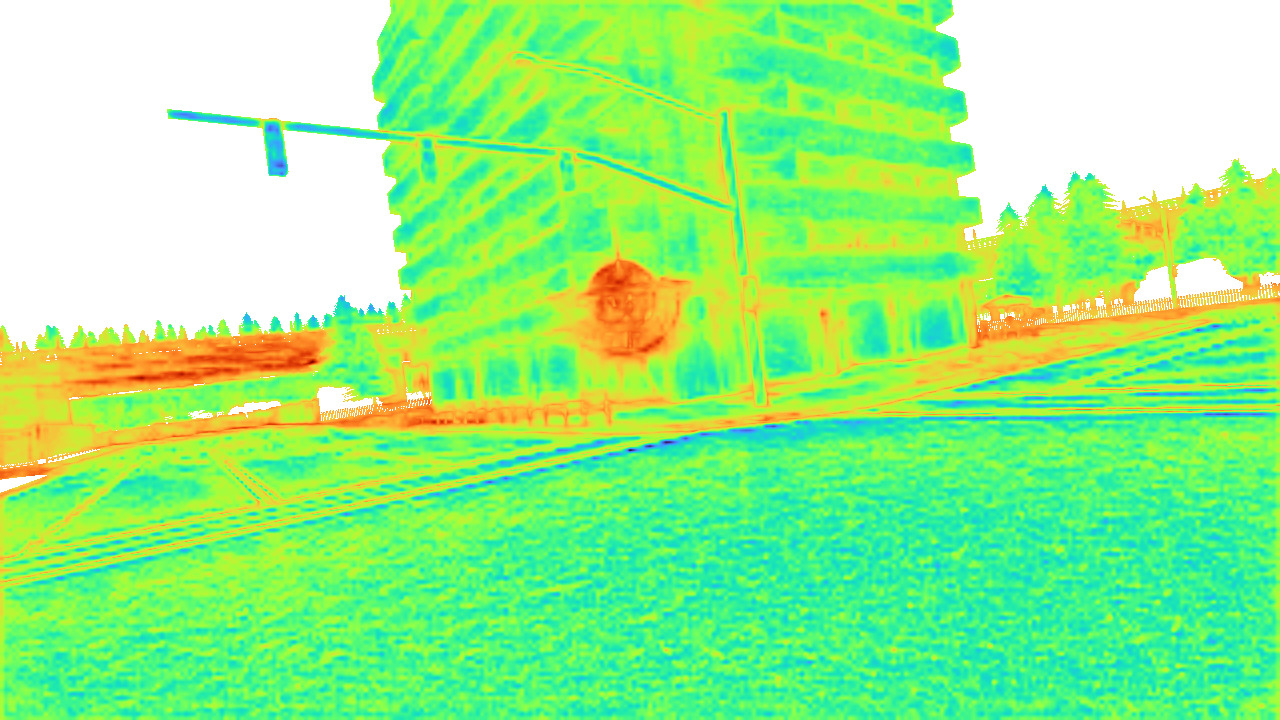}
        \caption*{\scriptsize Param.-CNXT}
    \end{subfigure}
    \begin{subfigure}[b]{0.11\textwidth}
        \centering
        \includegraphics[width=\textwidth]{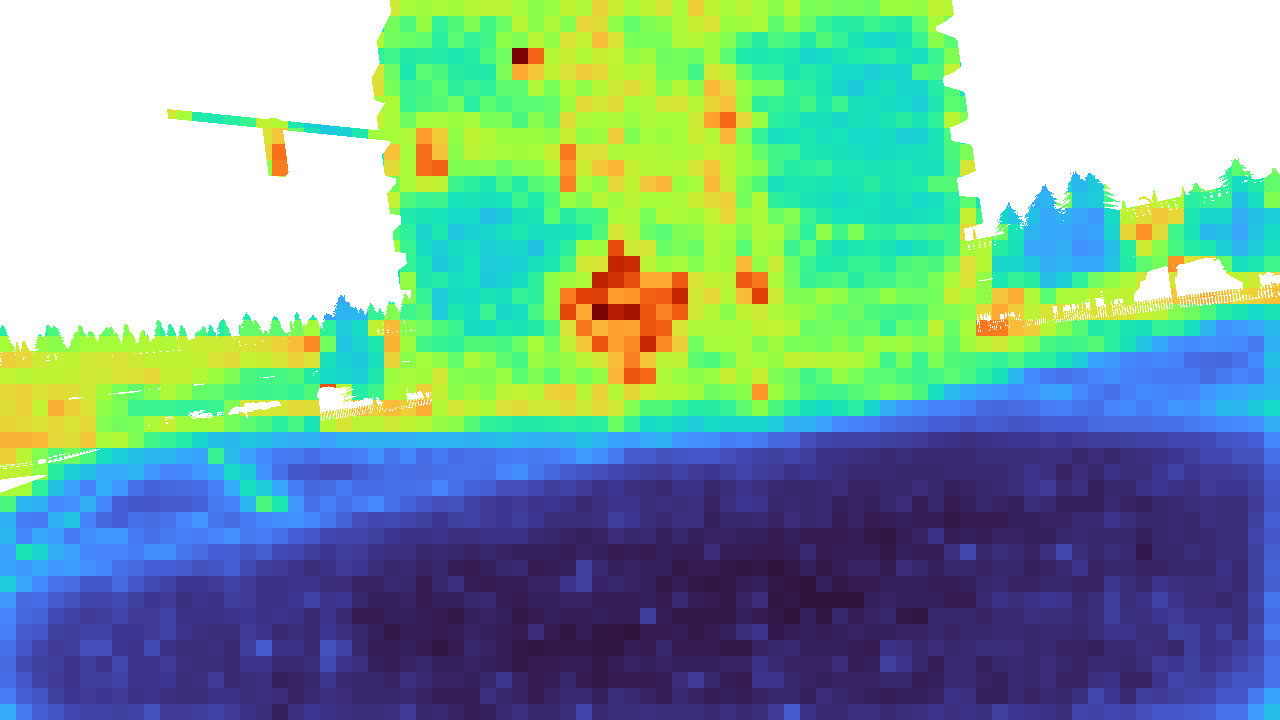}
        \caption*{\scriptsize DNP-CNXT}
    \end{subfigure}
    \begin{subfigure}[b]{0.11\textwidth}
        \centering
        \includegraphics[width=\textwidth]{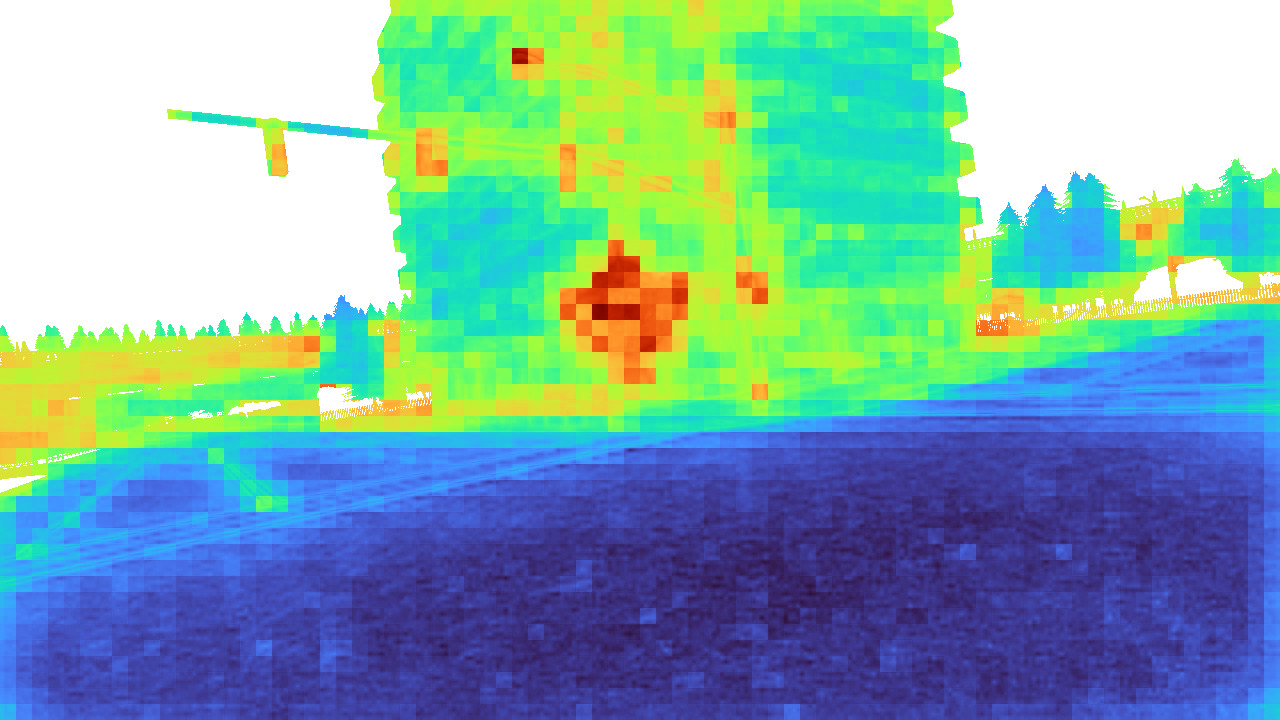}
        \caption*{\scriptsize cDNP-CNXT}
    \end{subfigure}
    \hspace{3px}
    \begin{subfigure}[b]{0.11\textwidth}
        \centering
        \includegraphics[width=\textwidth]{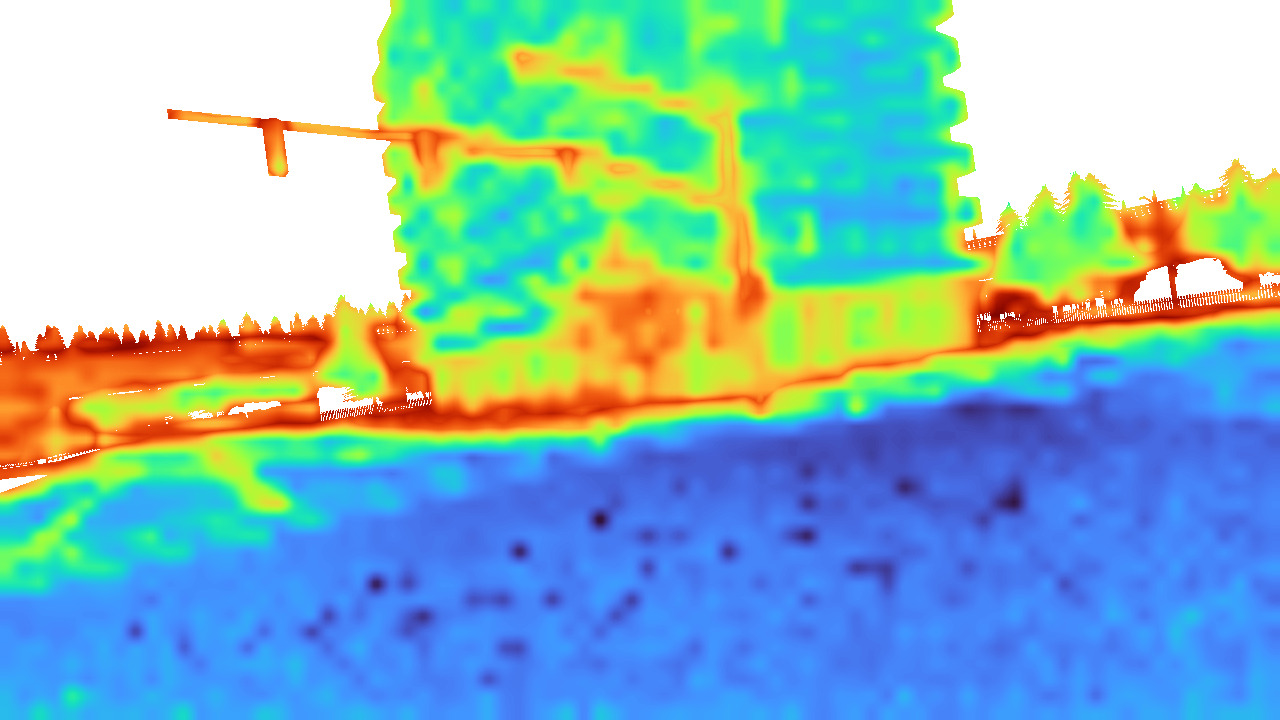}
        \caption*{\scriptsize Param.-ViT}
    \end{subfigure}
    \begin{subfigure}[b]{0.11\textwidth}
        \centering
        \includegraphics[width=\textwidth]{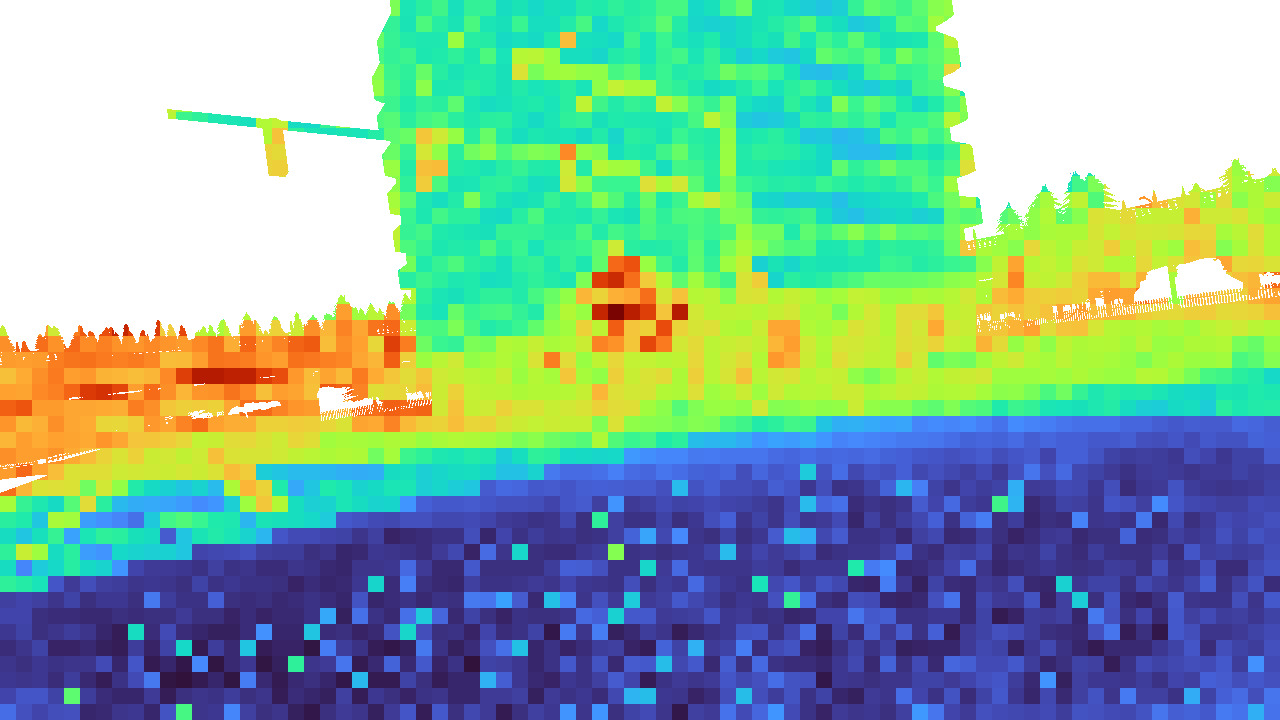}
        \caption*{\scriptsize DNP-ViT}
    \end{subfigure}
    \begin{subfigure}[b]{0.11\textwidth}
        \centering
        \includegraphics[width=\textwidth]{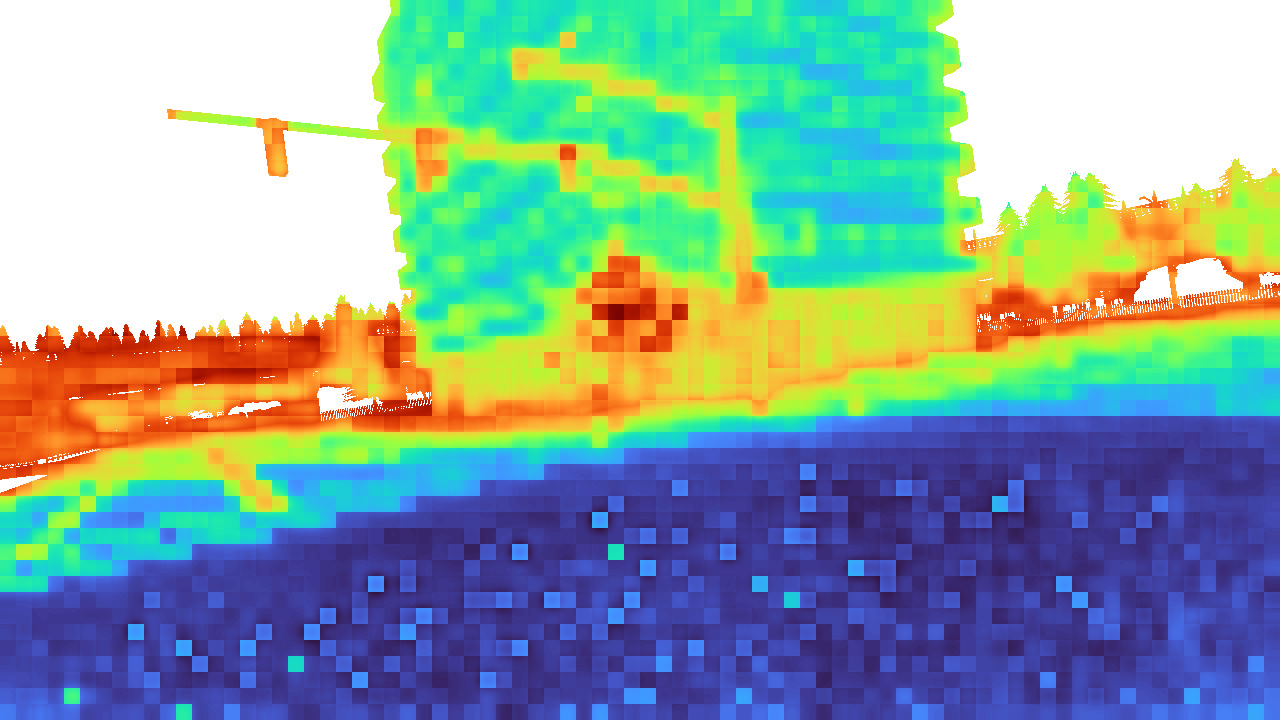}
        \caption*{\scriptsize cDNP-ViT}
    \end{subfigure}
    \caption{Qualitative results with UperNet-ConvNeXt-S (CNXT) and Segmenter-ViT-B. The top two examples are from RoadAnomaly, followed by two StreetHazards ones. The score maps show how the combination of parametric scores and DNP ones is an improvement over both, mostly through the removal/filtering of false positives. In the first and second row, ViT is clearly outperforming ConvNeXt, whereas in the last row it is the other way around.
    }
    \label{fig:quali_cnxt_vit}
\end{figure*}

More examples from both datasets, including results from both ViT-B and UPerNet-ConvNeXt-S are shown in Figure~\ref{fig:quali_cnxt_vit}. Here the benefits of the combined scores over parametric ones and DNP can be seen, especially for ConvNeXt.
While ViT generally performs better than ConvNeXt, the example in the third row is an exception.

A qualitative comparison with the RbA~\cite{RbA} approach is shown in Figure~\ref{fig:sota}. More qualitative results can be found in the Appendix.

\begin{figure}[h!]
    \centering
    \begin{subfigure}[b]{0.10\textwidth}
        \centering
        \includegraphics[width=\textwidth]{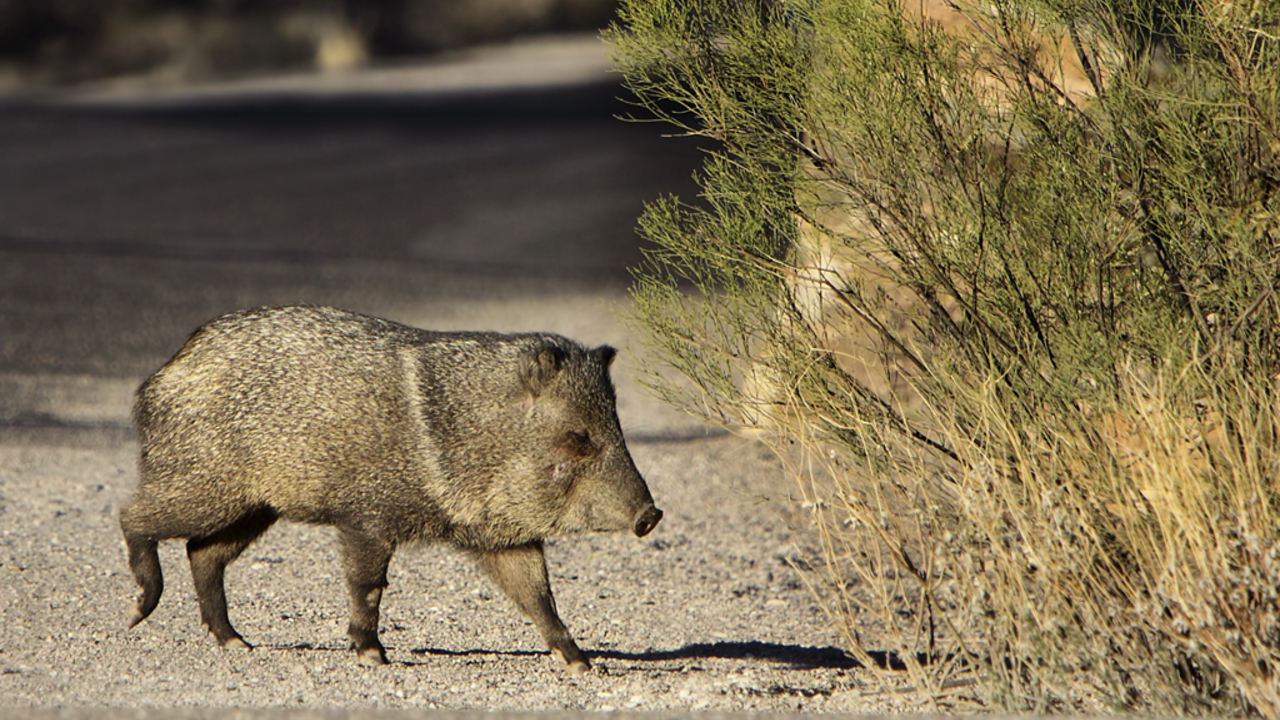}
    \end{subfigure}
    \begin{subfigure}[b]{0.10\textwidth}
        \centering
        \includegraphics[width=\textwidth]{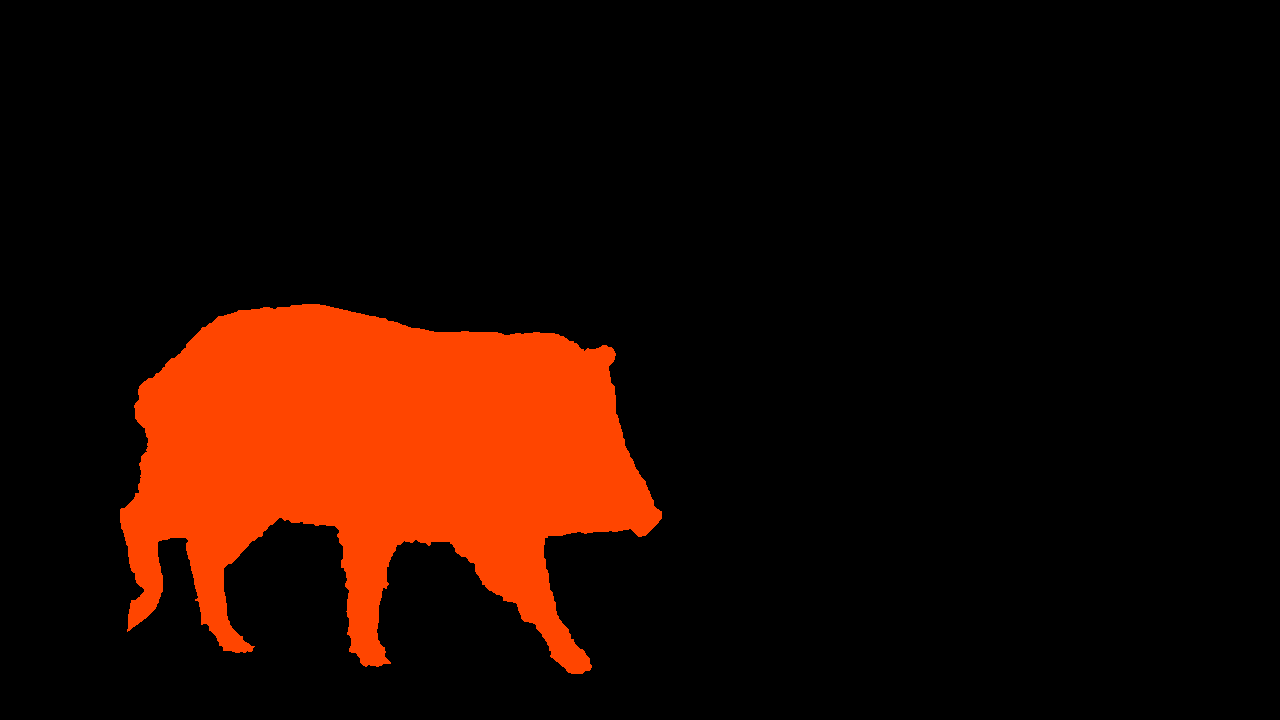}
    \end{subfigure}
    \begin{subfigure}[b]{0.10\textwidth}
        \centering
        \includegraphics[width=\textwidth]{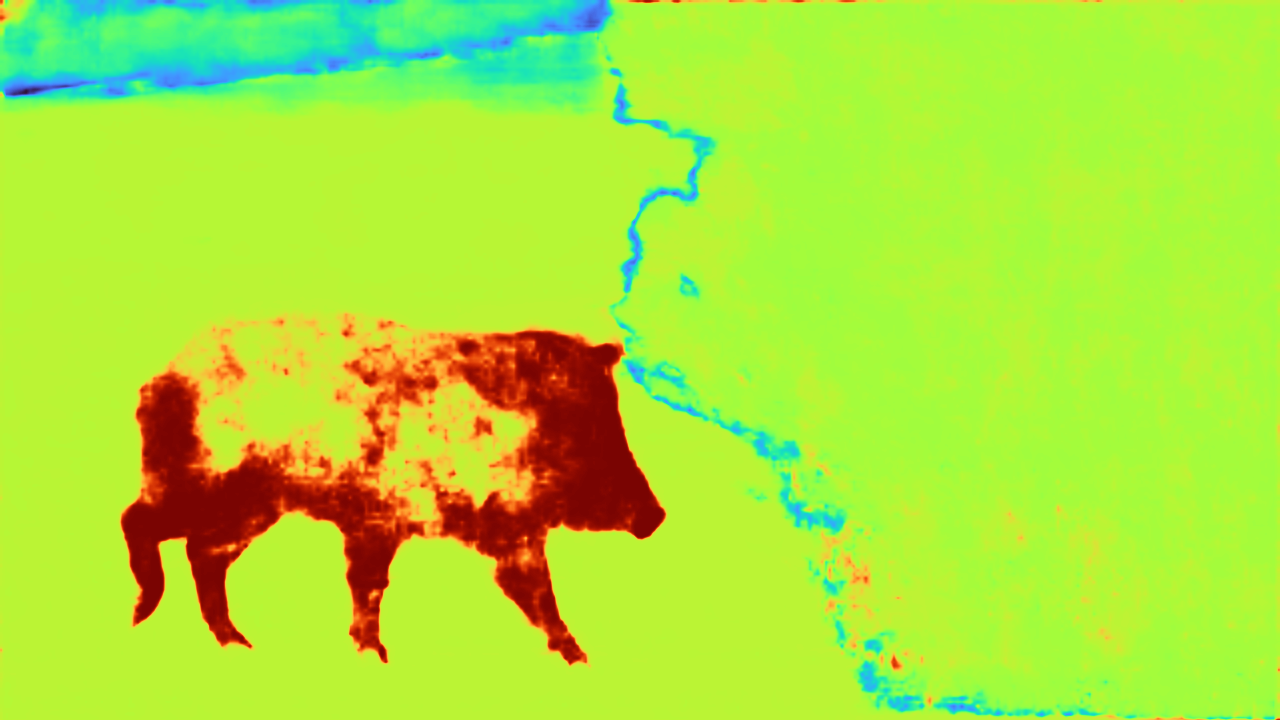}
    \end{subfigure}
    \begin{subfigure}[b]{0.10\textwidth}
        \centering
        \includegraphics[width=\textwidth]{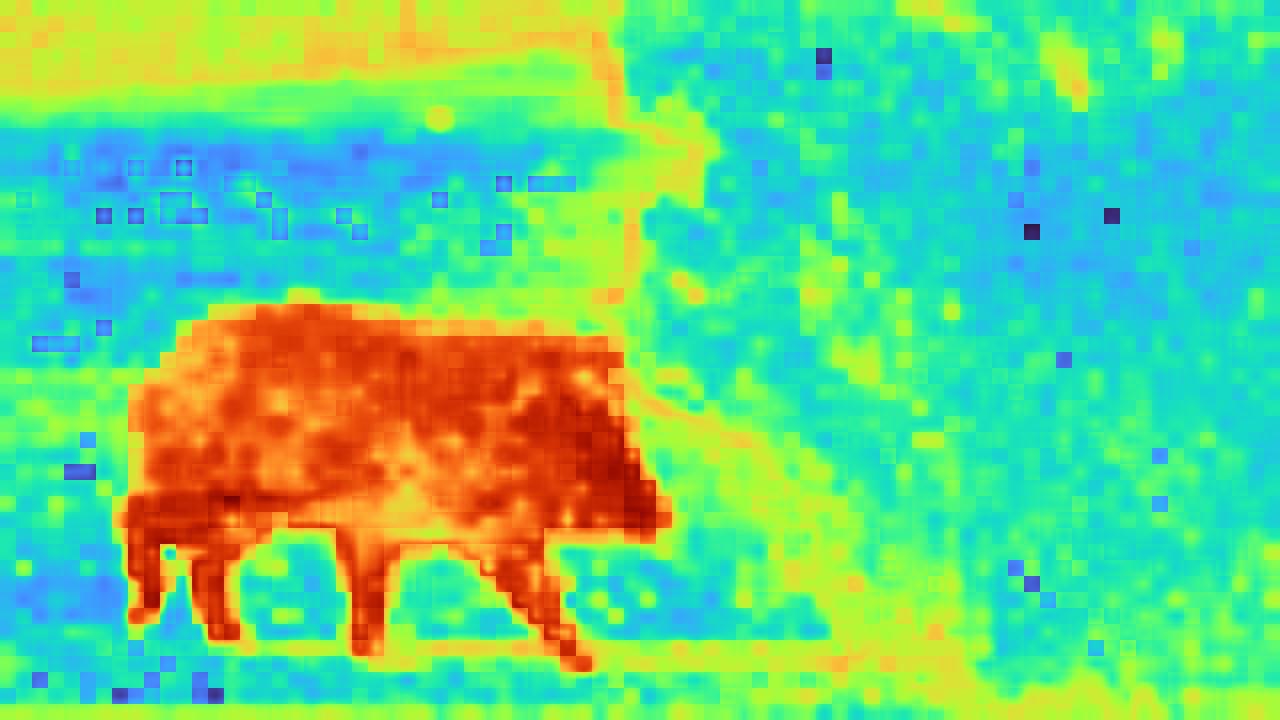}
    \end{subfigure}
    
    \begin{subfigure}[b]{0.10\textwidth}
        \centering
        \includegraphics[width=\textwidth]{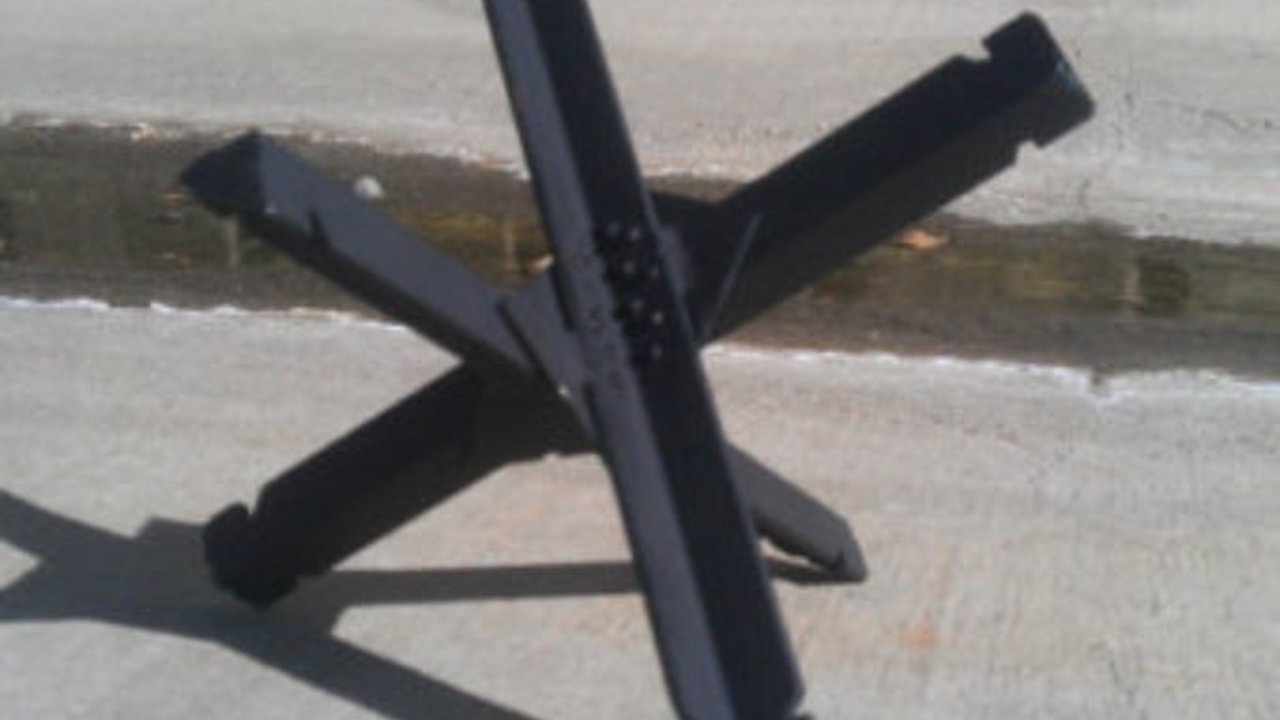}
    \end{subfigure}
    \begin{subfigure}[b]{0.10\textwidth}
        \centering
        \includegraphics[width=\textwidth]{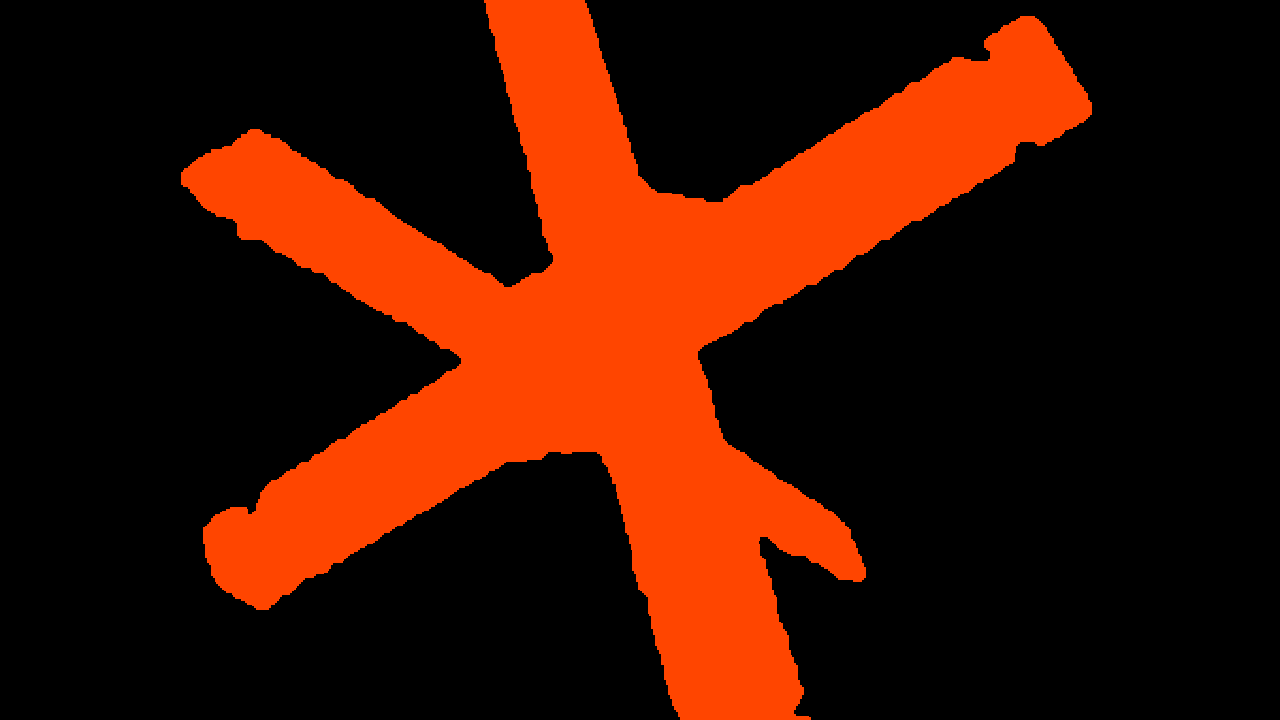}
    \end{subfigure}
    \begin{subfigure}[b]{0.10\textwidth}
        \centering
        \includegraphics[width=\textwidth]{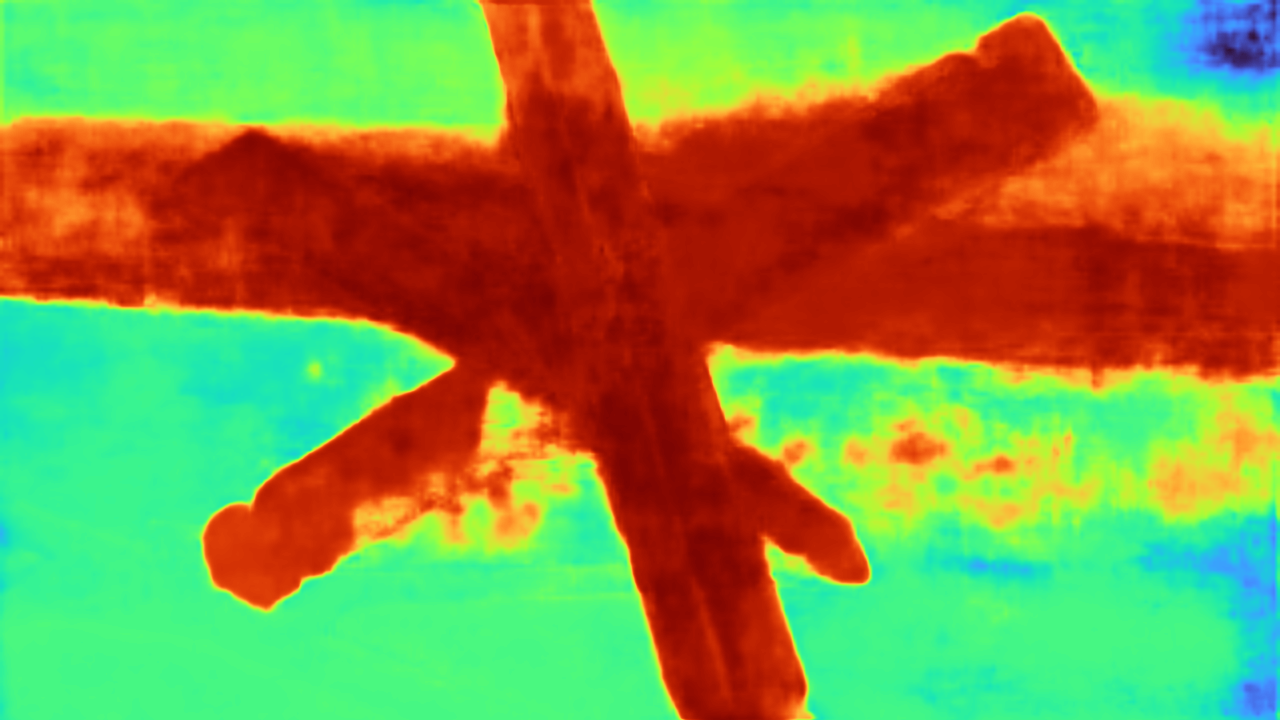}
    \end{subfigure}
    \begin{subfigure}[b]{0.10\textwidth}
        \centering
        \includegraphics[width=\textwidth]{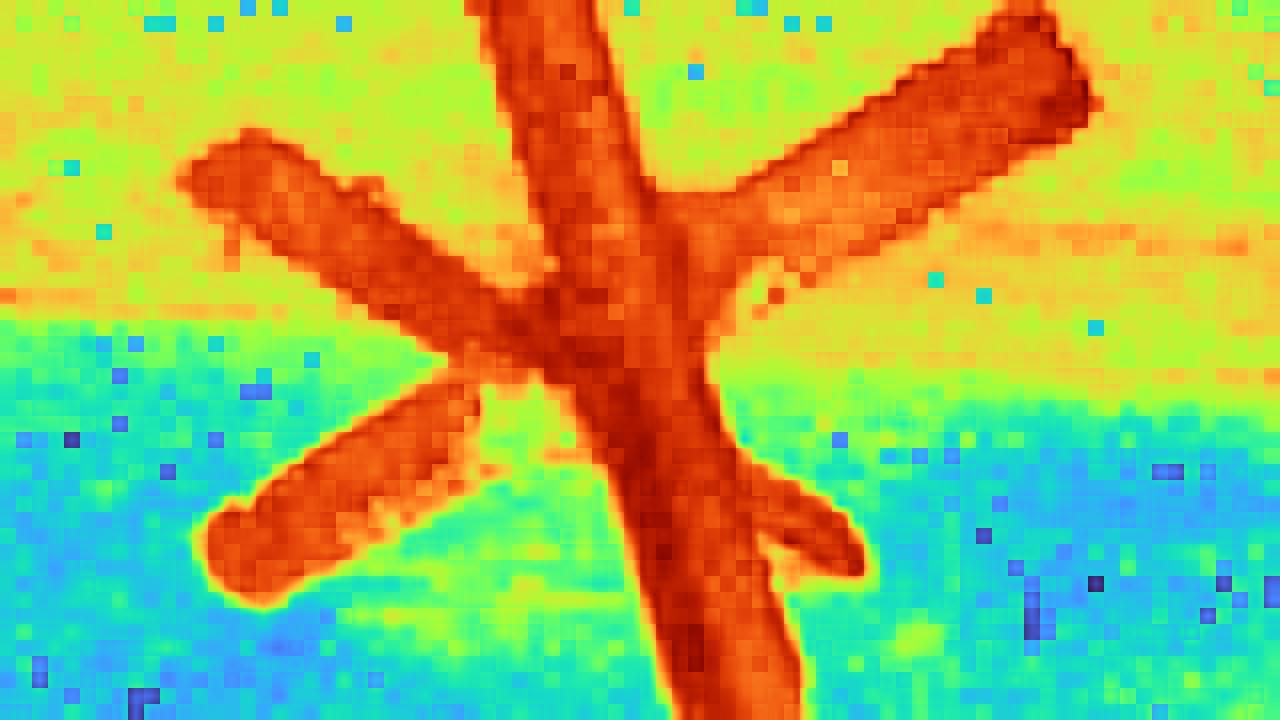}
    \end{subfigure}
    
    \begin{subfigure}[b]{0.10\textwidth}
        \centering
        \includegraphics[width=\textwidth]{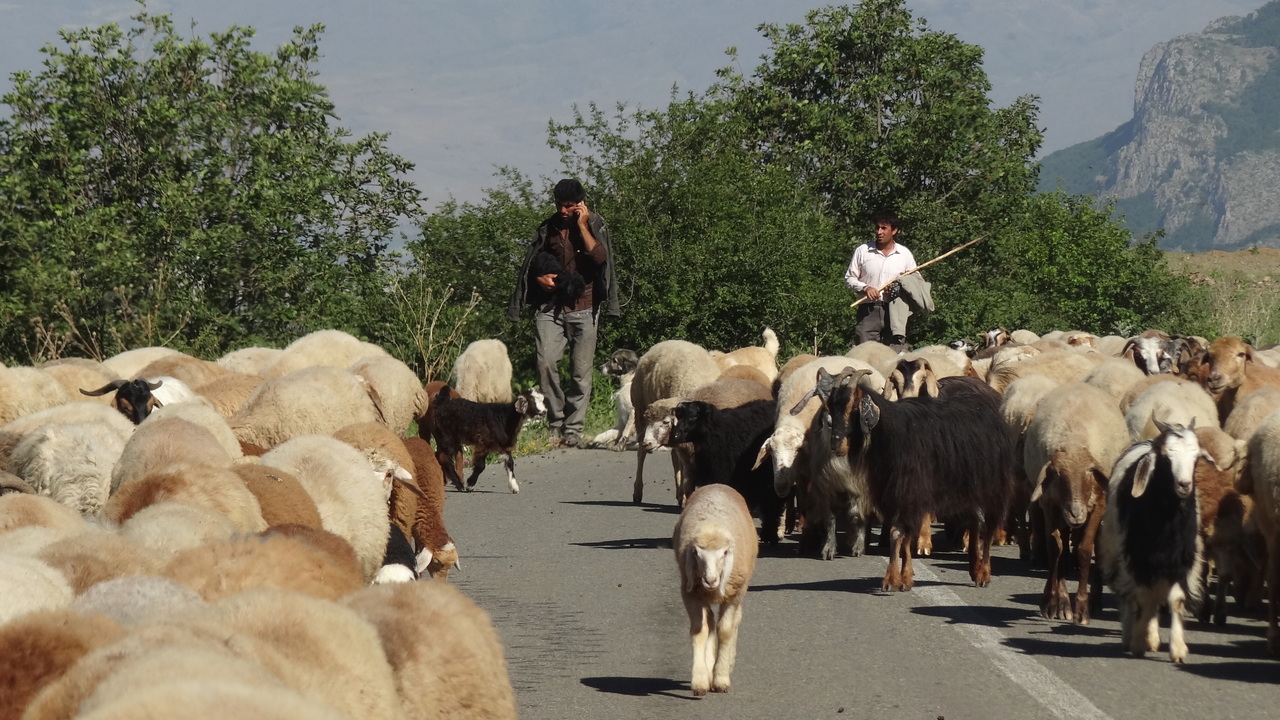}
    \end{subfigure}
    \begin{subfigure}[b]{0.10\textwidth}
        \centering
        \includegraphics[width=\textwidth]{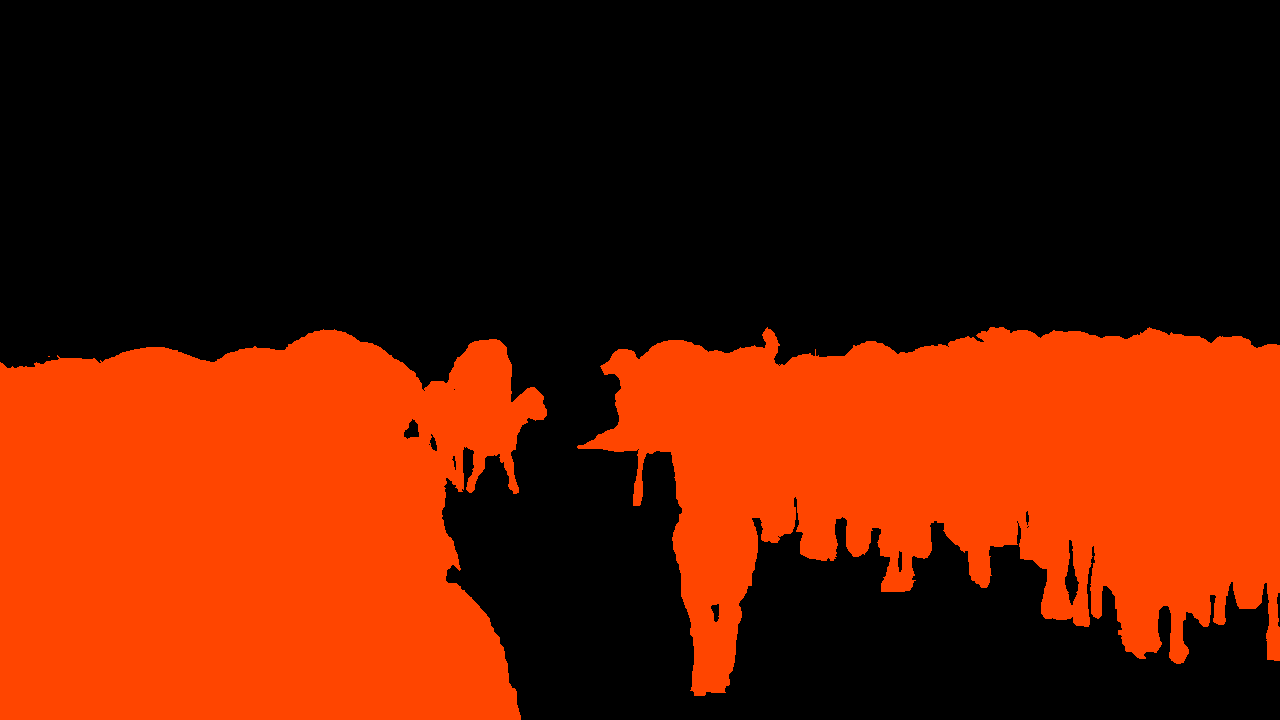}
    \end{subfigure}
    \begin{subfigure}[b]{0.10\textwidth}
        \centering
        \includegraphics[width=\textwidth]{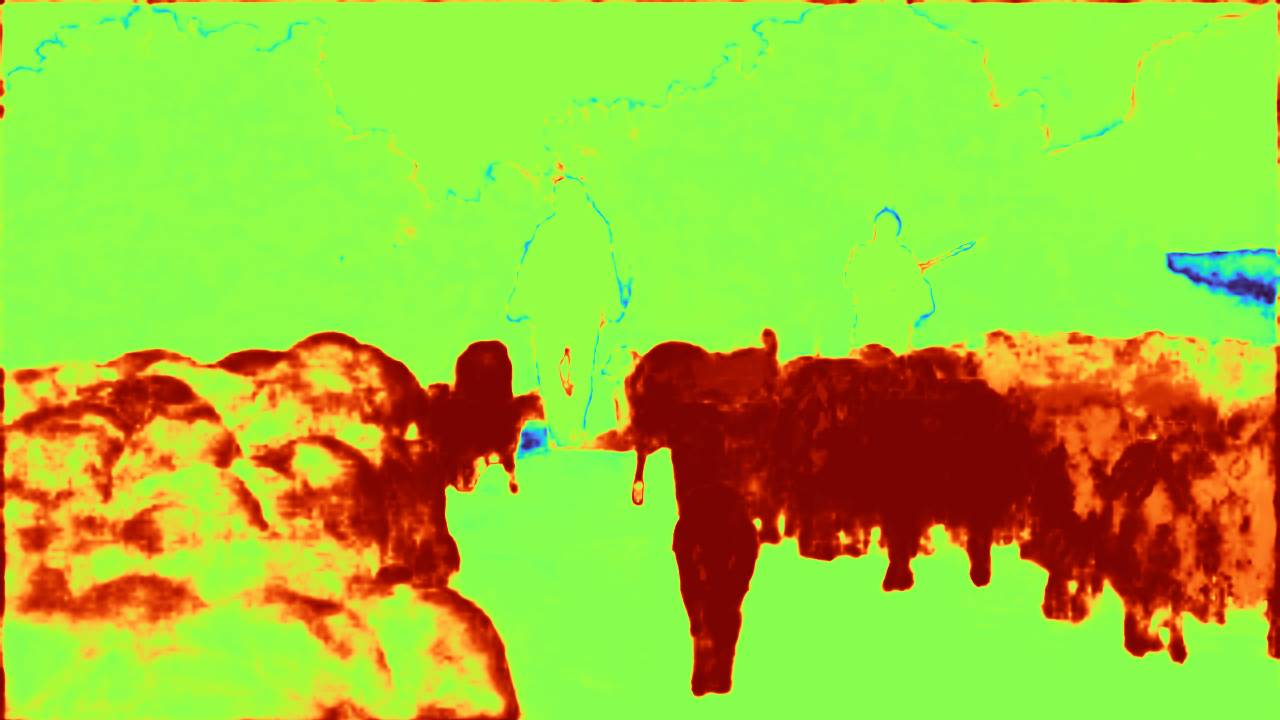}
    \end{subfigure}
    \begin{subfigure}[b]{0.10\textwidth}
        \centering
        \includegraphics[width=\textwidth]{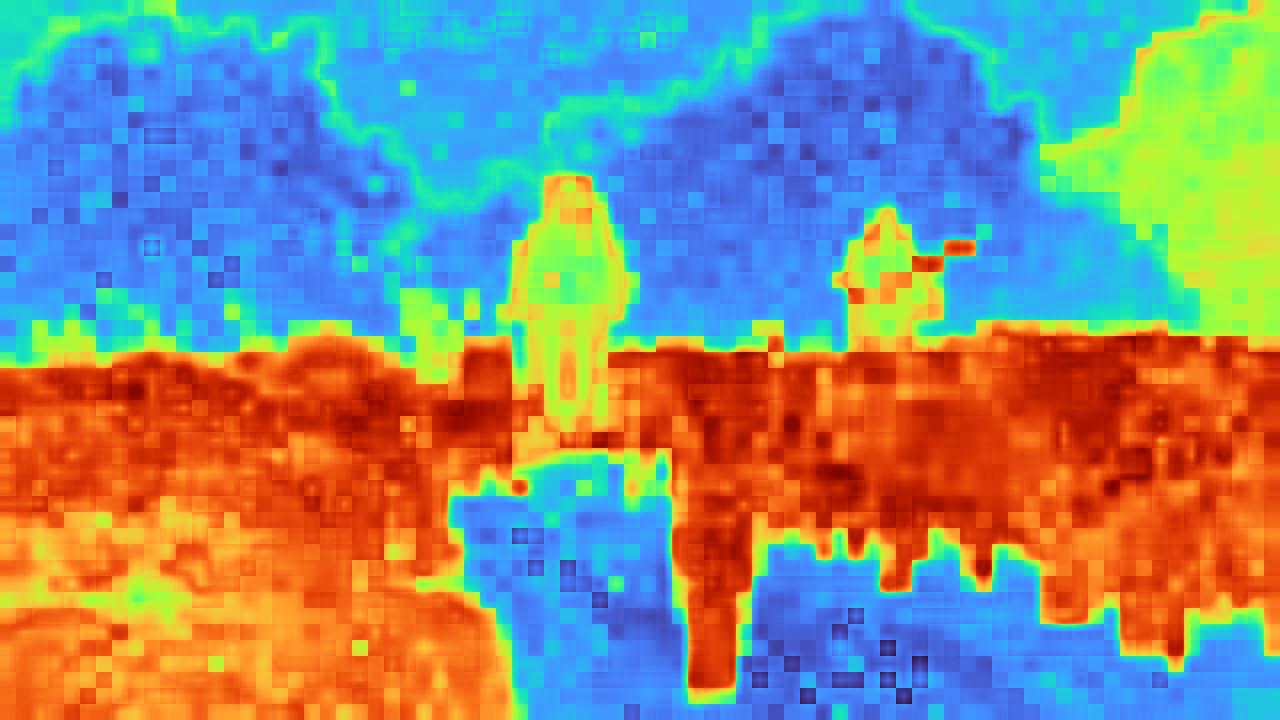}
    \end{subfigure}
    
    \begin{subfigure}[b]{0.10\textwidth}
        \centering
        \includegraphics[width=\textwidth]{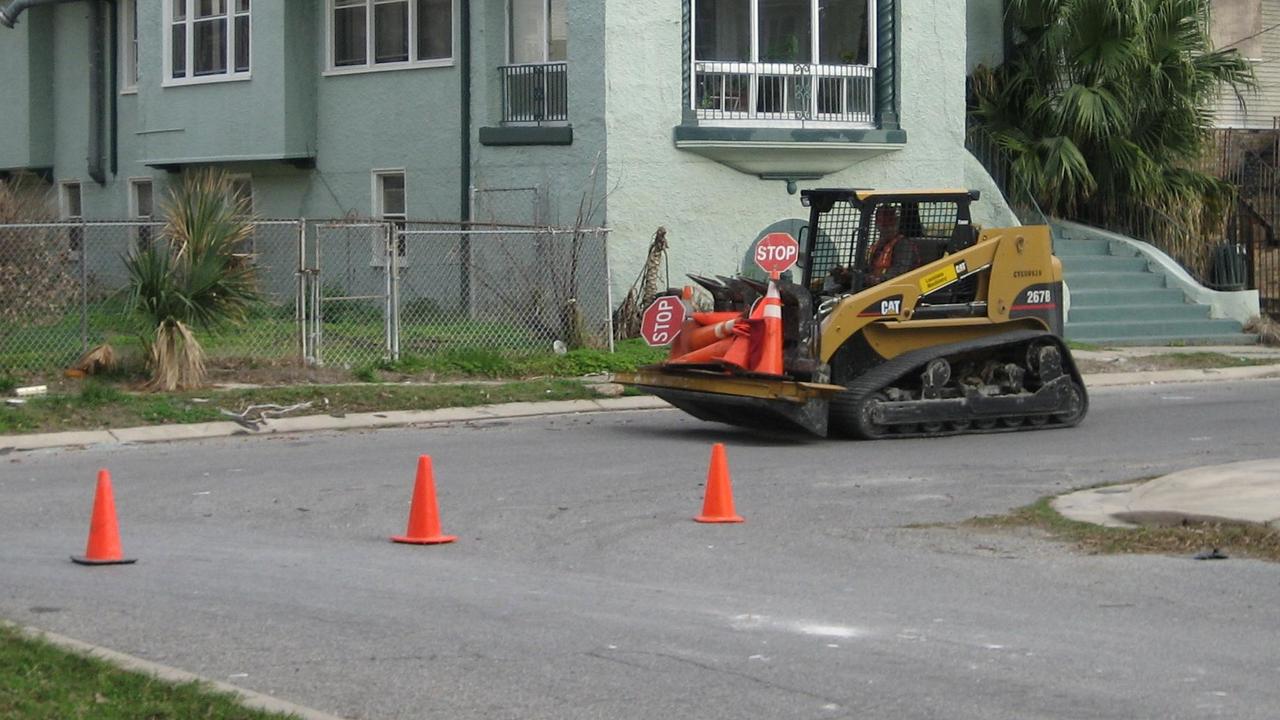}
    \end{subfigure}
    \begin{subfigure}[b]{0.10\textwidth}
        \centering
        \includegraphics[width=\textwidth]{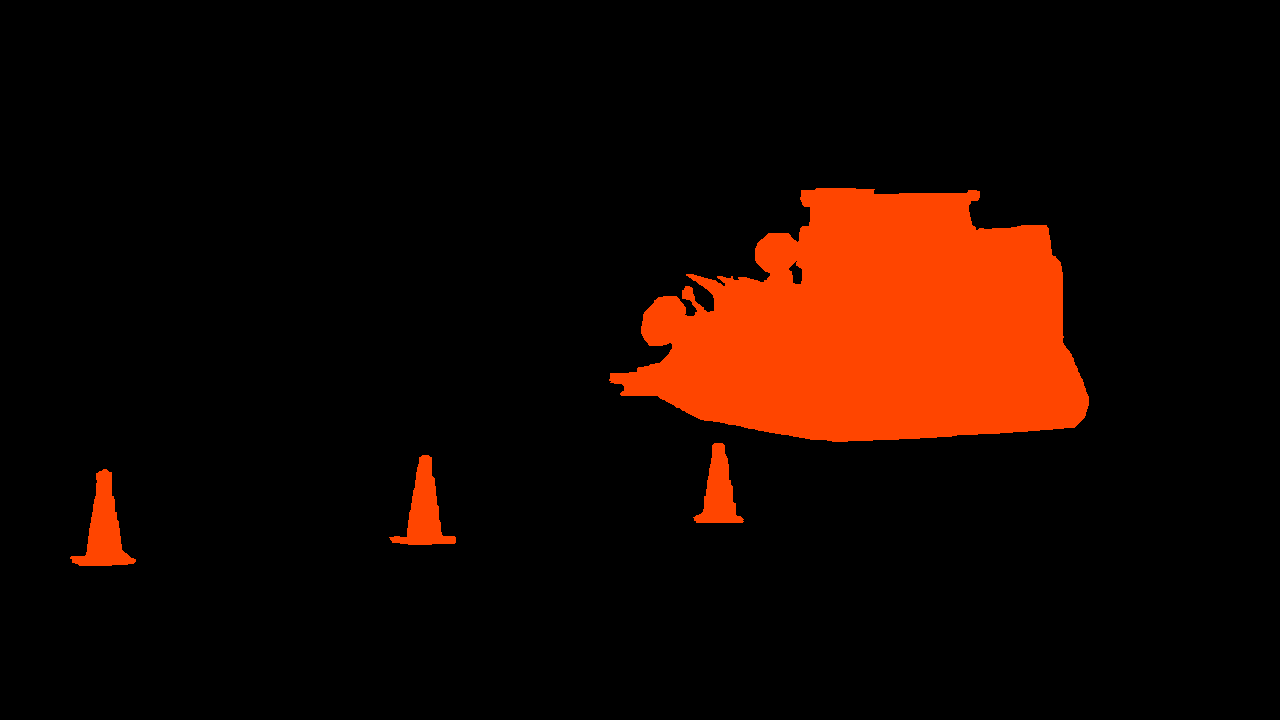}
    \end{subfigure}
    \begin{subfigure}[b]{0.10\textwidth}
        \centering
        \includegraphics[width=\textwidth]{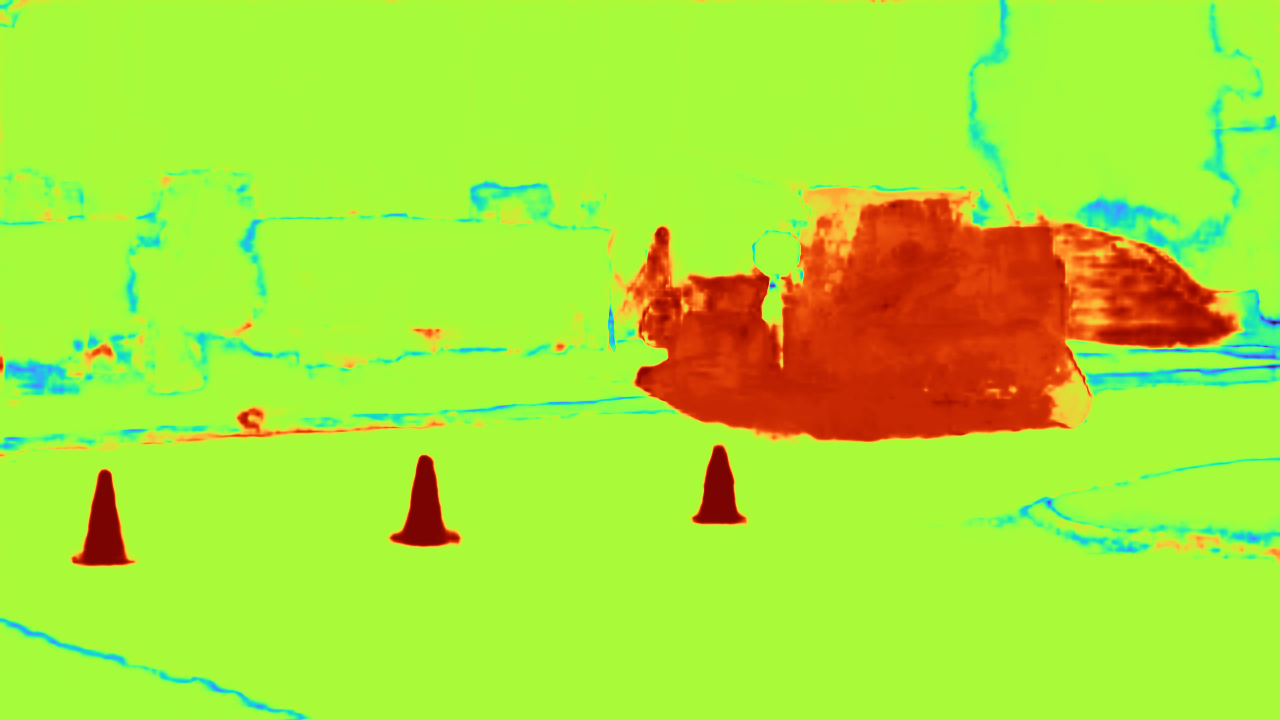}
    \end{subfigure}
    \begin{subfigure}[b]{0.10\textwidth}
        \centering
        \includegraphics[width=\textwidth]{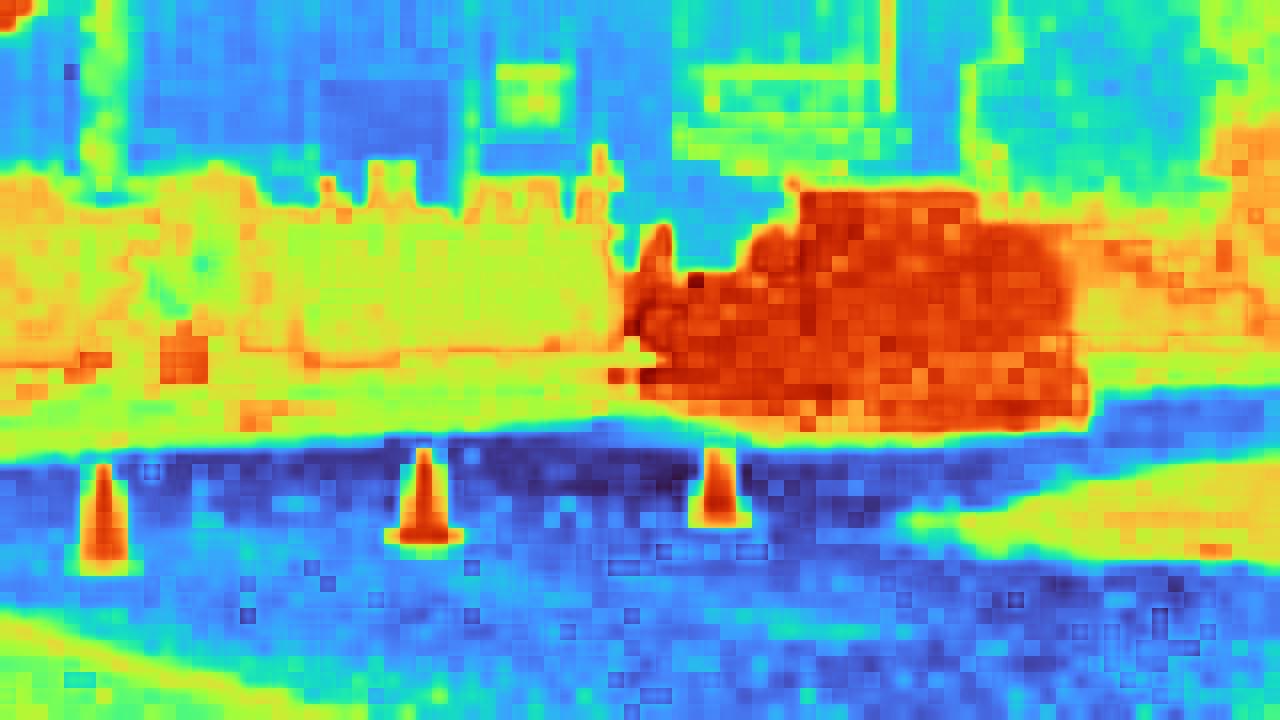}
    \end{subfigure}
    
    \begin{subfigure}[b]{0.10\textwidth}
        \centering
        \includegraphics[width=\textwidth]{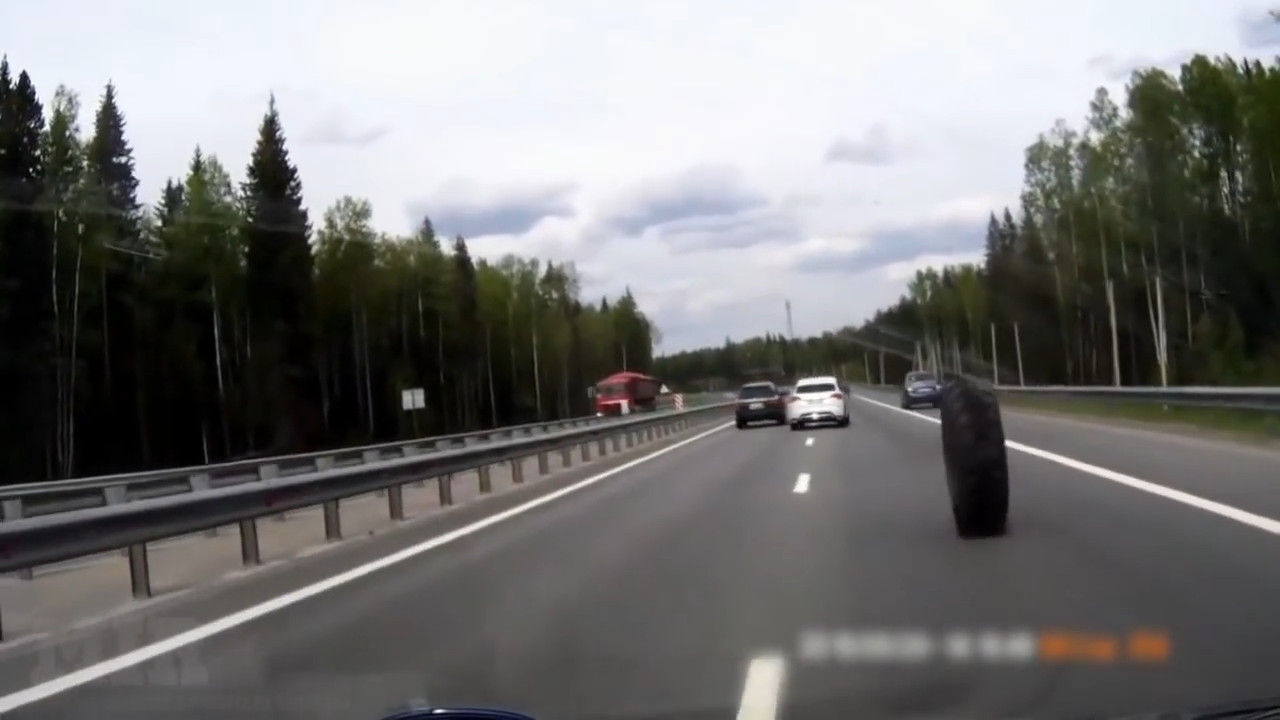}
        \caption*{\scriptsize Image}
    \end{subfigure}
    \begin{subfigure}[b]{0.10\textwidth}
        \centering
        \includegraphics[width=\textwidth]{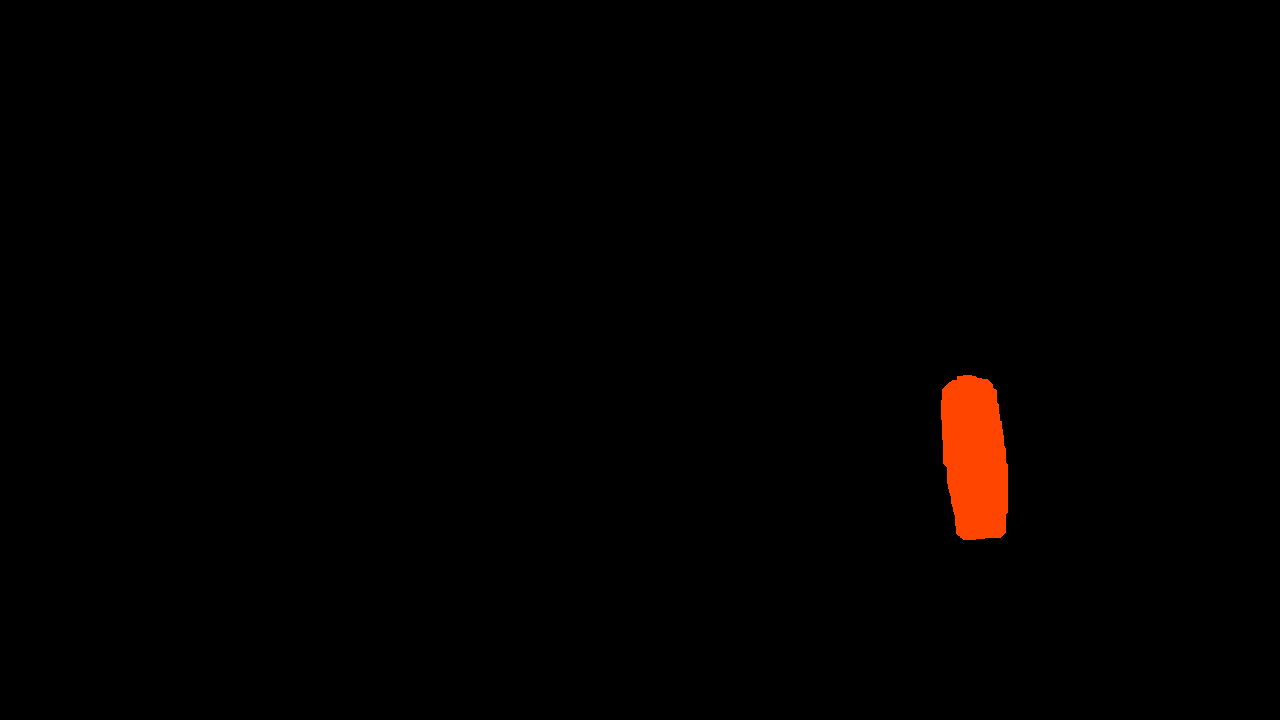}
        \caption*{\scriptsize Ground Truth}
    \end{subfigure}
    \begin{subfigure}[b]{0.10\textwidth}
        \centering
        \includegraphics[width=\textwidth]{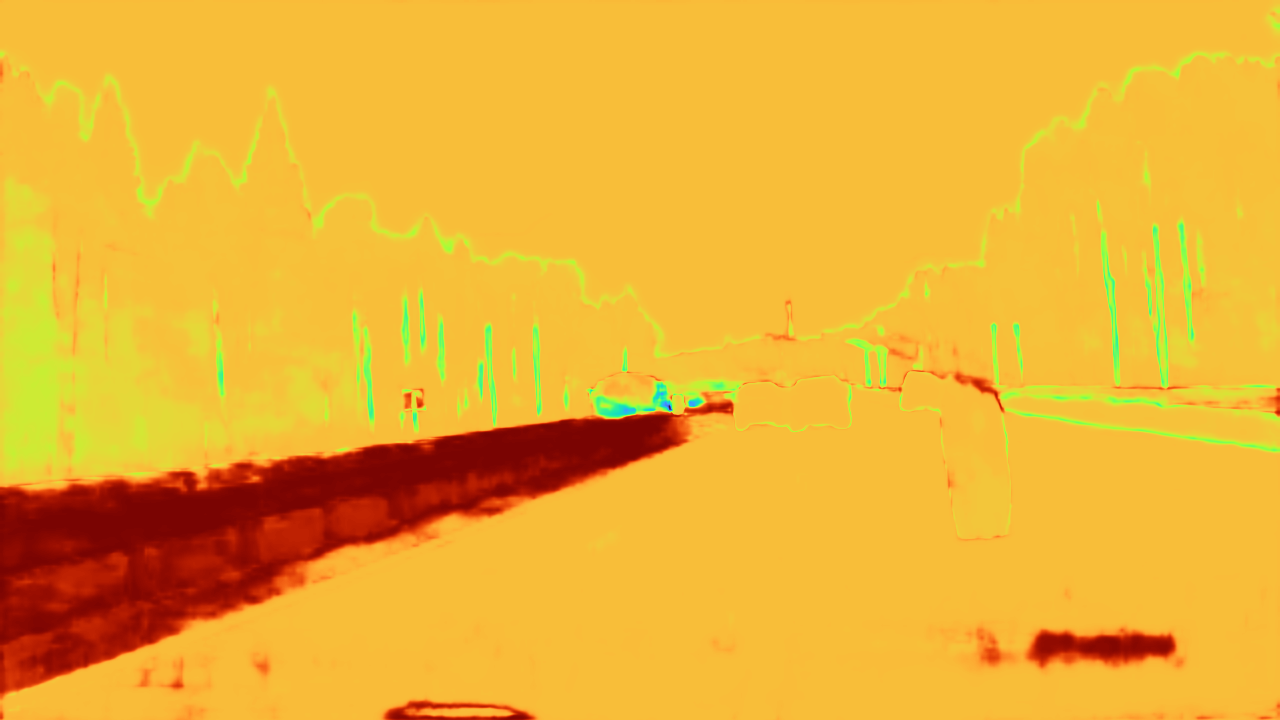}
        \caption*{\scriptsize RbA}
    \end{subfigure}
    \begin{subfigure}[b]{0.10\textwidth}
        \centering
        \includegraphics[width=\textwidth]{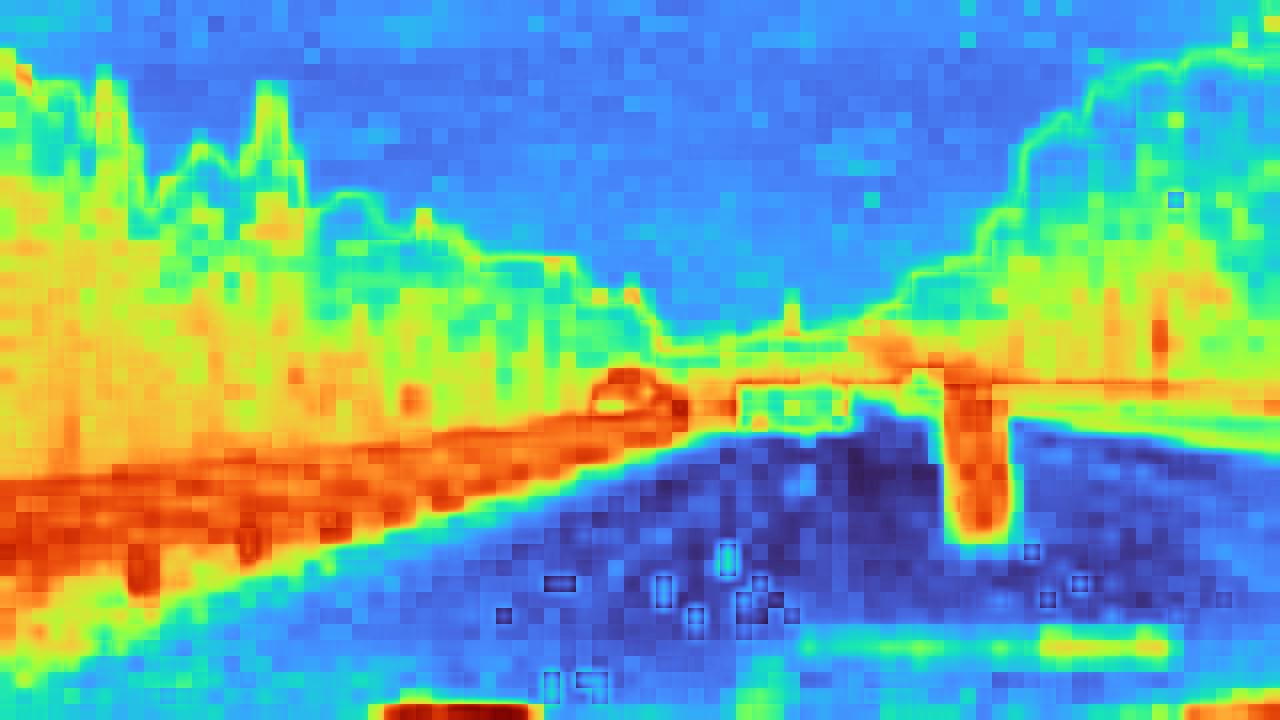}
        \caption*{\scriptsize cDNP (Ours)}
    \end{subfigure}
    
    \caption{Qualitative comparison between cDNP and RbA. The scoremaps from the latter have sharper contours, due to the mask based inference: this can lead to more accurate detection of smaller objects (fourth example), but also to stark false negatives/positives (first three examples). Both methods fail, albeit differently, on the last example.}
    \label{fig:sota}
\end{figure}

\subsection{Reference Feature Subsampling Methods}
\label{sec:ablation_subsampling}
\begin{figure}
    \centering
    \includegraphics[width=0.45\textwidth]{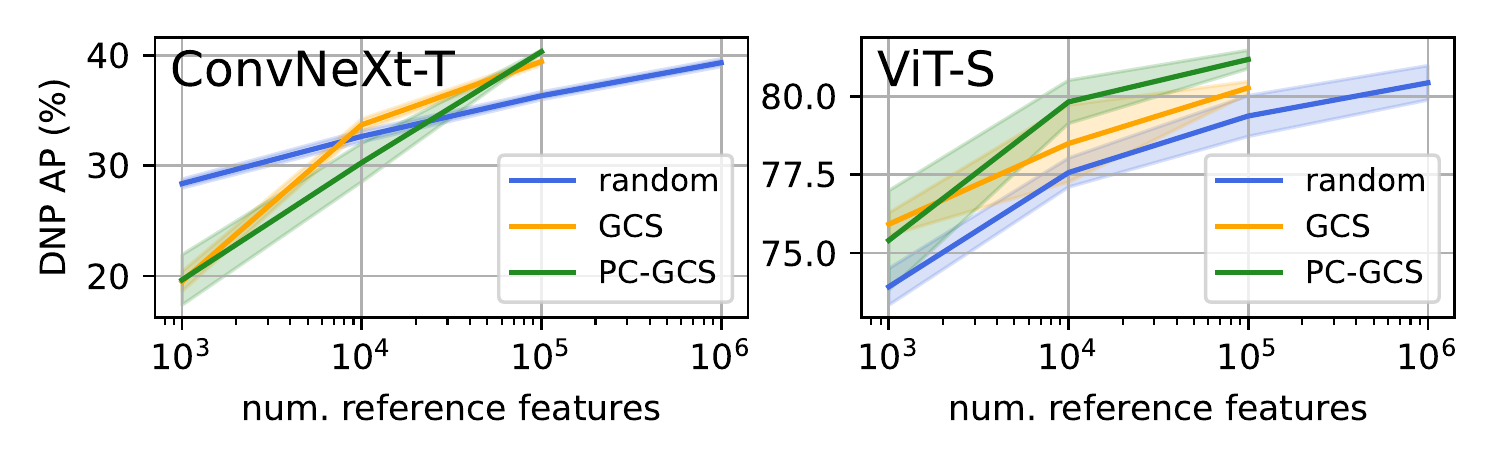}
    \caption{Results of different subsampling methods for ConvNeXt and ViT on RoadAnomaly. On the x-axes are the number of samples, on the y-axes the performance of DNP. All sampling methods perform well given enough reference features, with a slight superiority of PC-GCS.}
    \label{fig:ablation-sampling}
\end{figure}

Here we discuss the results on the choice of the subsampling method: random, GCS, PC-CGS, which are summarized in Figure~\ref{fig:ablation-sampling} for ConvNeXt-T and ViT-S on RoadAnomaly. We evaluate for different values of $N$: $1$k, $10$k, $100$k, and 1 million for random only, due to the prohibitive pre-processing costs of the coreset approaches. Each setup is evaluated with three random seeds.

The ranking between the methods changes with the architecture and the number of reference samples, but all methods are similarly competitive given enough features. The performance does not saturate with 1M reference samples, however the inference costs in that regime make the approach less attractive.
Throughout the following experiments we use PC-GCS, based on its results and faster pre-processing times than GCS.

\subsection{Computational Costs}
A major point of concern with k-nearest-neighbors is the computational cost due to the distance computations and search on large feature sets. This depends on two major factors: the number of reference features $N$, and the network architecture, which determines the test feature resolution $H{\cdot}W$ and channel size $C$. We estimate the runtime for each k-nearest-neighbors distance computation using the \texttt{IndexFlat} exact search index provided by the \texttt{faiss}~\cite{johnson2019billion} library, on an NVIDIA RTX 2080Ti card.

\begin{figure}[t]
    \centering
    \includegraphics[width=0.25\textwidth]{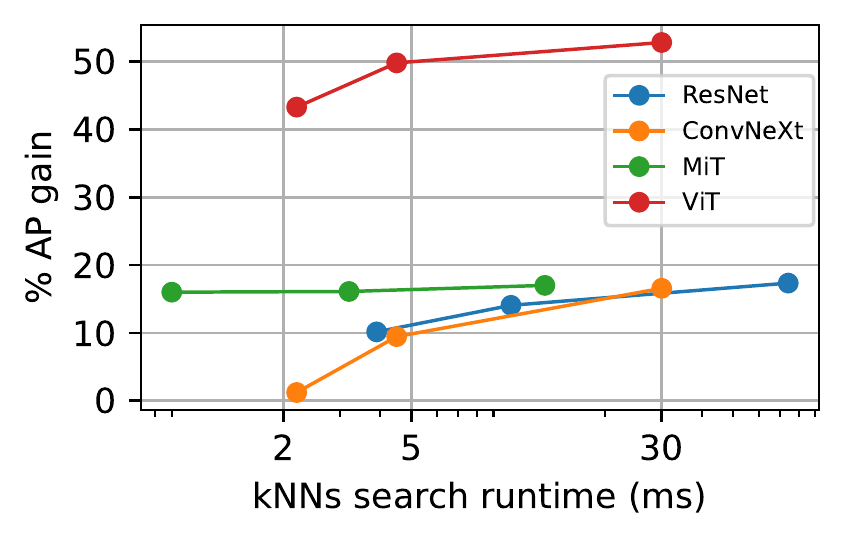}
    \vspace{-0.8em}
    \caption{
    Percentage performance gains in terms of AP (cDNP vs. parametric) over nearest-neighbors search time. For each of the four architectures, three reference set sizes are considered: $N{\in}\{10^3, 10^4, 10^5\}$.}
    \label{fig:overhead_vs_ap}
\end{figure}

The trade-off between the performance gain brought by our approach (cDNP) over the best performing parametric scores (LSE) and runtime is shown in Figure~\ref{fig:overhead_vs_ap}.
The x-axis shows the average kNNs search runtime, for images of  1280$\times$720px. We compare all architectures with reference feature set sizes 1k, 10k, 100k on RoadAnomaly.

ResNet50 is the most expensive architecture, due to its comparatively high feature size and resolution, while MiT, which has the lowest feature resolution, is the least expensive. MiT yields the same relative AP gain as the CNNs, but its absolute performance is much better, as per Table~\ref{tab:arch_scores_comparison}.
ConvNeXt and ViT have the same cost, but the latter offers the highest AP gain in return.

\subsection{Impact of the Number of Neighbors}
\label{sec:ablation_k}
\begin{figure}[h]
    \centering
    \includegraphics[width=0.45\textwidth]{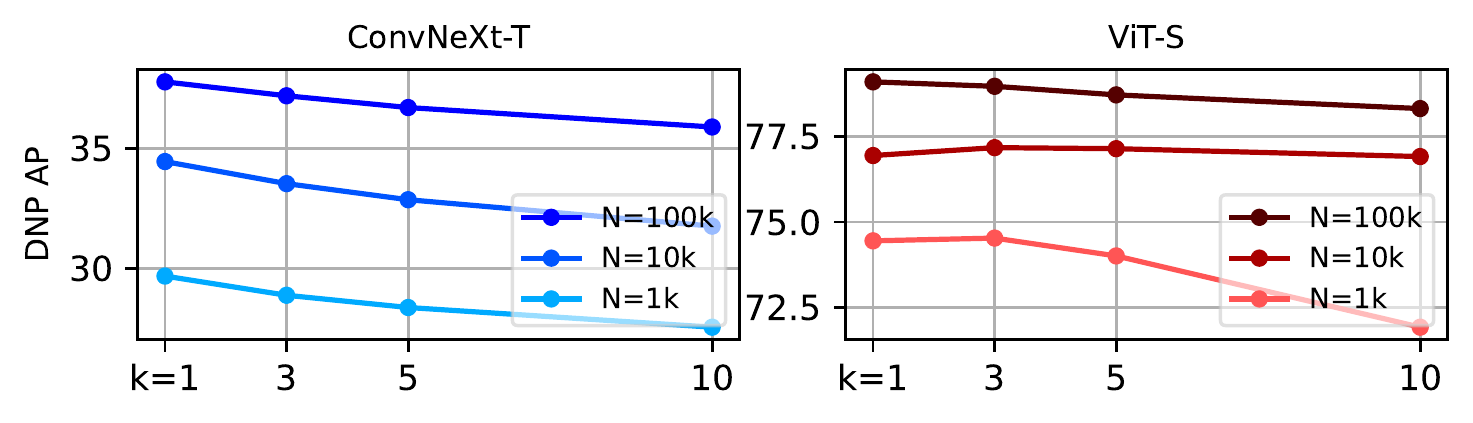}
    \caption{Effect of the number of neighbors~(k) on DNP scores for ConvNeXt-T and ViT-S, on RoadAnomaly. The optimal $k$ value depends on the number of reference features $N$ for ViT, but the difference is marginal.
    }
    \label{fig:k}
\end{figure}

The number of selected neighbors ($k$) is an important hyperparameter for any kNN-based approach, and typically depends on the size of the reference set~\cite{sun2022knnood} and on the source of the features.
An overview of the optimal $k$ values for our ConvNeXt-T and ViT-S models is presented in Figure~\ref{fig:k}, which shows that for $k>3$ the performance of both models decreases.

The choice of $k{=}3$ gives a near-optimal value and ensures a more robust distance expectation (Equation~\ref{eq:avg_nn_scores}).

\section{Feature Dimensionality and Partitioning}

A known limitation of non-parametric pattern recognition approaches, such as kNNs, is that they scale poorly to high dimensional problems. The term ``curse of dimensionality" refers to the observation that, as the data dimensionality increases, the data space becomes more sparsely populated and the distance to the nearest sample approaches the distance to the farthest one~\cite{Weber1998AQA, 10.1007/3-540-44503-X_27}. This section investigates the effect of this phenomenon in our setting and evaluates the role it plays in the large performance gap between different feature extractors observed in the previous experiments. 

Section~\ref{sec:groups_heads} explores the multi-head design as one of the reasons for the success of transformer features, and finds that a similar partitioning strategy can be used to boost OoD detection performance in CNNs.

Lower-order distance functions have also been tested to mitigate the curse of dimensionality, although with mixed success~\cite{10.1007/3-540-44503-X_27,e22101105}. See results in Appendix.

\begin{figure}
    \centering
    \includegraphics[width=0.37\textwidth]{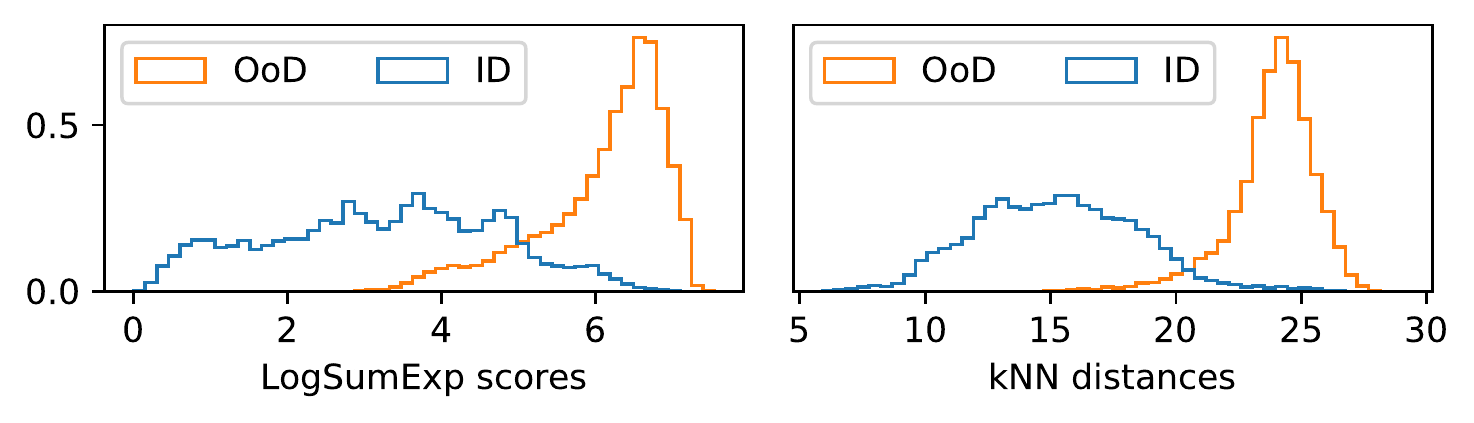}
    \caption{Density histograms of parametric scores and kNN distances, which do not collapse under the curse of dimensionality and show better ID/OoD separation.}
    \label{fig:knn_ml_density}
\end{figure}

\subsection{Curse of Dimensionality}
The results in Section~\ref{sec:experiments} leave no doubt that the representation size does not prevent the use of nearest neighbors. This is confirmed in Figure~\ref{fig:knn_ml_density}, where we compare the distributions of kNN distance and parametric scores.
Although the kNN distances do not collapse on a single value, it is still possible that their high dimensionality impairs performance. This would partially explain the performance difference between feature extractors reported in Table~\ref{tab:arch_scores_comparison}.

\begin{figure}
    \centering
    \includegraphics[width=0.43\textwidth]{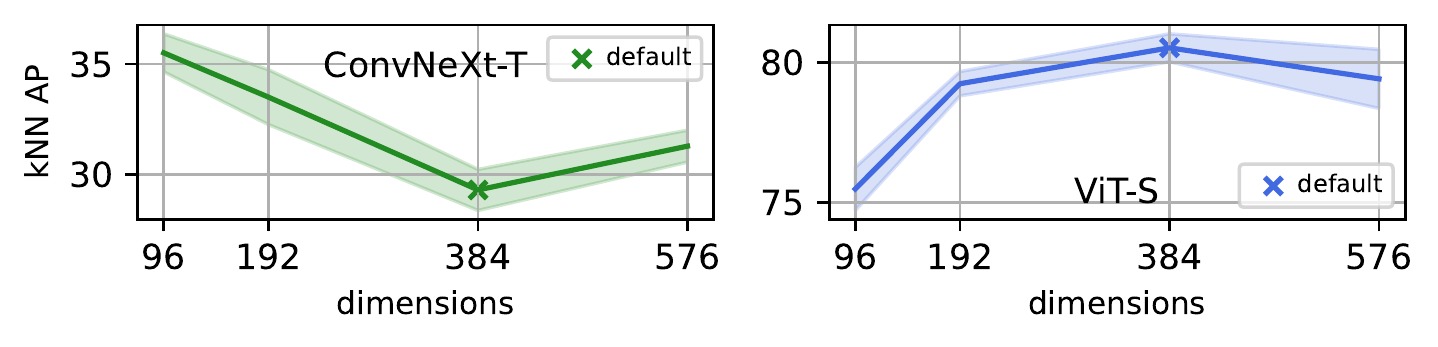}
    \caption{Effect of feature size on OoD detection performance using kNNs. The default size is the optimal for ViT, but smaller features perform better for ConvNeXt.}
    \label{fig:ft_size_ablation}
\end{figure}

To investigate this, we trained versions of UperNet-ConvNeXt-T and Segmenter-ViT-S varying the size of the features used for kNNs. Interestingly, according to the results reported in Figure~\ref{fig:ft_size_ablation}, ConvNeXt features benefit greatly from a reduced dimensionality, whereas ViT representations work best at the default size. This is despite the fact that the default feature size is the same in both cases.

\subsection{Feature Groups and kNNs}
\label{sec:groups_heads}
One major difference between ConvNeXt and ViT is that the latter computes its representations in a multi-head fashion, with each head being able to attend to different portions of the input and being responsible for a specific group of feature dimensions. This results in a functional partitioning of the representations.

We hypothesize that this design choice affects representation learning and impacts how the features behave with kNNs. We propose an ablation study in which the overall feature dimensionality is fixed to its default value (384), but the number of independent feature groups changes.

We achieve feature partitioning in ConvNeXt by using depthwise-separable convolutions~\cite{Chollet_2017_CVPR}, these divide the input features into equally sized groups, and process them independently into separate outputs before concatenating.

\begin{figure}
    \centering
    \includegraphics[width=0.43\textwidth]{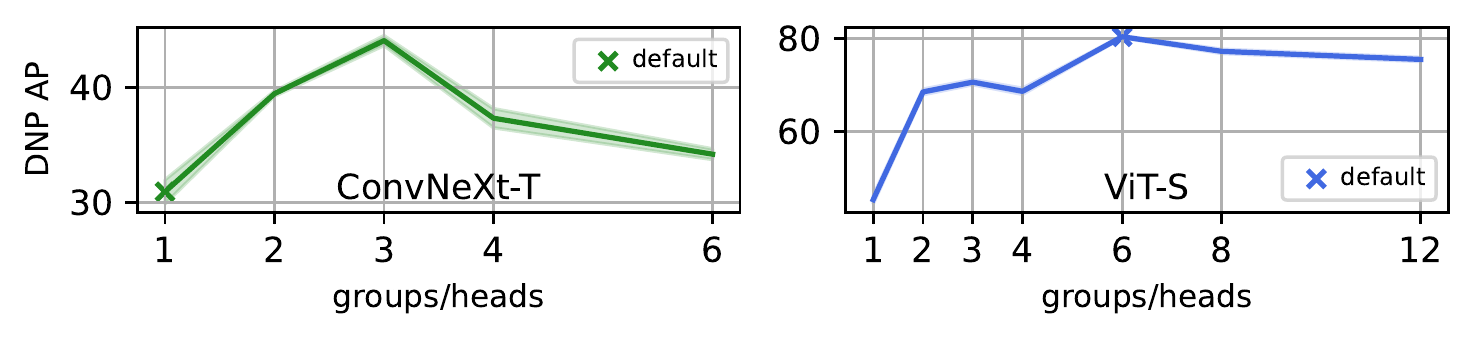}
    \caption{Effect of the number of groups (ConvNeXt) and heads (ViT) on OoD detection performance using kNNs. The behavior is similar for the two architectures, although the optimal number of feature groups is different.}
    \label{fig:groups_heads_ablation}
\end{figure}

The results of the ablation study are shown in Figure~\ref{fig:groups_heads_ablation}, and show that the two networks behave similarly. For both architectures a clear optimal number of groups exists: for ViT-S it happens to coincide with the default number of heads, whereas for ConvNeXt it's three. Fewer groups result in a drastic performance reduction, whereas the performance with more groups decreases more slowly, likely as a result of reduced representational power per group.

\section{Conclusion}

In this work we presented combined Deep Neighbor Proximity (cDNP), an approach for dense out-of-distribution detection based on k nearest neighbors, which is simple, cost-effective, and achieves state-of-the-art performance on common driving-focused anomaly detection benchmarks.
The method is easily combined with standard parametric scores for a performance boost, but also delivers exceptional standalone performance when paired with attention-based models.
We conducted a thorough comparative study to verify that the approach performs well on various datasets and is robust to parameter changes.

The large boost in performance we discovered with transformer-based representations is in line with the good feature clustering properties of transformers ~\cite{caron2021emerging,melas2022deep,zadaianchuk2022unsupervised}. 
We believe that the self-attention mechanism, based on similarity between feature tokens, is a major reason for these advantageous feature properties in similarity-centric approaches, although this is hard to prove. We found clear evidence that the multiple heads have a positive influence on the similarity metric, and we were even able to transfer this advantage to CNNs via corresponding group structures. 
Our results indicate that this effect is tied to the ``curse of dimensionality'', which -- although not catastrophically -- harms the performance of k-nearest-neighbors.
A deeper understanding of these effects holds the promise of principled progress in the field.

\blfootnote{Funded by the Deutsche Forschungsgemeinschaft (DFG, German Research Foundation) - 417962828}

A limitation of the current approach is its lower resolution, which depends on the encoder architecture and can harm performance in the presence of very small anomalies. Upsampling strategies or alternative architectures can be explored for whenever small-sized objects matter.


\FloatBarrier

{\small
\bibliographystyle{ieee_fullname}
\bibliography{egbib}
}

\newpage
\section{Ablation: Feature Selection for kNNs}
Here we present the results of our approach on different types of features, as summarized in the main paper.

In Table~\ref{tab:feature_selection} we report feature size (number of channels), resolution, and respective DNP performance for different feature options within the four encoders (ResNet, ConvNeXt, MiT, ViT). Most importantly, the results confirm the superiority of self-attention features (queries/keys/values) for MiT and ViT.

\begin{table}[h]
\small
\centering
\setlength\tabcolsep{5pt} 
    \begin{tabular}{llrc|c}
        \toprule
        Model & Features & Ft. size & Ft. res. & AP \\
        & & & & (DNP) \\
        \midrule
        \multirow{4}{*}{ResNet50}   & Stage 1 & 256  & 180$\times$320 & 16.9 \\
                                    & Stage 2 & 512  & \hphantom{0}90$\times$160 & 20.8 \\
                                    & Stage 3 & 1024 & 45$\times$80& 25.6 \\
                                    & Stage 4 & 2048 & 22$\times$40& 21.2 \\
        \midrule
        \multirow{4}{*}{ConvNeXt-T} & Stage 1 & 96   & 180$\times$320 & 17.0 \\
                                    & Stage 2 & 192  & \hphantom{0}90$\times$160 & 17.2 \\ 
                                    & Stage 3 & 384  & 45$\times$80& 32.5 \\ 
                                    & Stage 4 & 768  & 22$\times$40& 27.9 \\
        \midrule                            
        \multirow{4}{*}{MiT-B2} & Stage 4 - Q & 512 & 23$\times$40 & 79.4 \\
                                & Stage 4 - K & 512 & 23$\times$40 & 76.0 \\
                                & Stage 4 - V & 512 & 23$\times$40 & 77.9 \\
                                & Stage 4     & 512 & 23$\times$40 & 48.6 \\
        \midrule 
        \multirow{4}{*}{ViT-B}  & Layer 12 - Q & 768 & 45$\times$80 & 78.1 \\
                                & Layer 12 - K & 768 & 45$\times$80 & 85.4 \\
                                & Layer 12 - V & 768 & 45$\times$80 & 77.7 \\
                                & Layer 12     & 768 & 45$\times$80 & 71.6 \\
        \bottomrule
    \end{tabular}
    \caption{Overview of the feature selection results, including feature sizes and resolutions, as well as the performance of DNP on each, in terms of AP on RoadAnomaly.}
    \label{tab:feature_selection}
\end{table}

\begin{table}
\centering
\small
\begin{tabular}{l|ccr}
\toprule
      & \multicolumn{2}{c}{RoadAnomaly}  \\
      & AP   & FPR$_{95}$         \\
        \midrule   
        DNP ViT-B iBOT & 55.28 & 19.72         \\
        DNP ViT-B DINO & 67.83 & 18.99         \\
\bottomrule
\end{tabular}
\caption{DNP performance on RoadAnomaly using representations from self-supervision approaches iBOT and DINO. While the supervised features (trained for semantic segmentation on Cityscapes) yield the best results, DNP with DINO features outperforms the other architectures.}
\label{tab:selfsup}
\end{table}

\section{DNP with Self-Supervised Representations}
Although in this work we mostly apply DNP to feature representation which have been trained for semantic segmentation in a supervised way, the approach could in principle be applied to other types of representations. This is because it relies on a set of in-distribution reference features which carry information about the training data.

To explore the capabilities of our method in this direction, we combined it with feature extractors trained via self-supervision, using the popular iBOT~\cite{zhou2021ibot} and DINO~\cite{caron2021emerging} approaches, known to perform well on dense and local downstream tasks such as segmentation.

Table~\ref{tab:selfsup} shows the resulting DNP performances using parameters obtained from the respective iBOT and DINO official repositories, compared to the supervised features which we trained for semantic segmentation.
Although the supervised representations obtain the best results, DINO features perform well with an AP of 67.83, outperforming the CNN architectures.

\section{Additional Results}
\subsection{SegmentMeIfYouCan - Obstacle}
Here we report our results on the Onstacle track of the SegmentMeIfYouCan (SMIYC) benchmark, which considers obstacles on the road surface.

Table~\ref{tab:smiyc-obstacle} shows the official leaderboard results on the benchmark's test set (undisclosed ground truth). Excluding outlier exposure, there are two methods that currently outperform cDNP. The first is NFlowJS, which uses synthetic outliers from a generative model, and the second is DaCUP~\cite{Vojir_2023_WACV}, which is based on inpainting reconstruction error. It should be noted that NFlowJS performs worse than cDNP on SMIYC-Anomaly (oriented towards semantic anomalies, reported in the main paper), and DaCUP is specifically designed for road obstacle detection.

\begin{table}
    \centering
    \begin{tabular}{l|c|c|c}
    \toprule
        Method & OE & AP & FPR$_{95}$\\
        \midrule
        DaCUP~\cite{Vojir_2023_WACV} & & 81.50 & 1.13 \\
        NFlowJS & & 85.55 & 0.41 \\
        Maximized Entropy & \checkmark & 85.07 & 0.75 \\
        DenseHybrid & \checkmark & 87.08 & 0.24 \\
        \midrule
        cDNP-Segmenter-ViT-B & & 72.70 & 1.40 \\ 
    \bottomrule
    \end{tabular}
    \caption{Results on the SMIYC-Obstacle test benchmark. OE marks the use of outlier exposure.}
    \label{tab:smiyc-obstacle}
\end{table}

\subsection{Feature Partitioning - Fishyscapes Lost\&Found}
\begin{figure}
    \centering
    \includegraphics[width=0.47\textwidth]{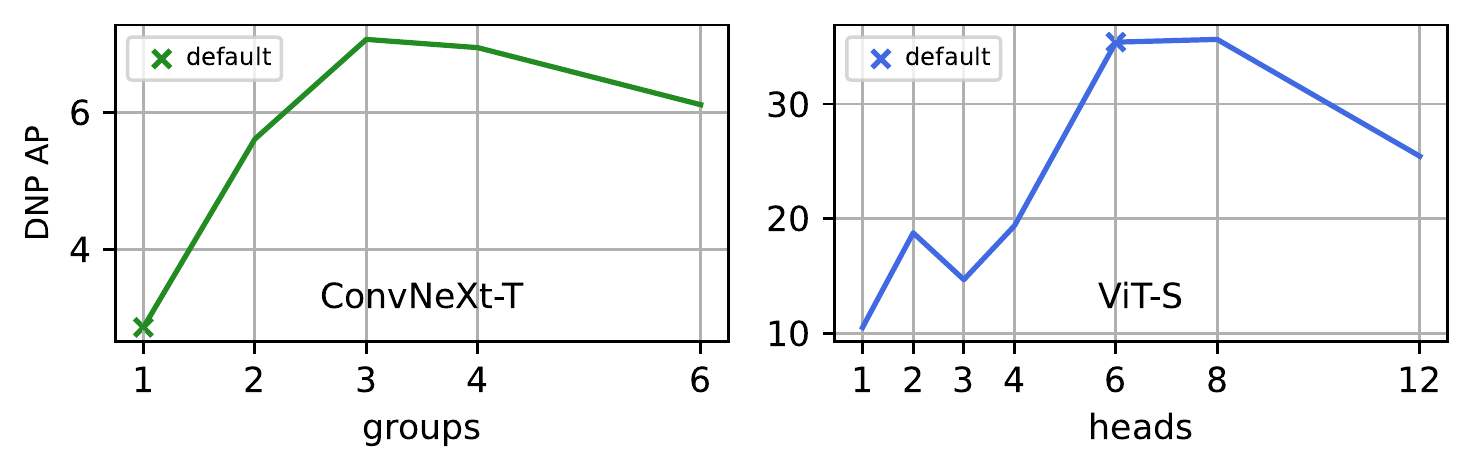}
    \caption{DNP performance of ConvNeXt and ViT features wit different numbers fo groups and heads respectively.}
    \label{fig:ablation_fslaf}
\end{figure}

In this section we extend the results of Section 5.2 of the main paper. In Figure~\ref{fig:ablation_fslaf} we report the performance on Fishyscapes Lost\&Found-val of the modified ConvNeXt-T and ViT-S backbones, with different numbers of convolutional groups and transformer heads respectively. The models evaluated here are the same as those evaluated in the main paper.

The average precision (AP) of DNP in both cases follows the same behavior as on RoadAnomaly, i.e. the same optimal number of groups/heads and performance that decreases rapidly when fewer greoups/heads are used.

\subsection{Feature Partitioning - Segmentation Performance}
In Table~\ref{tab:ablation_miou} we report the in-distribution segmentation performance of the models involved in the feature partitioning ablation (Section 5.2 of the main paper). The results show that the two architectures (ConvNeXt and ViT) behave differently in terms of mIoU, and confirm that there is no direct relation between segmentation and OoD detection performance.

It should also be noted that the segmentation and OoD detection performances are overall negative affected by the ablation protocol, which involves discarding the pre-trained weight initialization for the last stages.

\begin{table}
    \centering
    \small
    \begin{tabular}{c|c|c|c|c}
    \toprule
        Groups/ & \multicolumn{2}{c|}{ConvNeXt} & \multicolumn{2}{c}{ViT}\\
        Heads & mIoU & AP & mIoU & AP \\
        \midrule
        1 & 76 & 31 & 63 & 45\\
        2 & 76 & 39 & 66 & 69\\
        3 & 75 & 44 & 67 & 71\\
        4 & 75 & 37 & 69 & 69\\
        6 & 75 & 34 & 71 & 80\\
    \bottomrule
    \end{tabular}
    \caption{Segmentation (mIoU) and OoD detection (AP) performance for the models of the feature partitioning ablation, on Cityscapes/RoadAnomaly. For ConvNeXt the segmentation performance doesn't change substantially, and is inversely proportional to the number of groups. For ViT the segmentation performance increases with the number of heads.}
    \label{tab:ablation_miou}
\end{table}


\subsection{Qualitative Results}
\begin{figure}[h!]
    \centering
    
    \begin{subfigure}[b]{0.110\textwidth}
        \centering
        \includegraphics[width=\textwidth]{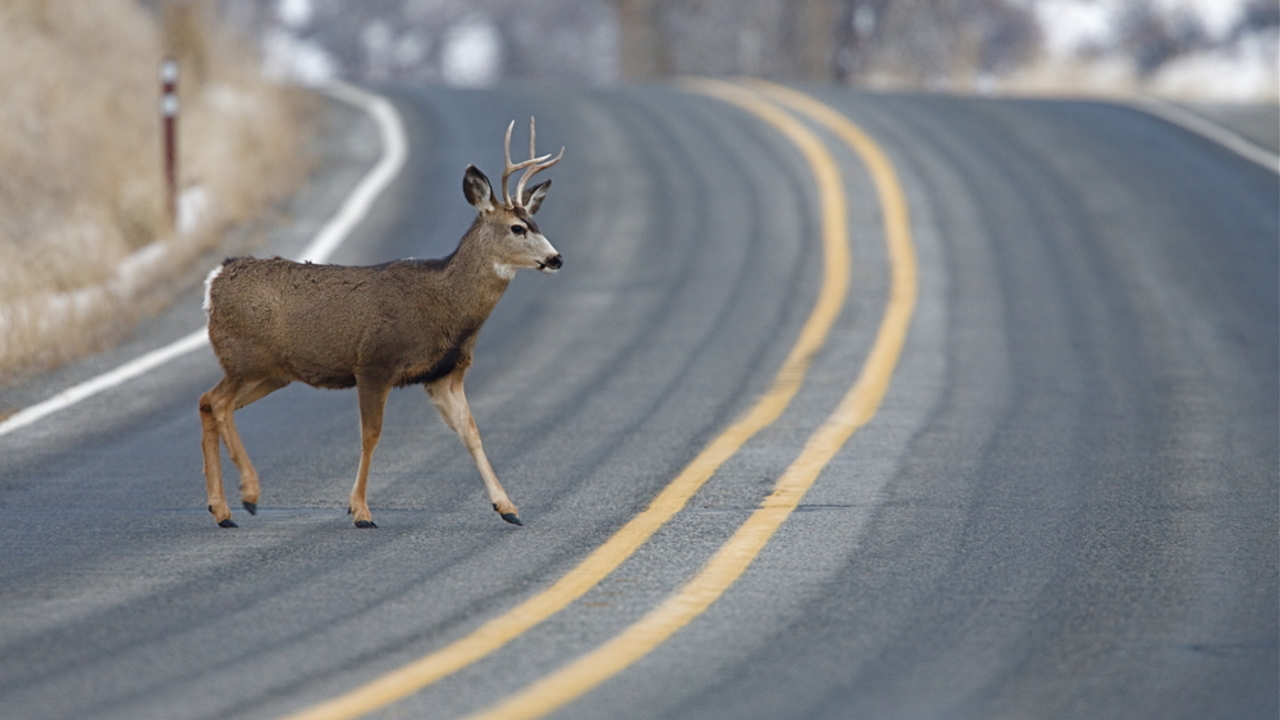}
    \end{subfigure}
    \begin{subfigure}[b]{0.110\textwidth}
        \centering
        \includegraphics[width=\textwidth]{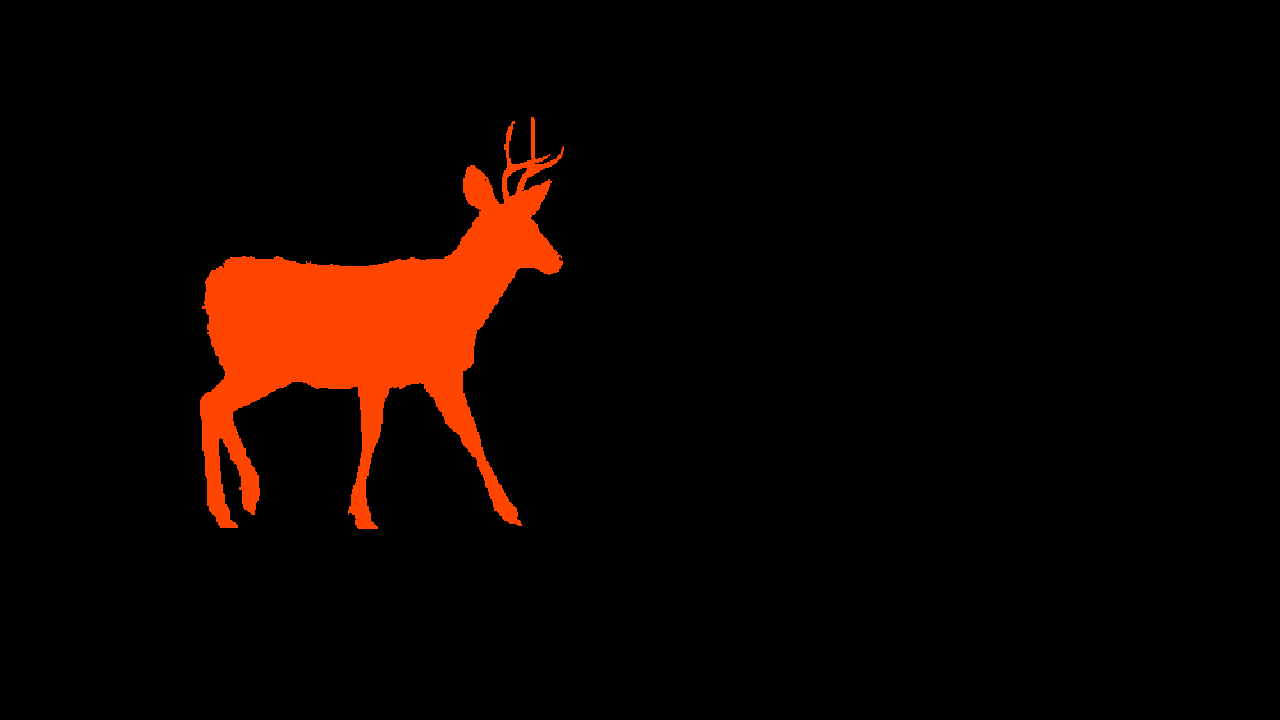}
    \end{subfigure}
    \begin{subfigure}[b]{0.110\textwidth}
        \centering
        \includegraphics[width=\textwidth]{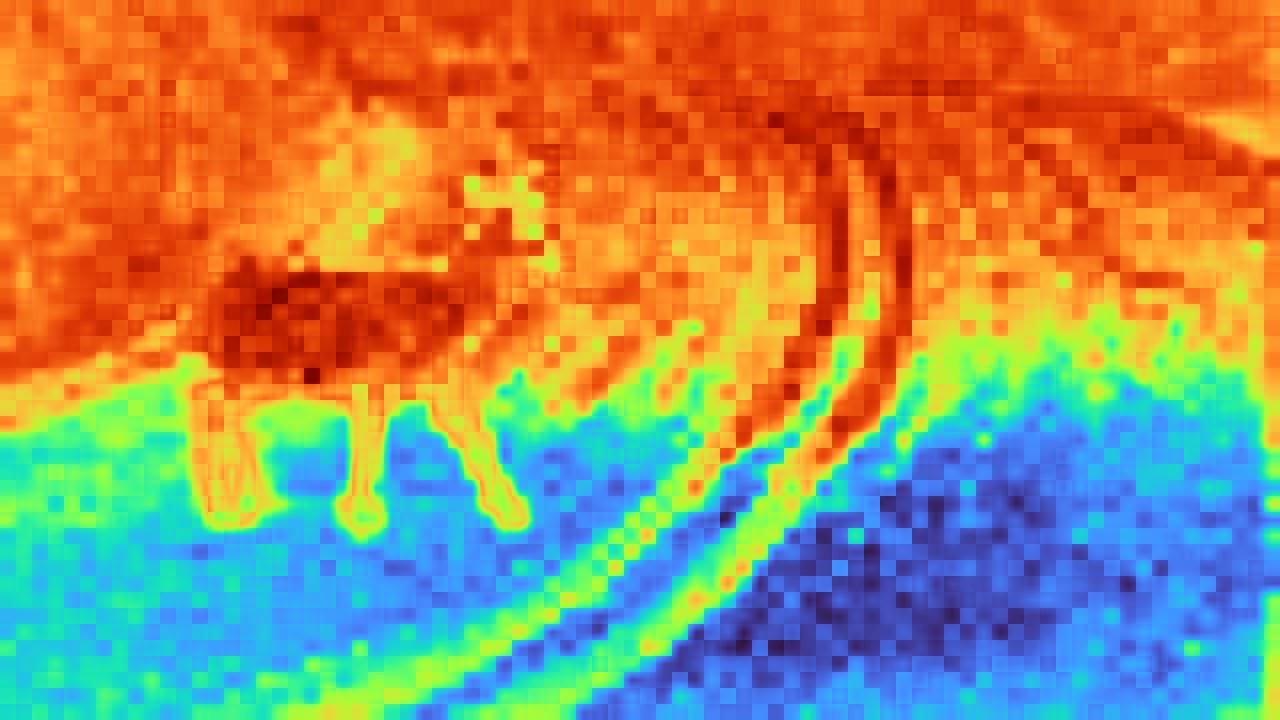}
    \end{subfigure}
    \begin{subfigure}[b]{0.110\textwidth}
        \centering
        \includegraphics[width=\textwidth]{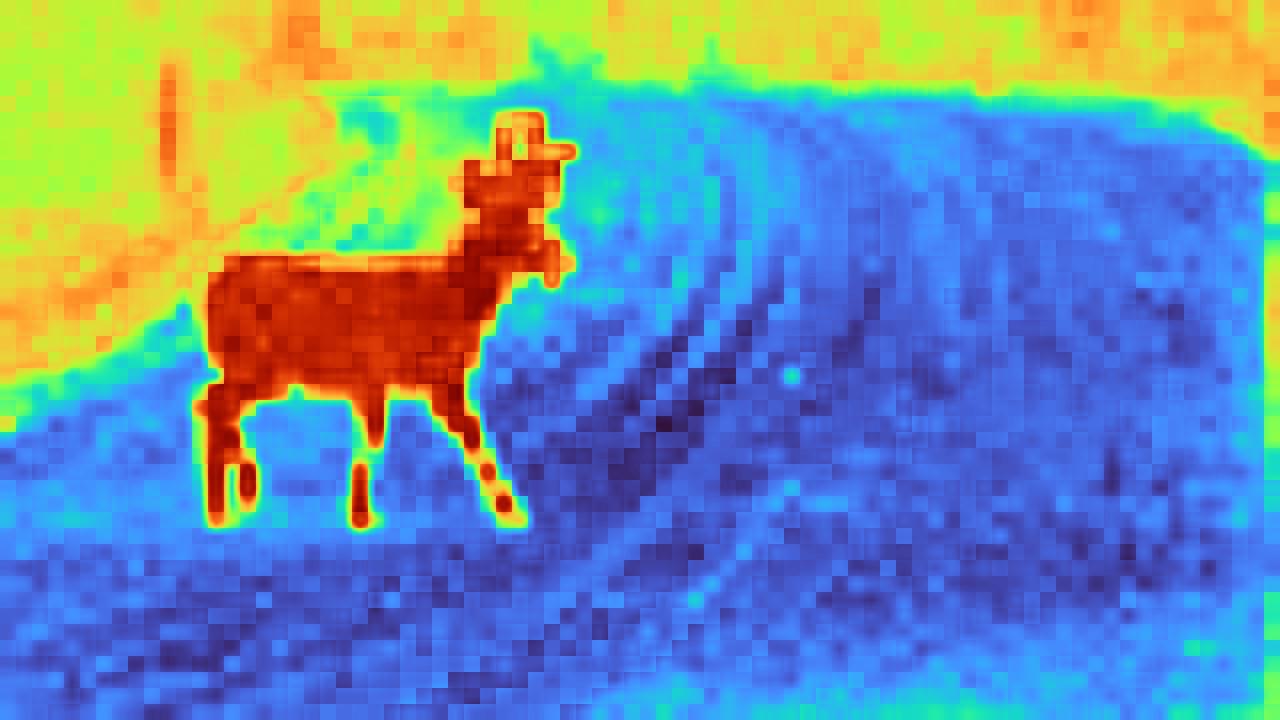}
    \end{subfigure}
    
    
    \begin{subfigure}[b]{0.110\textwidth}
        \centering
        \includegraphics[width=\textwidth]{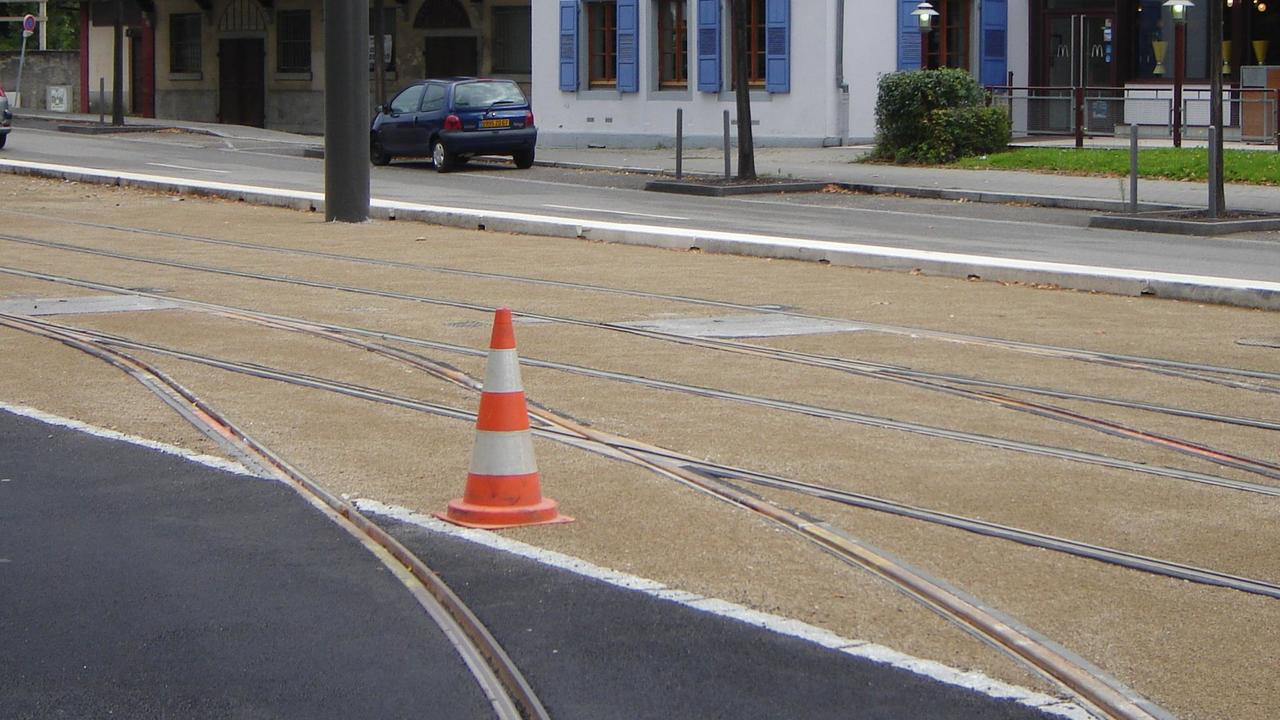}
    \end{subfigure}
    \begin{subfigure}[b]{0.110\textwidth}
        \centering
        \includegraphics[width=\textwidth]{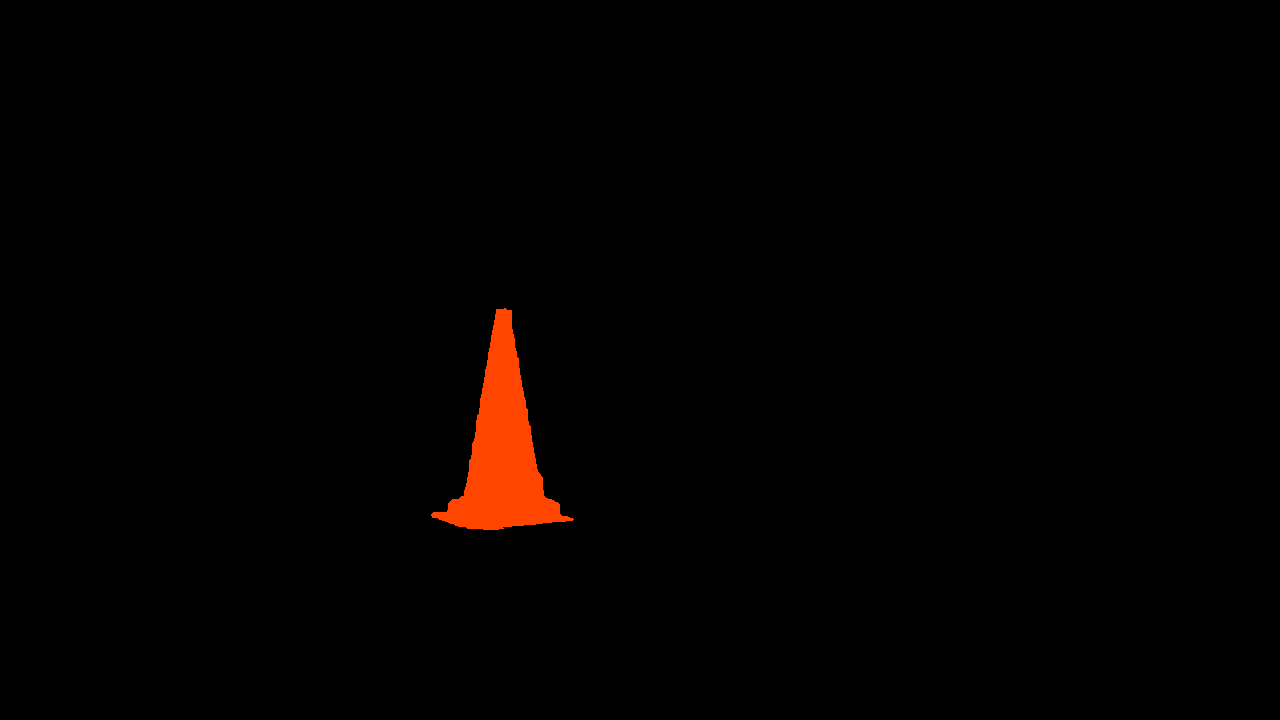}
    \end{subfigure}
    \begin{subfigure}[b]{0.110\textwidth}
        \centering
        \includegraphics[width=\textwidth]{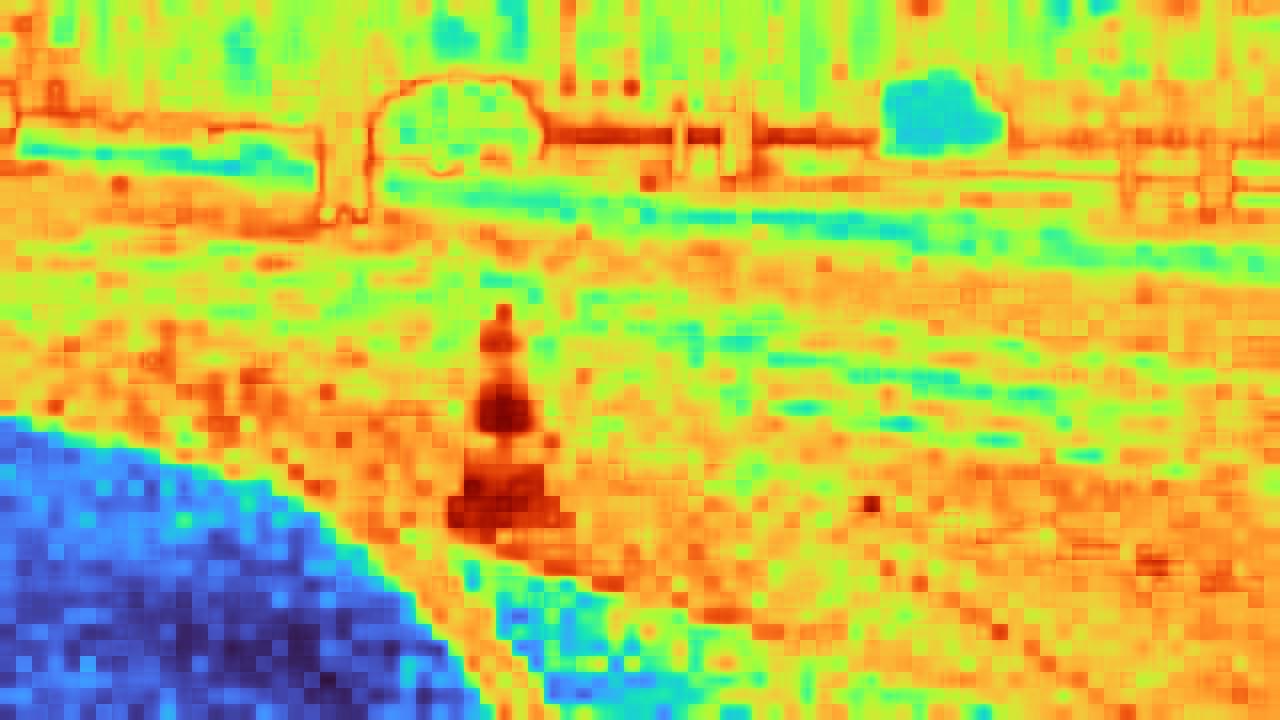}
    \end{subfigure}
    \begin{subfigure}[b]{0.110\textwidth}
        \centering
        \includegraphics[width=\textwidth]{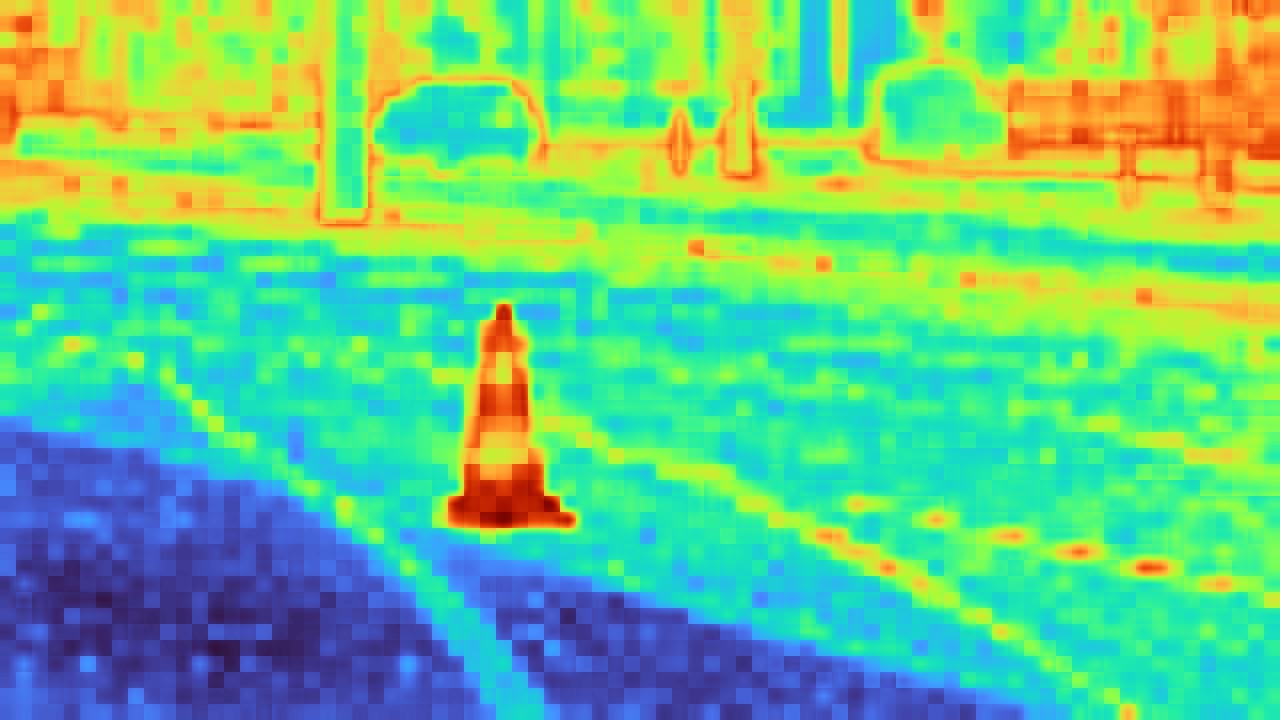}
    \end{subfigure}
    
    \begin{subfigure}[b]{0.110\textwidth}
        \centering
        \includegraphics[width=\textwidth]{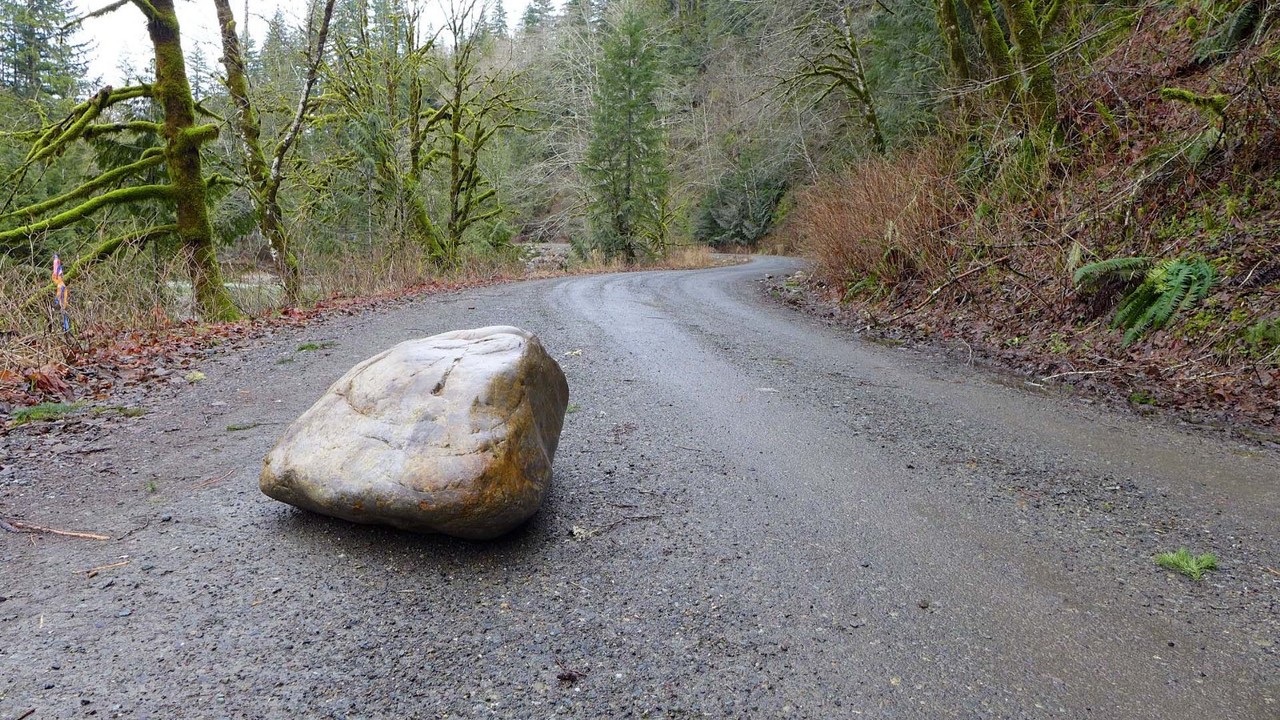}
        \caption*{\tiny Image}
    \end{subfigure}
    \begin{subfigure}[b]{0.110\textwidth}
        \centering
        \includegraphics[width=\textwidth]{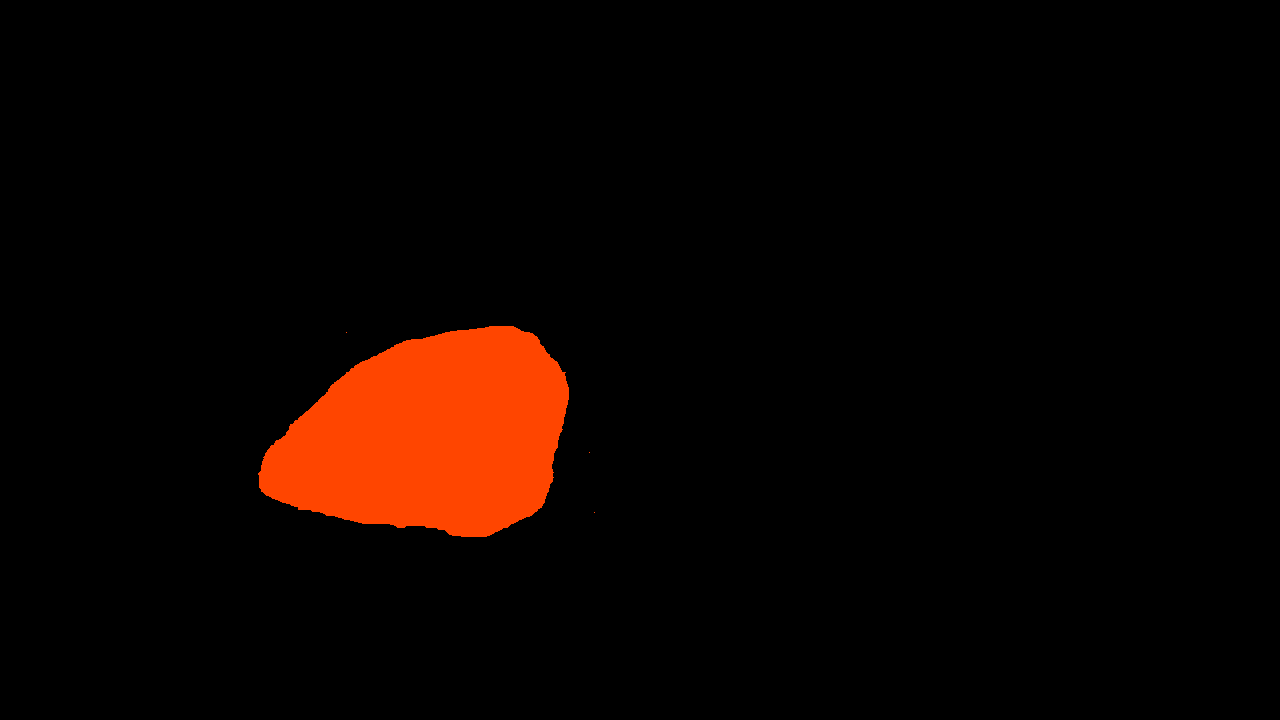}
        \caption*{\tiny Ground truth}
    \end{subfigure}
    \begin{subfigure}[b]{0.110\textwidth}
        \centering
        \includegraphics[width=\textwidth]{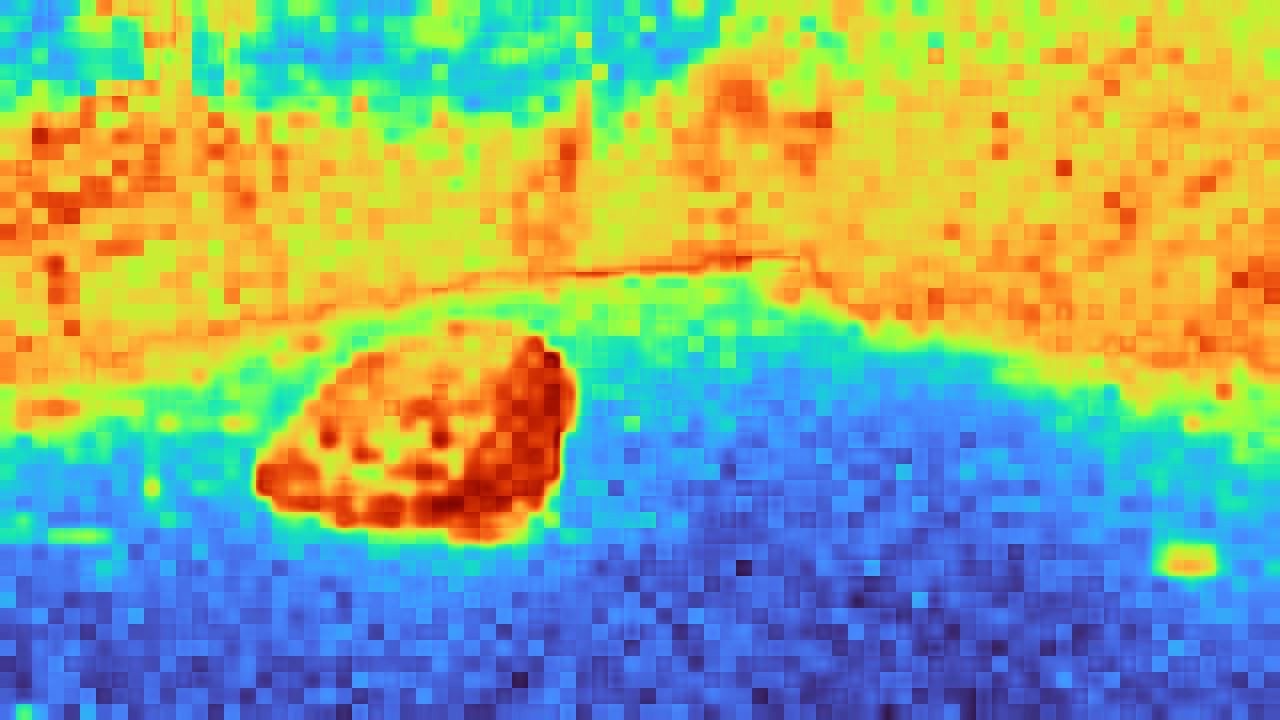}
        \caption*{\tiny cDNP ViT 1 head}
    \end{subfigure}
    \begin{subfigure}[b]{0.110\textwidth}
        \centering
        \includegraphics[width=\textwidth]{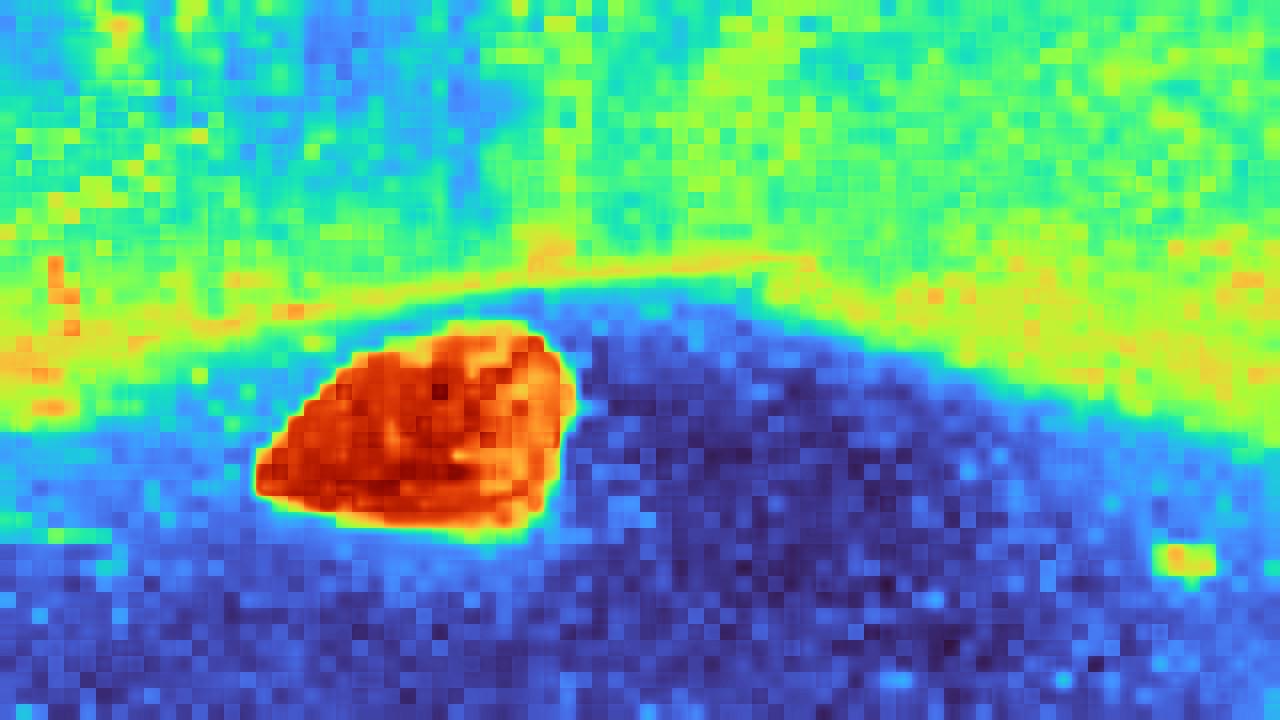}
        \caption*{\tiny cDNP ViT 6 heads}
    \end{subfigure}
    
    \caption{Qualitative examples for the feature partitioning ablation, showing results on RoadAnomaly examples, for ViT-S with 1 and 6 heads. Anomalous pixels are shown in red in the ground truth. The single head model struggles both with false negatives and false positives, and assigns a higher anomaly score to edges and backgrounds.}
    
    \label{fig:ablation_quali}
\end{figure}

\paragraph{Feature Partitioning}
Figure~\ref{fig:ablation_quali} shows a qualitative comparison between ViT with 1 head and 6 heads, respectively the worst and best versions of the network in the ablation study.

\begin{figure}[h!]
    \centering
    \begin{subfigure}[b]{0.110\textwidth}
        \centering
        \includegraphics[width=\textwidth]{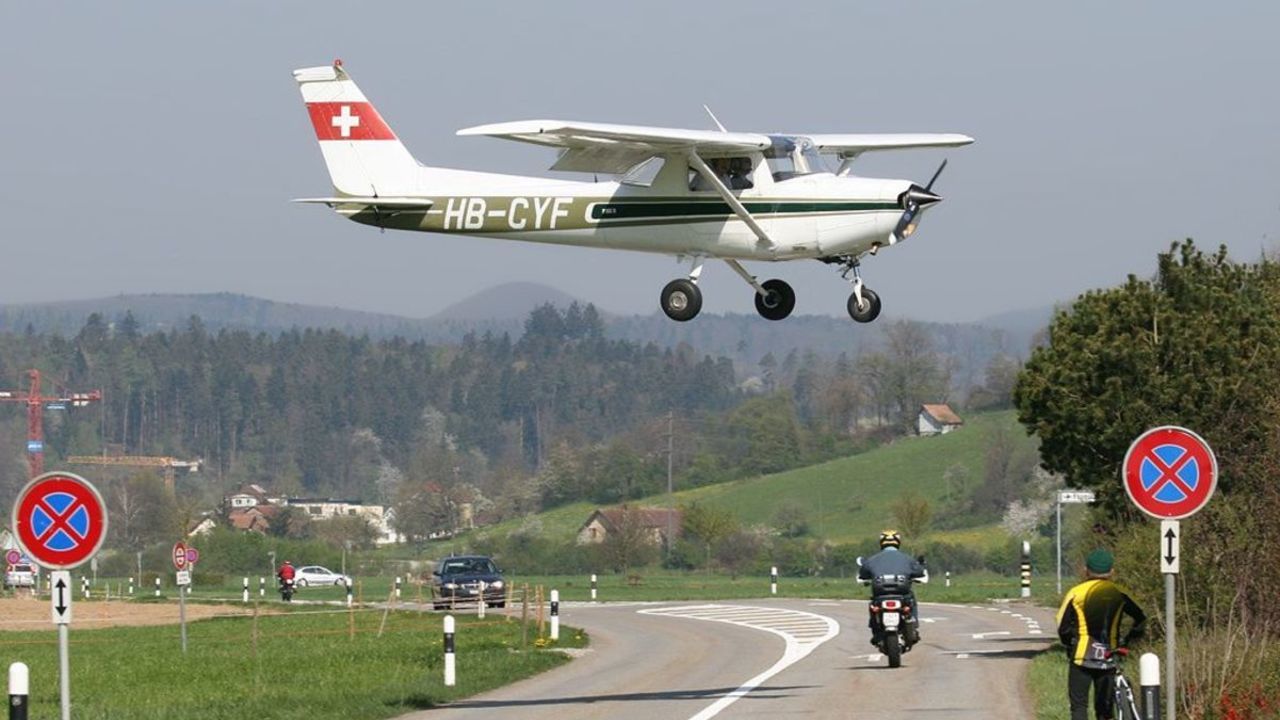}
    \end{subfigure}
    \begin{subfigure}[b]{0.110\textwidth}
        \centering
        \includegraphics[width=\textwidth]{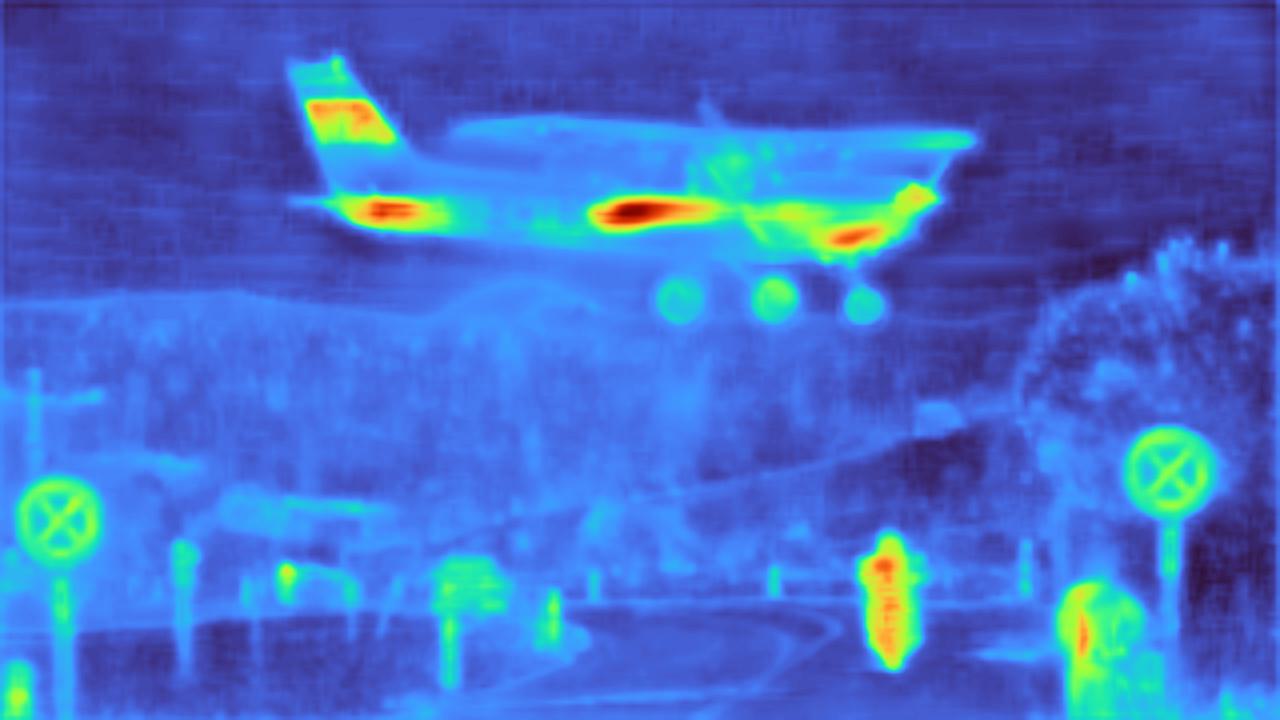}
    \end{subfigure}
    \begin{subfigure}[b]{0.110\textwidth}
        \centering
        \includegraphics[width=\textwidth]{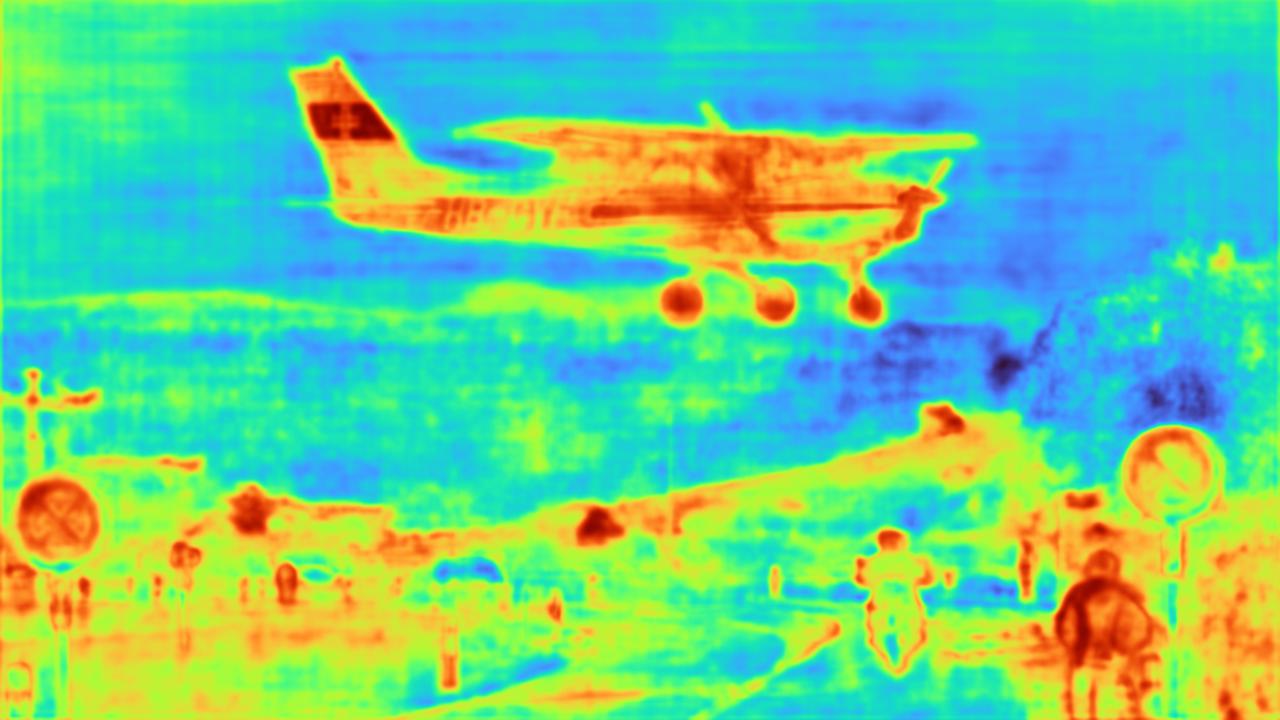}
    \end{subfigure}
    \begin{subfigure}[b]{0.110\textwidth}
        \centering
        \includegraphics[width=\textwidth]{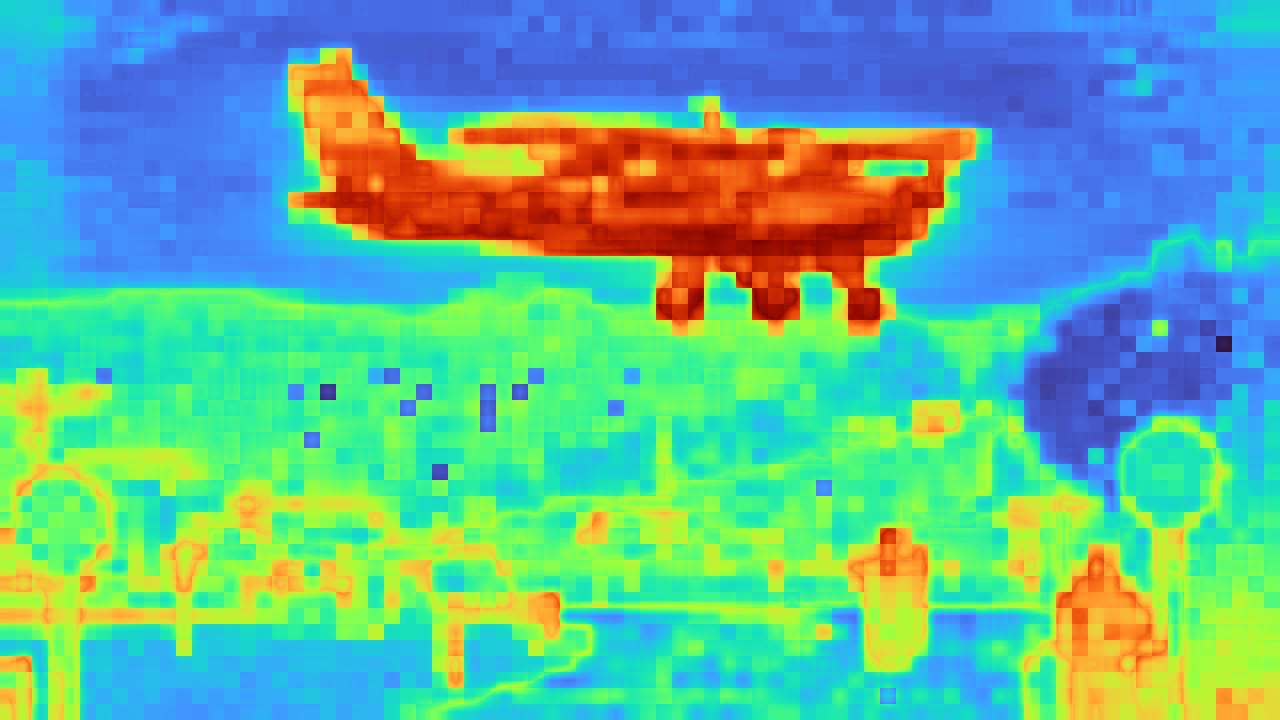}
    \end{subfigure}
    
    \begin{subfigure}[b]{0.110\textwidth}
        \centering
        \includegraphics[width=\textwidth]{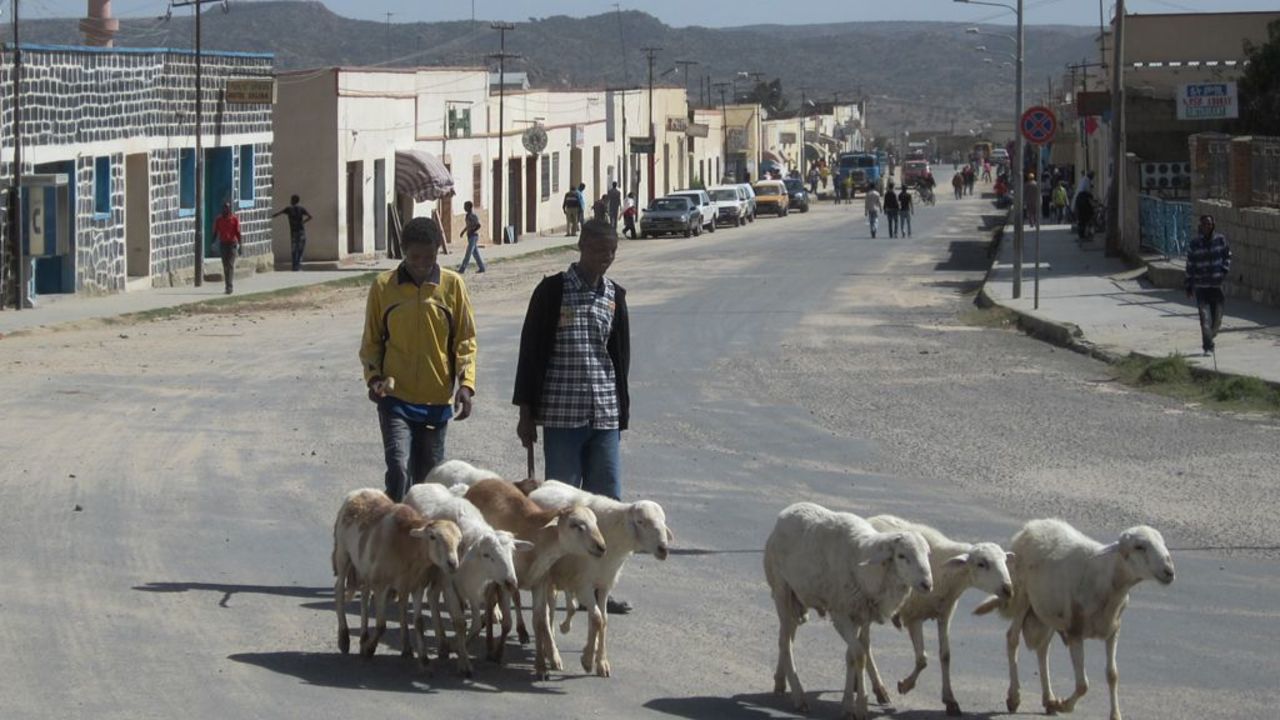}
    \end{subfigure}
    \begin{subfigure}[b]{0.110\textwidth}
        \centering
        \includegraphics[width=\textwidth]{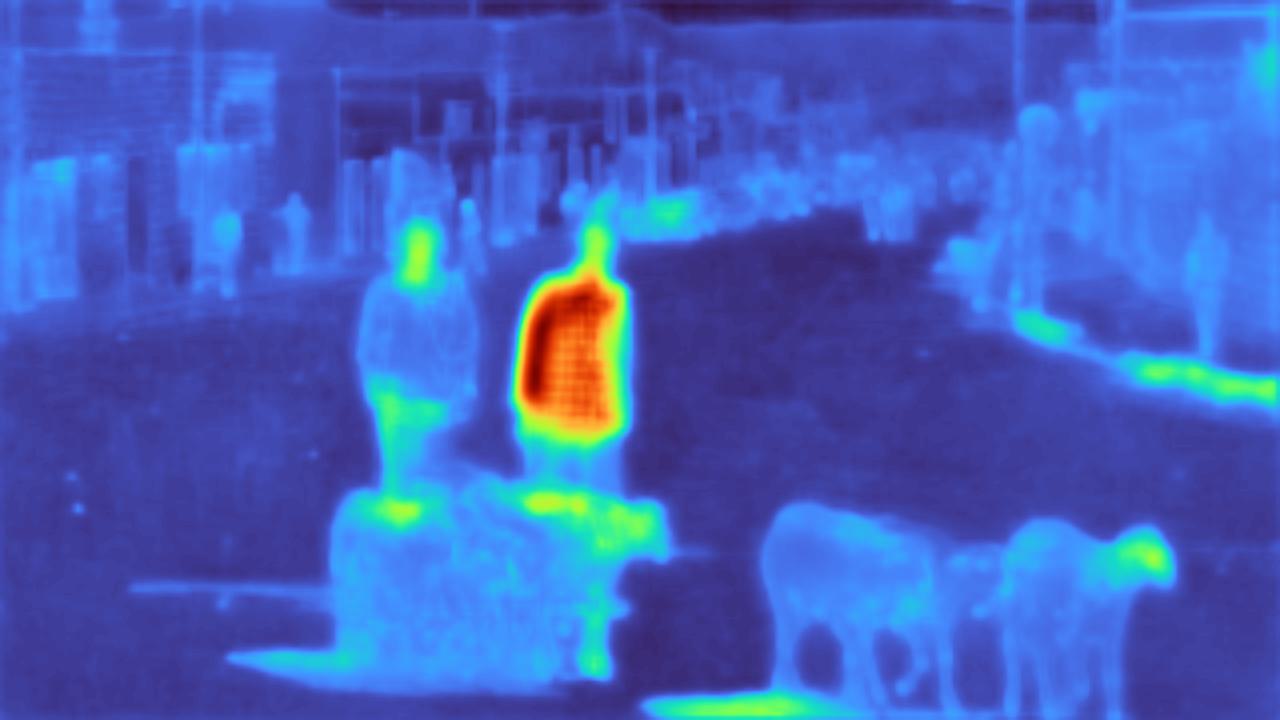}
    \end{subfigure}
    \begin{subfigure}[b]{0.110\textwidth}
        \centering
        \includegraphics[width=\textwidth]{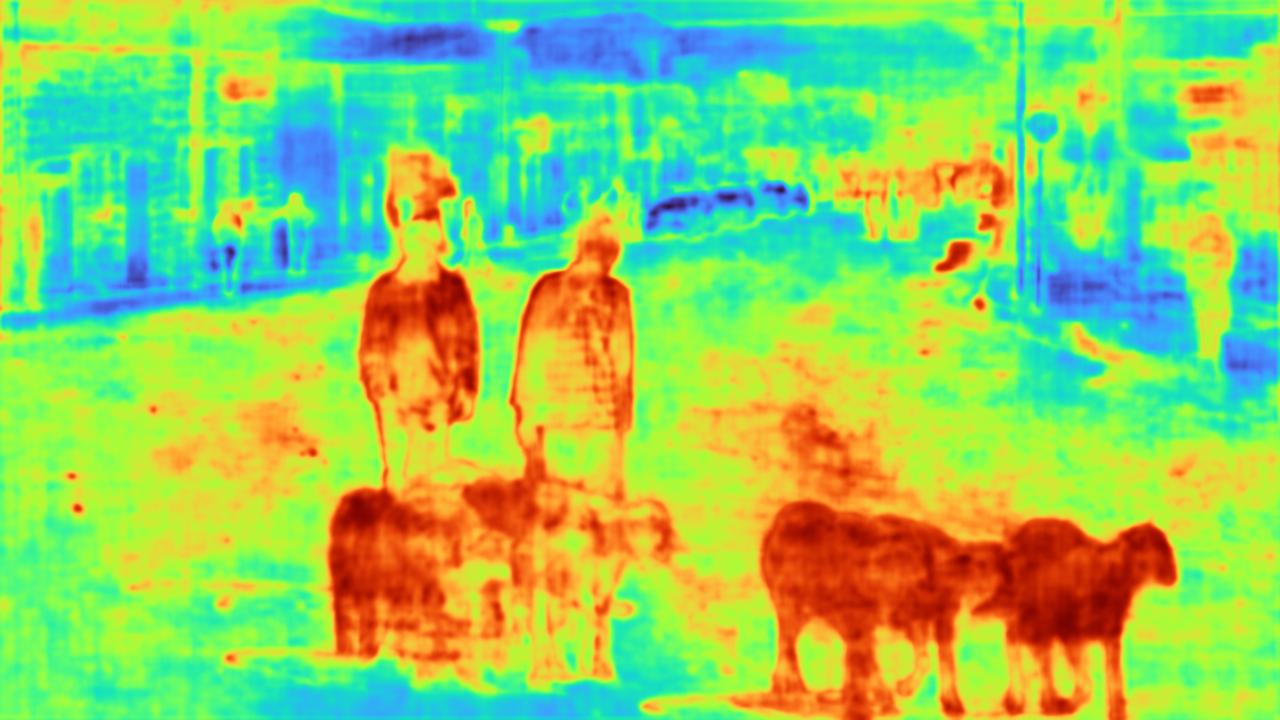}
    \end{subfigure}
    \begin{subfigure}[b]{0.110\textwidth}
        \centering
        \includegraphics[width=\textwidth]{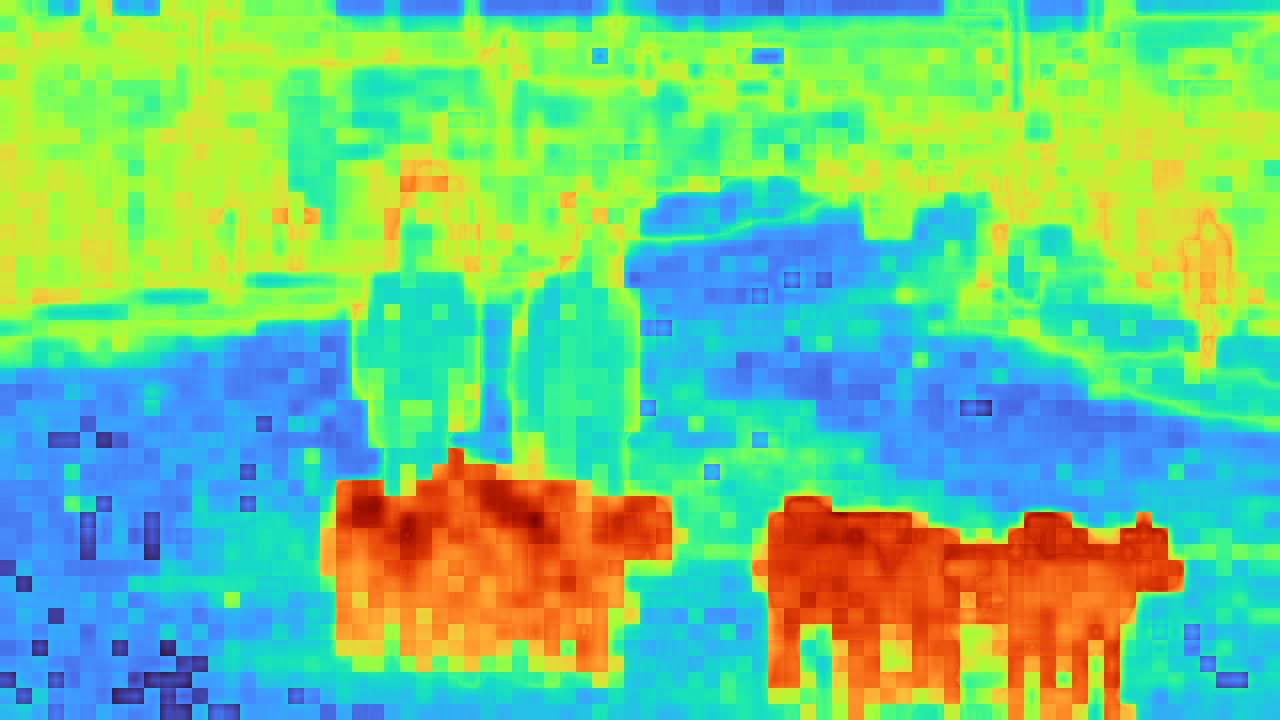}
    \end{subfigure}
    
    \begin{subfigure}[b]{0.110\textwidth}
        \centering
        \includegraphics[width=\textwidth]{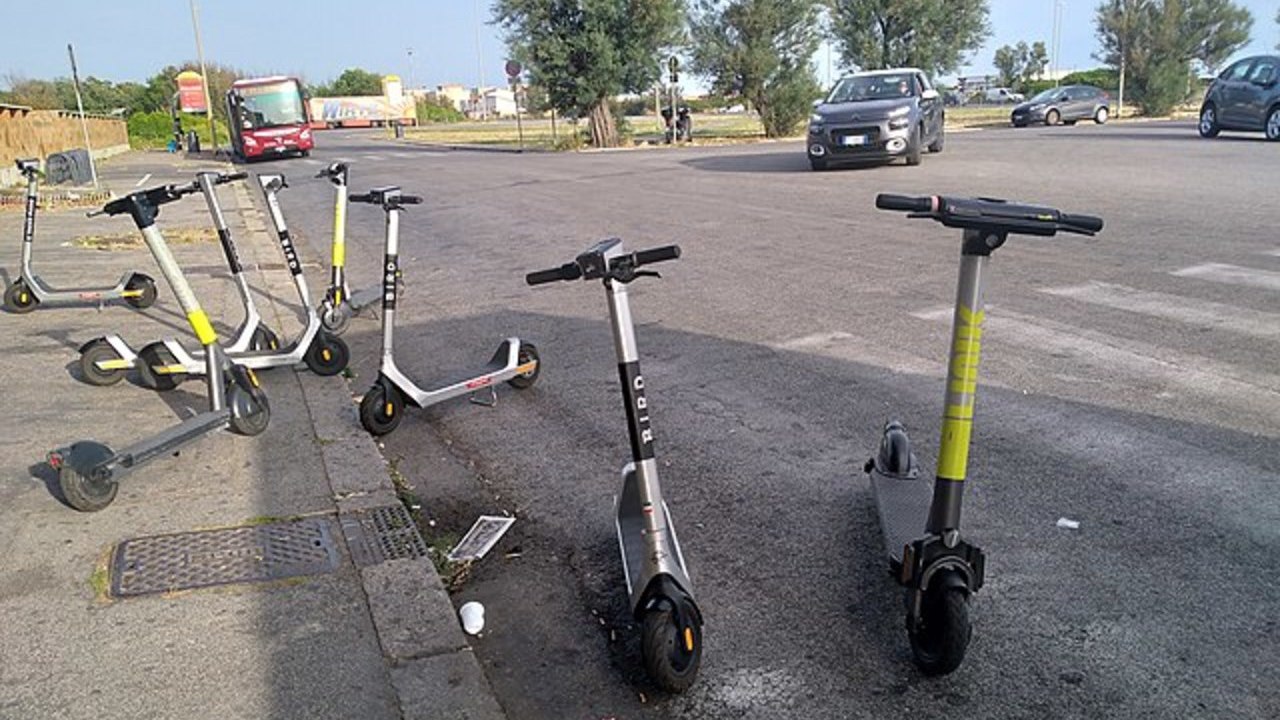}
    \end{subfigure}
    \begin{subfigure}[b]{0.110\textwidth}
        \centering
        \includegraphics[width=\textwidth]{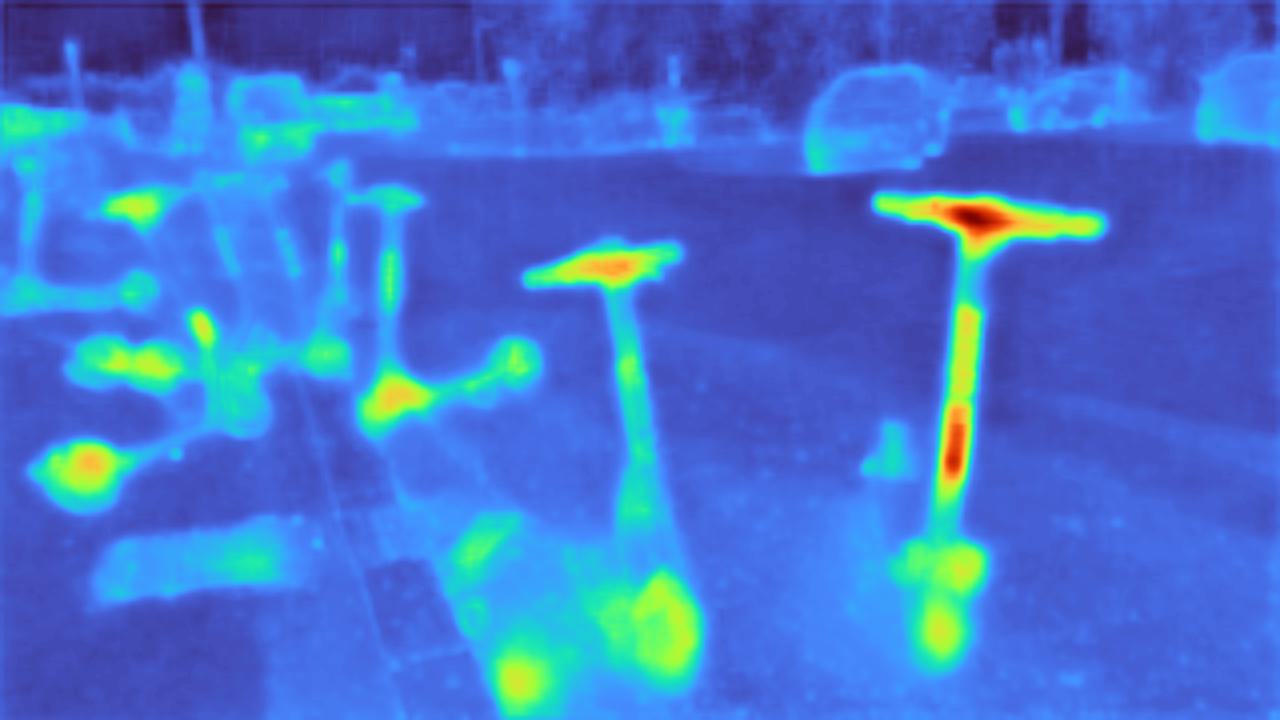}
    \end{subfigure}
    \begin{subfigure}[b]{0.110\textwidth}
        \centering
        \includegraphics[width=\textwidth]{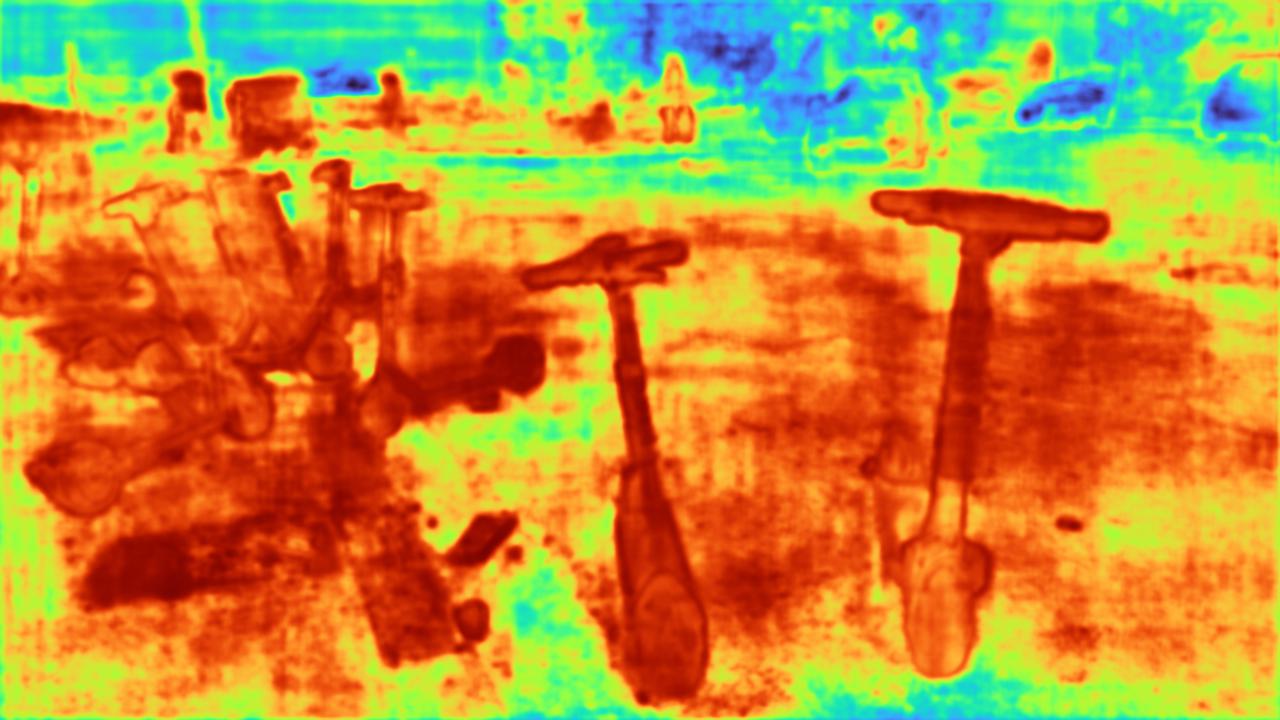}
    \end{subfigure}
    \begin{subfigure}[b]{0.110\textwidth}
        \centering
        \includegraphics[width=\textwidth]{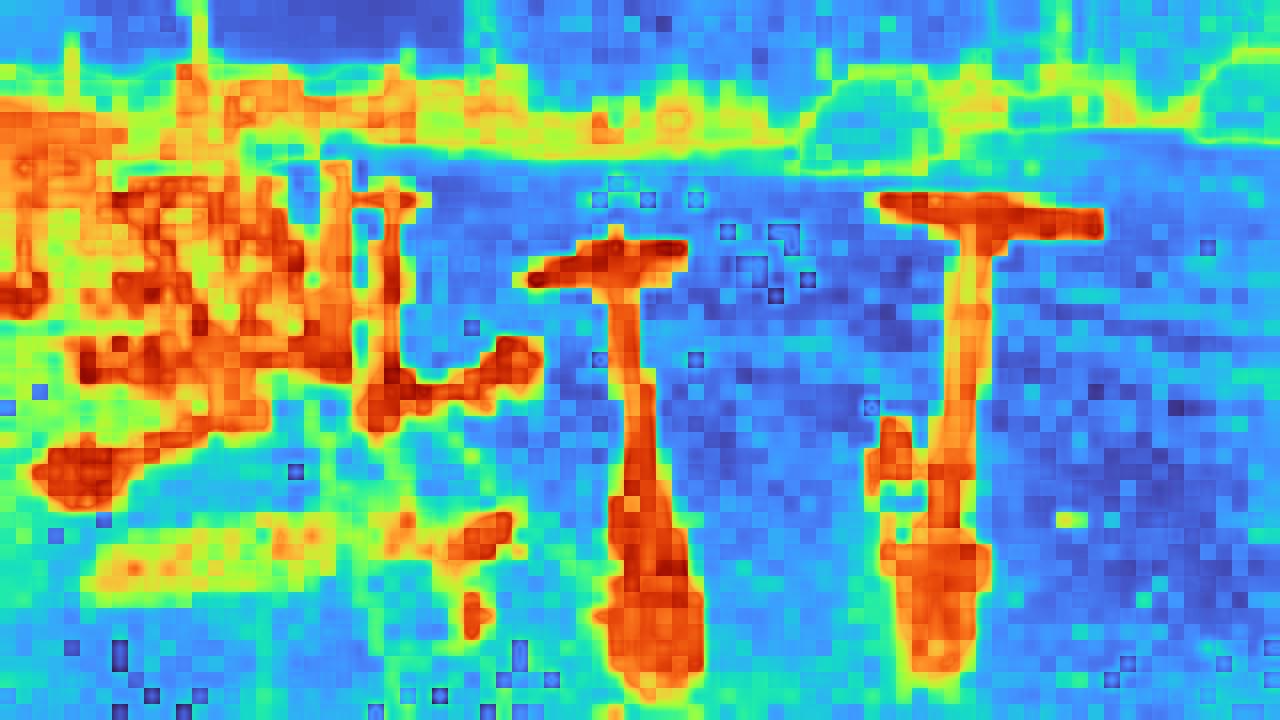}
    \end{subfigure}

    \begin{subfigure}[b]{0.110\textwidth}
        \centering
        \includegraphics[width=\textwidth]{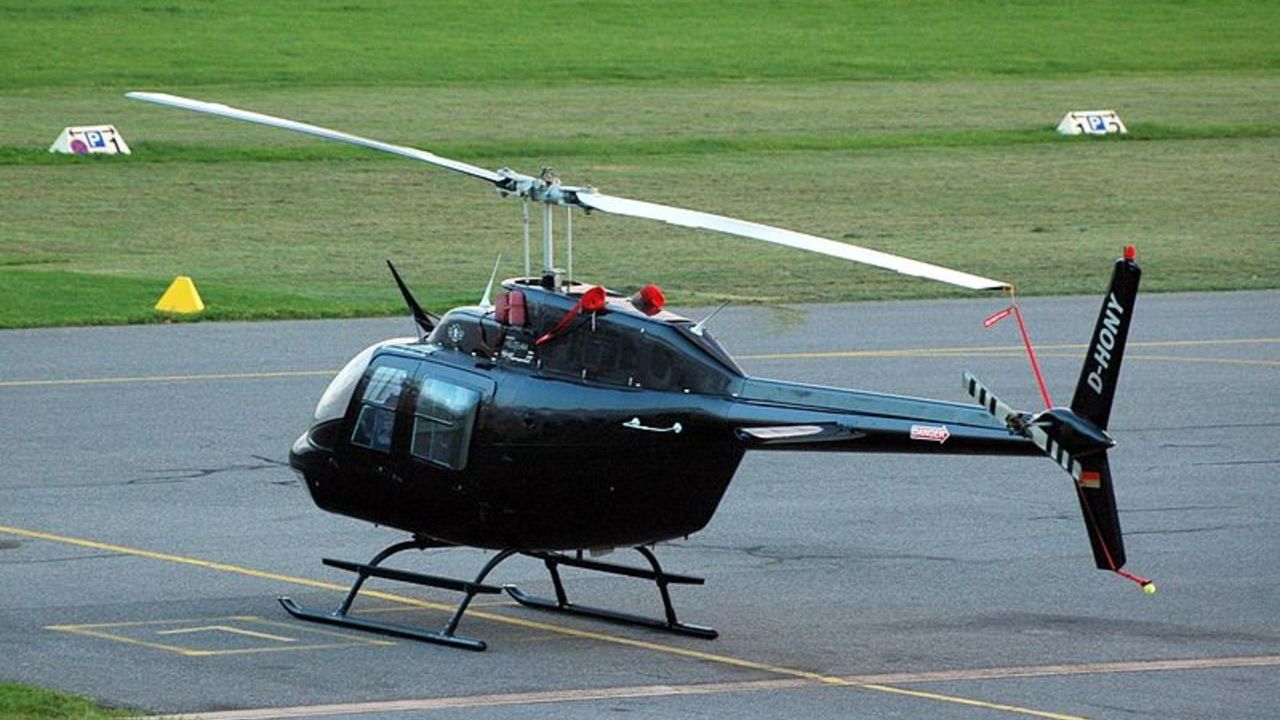}
        \caption*{Image}
    \end{subfigure}
    \begin{subfigure}[b]{0.110\textwidth}
        \centering
        \includegraphics[width=\textwidth]{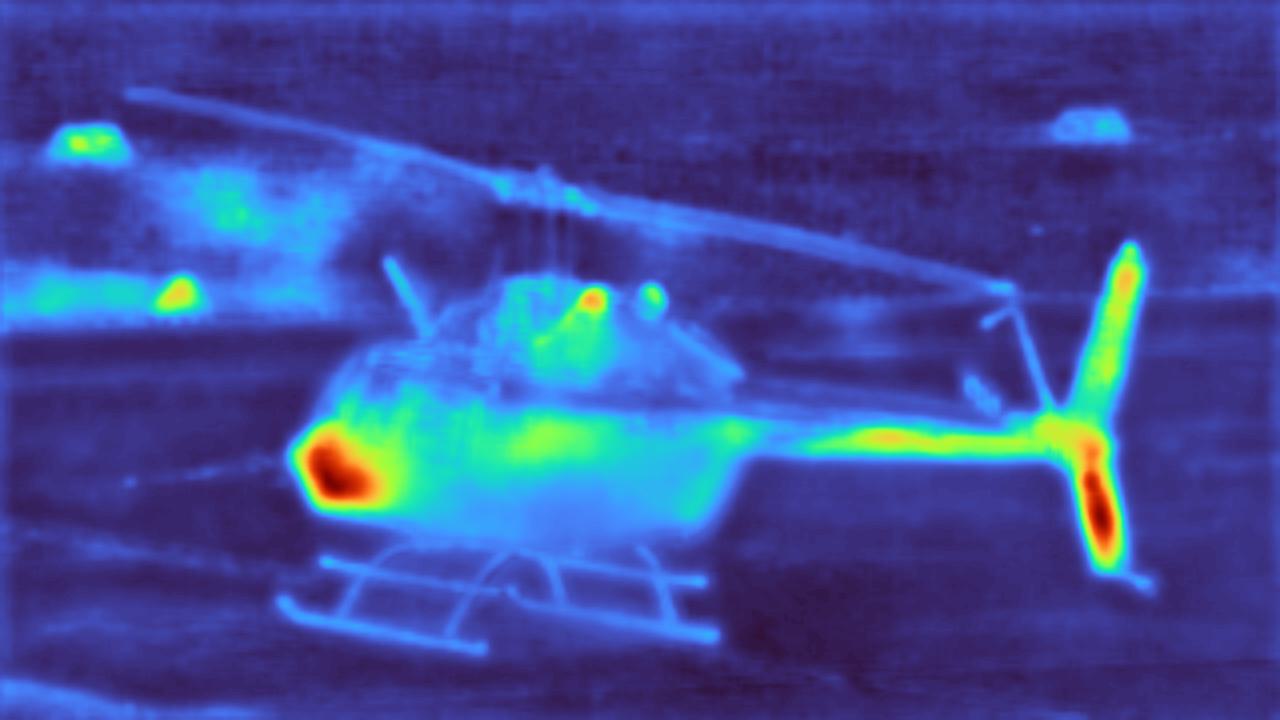}
        \caption*{DenseHybrid}
    \end{subfigure}
    \begin{subfigure}[b]{0.110\textwidth}
        \centering
        \includegraphics[width=\textwidth]{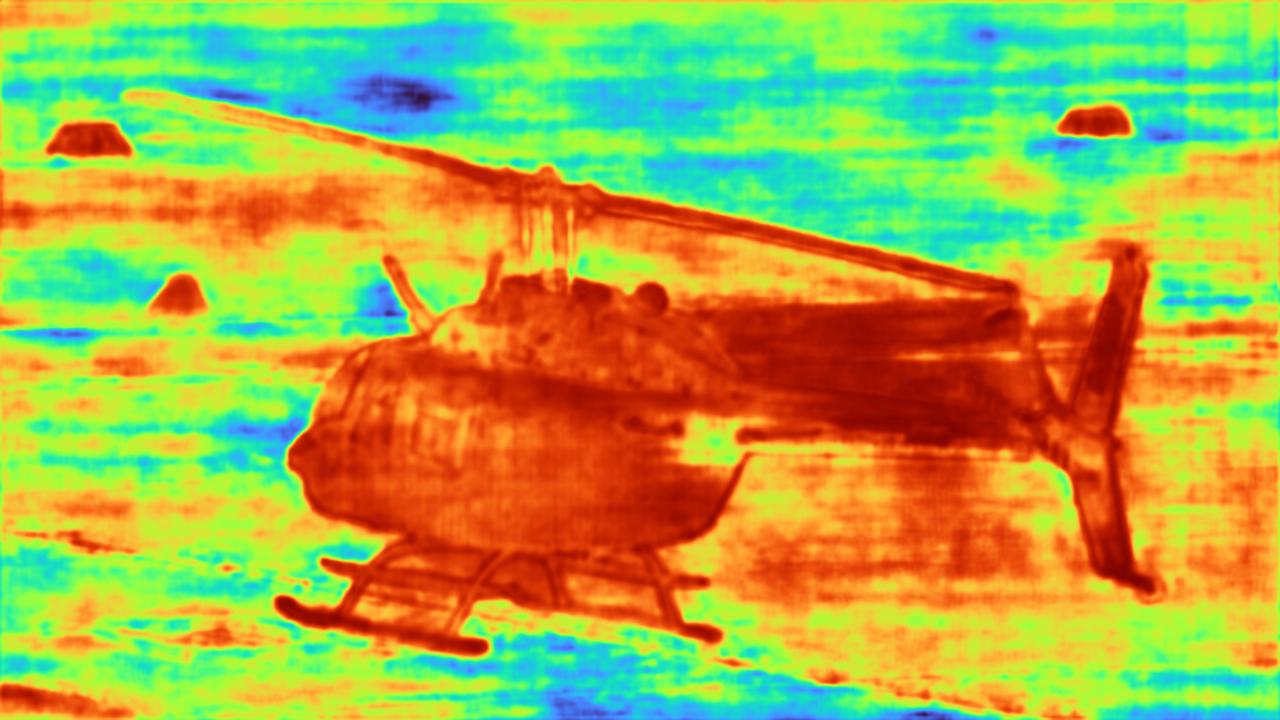}
        \caption*{PEBAL}
    \end{subfigure}
    \begin{subfigure}[b]{0.110\textwidth}
        \centering
        \includegraphics[width=\textwidth]{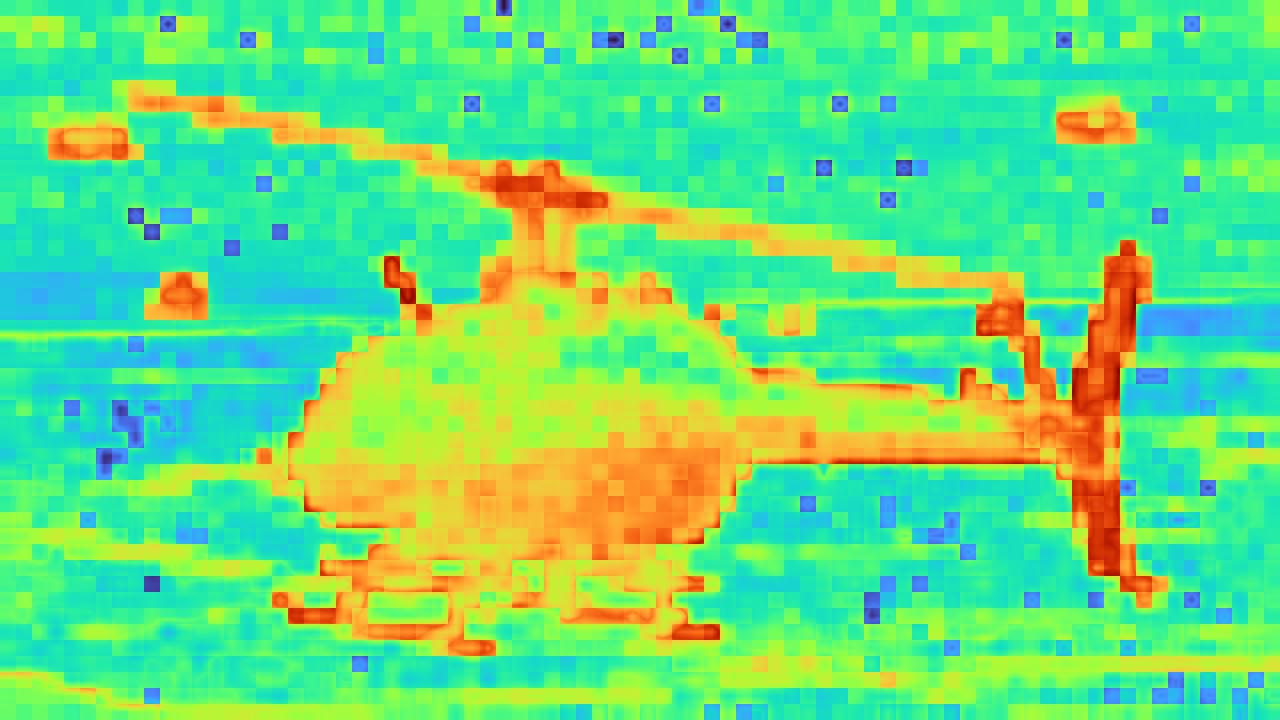}
        \caption*{cDNP}
    \end{subfigure}
    
    \caption{Qualitative comparison between cDNP and most recent state-of-the art methods DenseHybrid and PEBAL. On the first three examples, cDNP is the only method that correctly and entirely identifies the anomalous samples (airplane, sheep, and scooters): DenseHybrid misses many parts of the objects, and PEBAL suffers from false positives. The forth example is challenging for all approaches: PEBAL is the only approach to detect the helicopter entirely, still producing false positives.}
    
    \label{fig:sota_quali}
\end{figure}

\paragraph{State-Of-The-Art}
In Figure~\ref{fig:sota_quali} we show additional qualitative results for our approach, compared with recent approaches DenseHybrid and PEBAL.
In general, compared to cDNP, the other approaches suffer from more false negatives and false positives respectively.

\begin{figure*}[h!]
    \centering
    \begin{subfigure}[b]{0.175\textwidth}
        \centering
        \includegraphics[width=\textwidth]{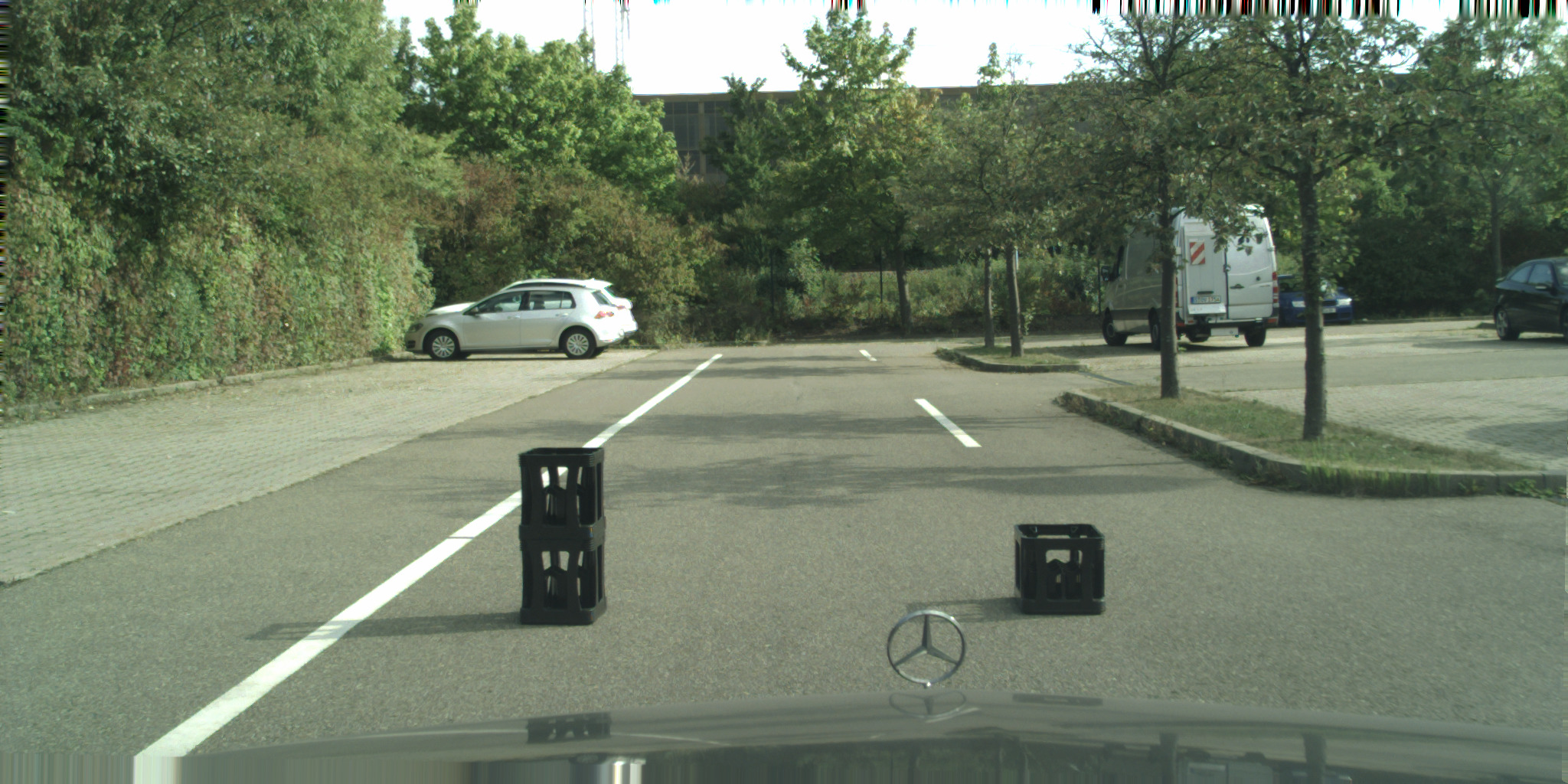}
    \end{subfigure}
    \begin{subfigure}[b]{0.175\textwidth}
        \centering
        \includegraphics[width=\textwidth]{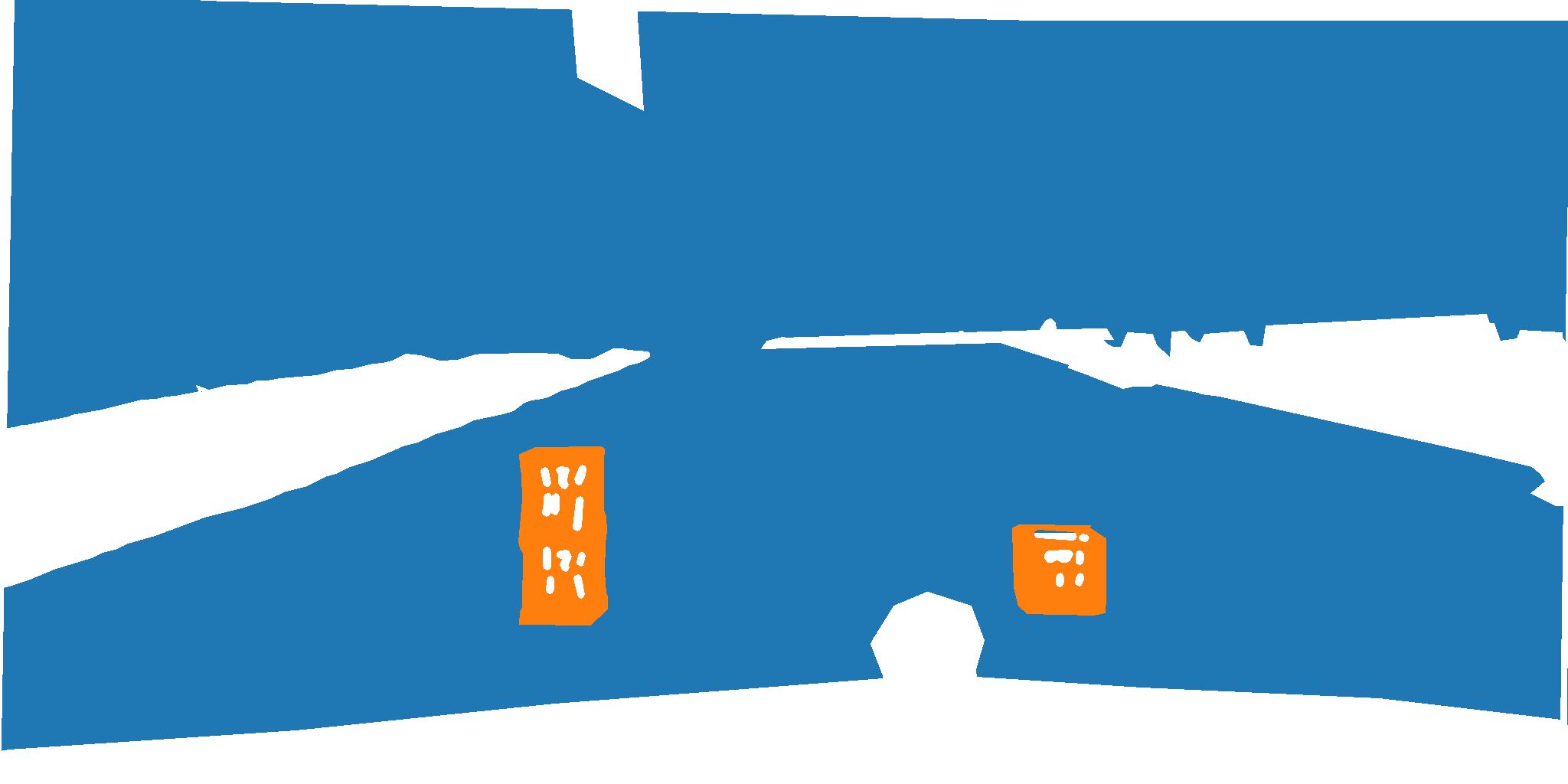}
    \end{subfigure}
    \begin{subfigure}[b]{0.175\textwidth}
        \centering
        \includegraphics[width=\textwidth]{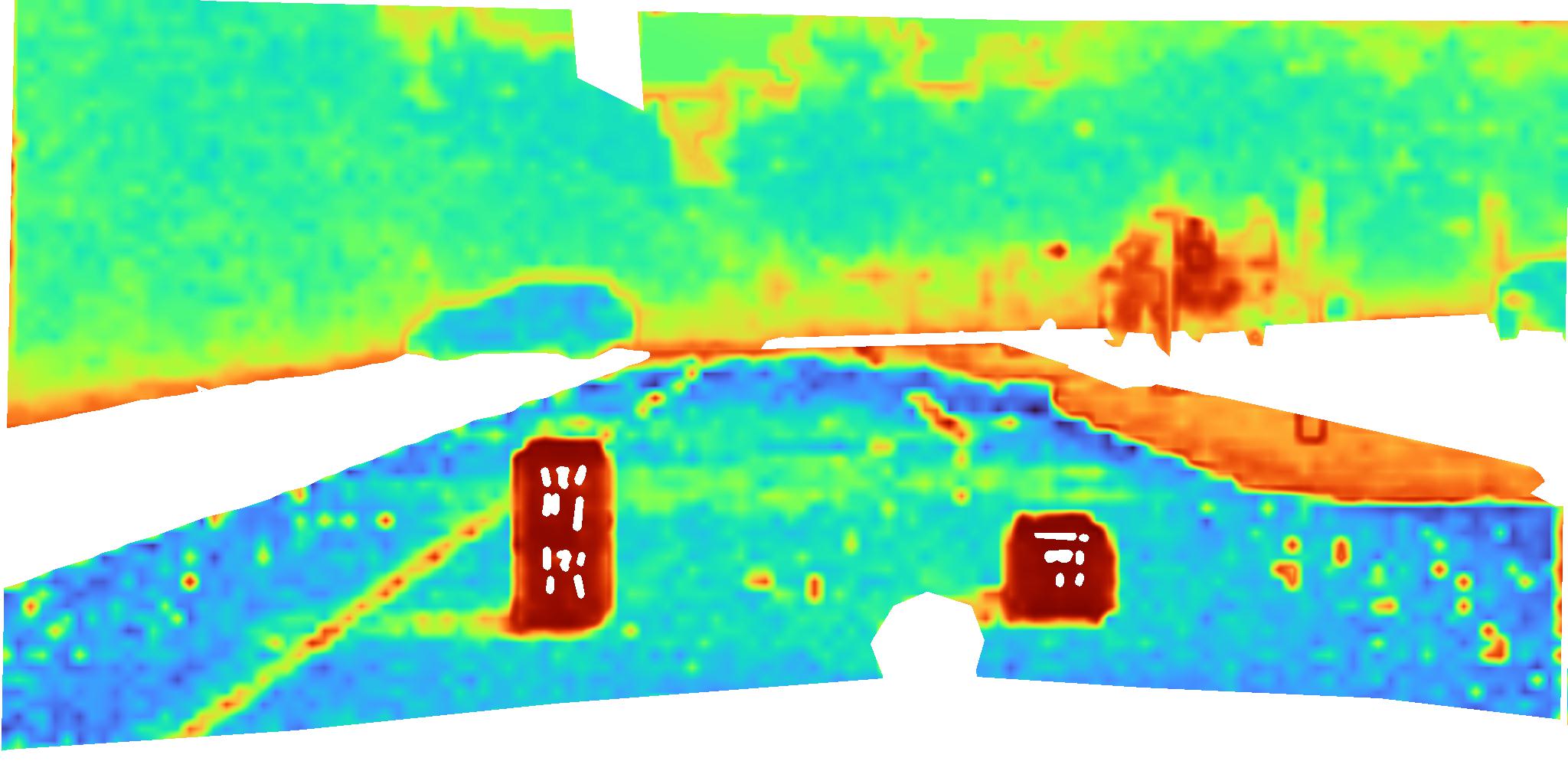}
    \end{subfigure}
    \begin{subfigure}[b]{0.175\textwidth}
        \centering
        \includegraphics[width=\textwidth]{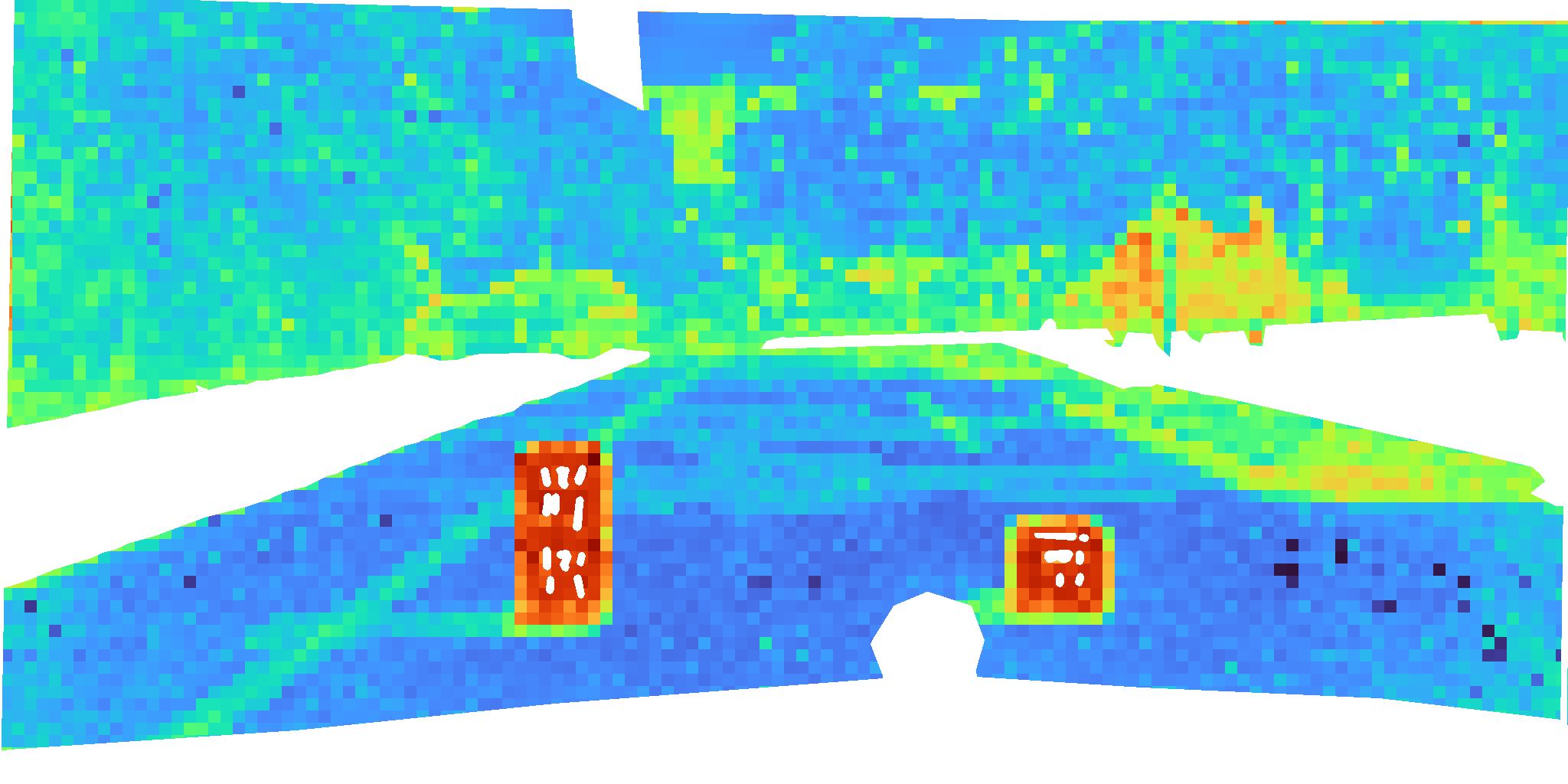}
    \end{subfigure}
    \begin{subfigure}[b]{0.175\textwidth}
        \centering
        \includegraphics[width=\textwidth]{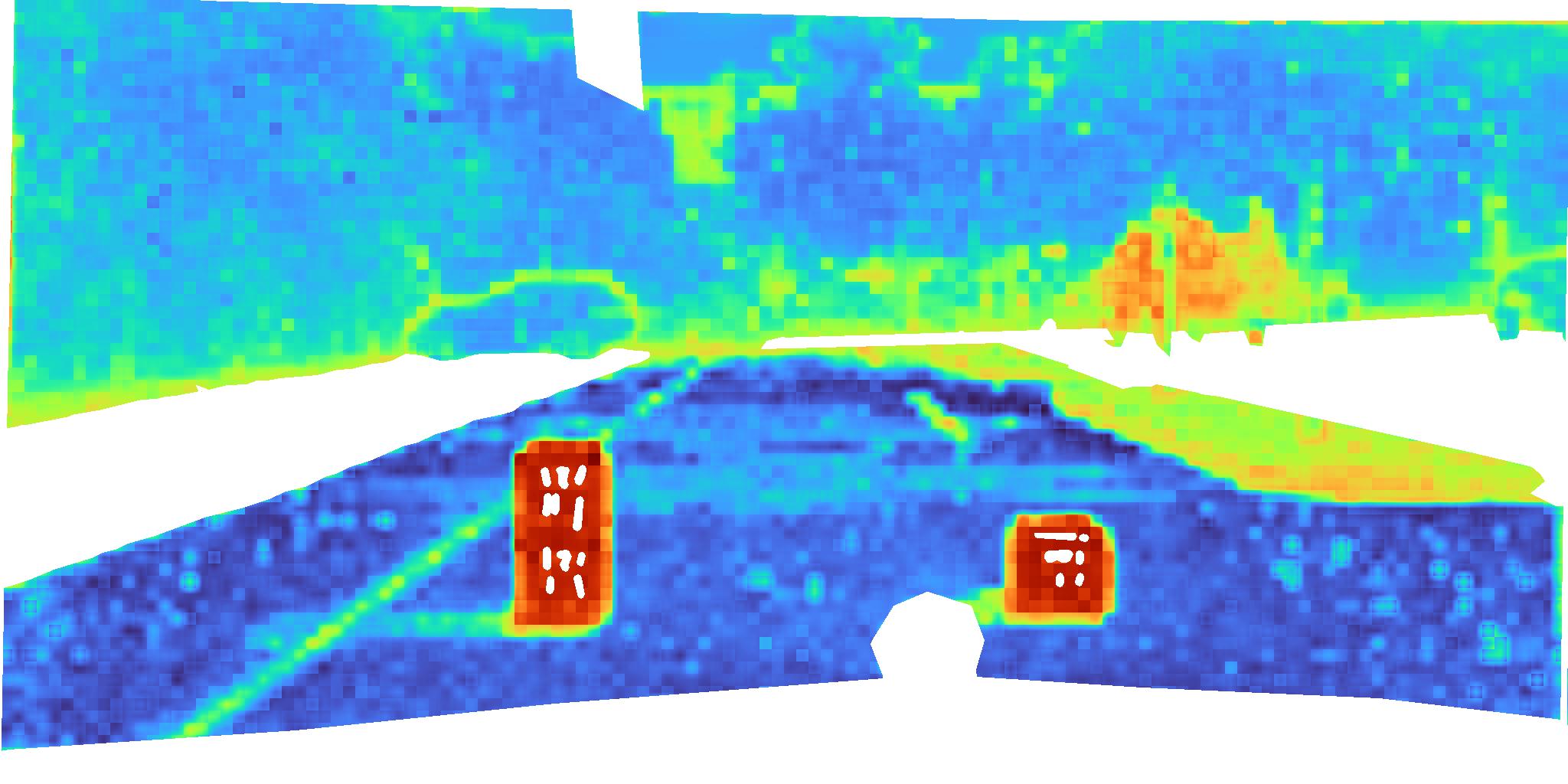}
    \end{subfigure}
    
    \begin{subfigure}[b]{0.175\textwidth}
        \centering
        \includegraphics[width=\textwidth]{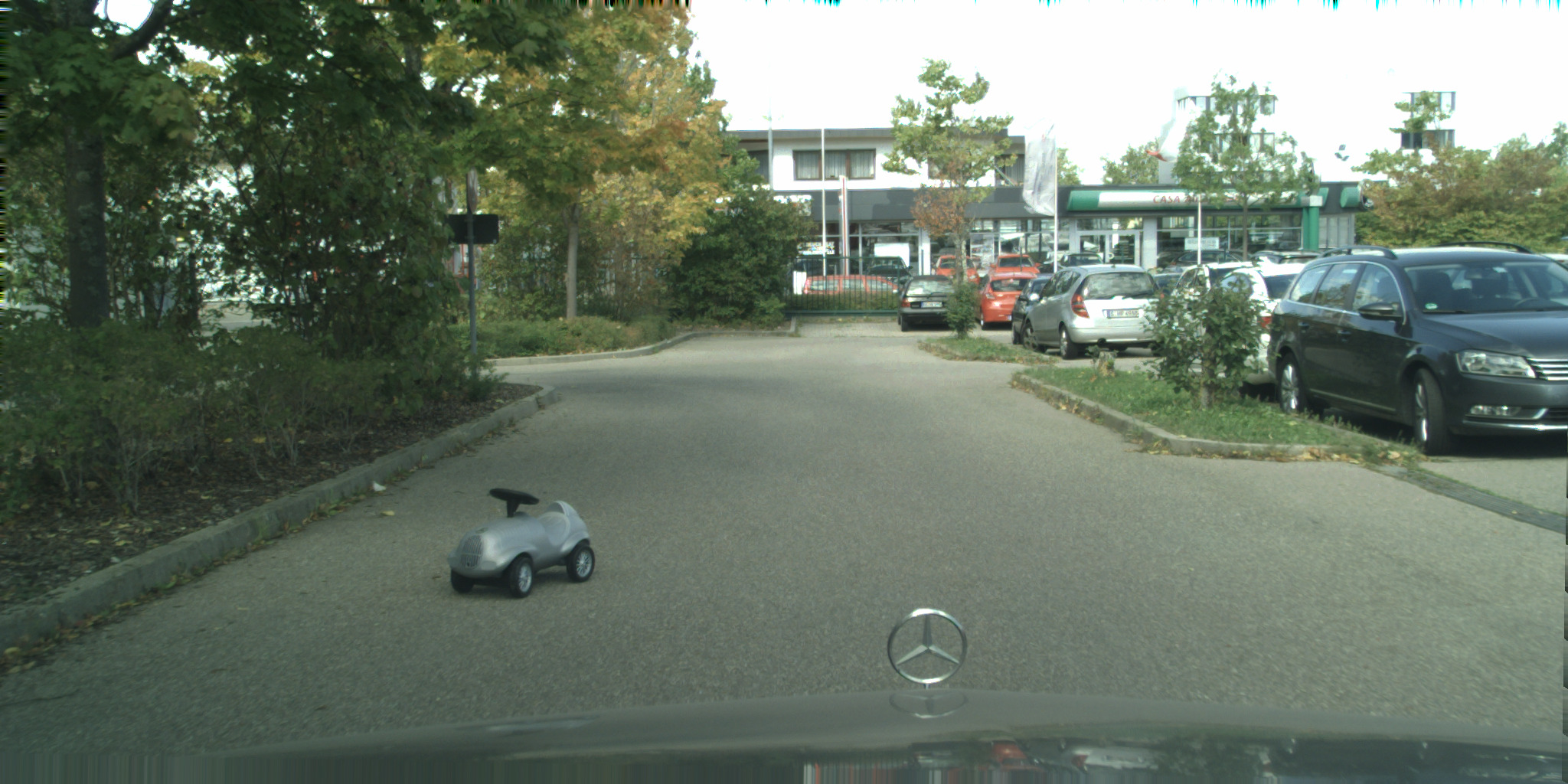}
    \end{subfigure}
    \begin{subfigure}[b]{0.175\textwidth}
        \centering
        \includegraphics[width=\textwidth]{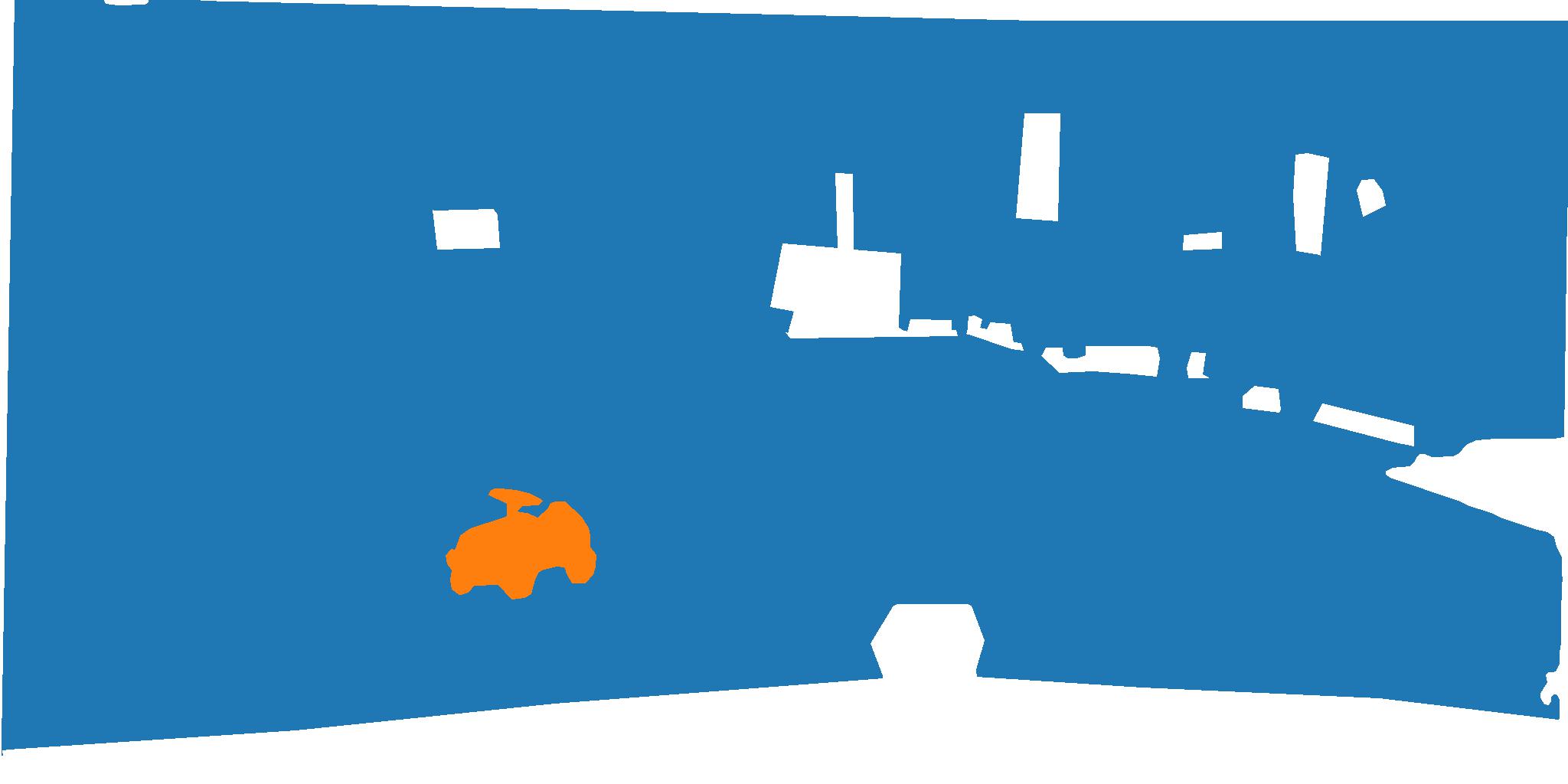}
    \end{subfigure}
    \begin{subfigure}[b]{0.175\textwidth}
        \centering
        \includegraphics[width=\textwidth]{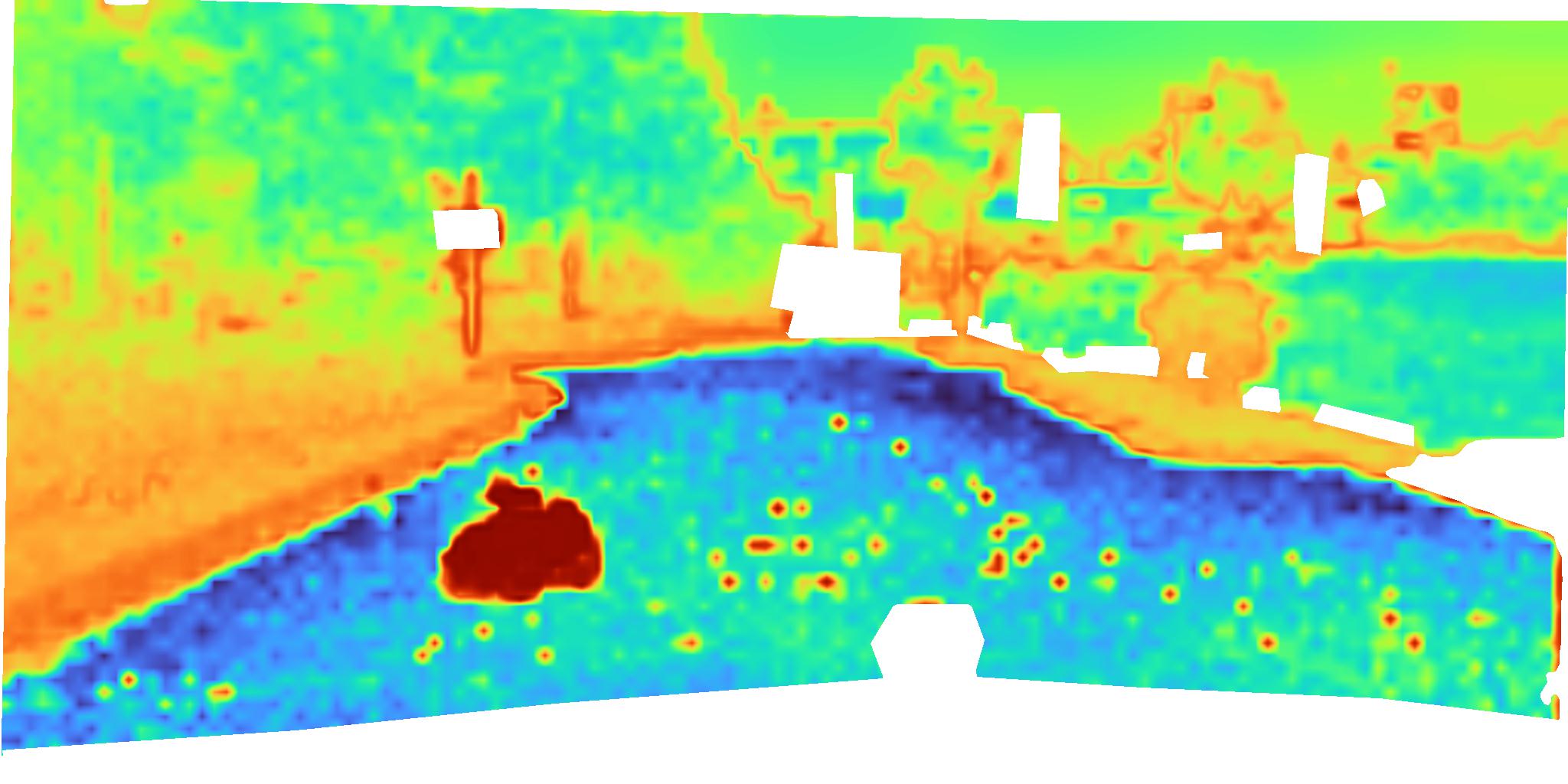}
    \end{subfigure}
    \begin{subfigure}[b]{0.175\textwidth}
        \centering
        \includegraphics[width=\textwidth]{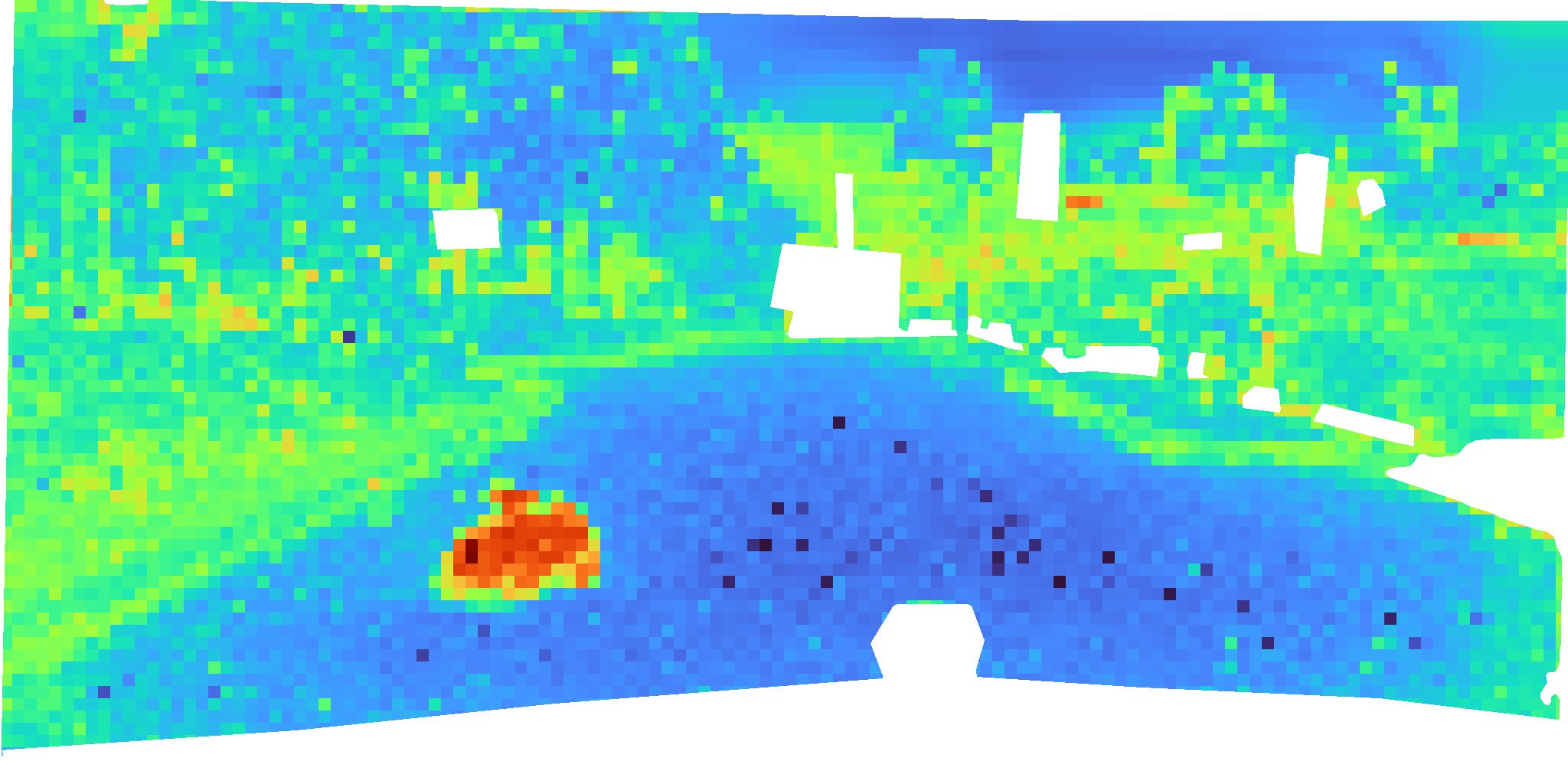}
    \end{subfigure}
    \begin{subfigure}[b]{0.175\textwidth}
        \centering
        \includegraphics[width=\textwidth]{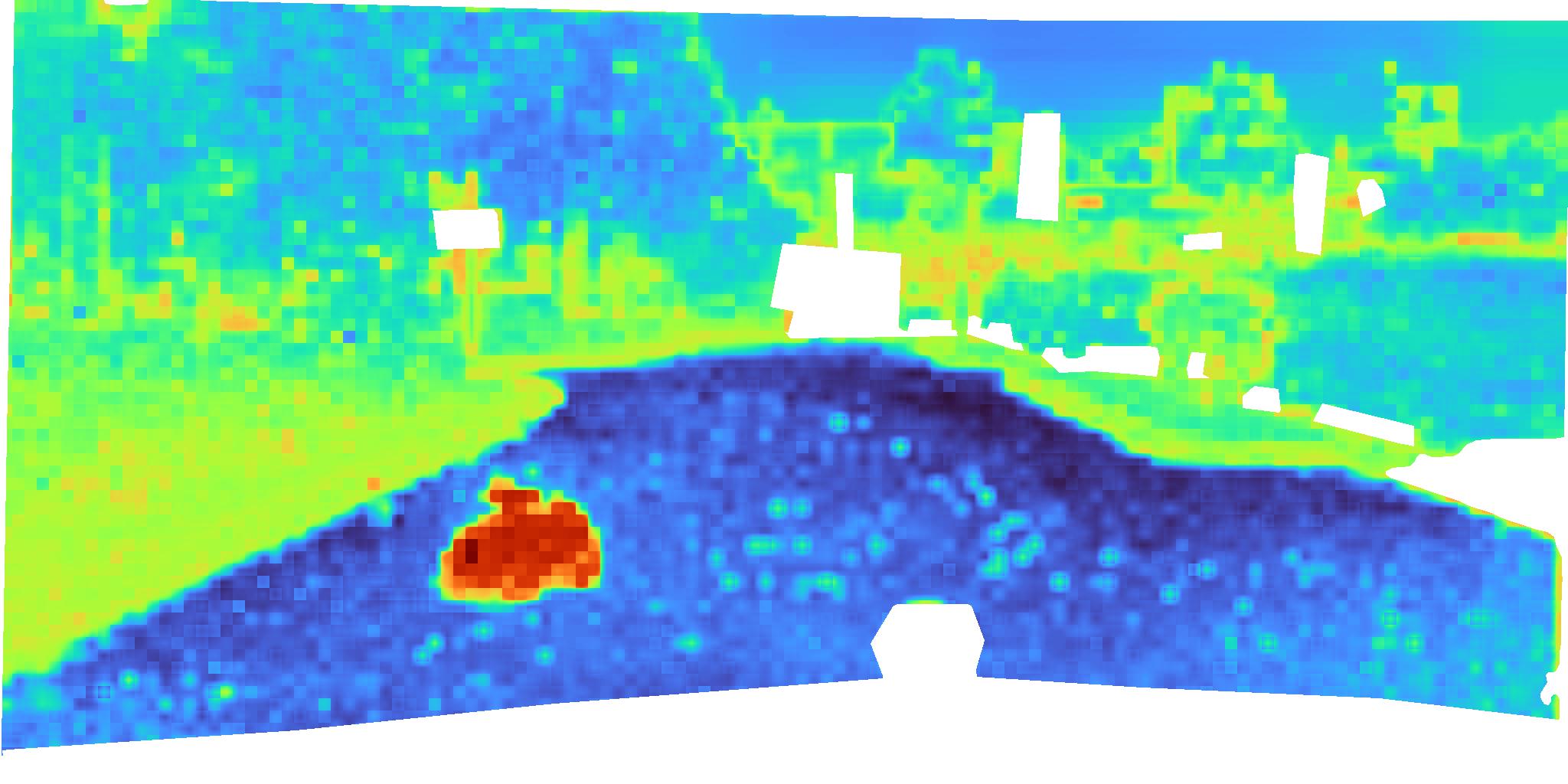}
    \end{subfigure}
    
    \begin{subfigure}[b]{0.175\textwidth}
        \centering
        \includegraphics[width=\textwidth]{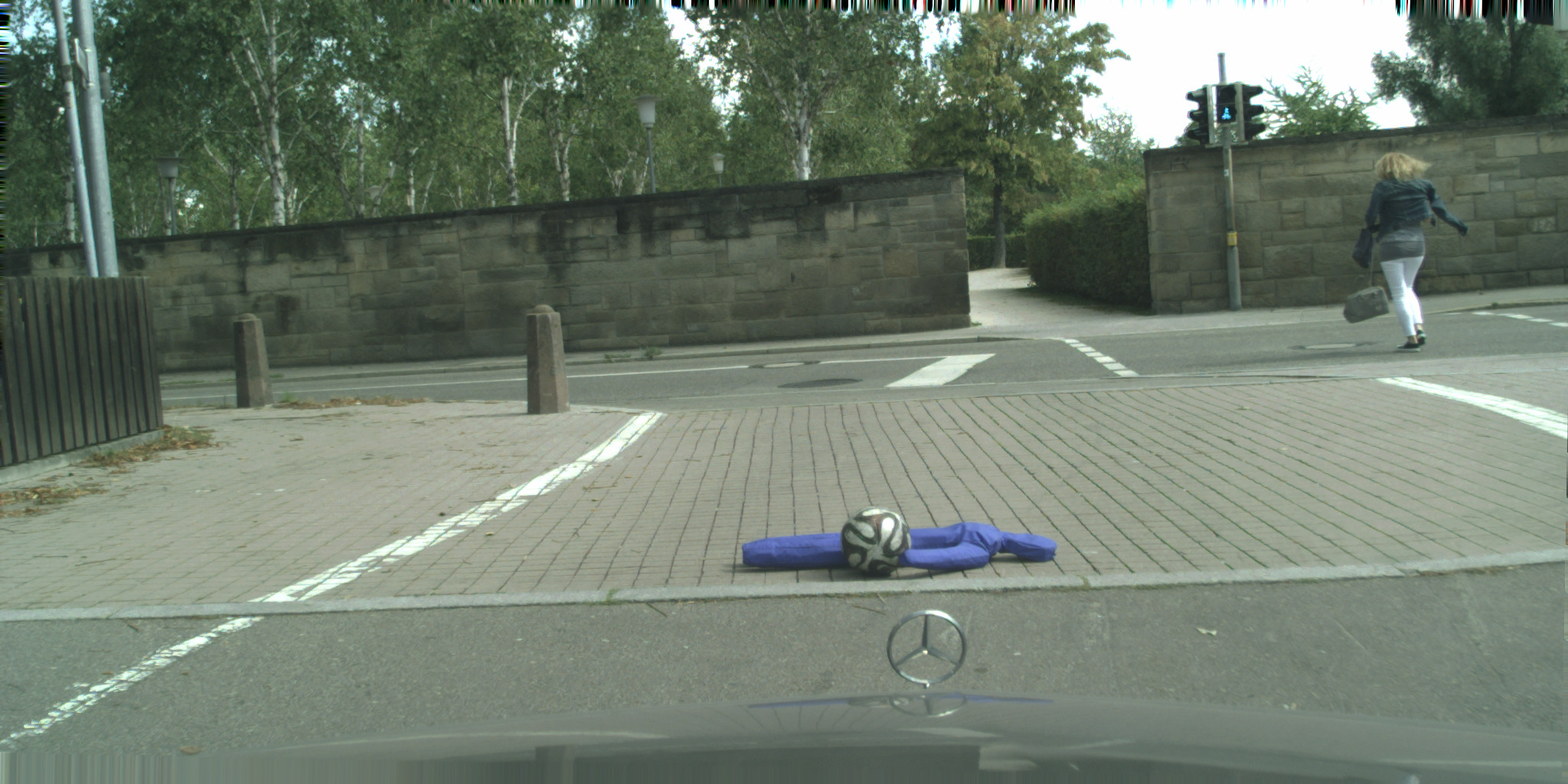}
    \end{subfigure}
    \begin{subfigure}[b]{0.175\textwidth}
        \centering
        \includegraphics[width=\textwidth]{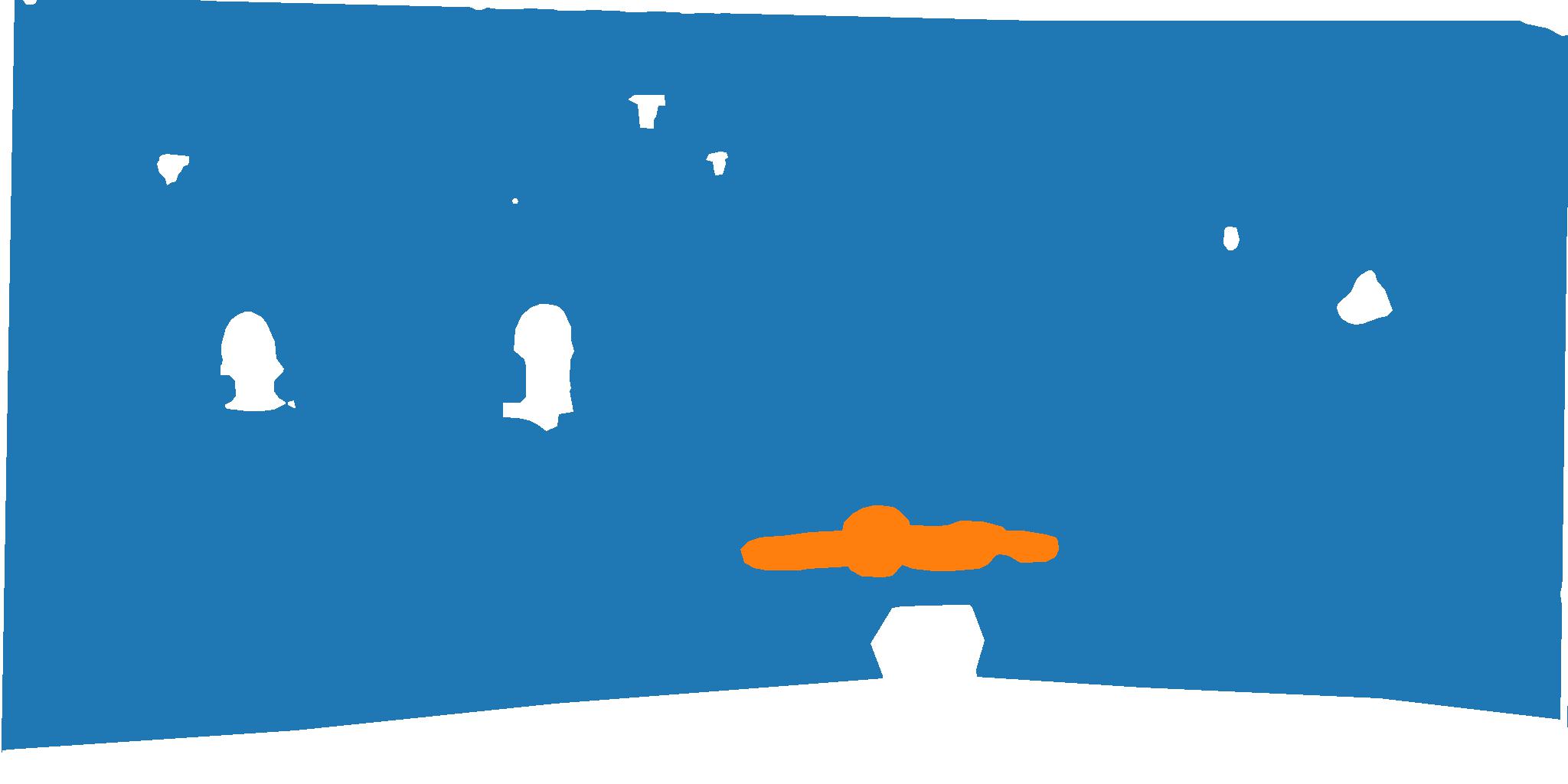}
    \end{subfigure}
    \begin{subfigure}[b]{0.175\textwidth}
        \centering
        \includegraphics[width=\textwidth]{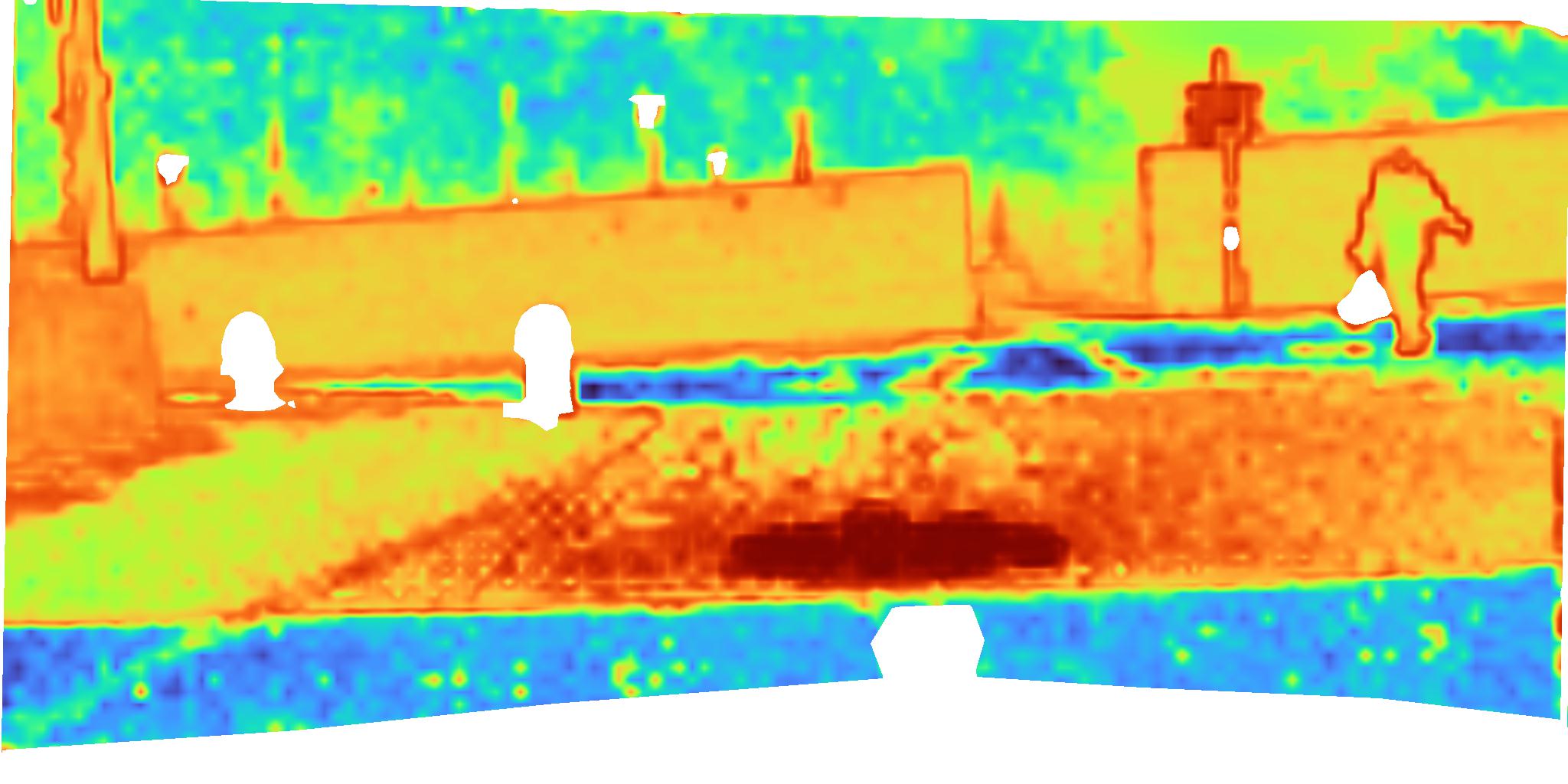}
    \end{subfigure}
    \begin{subfigure}[b]{0.175\textwidth}
        \centering
        \includegraphics[width=\textwidth]{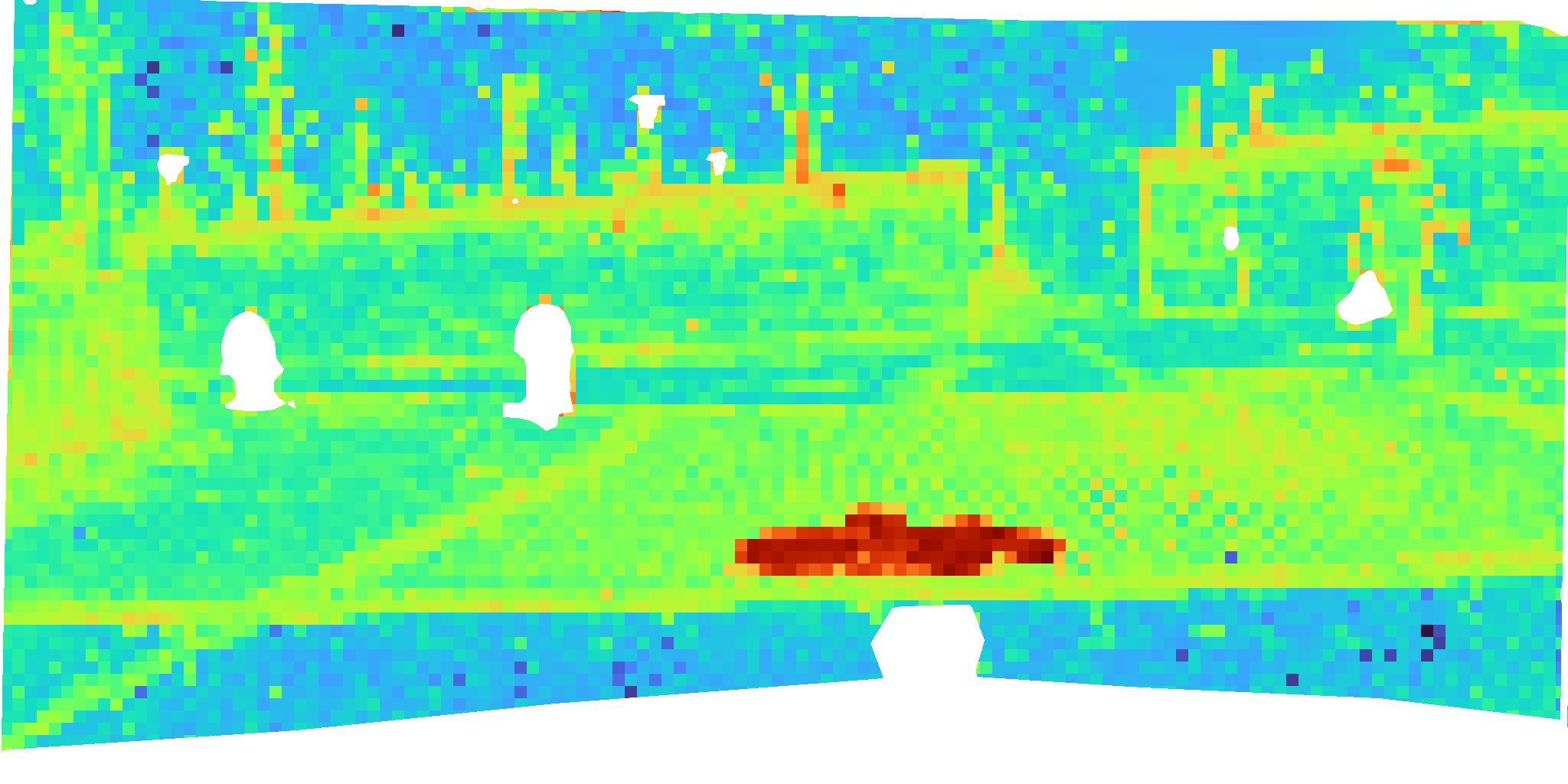}
    \end{subfigure}
    \begin{subfigure}[b]{0.175\textwidth}
        \centering
        \includegraphics[width=\textwidth]{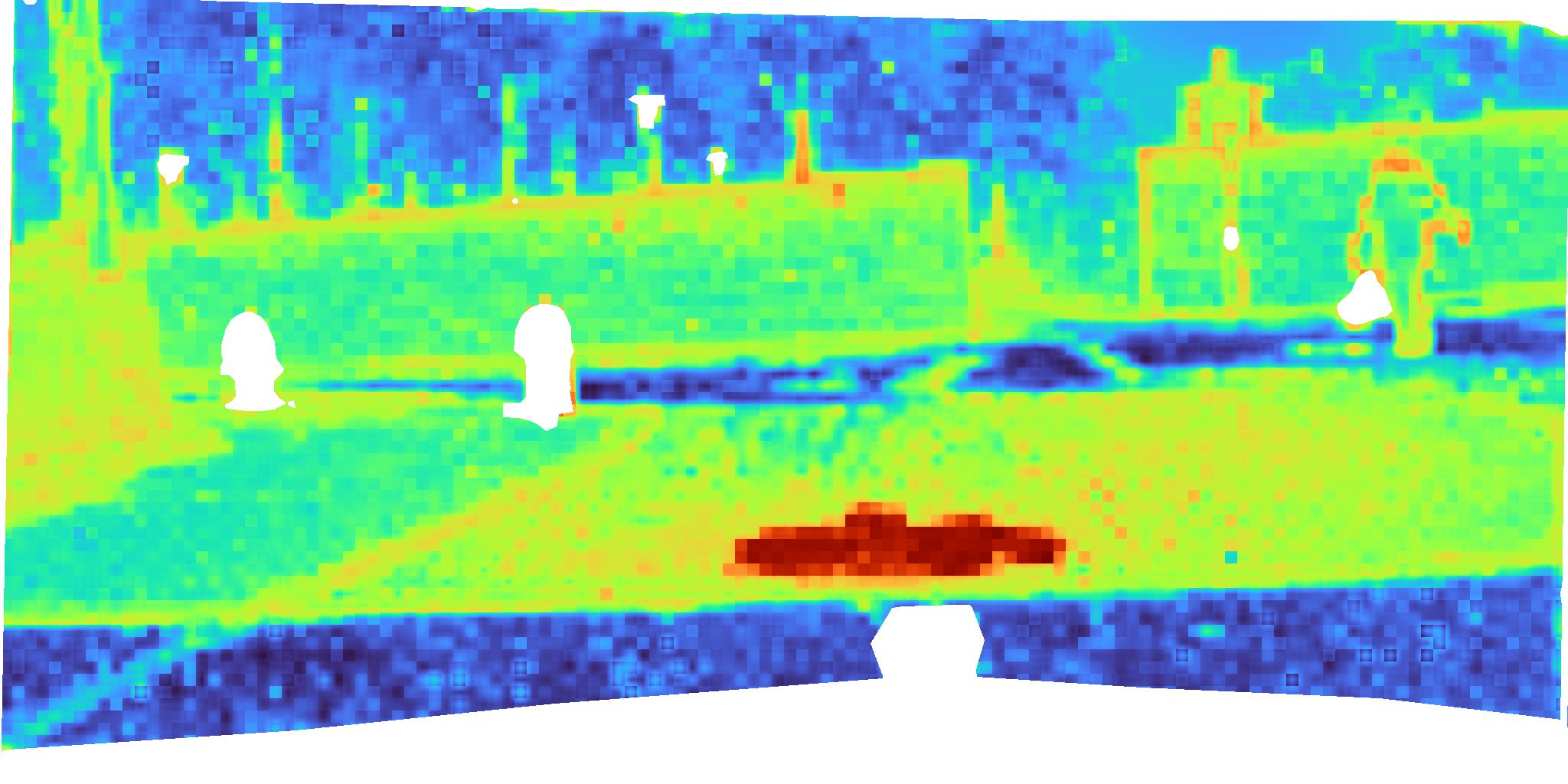}
    \end{subfigure}
    
    
    \begin{subfigure}[b]{0.175\textwidth}
        \centering
        \includegraphics[width=\textwidth]{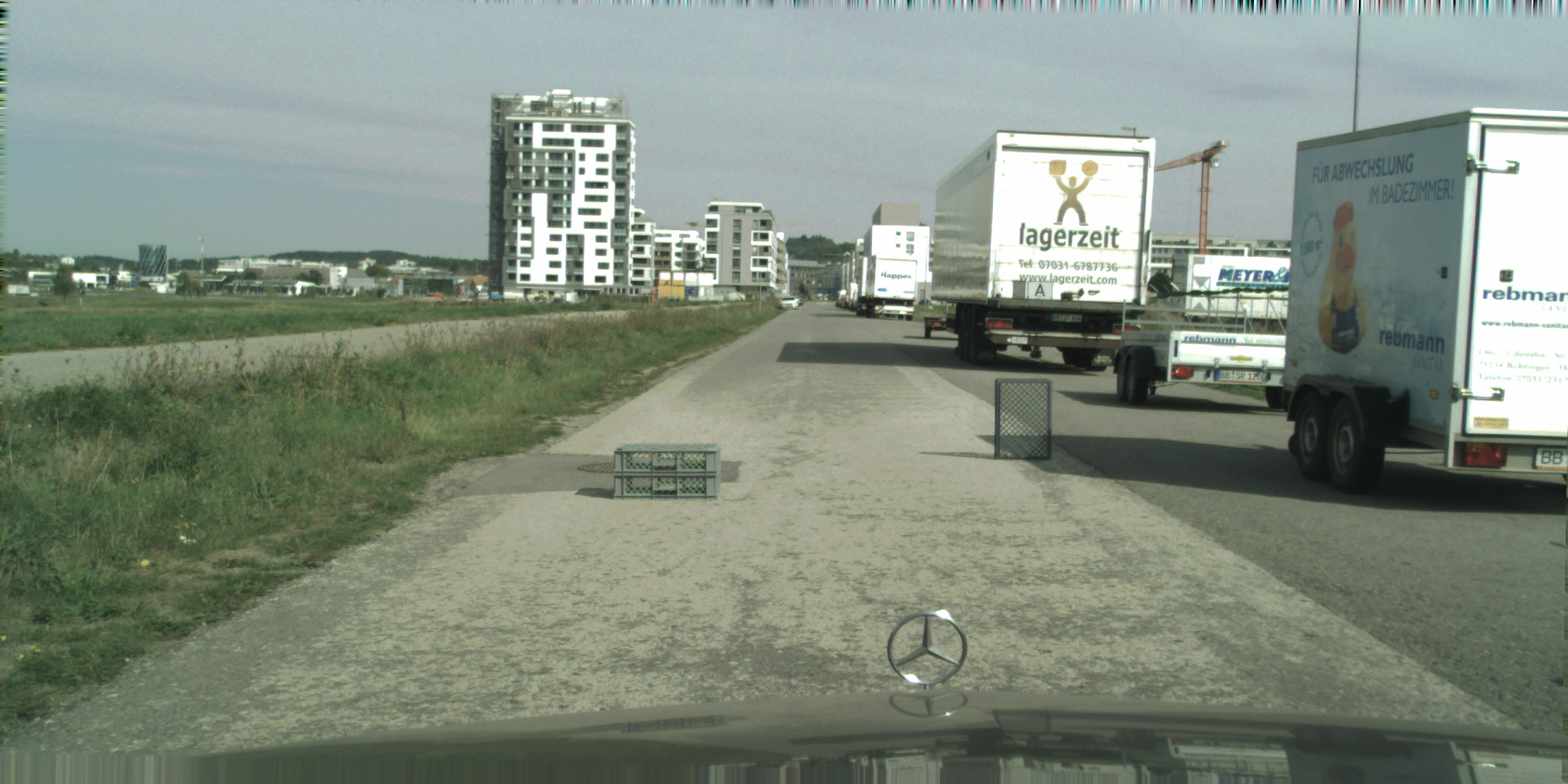}
    \end{subfigure}
    \begin{subfigure}[b]{0.175\textwidth}
        \centering
        \includegraphics[width=\textwidth]{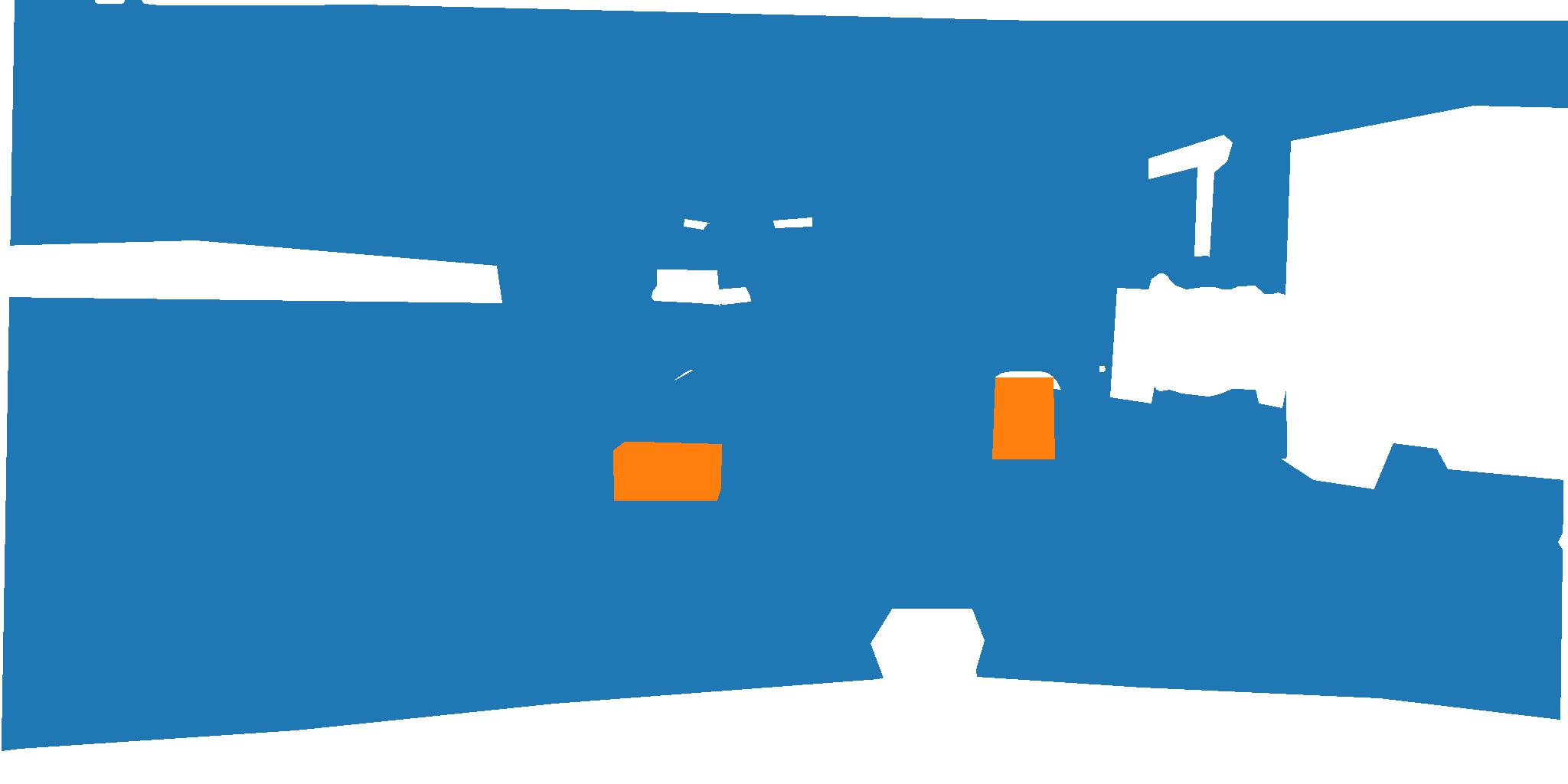}
    \end{subfigure}
    \begin{subfigure}[b]{0.175\textwidth}
        \centering
        \includegraphics[width=\textwidth]{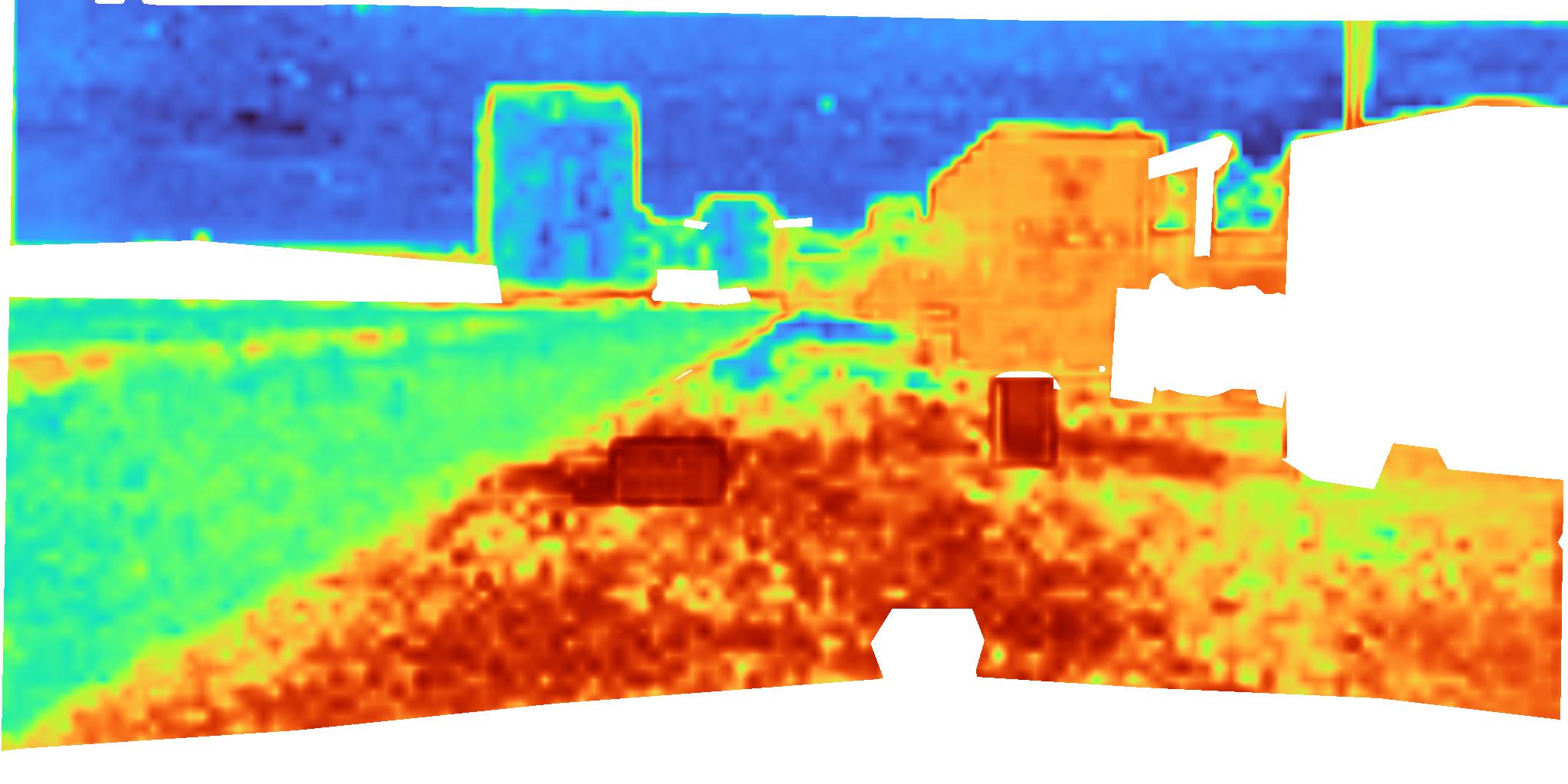}
    \end{subfigure}
    \begin{subfigure}[b]{0.175\textwidth}
        \centering
        \includegraphics[width=\textwidth]{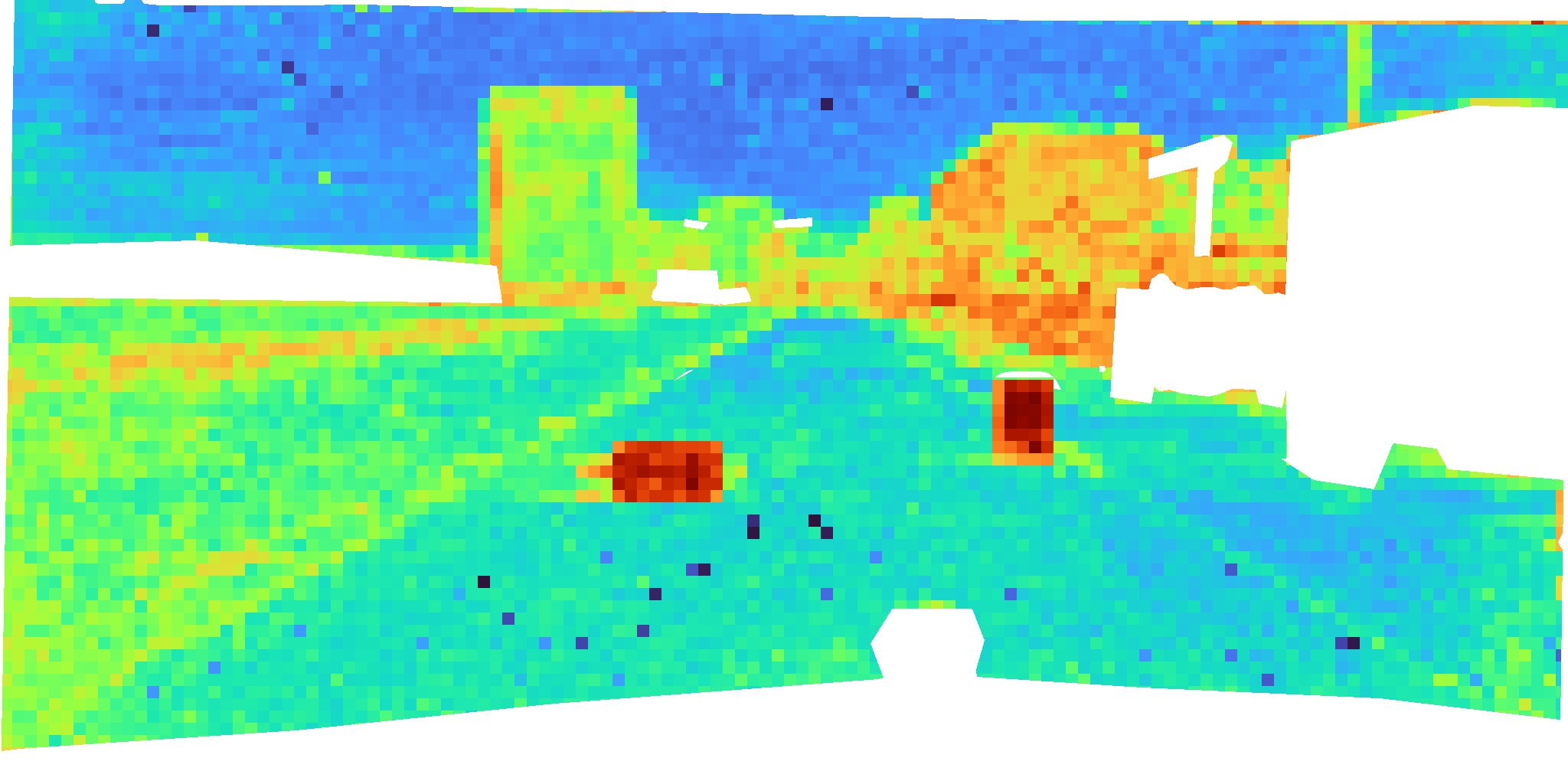}
    \end{subfigure}
    \begin{subfigure}[b]{0.175\textwidth}
        \centering
        \includegraphics[width=\textwidth]{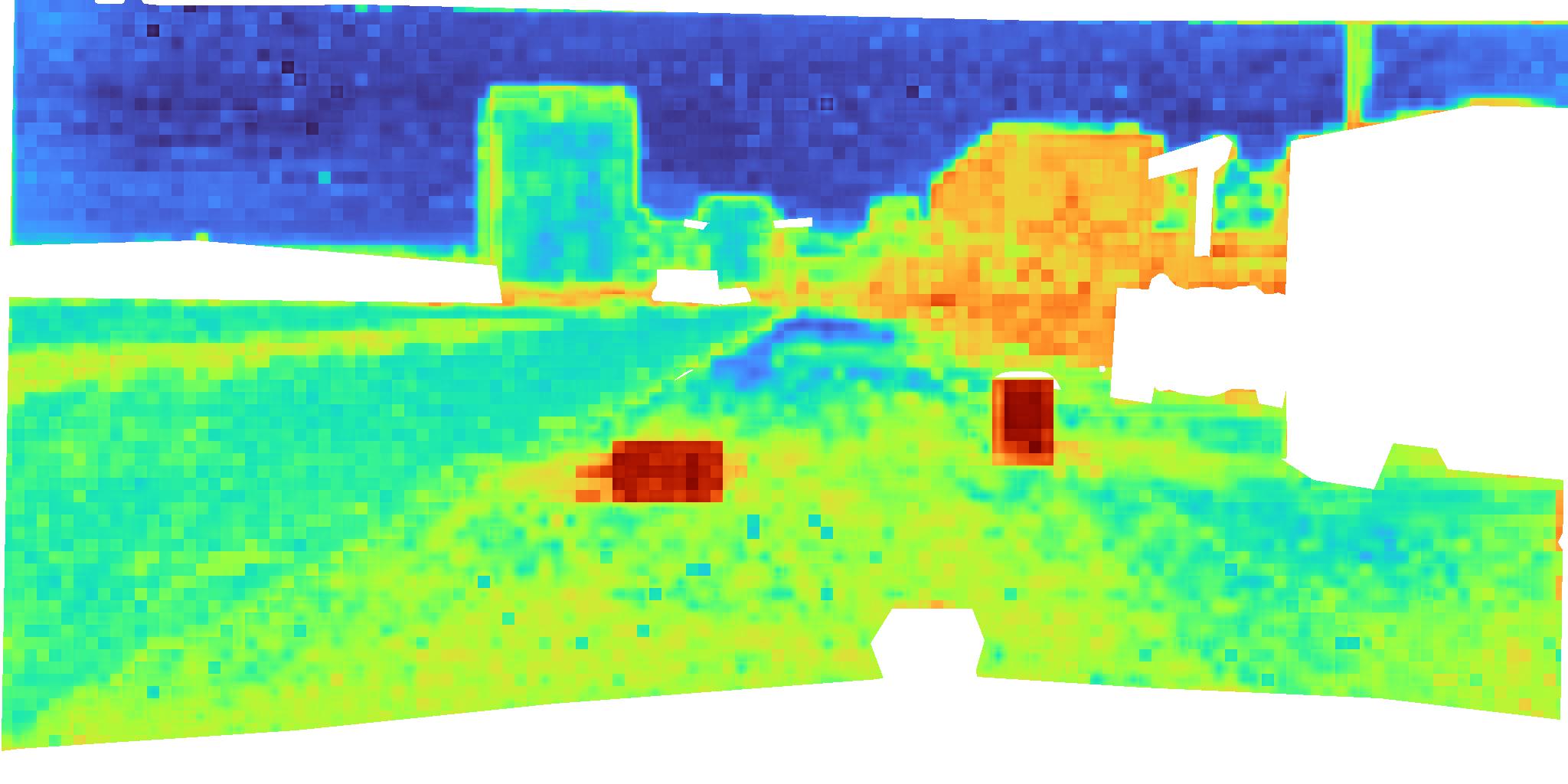}
    \end{subfigure}
    
    \begin{subfigure}[b]{0.175\textwidth}
        \centering
        \includegraphics[width=\textwidth]{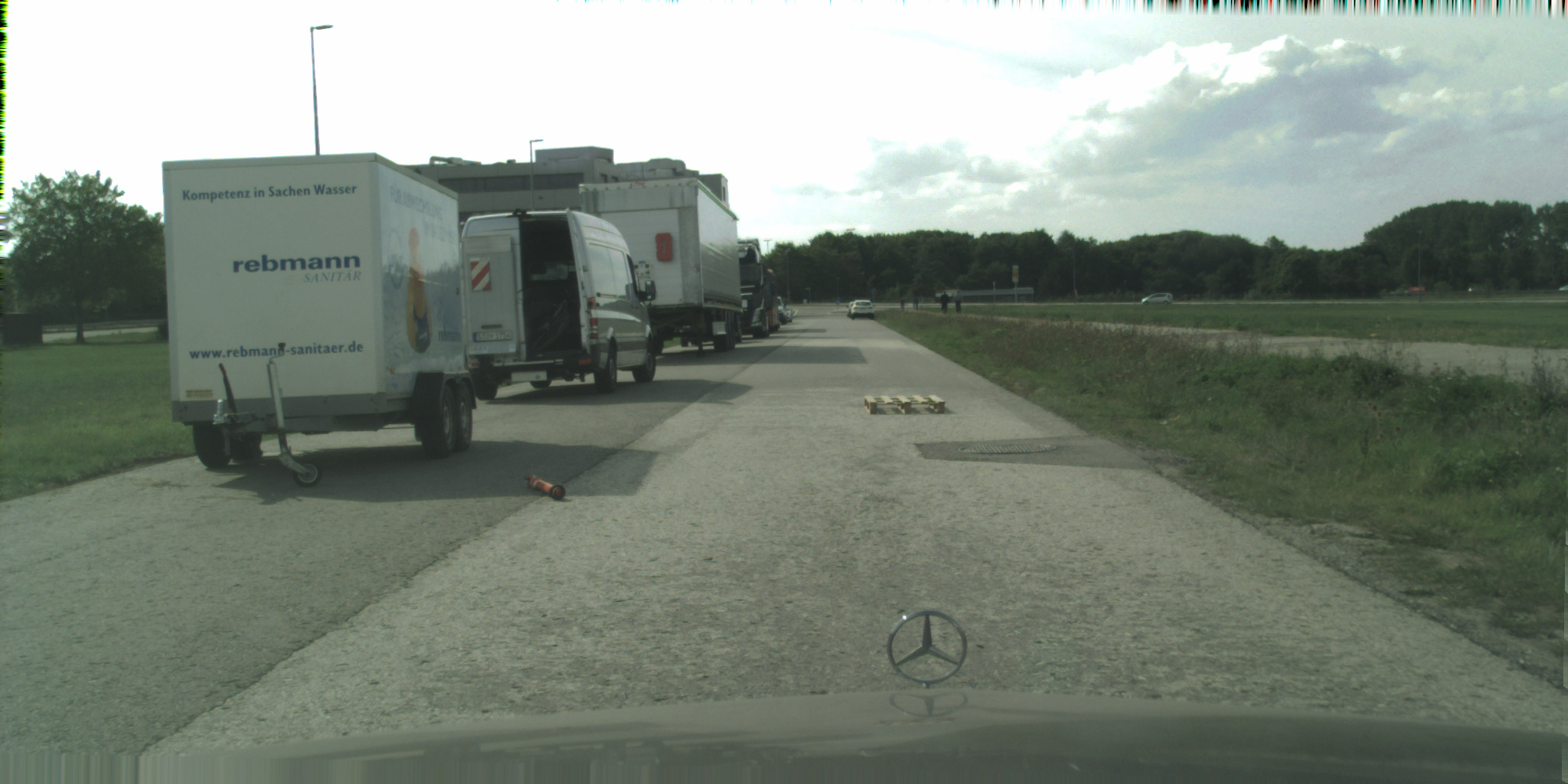}
    \end{subfigure}
    \begin{subfigure}[b]{0.175\textwidth}
        \centering
        \includegraphics[width=\textwidth]{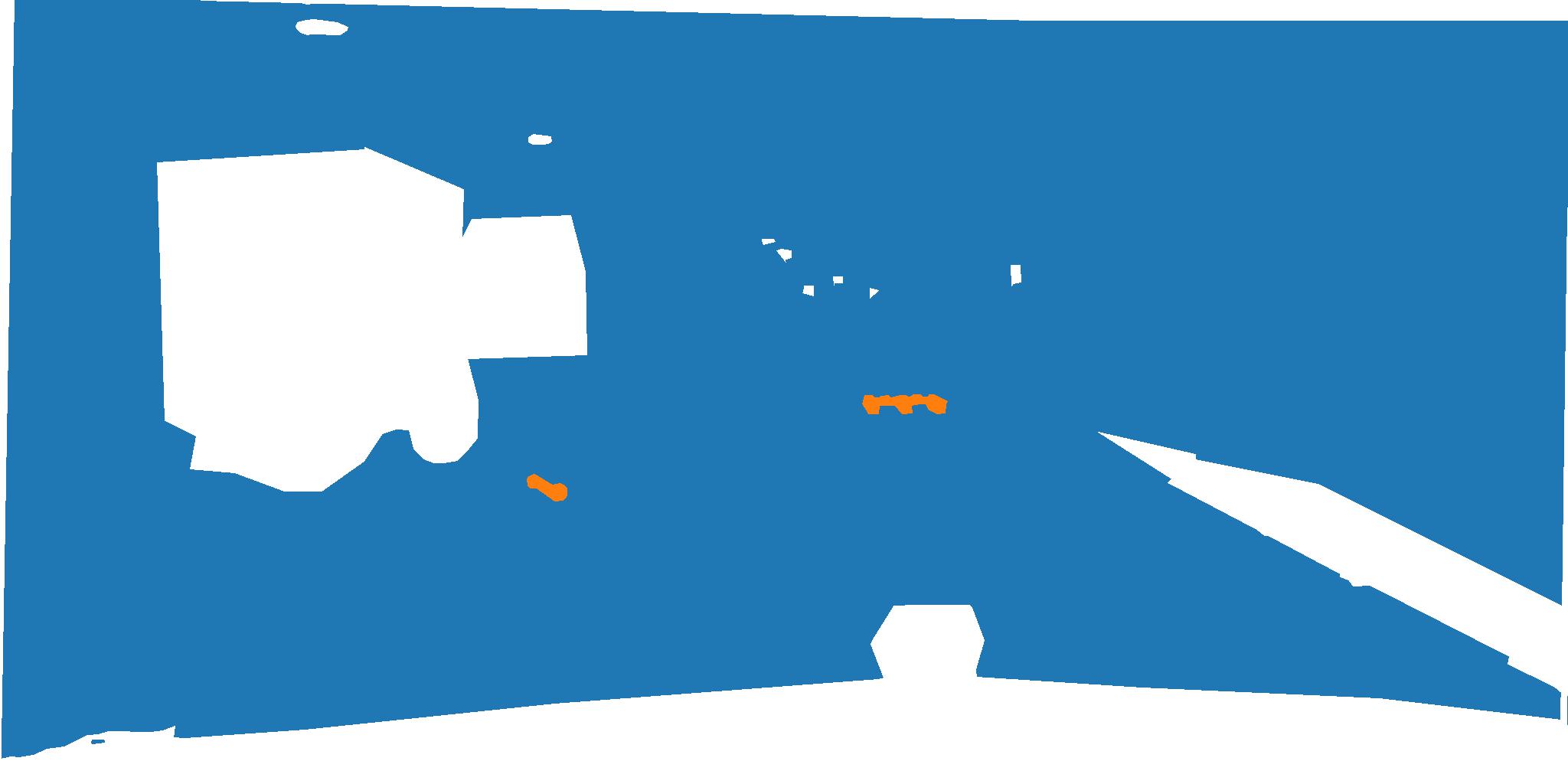}
    \end{subfigure}
    \begin{subfigure}[b]{0.175\textwidth}
        \centering
        \includegraphics[width=\textwidth]{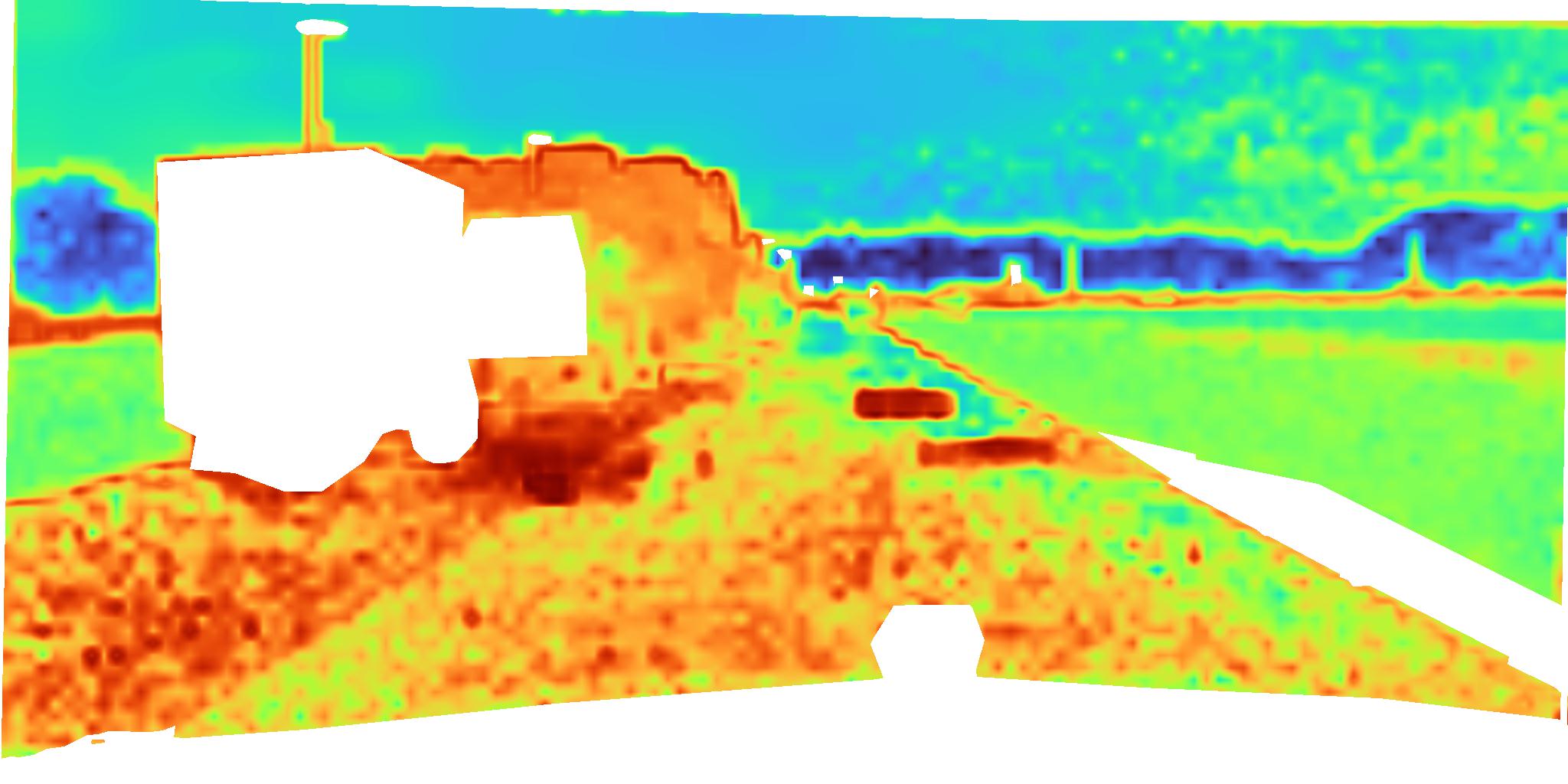}
    \end{subfigure}
    \begin{subfigure}[b]{0.175\textwidth}
        \centering
        \includegraphics[width=\textwidth]{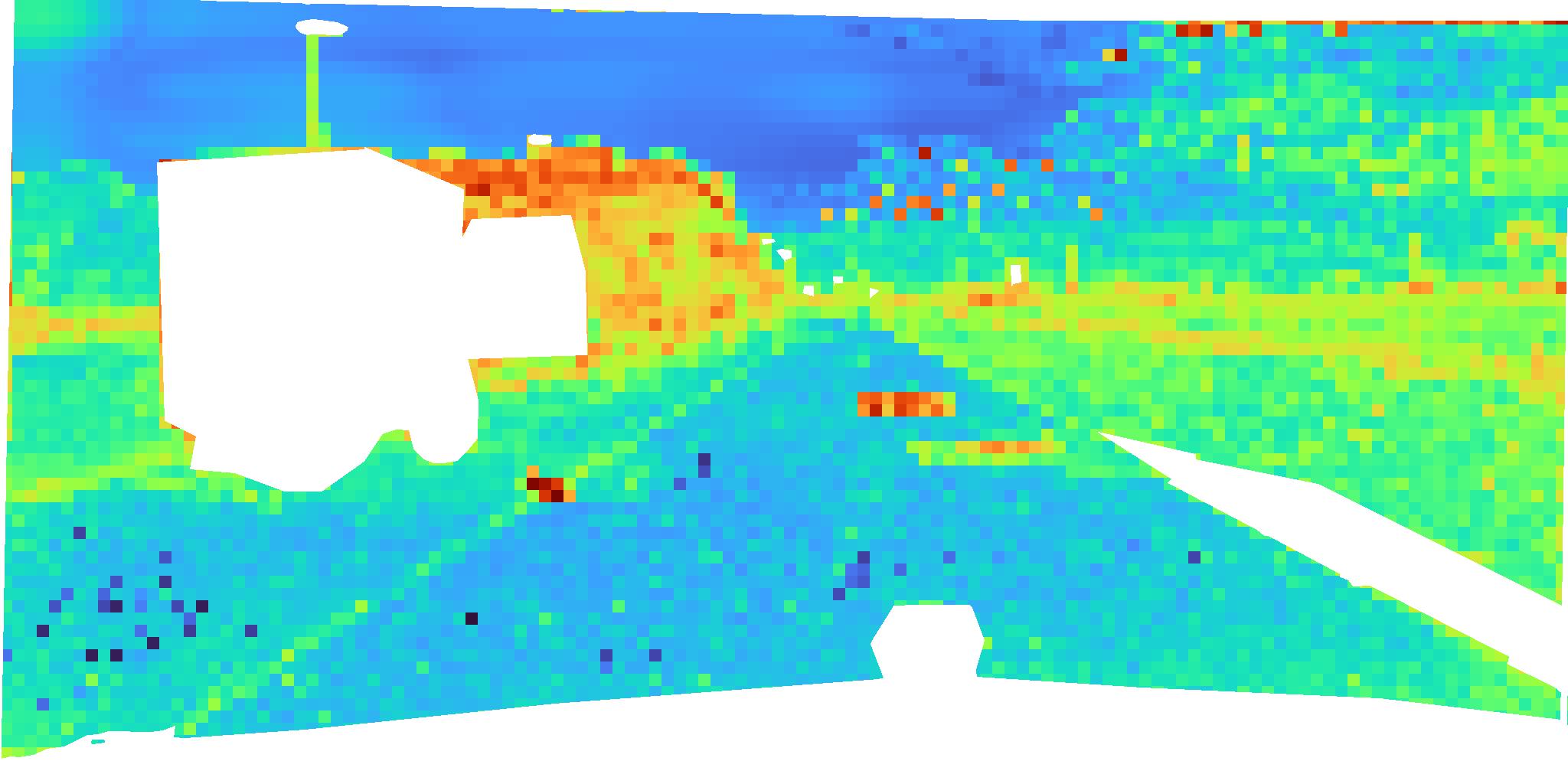}
    \end{subfigure}
    \begin{subfigure}[b]{0.175\textwidth}
        \centering
        \includegraphics[width=\textwidth]{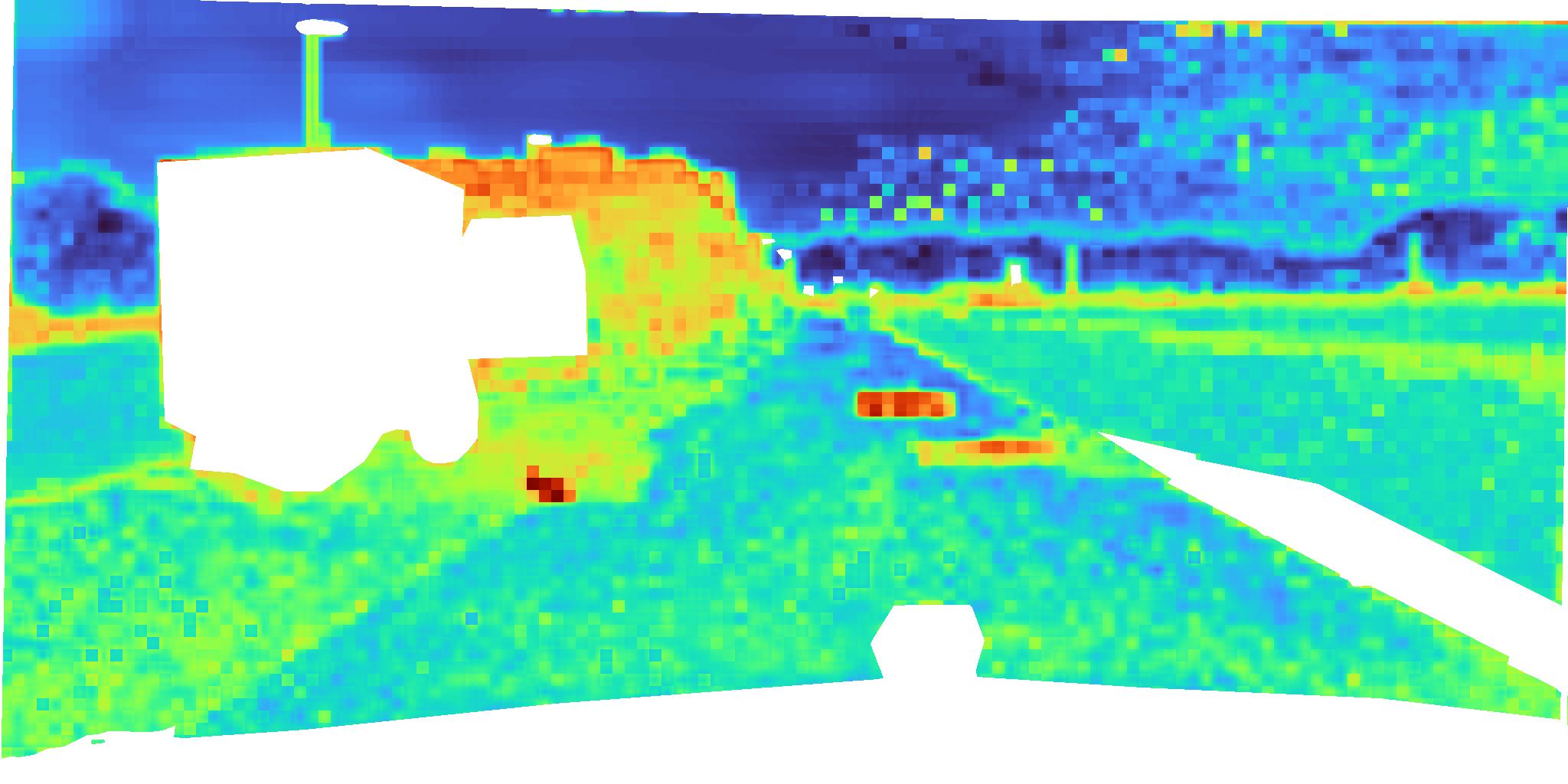}
    \end{subfigure}
    
    \begin{subfigure}[b]{0.175\textwidth}
        \centering
        \includegraphics[width=\textwidth]{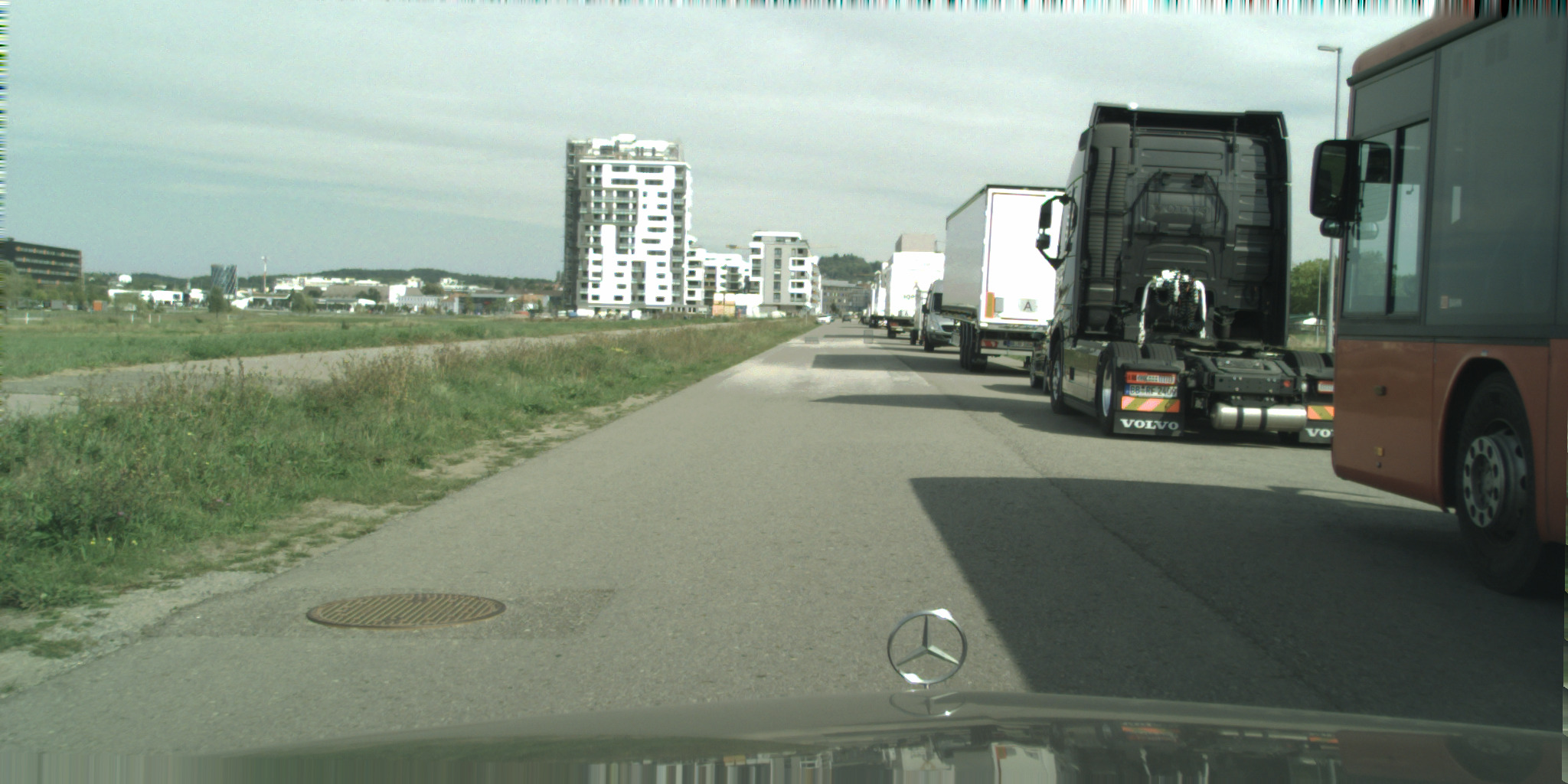}
        \caption*{Image}
    \end{subfigure}
    \begin{subfigure}[b]{0.175\textwidth}
        \centering
        \includegraphics[width=\textwidth]{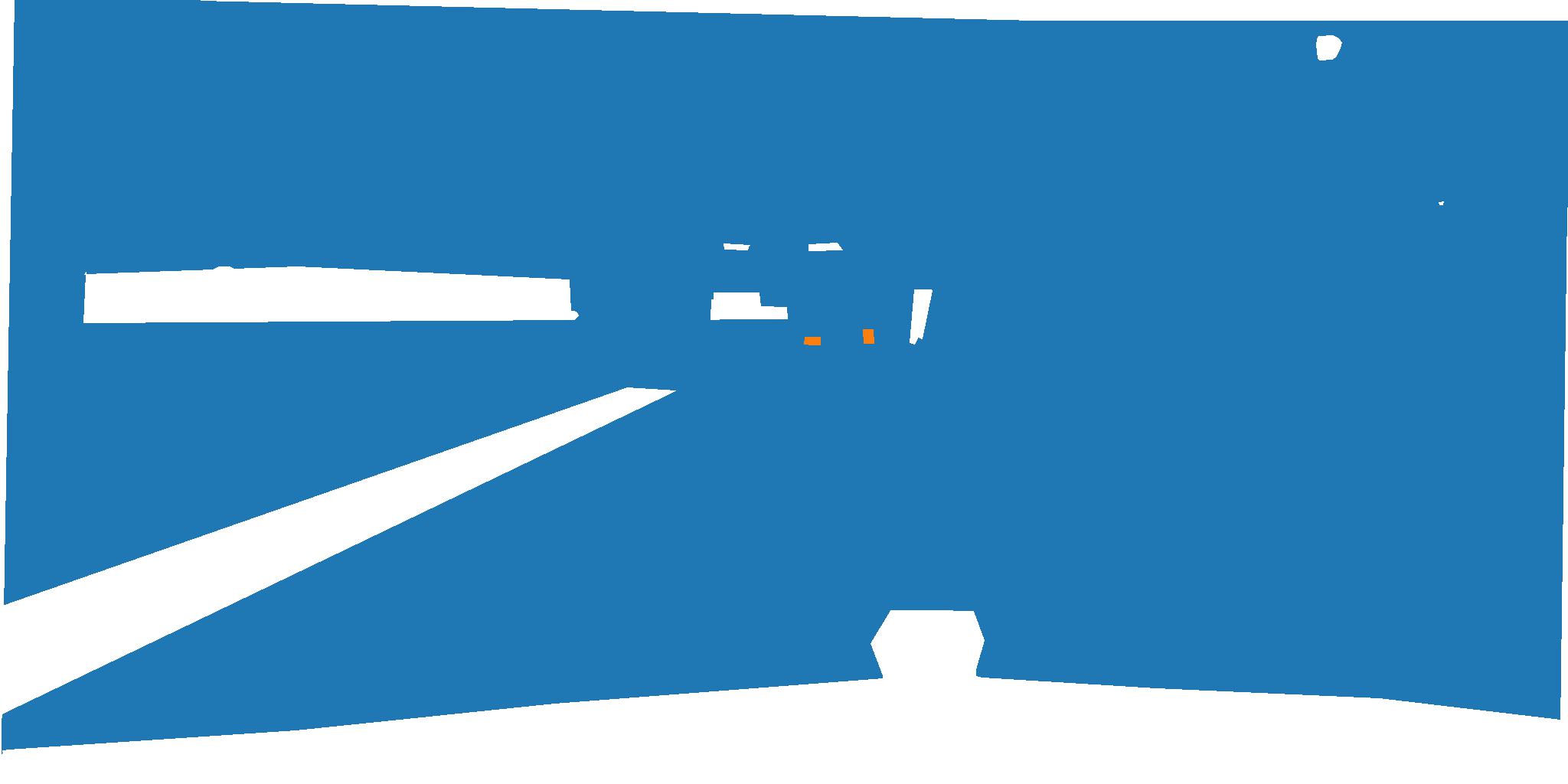}
        \caption*{Ground truth}
    \end{subfigure}
    \begin{subfigure}[b]{0.175\textwidth}
        \centering
        \includegraphics[width=\textwidth]{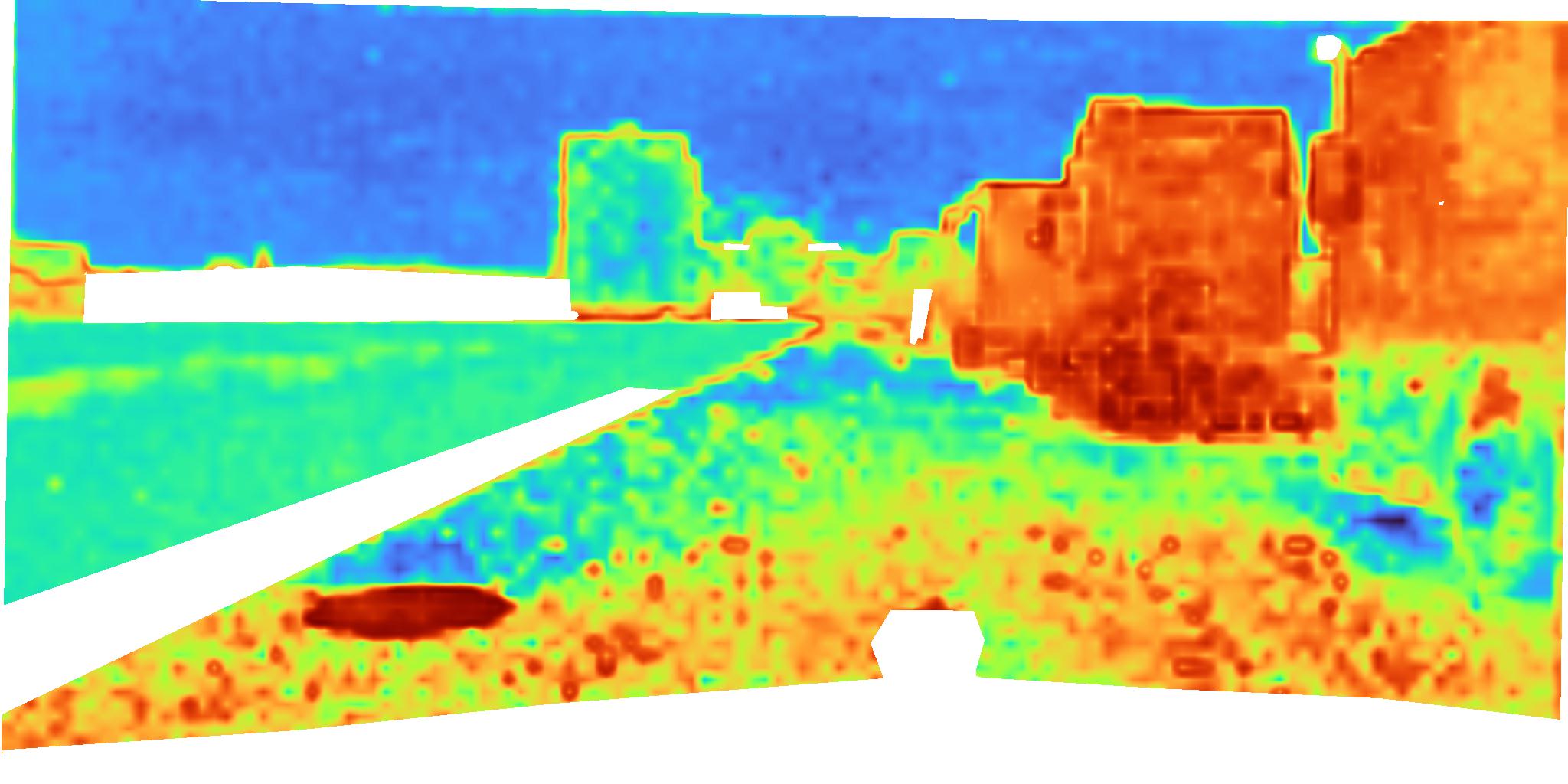}
        \caption*{Parametric}
    \end{subfigure}
    \begin{subfigure}[b]{0.175\textwidth}
        \centering
        \includegraphics[width=\textwidth]{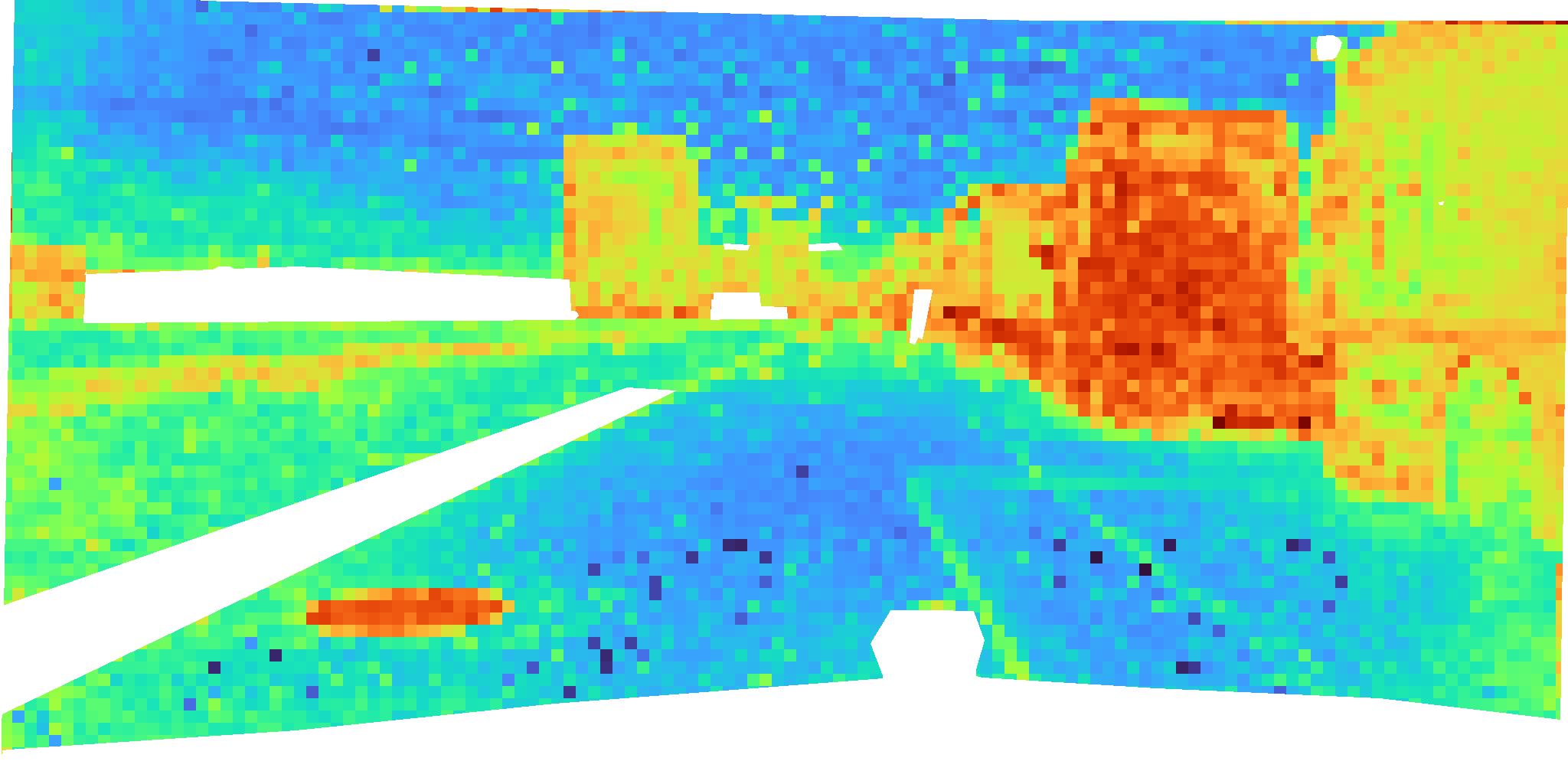}
        \caption*{DNP}
    \end{subfigure}
    \begin{subfigure}[b]{0.175\textwidth}
        \centering
        \includegraphics[width=\textwidth]{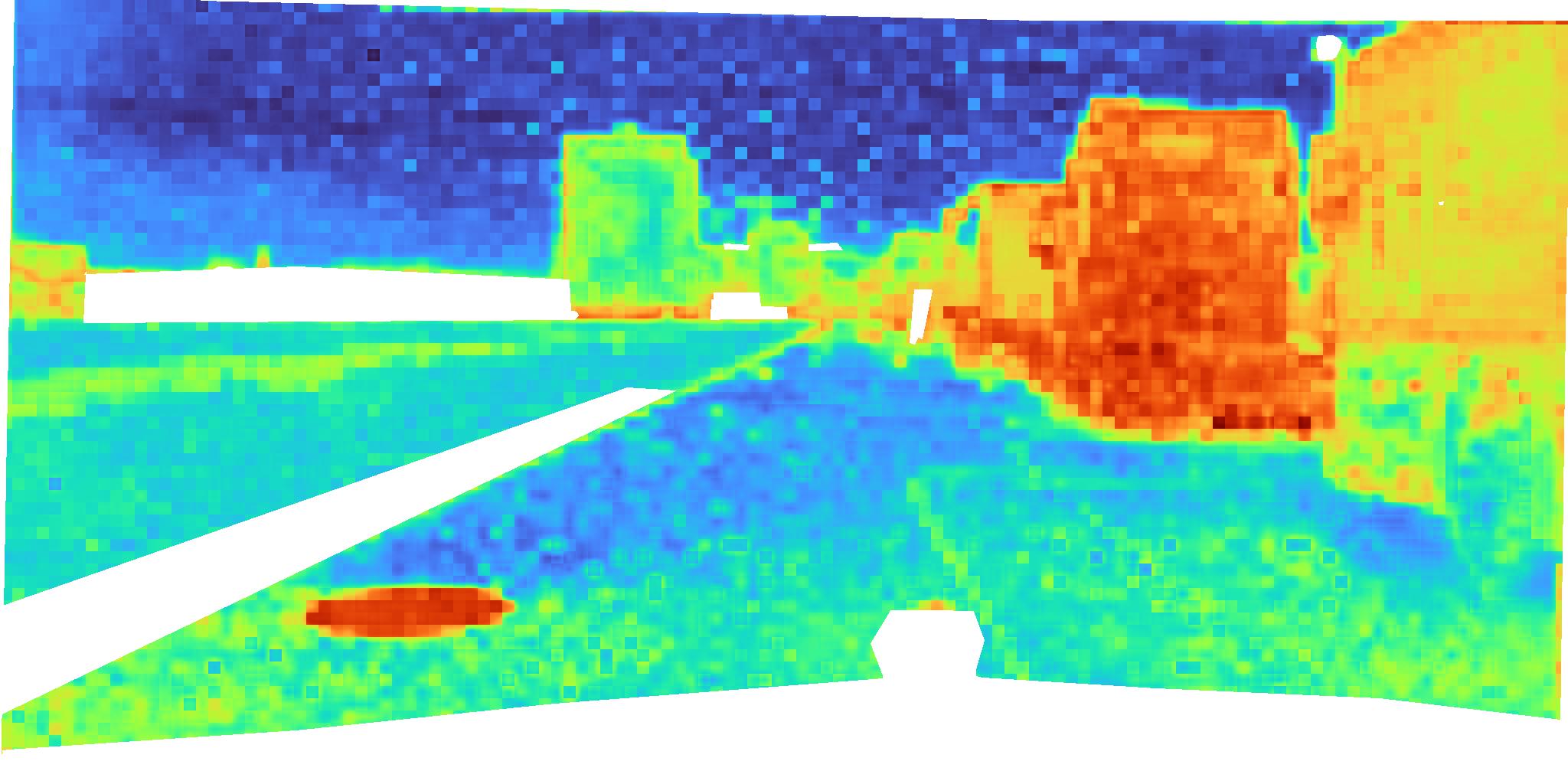}
        \caption*{cDNP}
    \end{subfigure}
    
    \caption{Qualitative examples of our approach on Fishyscapes Lost\&Found, showing the parametric (LogSumExp), DNP and cDNP scores for the best model reported in Table 2c of the main paper. The ground truth shows the valid in and out of distribution pixels in blue and orange respectively. DNP and cDNP exhibit fewer false positives than LogSumExp on all examples, especially on unusual terrains (rows 3, 4, and 5). DNP/cDNP can also successfully identify the small obstacles in the 5th row example. All methods fail on the example in the last row.}
    
    \label{fig:laf_quali}
\end{figure*}

\paragraph{Lost\&Found}
Figure~\ref{fig:laf_quali} contains qualitative examples of parametric, DNP, and cDNP scores on Fishyscapes Lost\&Found samples. As seen in other examples, the parametric (LogSumExp) scores suffer from frequent false positives, especially in correspondence of unusual terrain textures, which don't affect the nearest-neighbor based scores.

\section{Comparison of Parametric OoD Scoring Functions}
Here we compare the known parametric scoring functions which are available in the literature for dense OoD detection: maximum-softmax-probability~\cite{hendrycks2017a} (MSP), prediction entropy~\cite{hendrycks2019benchmark} (H), maximum-logit~\cite{hendrycks2019benchmark} (ML), and LogSumExp~\cite{Grathwohl2020Your,Tian2021} (LSE).

The results of the comparison, reported in Table~\ref{tab:parametric_scores}, reveal the superiority of maximum-logit and LogSumExp scores, with the latter outperforming the former.

\begin{table}
    \centering
    \scriptsize
    \begin{tabular}{l|l|c|c|c|c}
    \toprule
    Model & Scores & \multicolumn{2}{c|}{Parametric} & \multicolumn{2}{c}{cDNP} \\
    & & AP & FPR$_{95}$ & AP & FPR$_{95}$ \\
    \midrule
    UperNet-ConvNexT-T & MSP &  23.53 & 66.05 & 29.54 & 45.52 \\
    UperNet-ConvNexT-T & H &   28.34 & 65.79 & 34.33 & 45.67 \\
    UperNet-ConvNexT-T & ML &   39.31 & 59.50 & 43.53 & 41.12\\
    UperNet-ConvNexT-T & LSE &  40.04 & 59.43 & 44.02 & 40.83\\
    \midrule
    Segmenter-ViT-S & MSP &  35.23 & 41.63 & 63.22 & 27.18\\
    Segmenter-ViT-S & H &  45.10 & 40.26 & 70.44 & 25.76\\
    Segmenter-ViT-S & ML &  51.58 & 35.16 & 78.25 & 20.59\\
    Segmenter-ViT-S & LSE &  56.39 & 34.54 & 79.42 & 19.74\\
    \bottomrule
    \end{tabular}
    \caption{Comparison results for different parametric scoring functions on RoadAnomaly. We report the parametric performance and the final combined one (cDNP). LogSumExp (LSE) performs best, followed by maximum-logit (ML).}
    \label{tab:parametric_scores}
\end{table}

\section{Alternative Distance Functions for kNNs}
In this section we report the performance of kNNs/DNP using other distance functions than the $L_2$/Euclidean used throughout the paper. In particular, we consider $L_1$ distance and cosine similarity. We choose the former since low-order distance functions have been reported to mitigate the effects of the curse of dimensionality, and the latter because the transformer features we consider are part of the scaled-dot-product attention mechanism.

The results, shown in Table~\ref{tab:dist_sim_fns} for Segmenter-ViT-B on RoadAnomaly, reveal that $L_1$ and $L_2$ perform very similarly, both significantly better than cosine similarity. 

\begin{table}[]
    \centering
    \begin{tabular}{l|c|c}
        Dist./sim. & AP & FPR$_{95}$ \\
        \midrule
        cosine & 80.71	& 13.93 \\
        $L_1$ & 85.29	& 8.32 \\
        $L_2$ & 85.83	& 8.26 \\
    \end{tabular}
    \caption{Results for DNP on RoadAnomaly, using Segmenter-ViT-B features and different distance/similarity functions in the embedded space.}
    \label{tab:dist_sim_fns}
\end{table}

\section{Training Details}
For the experiments on Cityscapes we use a batch size of 8 and randomly crop the input image and ground truth to 769$\times$769 pixels. For StreetHazards we use a batch size of 4 and a crop size of 512.

The optimization algorithms and learning rates are the same for both datasets, and are listed in Table~\ref{tab:optim}, along with the segmentation performance of each architecture on the in-distribution validation sets.
UperNet-ConvNeXt performs best on semantic segmentation. While SegFormer-MiT and Segmenter-ViT have inferior mIoUs, the other transformer-based model (SETR) has a competitive segmentation performance and a state-of-the-art OoD detection performance with cDNP.

\begin{table}[]
    \centering
    \scriptsize
    \begin{tabular}{l|c|c|c|c|c|c}
    \toprule
    Model & optimizer & LR & \multicolumn{2}{c|}{mIoU} & \multicolumn{2}{c}{cDNP-AP} \\
    & & & CS & SH & RA & SH \\
        \midrule
        UperNet-ResNet50      & SGD   & $10^{-2}$ & 78 & 66 & 34 & 25\\
        UperNet-ConvNeXt-T    & AdamW & $10^{-4}$ & 81 & 72 & 47 & 27\\
        SegFormer-MiT-B3      & AdamW & $10^{-4}$ & 72 & 69 & 78 & 37\\
        Segmenter-ViT-S       & SGD   & $10^{-3}$ & 72 & 61 & 80 & 44\\
        SETR-Naive-ViT-L\tablefootnote{From: \texttt{https://github.com/open-mmlab/}\\\texttt{mmsegmentation/blob/master/configs/setr/README.md}}      & SGD   & $10^{-2}$ & 80 & - & 86 & -\\
        \bottomrule
    \end{tabular}
    \caption{Overview of the optimization details -- algorithm and learning rate -- and semantic segmentation (in-distribution) performance for the considered architectures in terms of mIoU on Cityscapes (CS) and StreetHazards validation (SH). We also report the out-of-distribution detection performance on RoadAnomaly (RA) and StreetHazards test (SH).}
    \label{tab:optim}
\end{table}

\end{document}